# Forward Super-Resolution: How Can GANs Learn Hierarchical Generative Models for Real-World Distributions

Zeyuan Allen-Zhu
zeyuan2023@allen-zhu.com
Allen-Zhu Research

Yuanzhi Li
Yuanzhi.Li@mbzuai.ac.ae
Mohamed bin Zayed University of AI

**Abstract**

Generative adversarial networks (GANs) are among the most successful models for learning high-complexity, real-world distributions. However, in theory, due to the highly non-convex, non-concave landscape of the minmax training objective, GAN remains one of the least understood deep learning models. In this work, we formally study how GANs can efficiently learn certain hierarchically generated distributions that are close to the distribution of real-life images. We prove that when a distribution has a structure that we refer to as *forward super-resolution*, then simply training generative adversarial networks using stochastic gradient descent ascent (SGDA) can learn this distribution efficiently, both in sample and time complexities. We also provide empirical evidence that our assumption "forward super-resolution" is very natural in practice, and the underlying learning mechanisms that we study in this paper (to allow us efficiently train GAN via SGDA in theory) simulates the actual learning process of GANs on real-world problems.

*V1 appeared on this date and we polished writing in V2. An extended abstract of this paper has appeared in ICLR 2023.

# 1 Introduction

Generative adversarial networks (GANs) [52] are among the successful models for learning high-complexity, real-world distributions. In practice, by training a *min-max* objective with respect to a generator and a discriminator consisting of multi-layer neural networks, using simple local search algorithms such as stochastic gradient descent ascent (SGDA), the *generator* can be trained *efficiently* to generate samples from complicated distributions (such as the distribution of images). But, from a theoretical perspective, how can GANs learn these distributions efficiently given that learning much simpler ones are already computationally hard [33]?

Answering this in full can be challenging. However, following the tradition of learning theory, one may hope for discovering some concept class consisting of non-trivial target distributions, and showing that using SGDA on a min-max generator-discriminator objective, not only the training converges in poly-time (a.k.a. trainability), but more importantly, the generator learns the target distribution to good accuracy (a.k.a. learnability). To this extent, we believe prior theory works studying GANs may still be somewhat inadequate.

- Some existing theories focus on properties of GANs at the *global-optimum* [15, 16, 20, 91]; while it remains unclear how the training process can find such global optimum efficiently.
- Some theories focus on the trainability of GANs, in the case when the loss function is convex-concave (so a global optimum can be reached), or when the goal is only to find a critical point [39, 40, 51, 56, 74, 76, 78, 81, 82]. Due to non-linear neural networks used in practical GANs, it is highly unlikely that the min-max training objective is convex-concave. Also, it is unclear whether such critical points correspond to learning certain non-trivial distributions (like image distributions).
- Even if the generator and the discriminator are linear functions over prescribed feature mappings— such as the neural tangent kernel (NTK) feature mappings [3, 8, 9, 17, 18, 32, 37, 42, 43, 50, 54, 57, 68, 72, 96, 101] — the training objective can still be non-convex-concave. [1]
- Some other works introduced notions such as proximal equilibria [44] or added gradient penalty [79] to improve training convergence. Once again, they do not study the "learnability" aspect of GANs. In particular, Chen et al. [34] even explicitly argue that min-max optimality may not directly imply distributional learning for GANs.
- Even worse, unlike supervised learning where some non-convex learning problems can be shown to haveno bad local minima [47], to the best of our knowledge, it still remains unclear what the qualities are of those critical points in GANs except in the most simple setting when the generator is a one-layer neural network [45, 65].

(We discuss some other related works in distributional learning, in Appendix A.)

Motivate by this *huge gap* between theory and practice, in this work, we make a preliminary step by showing that, when an image-like distribution is hierarchically generated (using an unknown $O(1)$-layered target generator) with a structural property that we refer to as *forward super-resolution*, then under certain mild regularity conditions, such distribution can be *efficiently* learned— both in sample and time complexity— by applying SGDA on a GAN objective.[2] Moreover, to justify the scope of our theorem, we provide empirical evidence that forward super-resolution *holds for practical image distributions*, and most of our regularity conditions hold in practice as well.

[1] Indeed, the discriminator takes the generator's output as input; although the NTK function is linear in weight parameters, it is extremely non-linear over the input space.

[2] Plus a simple SVD warmup initialization that is easily computable from the covariance of image patches.

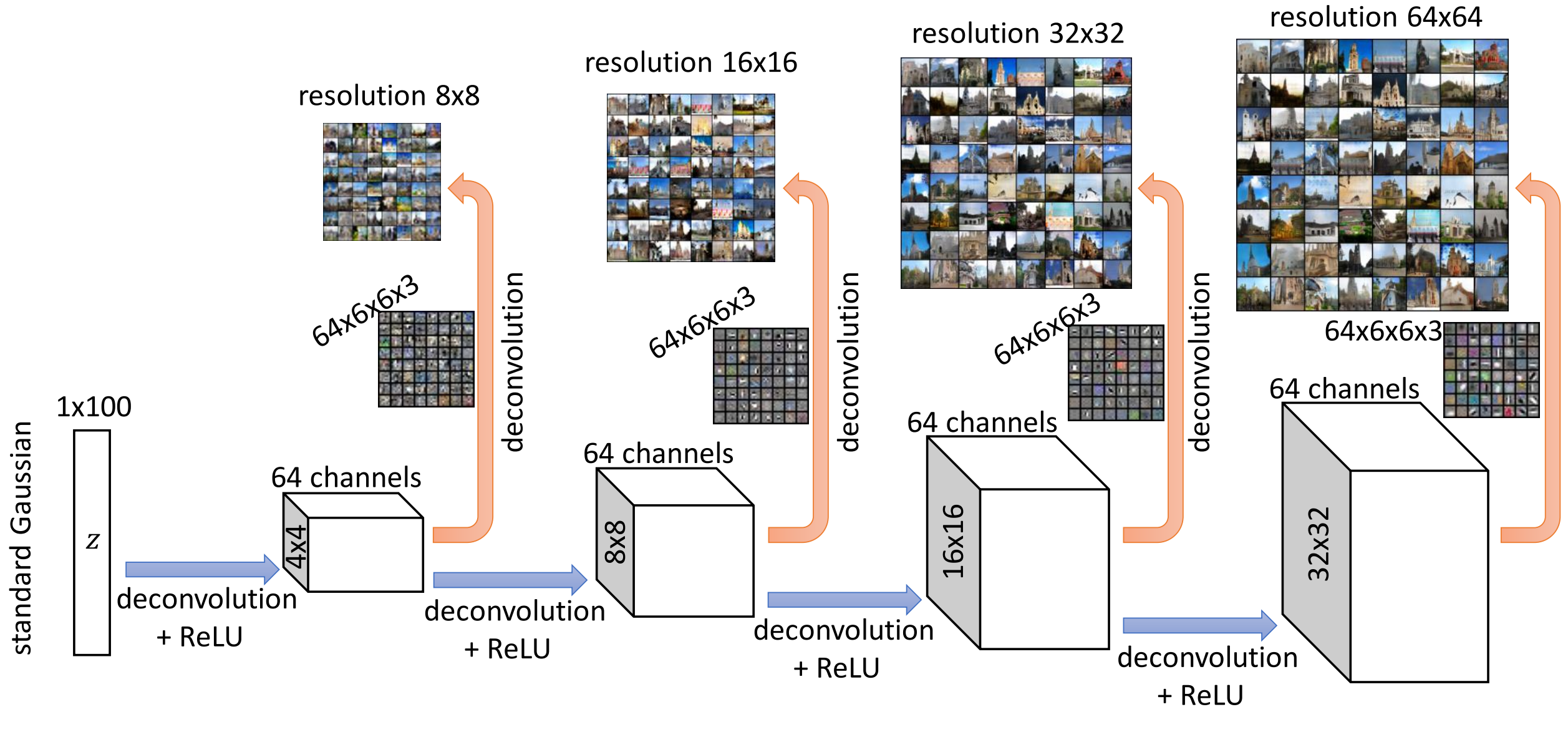

Figure 1: Illustration of the **forward super-resolution structure**. Church images generated by 4-hidden-layer deconvolution network (DCGAN), trained on LSUN Church data set using multi-scaled gradient [60]. The structure of the generator is shown as above, and there is a ReLU activation between each layers. We use simple average pooling to construct low resolution images from the original training images.

We believe our work extends the scope of traditional distribution learning theory to the regime of learning continuous, complicated real-world distributions such as the distribution of images, which are often generated through some *hierarchical generative models*. We draw connections between traditional distribution learning techniques such as method of moments to the generator-discriminator framework in GANs, and shed lights on what GANs are doing beyond these techniques.

## 1.1 Forward Super-Resolution: A Special Property of Images

Real images can be viewed in multiple resolutions without losing the semantics. In other words, the resolution of an image can be greatly reduced (e.g. by taking the average of nearby pixels), while still keeping the structure of the image. Motivated by this observation, the seminal work of Karras et al. [61] proposes to train a generator progressively: the lower levels of the generator are trained first to generate the lower-resolution version of images, and then the higher levels are gradually trained to generate higher and higher resolution images. In our work, we formulate this property of images as what we call *forward super-resolution*:

**Forward super-resolution property (mathematical statement see Section 2.1):**

There exists a generator $G$ as an $L$-hidden-layer neural network with ReLU activation, where each $G_\ell$ represent the hidden neuron values at layer $\ell$, and there exists matrices $\mathbf{W}_\ell$ such that

$$\text{the distribution of images at resolution level } \ell \text{ is given by } \mathbf{W}_\ell G_\ell$$

and the randomness is taken over the randomness of the input to $G$ (usually standard Gaussian).

In plain words, we assume there is an (unknown) neural network $G$ whose hidden layer $G_\ell$ can be used to generate images of resolution level $\ell$ (larger $\ell$ means better resolution) via a linear transformation, typically a deconvolution. We illustrate that this assumption holds on practical GAN training in Figure 1. This assumption is also made in the practical work [61]. Moreover, there is a body of works that directly use GANs or deconvolution networks for super-resolution [31, 64, 75, 94, 98].

# 2 Problem Setup

Throughout this paper, we use $a = \mathsf{poly}(b)$ for $a > 0, b > 1$ to denote that there are absolute constants $C_1 > C_2 > 0$ such that $b^{C_2} < a < b^{C_1}$. For a target learning error $\varepsilon \in [\frac{1}{d^{\omega(1)}}, \frac{1}{\mathsf{poly}(d)}]$, we use "w.h.p." to indicate with probability $\geq 1 - \frac{1}{(d/\varepsilon)^{\omega(1)}}$. Recall $\mathsf{ReLU}(z) = \max\{z, 0\}$. In this paper, for theoretical purpose we consider a smoothed version $\widetilde{\mathsf{ReLU}}(z)$ and a leaky version $\mathsf{LeakyReLU}(z)$. We give their details in Appendix B, and they are different from $\mathsf{ReLU}(z)$ only by a sufficiently small quantity $1/\mathsf{poly}(d/\varepsilon)$.

## 2.1 The Target Distribution: Forward Super-Resolution Structure

We consider outputs (think of them as images) $\{X_\ell^\star\}_{\ell \in [L]}$, where $X_L^\star$ is the final output, and $X_\ell^\star$ is the "low resolution" version of $X_L^\star$, with $X_1^\star$ having the lowest resolution. We think of each $\ell$-resolution image $X_\ell^\star$ consists of $d_\ell$ patches (for example, an image of size $36 \times 36$ contains 36 patches of size $6 \times 6$), where $X_\ell^\star = (X_{\ell,j}^\star)_{j \in [d_\ell]}$ and each $X_{\ell,j}^\star \in \mathbb{R}^d$. Typically, such "resolution reduction" from $X_L^\star$ to $X_\ell^\star$ can be given by sub-sampling, average pooling, Laplacian smoothing, etc., but we do not consider any specific form of resolution reduction in this work, as it does not matter for our main result to hold.

Formally, we define the **forward super-resolution** property as follows. We are given samples of the form $G^\star(z) = (X_1^\star, X_2^\star, \cdots, X_L^\star)$, where each $X_\ell^\star$ is generated by an **unknown** target neural network $G^\star(z)$ at layer $\ell$, with respect to a standard Gaussian $z \sim \mathcal{N}(0, \mathbf{I}_{m_0 \times m_0})$.

- The basic resolution: for every $j \in [d_1]$,

$$X_{1,j}^\star = \mathbf{W}_{1,j}^\star \mathcal{S}_{1,j}^\star \in \mathbb{R}^d \quad \text{for} \quad \mathcal{S}_{1,j}^\star = \mathcal{S}_{1,j}^\star(z) = \mathsf{ReLU}(\mathbf{V}_{1,j}^\star z - b_{1,j}^\star) \in \mathbb{R}_{\geq 0}^{m_1}$$

  where $\mathbf{V}_{1,j}^\star \in \mathbb{R}^{m_1 \times m_0}$, $b_{1,j}^\star \in \mathbb{R}^{m_1}$ and for simplicity we assume $\mathbf{W}_{1,j}^\star \in \mathbb{R}^{d \times m_1}$ is column orthonormal.

- For every $\ell > 1$, the image patches at resolution level $\ell$ are given as: for every $j \in [d_\ell]$,

$$X_{\ell,j}^\star = \mathbf{W}_{\ell,j}^\star \mathcal{S}_{\ell,j}^\star \in \mathbb{R}^d \quad \text{for} \quad \mathcal{S}_{\ell,j}^\star = \mathsf{ReLU}\left(\sum_{j' \in \mathcal{P}_{\ell,j}} \mathbf{V}_{\ell,j,j'}^\star \mathcal{S}_{\ell-1,j'}^\star - b_{\ell,j}^\star\right) \in \mathbb{R}_{\geq 0}^{m_\ell}$$

  where $\mathbf{V}_{\ell,j,j'}^\star \in \mathbb{R}^{m_\ell \times m_{\ell-1}}$, $b_{\ell,j}^\star \in \mathbb{R}^{m_\ell}$, and for simplicity we assume $\mathbf{W}_{\ell,j}^\star \in \mathbb{R}^{d \times m_\ell}$ is column orthonormal. Here, $\mathcal{P}_{\ell,j} \subseteq [d_{\ell-1}]$ can be any subset of $[d_{\ell-1}]$ to describe the connection graph.

**Remark.** For every layer $\ell$, $j \in [d_\ell]$, $r \in [m_\ell]$, one should view of each $[\mathcal{S}_{\ell,j}^\star]_r$ as the **$r$-th channel** in the **$j$-th patch** at layer $\ell$.

One should think of $\sum_{j' \in \mathcal{P}_{\ell,j}} \mathbf{V}_{\ell,j,j'}^\star \mathcal{S}_{\ell-1,j'}^\star$ as the linear "deconvolution" operation over hidden layers. When the network is a deconvolutional network such as in DCGAN [85], we have all $\mathbf{W}_{\ell,j}^\star = \mathbf{W}_\ell^\star$; but we do not restrict ourselves to this case. As illustrated in Figure 2, we should view $\mathbf{W}_{\ell,j}^\star$ as a matrix consisting of the "edge-color" features to generate image patches. Crucially, when we get a data sample $G^\star(z) = (X_1^\star, X_2^\star, \cdots, X_L^\star)$, the learning algorithm **does not know** the underlying $z$ used for this sample.

Although our analysis holds in many settings, for simplicity, in this paper we focus on the following parameter regime (for instance, $d_\ell$ can be $d^\ell$):

**Setting 2.1.** *$L = O(1)$, each $m_\ell = \mathsf{poly}(d)$, each $d_\ell = \mathsf{poly}(d)$, and each $\|\mathbf{V}_{\ell,j,j'}^\star\|_F \leq \mathsf{poly}(d)$.*

To efficient learn a distribution with the "forward super-resolution" structure, we assume that the true distribution in each layer of $G^\star$ satisfies the following "sparse coding" structure:

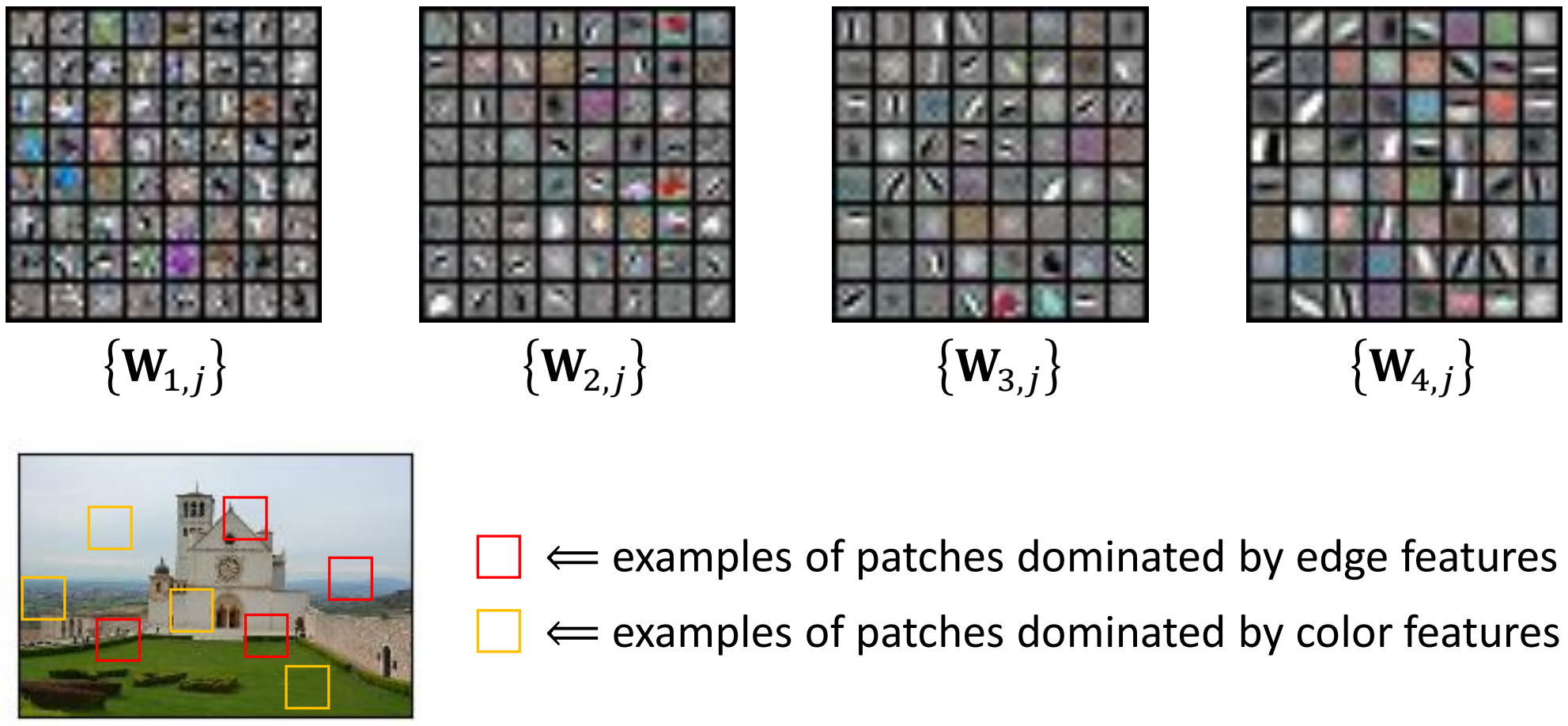

Figure 2: Visualization of the ***edge-color features*** learned in the output layers of $G^\star$. Each $\mathbf{W}_{\ell,j}$ is of dimension $m_\ell \times d = 64 \times 108 = 64 \times (6 \times 6 \times 3)$. The network is trained as in Figure 1. Note: For a deconvolutional output layer, all $\mathbf{W}_{\ell,j}$'s are equal for all $j \in [m_\ell]$.

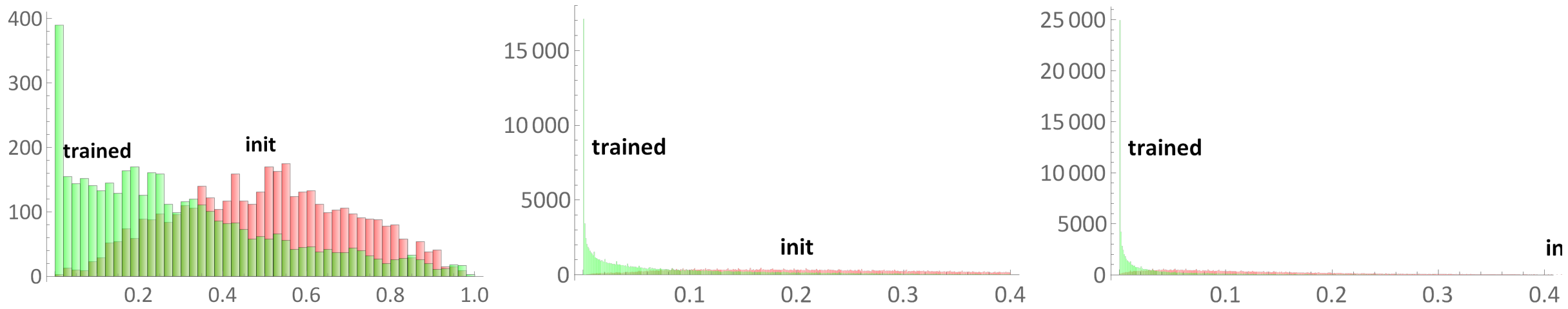

histogram of $\mathbf{Pr}[[\mathcal{S}^\star_{2,j}]_p > 0]$ of $\mathbf{Pr}[[\mathcal{S}^\star_{2,j}]_p > 0, [\mathcal{S}^\star_{2,j}]_q > 0]$ histogram of $\mathbf{Pr}[[\mathcal{S}^\star_{2,j}]_p > 0, [\mathcal{S}^\star_{2,j}]_q > 0, [\mathcal{S}^\star_{2,j}]_r > 0]$

Figure 3: Histograms at random init vs. after training for layer $\ell = 2$ of the architecture in Figure 1. Experiments for other layers can be found in Figure 9. It shows the learned network has sparse, not-too-positively correlated hidden activations (we did not regularize sparsity or correlation during training). Thus, it can be reasonable to assume that the activations of the *target network* are also sparse.

**Assumption 2.2** (sparse coding structure)**.** *For every $\ell \in [L], j \in [d_\ell], p \in [m_\ell]$, there exists some $k_\ell \ll m_\ell$ with $k_\ell \in \left[\Omega(\log m_\ell), m_\ell^{o(1)}\right]$ such that— recalling $\mathcal{S}^\star_{\ell,j} \geq 0$ is a non-negative vector:*[3]

$$\mathbf{Pr}_{z\sim\mathcal{N}(0,\mathbf{I})}\left[[\mathcal{S}^\star_{\ell,j}]_p > 0\right] \leq \frac{\mathsf{poly}(k_\ell)}{m_\ell}, \quad \mathbb{E}_{z\sim\mathcal{N}(0,\mathbf{I})}\left[[\mathcal{S}^\star_{\ell,j}]_p\right] \geq \frac{1}{\mathsf{poly}(k_\ell) m_\ell}$$
$$\textit{w.h.p. over } z: \quad \|\mathcal{S}^\star_{\ell,j}\|_\infty \leq \mathsf{poly}(k_\ell), \quad \|\mathcal{S}^\star_{\ell,j}\|_0 \leq k_\ell$$

*Moreover, we within the same patch, the channels are pair-wise and three-wise "not-too-positively correlated": $\forall p, q, r \in [m_\ell]$, $p \neq q \neq r$:*

$$\mathbf{Pr}_z\left[[\mathcal{S}^\star_{\ell,j}]_p > 0, [\mathcal{S}^\star_{\ell,j}]_q > 0\right] \leq \varepsilon_1 = \frac{\mathsf{poly}(k_\ell)}{m_\ell^2}, \quad \mathbf{Pr}_z\left[[\mathcal{S}^\star_{\ell,j}]_p > 0, [\mathcal{S}^\star_{\ell,j}]_q > 0, [\mathcal{S}^\star_{\ell,j}]_r > 0\right] \leq \varepsilon_2 = \frac{1}{m_\ell^{2.01}}$$

*Remark* 2.3. Although we have borrowed the notion of sparse coding, our task is very different from traditional sparse coding. We discuss more in Appendix A.

**Sparse coding structure in practice.** The sparse coding structure is very natural in practice for generating images [53, 99]. As illustrated in Figure 2, typically, after training, the output layer of the generator network $\mathbf{W}_{\ell,j}$ forms edge-color features. It is known that such edge-color features are indeed a (nearly orthogonal) basis for images, under which the coefficients are indeed *very sparse*. We refer to [5] for concrete measurement of the sparsity and orthogonality. The "not-too-positive correlation" property is also very natural: for instance, in an image patch if an edge feature is used, it is less likely that a color feature shall be used (see Figure 2). In Figure 3, we demonstrate that for some learned generator networks, the activations indeed become sparse and "not-too-positively correlated" after training.

[3]Here, $\mathsf{poly}(k_\ell)$ can be an arbitrary polynomial such as $(k_\ell)^{100}$, and our final theorem holds for sufficiently large $d$ because $d^{o(1)} > \mathsf{poly}(k_\ell)$.

Crucially, we have *only* assumed that channels are not-too-positively correlated *within a single patch*, and channels across different patches (e.g $\mathcal{S}^\star_{\ell,1}$ and $\mathcal{S}^\star_{\ell,2}$) can be arbitrarily dependent. This makes sure the global structure of the images can still be quite arbitrary, so Assumption 2.2 can *indeed be reasonable.*[4]

**Missing details.** We also make mild non-degeneracy and anti-concentration assumptions, and give examples for networks satisfying our assumptions. We defer them to Appendix B on Page 19.

## 2.2 Learner Network (Generator)

We use a learner network (generator) that has the same structure as the (unknown) target network:

- The image of the first resolution is given by:

$$X_{1,j} = \mathbf{W}_{1,j}\mathcal{S}_{1,j} \in \mathbb{R}^d \quad \text{for} \quad \mathcal{S}_{1,j} = \mathsf{LeakyReLU}(\mathbf{V}_{1,j} z - b_{1,j}) \in \mathbb{R}^{m_1}$$

  for $\mathbf{W}_{1,j} \in \mathbb{R}^{d\times m_1}$, $\mathbf{V}_{1,j} \in \mathbb{R}^{m_1 \times m_0'}$ with $m_0' \geq 2 d_1 m_1$.

- The image of higher resolution is given by:

$$X_{\ell,j} = \mathbf{W}_{\ell,j}\mathcal{S}_{\ell,j} \in \mathbb{R}^d \quad \text{for} \quad \mathcal{S}_{\ell,j} = \mathsf{LeakyReLU}\left(\sum_{j'\in\mathcal{P}_{\ell,j}} \mathbf{V}_{\ell,j,j'}\mathcal{S}_{\ell-1,j'} - b_{\ell,j}\right) \in \mathbb{R}^{m_\ell}$$

  for $\mathbf{W}_{\ell,j} \in \mathbb{R}^{d\times m_\ell}$ and $\mathbf{V}_{\ell,j} \in \mathbb{R}^{m_\ell \times m_{\ell-1}}$.

One can view $\mathcal{S}_\ell$ as the $\ell$-th hidden layer. We use $G_\ell(z)$ to denote $(X_{\ell,j})_{j\in[d_L]}$. We point out both the target and the learner network we study here are standard deconvolution networks widely used in practice (such as the generator network for GANs).

## 2.3 Theorem Statement

This papers proves that by applying SGDA on a generator-discriminator objective (algorithm to be described in Section 3), we can learn the target distribution using the above generator network.

**Theorem 4.1.** *For every $d > 0$, every $\varepsilon \in [\frac{1}{d^{\omega(1)}}, \frac{1}{2}]$, letting $G(z) = (X_1(z), \dots, X_L(z))$ be the generator learned after running Algorithm 4 (which runs in time/sample complexity $\mathsf{poly}(d/\varepsilon)$), then w.h.p. there is a column orthonormal matrix $\mathbf{U} \in \mathbb{R}^{m_0 \times m_0'}$ such that*

$$\mathbf{Pr}_{z\sim\mathcal{N}(0,\mathbf{I}_{m_0'\times m_0'})}\left[\left\|G^\star(\mathbf{U}z) - G(z)\right\|_2 \leq \varepsilon\right] \geq 1 - \frac{1}{(d/\varepsilon)^{\omega(1)}} \ .$$

*In particular, this implies the 2-Wasserstein distance $\mathcal{W}_2(G(\cdot), G^\star(\cdot)) \leq \varepsilon$.*

[4] Within a patch, it is natural that the activations are not-too-positively correlated: for example, once a patch chooses to use a horizontal edge feature, it is *less likely* that it will pick up another vertical edge feature. We also point out that if $[\mathcal{S}^\star_{\ell,j}]_p$'s are all independent, then $\mathbf{Pr}[[\mathcal{S}^\star_{\ell,j}]_p > 0, [\mathcal{S}^\star_{\ell,j}]_q > 0] \approx \frac{1}{m_\ell^2} \leq \varepsilon_1$ and $\mathbf{Pr}[[\mathcal{S}^\star_{\ell,j}]_p > 0, [\mathcal{S}^\star_{\ell,j}]_q > 0, [\mathcal{S}^\star_{\ell,j}]_r > 0] \approx \frac{1}{m_\ell^3} \ll \varepsilon_2$.

# 3 Learning Algorithm

In this section, we define the learning algorithm using min-max optimization. We assume one access polynomially many (i.e., $\mathsf{poly}(d/\varepsilon)$) i.i.d. samples from the true distribution $X^\star = (X_1^\star, X_2^\star, \cdots, X_L^\star)$, generated by the (unknown) target network defined in Section 2.1.

To begin with, we use a simple SVD warm start to initialize (only) the output layers $\mathbf{W}_{\ell,j}$ of the network. It merely involves a simple estimator of certain truncated covariance of the data. We defer it to Algorithm 5 in Section C.1. Also, we refer stochastic gradient descent ascent SGDA (on the GAN objective) to an algorithm to optimize $\min_x \max_y f(x,y)$, where the inner maximization is trained at a faster frequency. For completeness' sake, see Algorithm 6 in Section C.2.

To make the learning process more clear, we *break the learning into multiple parts* and introduce them separately in this section:

- GAN_OutputLayer: to learn output matrices $\{\mathbf{W}_{\ell,j}\}$ per layer.
- GAN_FirstHidden: to learn hidden matrices $\{\mathbf{V}_{1,j}\}$ for the first layer.
- GAN_FowardSuperResolution: to learn higher-level hidden layers $\{\mathbf{V}_{\ell,j,j'}\}$.

We use different discriminators at different parts for our theory analysis, and shall characterize what discriminator does and how the generator can leverage the discriminator to learn the target distribution. We point out, although one can add up and mix those discriminators to make it a single one, how to use a same discriminator across the entire algorithm is an important open research direction.

At the end of this section, we shall explain how they are combined to give the final training process.

*Remark* 3.1. Although we apply an SVD algorithm to get a *warm start* on the output matrices $\mathbf{W}_{\ell,j}$, the majority of the learning of $\mathbf{W}_{\ell,j}$ (e.g., to any small $\varepsilon = \frac{1}{\mathsf{poly}(d)}$ error) is still done through gradient descent ascent. We point out that the seminal work on neurally plausible dictionary learning also considers such a warm start [13].

## 3.1 Learn the Output Layer

We first introduce the discriminator for learning the output layer. For each resolution $\ell \in [L]$ and patch $j \in [d_\ell]$, we consider a one-hidden-layer discriminator

$$D_{\ell,j}^{(1)}(Y) \stackrel{\text{def}}{=} \sum_{r\in[m_\ell]} \Big( \mathsf{ReLU}'([(\mathbf{W}_{\ell,j}^D)^\top Y_j]_r - \mathbb{b}) \langle Y_j, V_{\ell,j,r}^D \rangle \Big) \enspace ,$$

where the input is either $Y = X_\ell^\star$ (from the true distribution) or $Y = X_\ell$ (from the generator).

Above, on the discriminator side, we have default parameter $\mathbf{W}_{\ell,j}^D, \mathbb{b}$ and trainable parameters $V_{\ell,j}^D = (V_{\ell,j,r}^D)_{r\in[m_\ell]}$ where each $V_{\ell,j,r}^D \in \mathbb{R}^d$. On the generator side, we have trainable parameters $\mathbf{W}_{\ell,j}$ (which are used to calculate $X_\ell$). (We use superscript $^D$ to emphasize $\mathbf{W}_{\ell,j}^D$ are the parameters for the discriminator, to distinguish it from $\mathbf{W}_{\ell,j}$.)

In our pseudocode GAN_OutputLayer (see Algorithm 1), for fixed $\mathbf{W}_{\ell,j}^D, \mathbb{b}$, we perform gradient descent ascent on the GAN objective with discriminator $D_{\ell,j}^{(1)}$, to minimize over $V_{\ell,j}^D$ and maximize over $\mathbf{W}_{\ell,j}$. In our final training process (to be given in full in Algorithm 4), we shall start with some $\mathbb{b} \ll 1$ and periodically decrease it; and we shall periodically set $\mathbf{W}_{\ell,j}^D = \mathbf{W}_{\ell,j}$ to be the same as the generator from a previous check point.

- Simply setting $\mathbf{W}_{\ell,j}^D = \mathbf{W}_{\ell,j}$ involves *no additional learning*, as all the learning is still being done using gradient descent ascent.

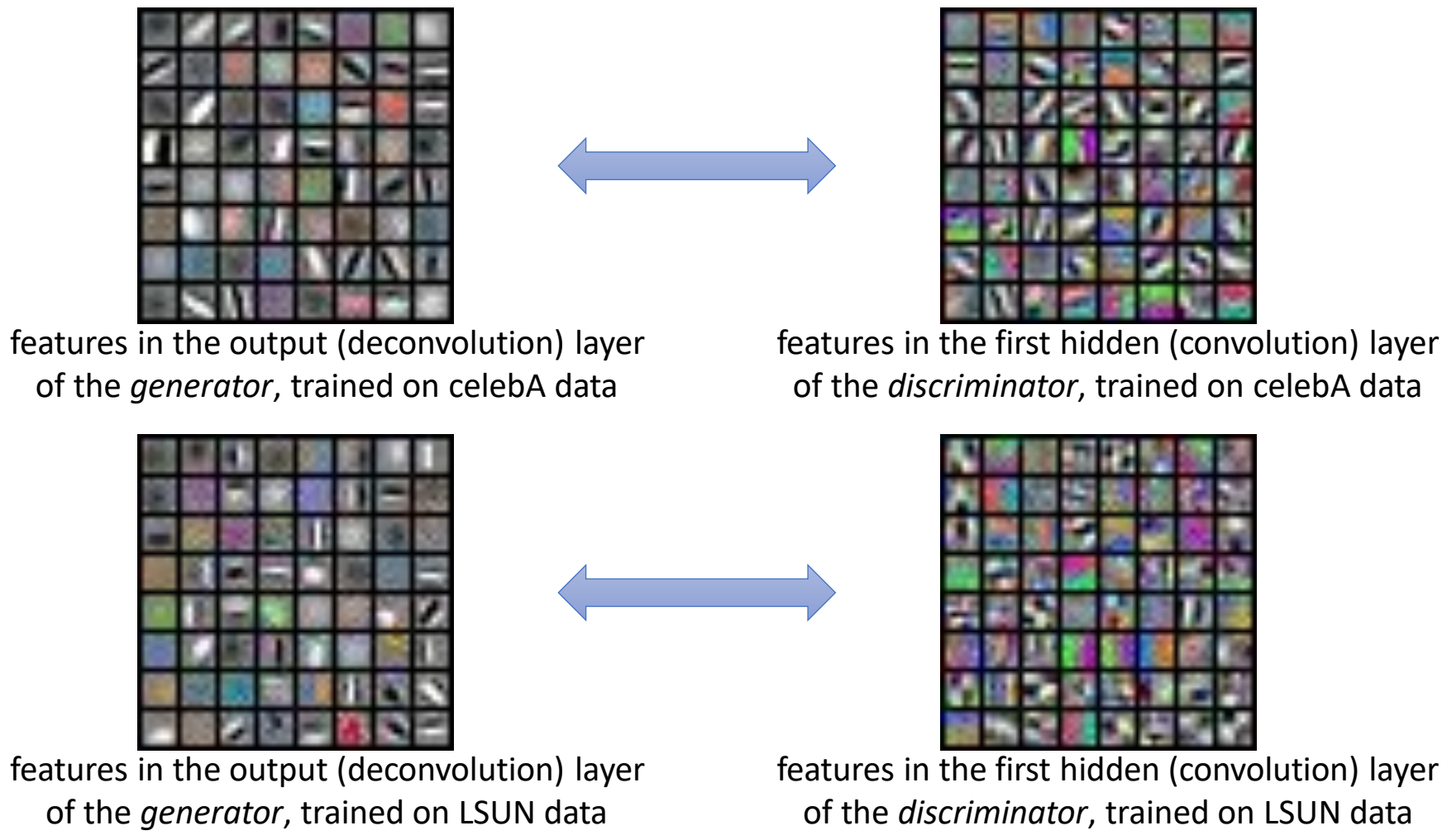

Figure 4: The first hidden layer in the discriminator is indeed learning edge color detectors, while the output layer of the generator is also learning edge color features. Here, by features we simply mean the hidden weights $\mathbf{W}_{\ell,j}$ and $\mathbf{W}^D_{\ell,j}$. In our DCGAN experiment, they are of size $64 \times 6 \times 6 \times 3$.

- In practice, the first hidden layer of the discriminator indeed learns the edge-color detectors (see Figure 4), similar to the edge-color features in the output layer of the generator. Thus, setting $\mathbf{W}^D_{\ell,j} = \mathbf{W}_{\ell,j}$ is *a reasonable approximation*. As we pointed out, how to analyze a discriminator that exactly matches practice is an important open theory direction.

---

**Algorithm 1** (GAN_OutputLayer) method of moments

---

**Input:** $\mathbf{W}^{(0)}_{\ell,j}, b, \ell, j$

1: Set default parameters $\mathbf{W}^D_{\ell,j} \leftarrow \mathbf{W}^{(0)}_{\ell,j}$; $\mathbb{b} \leftarrow bm^{0.152}$; $N \leftarrow \frac{1}{\mathsf{poly}(d/\varepsilon)}$, $\eta \leftarrow \frac{1}{\mathsf{poly}(d/\varepsilon)}$, $T \leftarrow \frac{\mathsf{poly}(d/\varepsilon)}{\eta}$

2: Set initialization $\mathbf{W}_{\ell,j} \leftarrow \mathbf{W}^{(0)}_{\ell,j}$ and $V^D_{\ell,j} \leftarrow 0$.

3: Apply SGDA (Algorithm 6) with $N$ samples, learning rate $\eta$ for $T$ steps on the following GAN objective (with $c$ being a small constant such as 0.001):

$$\min_{\mathbf{W}_{\ell,j}} \max_{V^D_{\ell,j}} \left( \left( \mathbb{E}[D^{(1)}_{\ell,j}(X^\star_\ell)] - \mathbb{E}[D^{(1)}_{\ell,j}(X_\ell)] \right) - \sum_{r \in [m_\ell]} \|V^D_{\ell,j,r}\|_2^{1+c} \right)$$

$\diamond$ $\|V^D_{\ell,j,r}\|_2^{1+c}$ *is an analog of the weight decay, which people use widely in practice*

4: $[\mathbf{W}_{\ell,j}]_p \leftarrow [\mathbf{W}_{\ell,j}]_p / \|[\mathbf{W}_{\ell,j}]_p\|_2$

---

**Intuition: what does the discriminator do?** To further understand the algorithm, we can see that for each $V^D_{\ell,j,r}$, when its norm is fixed, then the maximizer is obtained at

$$V^D_{\ell,j,r} \propto \left( \mathbb{E}[\mathsf{ReLU}'([(\mathbf{W}^D_{\ell,j})^\top X^\star_{\ell,j}]_r - b) X^\star_{\ell,j}] - \mathbb{E}[\mathsf{ReLU}'([(\mathbf{W}^D_{\ell,j})^\top X_{\ell,j}]_r - b) X_{\ell,j}] \right)$$

Thus, for the generator to further minimize the objective, the generator will learn to *match the moments of the true distribution*. In other words, generator wants to ensure

$$\mathbb{E}[\mathsf{ReLU}'([(\mathbf{W}^D_{\ell,j})^\top X_{\ell,j}]_r - b) X_{\ell,j}] \approx \mathbb{E}[\mathsf{ReLU}'([(\mathbf{W}^D_{\ell,j})^\top X^\star_{\ell,j}]_r - b) X^\star_{\ell,j}]$$

In this paper, we prove that such a truncated moment can be matched efficiently simply by running gradient descent ascent. Moreover, we empirically observe that *GANs can indeed do moment matching within each patch* even at the earlier stage of training, consistent with the work [66].

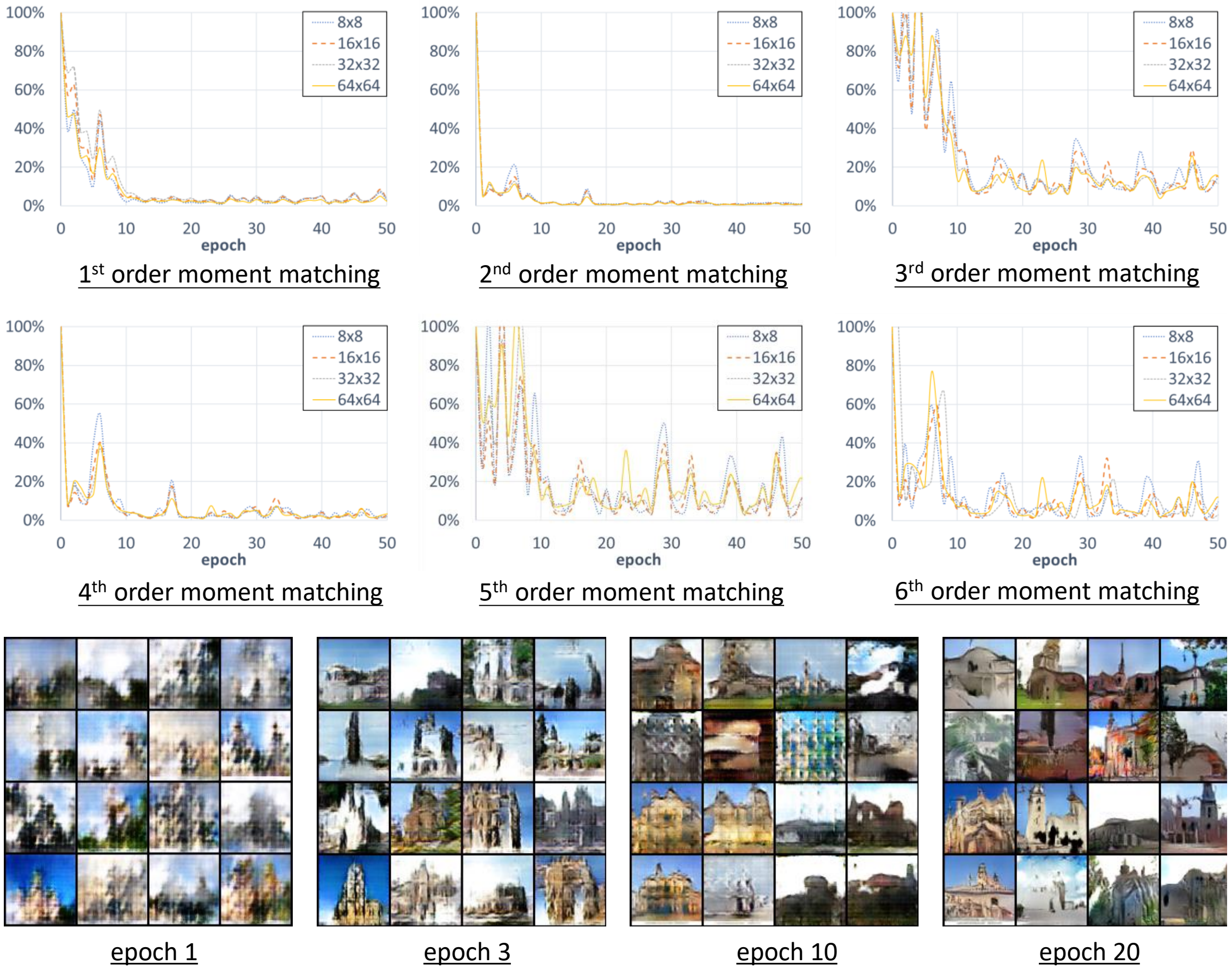

Figure 5: The difference between the moments of a generator's output and the true distribution, within each patch of size $6 \times 6$. Each $r$-th order moment of a patch $j$ at resolution level $\ell$ is given by: for $i_1 > i_2 > \cdots > i_r \in [m_\ell]$, $\mathbb{E}\left[\prod_{s \in [r]} \left([X_{\ell,j}]_{i_s} - \mathbb{E}[[X_{\ell,j}]_{i_s}]\right)\right]$. In the plots, we have averaged over each of these $\{i_s\}_{s \in [r]}$ in each patch, and computed the "squared difference" between the moments of the generator's output and the true distribution. In the plots, the $x$-axis is the number of epochs, and the $y$-axis represents this "squared difference" comparing to initialization. The generator network is trained as in Figure 1.
**Observation.** From these plots, one can see that the moments begin to match after epoch 10, so GAN's discriminator-generator framework is indeed doing moment matching at the earlier stage of training.

We plot the difference between the moments of the output of the generator vs. the moment of the true distribution in Figure 5. Method of moment is a classical technique used in distribution learning [10–12, 68, 80, 92], and as we show empirically, part of GANs training is indeed to match moments.

## 3.2 Learn the First Hidden Layer

To learn the first hidden layer weights $\mathbf{V}_1$ of the learner (generator) network, for simplicity, we re-parameterize $\mathbf{V}_{1,j}$ as $[\mathbf{V}_{1,j}]_r = \alpha_{1,j,r} \frac{[\mathbf{V}_{1,j}]_r}{\|[\mathbf{V}_{1,j}]_r\|_2}$, and learn $\alpha_{1,j,r}$ and $\frac{[\mathbf{V}_{1,j}]_r}{\|[\mathbf{V}_{1,j}]_r\|_2}$ separately using two discriminators $D^{(4)}$ and $D^{(5)}$.

Recall a discriminator takes as input an image $Y$: it can be either $Y = X_1^\star$ from the true distribution or $Y = X_1$ from the fake distribution. Like in the previous subsection, we fix default parameters $\{\mathbf{W}^D_{1,j}\}_{j \in [d_1]}$ and $b \ll 1$ in these discriminators (but shall update them periodically in the final algorithm).

For every $j \in [d_1], r \in [m_1]$, discriminator

$$D^{(4)}_{j,r}(Y) \stackrel{\text{def}}{=} \widetilde{\mathsf{ReLU}}\left([(\mathbf{W}^D_{1,j})^\top Y_j]_r - 2b\right) o_k(v^D_{j,r}) + \widetilde{\mathsf{ReLU}}'\left([(\mathbf{W}^D_{1,j})^\top Y_j]_r - b\right) o_k(w^D_{j,r})$$

where the trainable parameters are $v^D = (v^D_{j,r})_{j\in[d_1],r\in[m_1]}, w^D = (w^D_{j,r})_{j\in[d_1],r\in[m_1]}$; and $o_k(z) := \mathsf{poly}(k_1)\mathsf{ReLU}(z) - \mathsf{ReLU}(-z)$.

The discriminator

$$D^{(5)}(Y) \stackrel{\text{def}}{=} \sum_{(j,r)\neq(j',r')\in[d_1]\times[m_1]} C_{j,j',r,r'} o_d(u^D_{j,j',r,r'}) \widetilde{\mathsf{ReLU}}\left([(\mathbf{W}^D_{1,j})^\top Y_j]_r - b\right) \widetilde{\mathsf{ReLU}}\left([(\mathbf{W}^D_{1,j'})^\top Y_{j'}]_{r'} - b\right)$$

where the trainable parameters are $u^D = (u^D_{j,j',r,r',+}, u^D_{j,j',r,r',-})_{j,j',r,r'}$; $o_d(z) := z_+ - z_-$; and constants $C_{j,j',r,r'}$.[5]

We make a simplification during the training:[6]

- When we perform gradient descent ascent on each $D^{(4)}_{j,r}$, only $\alpha_{1,j,r}, b_{1,j}$ are trainable parameters on the generator side, and $v^D_{j,r}, w^D_{j,r}$ are the trainable parameters on the discriminator side.
- When we perform gradient descent ascent on $D^{(5)}$, only these $\frac{[\mathbf{V}_{1,j}]_r}{\|[\mathbf{V}_{1,j}]_r\|_2}$ unit vectors will be updated on the generator side, and $u^D$ gives the trainable parameters of the discriminator.

We also add $L_2$ regularizer on the discriminator side for $D^{(4)}$, and two barrier regularizers $H(u^D)$ and $R^{(5)}(\mathbf{V}_1)$ on the discriminator and generator side respectively for $D^{(5)}$ to ensure the parameters do not touch the "boundary" (details in Appendix F.4).

**High-Level Intuitions.** In the process of learning the lowest-resolution images $X^\star_1$, one cannot hope for (even approximately) learning the exact matrices $\mathbf{V}^\star_{1,j}$, or the exact function that maps from $z \mapsto X^\star_1$ (because $z$ is unknown during the training). Instead, the task is for learning the *distribution* of $X^\star_{1,j} = \mathbf{W}^\star_{1,j}\mathsf{ReLU}(\mathbf{V}^\star_{1,j}z - b^\star_{1,j})$.

Suppose for a moment that $\mathbf{W}^\star_{1,j}$ are already fully learned; then, it is perhaps not surprising that for the remaining part $\mathcal{S}^\star_{1,j} = \mathsf{ReLU}(\mathbf{V}^\star_{1,j}z - b^\star_{1,j})$, if we can somehow

1. learn the marginal distribution of $[\mathcal{S}^\star_{1,j}]_r$ for each $j, r$, and
2. learn the joint distribution of $\left([\mathcal{S}^\star_{1,j}]_r, [\mathcal{S}^\star_{1,j'}]_{r'}\right)$ for each pair $(j,r) \neq (j',r')$, then, we can recover the joint distribution of $\{[\mathcal{S}^\star_{1,j}]_r\}_{j,r}$. (As an analogy, for a joint Gaussian distribution, it suffices to learn the pair-wise correlation.)

The roles of $D^{(4)}$ and $D^{(5)}$ are precisely for the purpose of (1) and (2). In particular,

- $D^{(4)}$ discriminates the mismatch from one single neuron (e.g. sparsity, mean) by ensuring

$$\mathbb{E}\,\widetilde{\mathsf{ReLU}}\left([(\mathbf{W}^D_{1,j})^\top X_{1,j}]_r - b\right) \approx \mathbb{E}\,\widetilde{\mathsf{ReLU}}\left([(\mathbf{W}^D_{1,j})^\top X^\star_{1,j}]_r - b\right)$$
$$\mathbb{E}\,\widetilde{\mathsf{ReLU}}'\left([(\mathbf{W}^D_{1,j})^\top X_{1,j}]_r - b\right) \approx \mathbb{E}\,\widetilde{\mathsf{ReLU}}'\left([(\mathbf{W}^D_{1,j})^\top X^\star_{1,j}]_r - b\right)$$

  Furthermore, as long as $\mathbf{W}^D_{1,j}$ is moderately learned, the sparse coding structure shall ensure $(\mathbf{W}^D_{1,j})^\top X_{1,j} \approx \mathcal{S}_{1,j}$ and $(\mathbf{W}^D_{1,j})^\top X^\star_{1,j} \approx \mathcal{S}^\star_{1,j}$. For such reason, and using $b \ll 1$, applying gradient descent ascent using discriminator $D^{(4)}_{j,r}$, in fact guarantees

$$\mathbb{E}\,\widetilde{\mathsf{ReLU}}\left([\mathcal{S}_{1,j}]_r\right) \approx \mathbb{E}\,\widetilde{\mathsf{ReLU}}\left([\mathcal{S}^\star_{1,j}]_r\right) \quad \text{and} \quad \mathbb{E}\,\widetilde{\mathsf{ReLU}}'\left([\mathcal{S}_{1,j}]_r\right) \approx \mathbb{E}\,\widetilde{\mathsf{ReLU}}'\left([\mathcal{S}^\star_{1,j}]_r\right)$$

[5] We shall choose $C_{j,j,r,r'} = \frac{m_1^2}{b\mathsf{poly}(k_1)}$ for $r \neq r'$ and $C_{j,j',r,r'} = \frac{1}{\mathsf{poly}(m_1)}$ for $j \neq j'$.

[6] One can train them together, but we separate them to make the goal more clear and the analysis simpler.

Recall $[\mathcal{S}^\star_{1,j}]_r$ behaves as $\mathsf{ReLU}(g)$ for $g \sim \mathcal{N}(-\mu, \sigma^2)$ and has only 2 degrees of freedom; thus, matching moments on $\widetilde{\mathsf{ReLU}}$ and $\widetilde{\mathsf{ReLU}}'$ can learn the distribution of a single neuron $[\mathcal{S}^\star_{1,j}]_r$.

- $D^{(5)}$ discriminates the mismatch from the moments across two neurons, by ensuring

$$\begin{aligned}&\mathbb{E}\left[\widetilde{\mathsf{ReLU}}\left([(\mathbf{W}^D_{1,j})^\top X_{1,j}]_r - b\right)\widetilde{\mathsf{ReLU}}\left([(\mathbf{W}^D_{1,j'})^\top X_{1,j'}]_{r'} - b\right)\right]\\ \approx\ &\mathbb{E}\left[\widetilde{\mathsf{ReLU}}\left([(\mathbf{W}^D_{1,j})^\top X^\star_{1,j}]_r - b\right)\widetilde{\mathsf{ReLU}}\left([(\mathbf{W}^D_{1,j'})^\top X^\star_{1,j'}]_{r'} - b\right)\right]\end{aligned}$$

For similar reason, gradient descent ascent learns to match moments on the cross terms:

$$\mathbb{E}\,\widetilde{\mathsf{ReLU}}\left([\mathcal{S}_{1,j}]_r\right)\widetilde{\mathsf{ReLU}}\left([\mathcal{S}_{1,j'}]_{r'}\right) \approx \mathbb{E}\,\widetilde{\mathsf{ReLU}}\left([\mathcal{S}^\star_{1,j}]_r\right)\widetilde{\mathsf{ReLU}}\left([\mathcal{S}^\star_{1,j'}]_{r'}\right)$$

We show this corresponds to learning $\langle [\mathbf{V}^\star_{1,j}]_r, [\mathbf{V}^\star_{1,j'}]_{r'}\rangle$ to a moderate accuracy.

In sum, if we apply $\mathsf{SGDA}$ on $D^{(4)}$ and $D^{(5)}$ together, we can hope for learning $\mathbf{V}_1$ up to a unitary transformation (see Lemma F.18). This ensures that we learn the distribution of $X^\star_1$.

*Remark* 3.2. The purpose to have the $o_k(z) := \mathsf{poly}(k_1)\mathsf{ReLU}(z) - \mathsf{ReLU}(-z)$ function in $D^{(4)}$ is to ensure during the training process, it is more likely to have $\widetilde{\mathsf{ReLU}}\left([(\mathbf{W}^D_{1,j})^\top X_{1,j}]_r - b\right)$ *greater than* (as opposed to less than) $\widetilde{\mathsf{ReLU}}\left([(\mathbf{W}^D_{1,j})^\top X^\star_{1,j}]_r - b\right)$; this ensures the neurons $[\mathcal{S}_{1,j}]_r$ in the generators do not die. (For instance, if the generator gives $X_{1,j} = 0$ the training will get stuck.

*Remark* 3.3. The purpose to have $C_{j,j',r,r'} \ll C_{j,j,r,r'}$ in $D^{(5)}$ is to ensure that the moments of cross terms within a single patch $j = j'$ are learned first. We want to learn them first because the "not-so-correlated" assumption in Assumption 2.2 shall ensure the learning process is time efficient.

---

**Algorithm 2** ($\mathsf{GAN_FirstHidden}$) method of moments for first hidden layer

---

**Input:** $\mathbf{W}_1^{(0)}, b$

Set default parameters $\mathbf{W}_1^D \leftarrow \mathbf{W}_1^{(0)}$; $N \leftarrow \frac{1}{\mathsf{poly}(d/\varepsilon)}$, $\eta \leftarrow \frac{1}{\mathsf{poly}(d/\varepsilon)}$, $T \leftarrow \frac{\mathsf{poly}(d/\varepsilon)}{\eta}$

Initialize $[\mathbf{V}_{1,j}]_r \leftarrow \alpha \mathbf{e}_{j\cdot m_\ell + r}$ and $[b_{1,j}]_r = \beta$ for some $\alpha, \beta = \mathsf{poly}(k_1)$.[7]

**for** $j \in [d_1], r \in [m_1]$ **do**

  Apply $\mathsf{SGDA}$ with $N$ samples, learning rate $\eta$ for $T$ steps on the following GAN objective

$$\min_{\alpha_{1,j,r},[b_{1,j}]_r} \max_{v^D_{j,r}, w^D_{j,r}} \left(\mathbb{E}[D^{(4)}_{j,r}(X^\star_1)] - \mathbb{E}[D^{(4)}_{j,r}(X_1)]\right) - |v^D_{j,r}|^2 - |w^D_{j,r}|^2$$

**end for**

Apply $\mathsf{SGDA}$ with $N$ samples, learning rate $\eta$ for $T$ steps on the following GAN objective

$$\min_{\{\mathbf{V}_{1,j}\}_{j\in[d_1]}} \max_{u^D} \left(\mathbb{E}[D^{(5)}(X^\star_1)] - \mathbb{E}[D^{(5)}(X_1)]\right) + R^{(5)}(\mathbf{V}_1) - H(u^D)$$

---

## 3.3 Learn Higher Hidden Layers

For resolution $\ell > 1$, patch $j \in [d_\ell]$, channel $r \in [m_\ell]$, to learn $[\mathbf{V}^\star_{\ell,j}]_r$, we introduce discriminator $D^{(2)}_{\ell,j,r}(Y_1, Y_2)$. It takes as input images of two resolutions: one should think of either $(Y_1, Y_2) =$

[7]We choose $\alpha = \mathsf{poly}(k_1)$ and $\beta = \alpha \cdot \sqrt{2\log(m_1 \mathsf{poly}(k_1))}$ and this ensures $\mathbb{E}[[\mathcal{S}_{1,j}]_r] = \frac{\mathsf{poly}(k_1)}{m_1}$ and $\mathbf{Pr}[[\mathcal{S}_{1,j}]_r > 0] = \frac{\mathsf{poly}(k_1)}{m_1}$. See Proposition F.3. This initialization also ensures $[\mathcal{S}_{1,j}]_r$ are independent for different $(j, r)$ pairs.

$(X^\star_\ell, X^\star_{\ell-1})$ comes from the true distribution, or $(Y_1, Y_2) = (X_\ell, X_{\ell-1})$ from the generator.

$$D^{(2)}_{\ell,j,r}(Y_1, Y_2) \stackrel{\text{def}}{=} \widetilde{\mathsf{abs}}\left(s_r - \mathsf{LeakyReLU}(\mathfrak{s}_r)\right)$$
$$\text{where} \quad \widetilde{\mathsf{abs}}(x) := \widetilde{\mathsf{ReLU}}(x-b) + \widetilde{\mathsf{ReLU}}(-x-b)$$
$$s_r := \left[\left([\mathbf{W}^D_{\ell,j}]^\top Y_{1,j}\right)\right]_r$$
$$\mathfrak{s}_r := \left(\sum_{j'\in\mathcal{P}_{\ell,j}} \mathbf{V}^D_{\ell,j,j'} \mathsf{LeakyReLU}\left([\mathbf{W}^D_{\ell-1,j'}]^\top Y_{2,j'}\right) - b^D_{\ell,j}\right)_r$$

Above, again $\mathbf{W}^D_{\ell,j}$,$\{\mathbf{W}^D_{\ell-1,j'}\}_{j'\in[d_{\ell-1}]}$, $b$ are default parameters (changed only periodically).

On the discriminator side, $\{[\mathbf{V}^D_{\ell,j,j'}]_r\}_{j'\in\mathcal{P}_{\ell,j}}, [b^D_{\ell,j}]_r$ are the actual trainable parameters; on the generator side, $\{[\mathbf{V}_{\ell,j,j'}]_r\}_{j'\in\mathcal{P}_{\ell,j}}, [b_{\ell,j}]_r$ as the trainable parameters. We note this discriminator $D^{(2)}$ is a **three-hidden layer neural network**. Yet, we show that such an network (together with the generator) can still be trained efficiently using gradient descent ascent.

---

**Algorithm 3** (GAN_FowardSuperResolution) using super-resolution to learn higher hidden layers

---

**Input:** $\mathbf{W}^{(0)}_\ell, \mathbf{W}^{(0)}_{\ell-1}, b, \ell, j$

1: Set default parameters $\mathbf{W}^D_{\ell,j} \leftarrow \mathbf{W}^{(0)}_{\ell,j}, \mathbf{W}^D_{\ell-1,j'} \leftarrow \mathbf{W}^{(0)}_{\ell-1,j'}$;

2: $N \leftarrow \frac{1}{\mathsf{poly}(d/\varepsilon)}, \eta \leftarrow \frac{1}{\mathsf{poly}(d/\varepsilon)}, T \leftarrow \frac{\mathsf{poly}(d/\varepsilon)}{\eta}; \lambda_G, \lambda_D \leftarrow \frac{1}{\mathsf{poly}(d/\varepsilon)}$

3: Initialize $\mathbf{V}_{\ell,j,j'} = \mathbf{V}^D_{\ell,j,j'} = \mathbf{I}$ for one of $j' \in \mathcal{P}_{\ell,j}$ and setting others as zero. Initialize $b_{\ell,j} = 0$.

4: **for** $r \in [m_\ell]$ **do**

5: Apply SGDA with $N$ samples, learning rate $\eta$ for $T$ steps on the following GAN objective

$$\min_{\{[\mathbf{V}^D_{\ell,j,j'}]_r\}_{j'\in\mathcal{P}_{\ell,j}},[b^D_{\ell,j}]_r;} \max_{\{[\mathbf{V}_{\ell,j,j'}]_r\}_{j'\in\mathcal{P}_{\ell,j}},[b_{\ell,j}]_r} \left(\mathbb{E}[D^{(2)}_{\ell,j,r}(X^\star_\ell, X^\star_{\ell-1})] - \mathbb{E}[D^{(2)}_{\ell,j,r}(X_\ell, X_{\ell-1})]\right) - \lambda_G\|\mathbf{V}_\ell\|_F^2 + \lambda_D\|\mathbf{V}^D_\ell\|_F^2$$

6: $[b_{1,j}]_r \leftarrow [b_{1,j}]_r + \mathsf{poly}(k_1)b$.

---

**Intuition: what does the discriminator do?** In this case, applying gradient descent ascent on $D^{(2)}$ actually learns *how to "super-resolute" the image from resolution level* $\ell - 1$ *to level* $\ell$. In particular, the discriminator wants to find a way where the patches $(X_{\ell,j}, X_{\ell-1,j'})$ differ statistically from the patches $(X^\star_{\ell,j}, X^\star_{\ell-1,j'})$. For example, it can discriminate when $X^\star_{\ell-1,j'} = v_1 \implies X^\star_{\ell,j} = v_2$, but $X_{\ell-1,j'} = v_1, X_{\ell,j} \neq v_2$. In essence, it is discriminating the way where the generator super-resolutes a patch $X^\star_{\ell,j}$ from lower resolution *differently* from that of the true distribution.

As we demonstrate in Figure 6, such "super-resolution" operation is local, meaning that the learning process can be *separated* to learning over *individual patches*. The global structure across different patches of the images are learned in lower resolutions. This makes the learning process much simpler comparing to learning the full image from scratch. [8] We also provide empirical justification of the power of this "forward super-resolution", as in Figure 8(top) on Page 16: higher layers can indeed learn to super-resolute from the lower resolution images, which makes the learning much easier comparing to learning from scratch.

[8]At resolution 1 the learning is global; in this case the one-hidden-layer generator can be trained via SGDA to capture the "global structure" of images (see Section **??** and Figure 1), with the help from properties of Gaussian random variable.

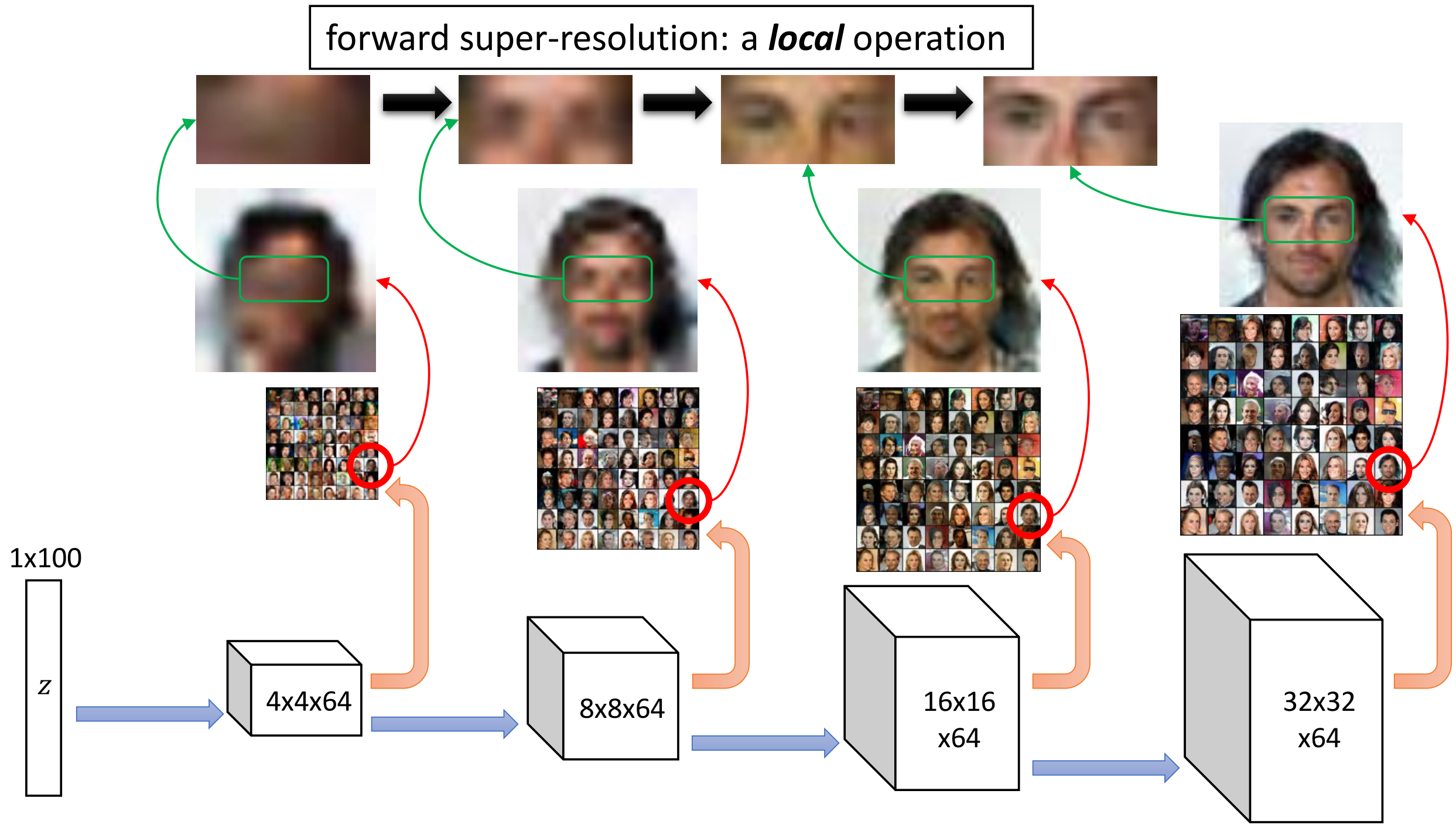

Figure 6: Forward super-resolution is a **local** operation, which makes the learning much simpler. In this experiment, generator is a 4-layer DCGAN (same structure as in Figure 1) trained on the CelebA data set.
**Remark:** When we train the generator using multi-scaled gradient, we did not explicitly ensure that the image $X_\ell$ at scale $\ell < L$ is a lower resolution image of $X_L$. Rather we simply train the generator to match the marginal distribution (i.e. each $X_\ell$ follows from the same distribution as $X^\star_\ell$). Still, **the generator learns to generate each $X_\ell$ using forward super-resolution**. Indicating that forward super-resolution is also potentially the easiest way to generate images.

## 3.4 Final Algorithm

We implement our full algorithm in Algorithm 4. It performs layer-wise training.

In each outer loop $\ell = 1, 2, \dots, L$, it first gives the output layer $\{\mathbf{W}_{\ell,j}\}_{j\in[d_\ell]}$ a very rough SVD warm start (details in Section C.1) — note this is merely a warm start so the weights are still very inaccurate.[9] Next, for this layer $\ell$, Algorithm 4 alternatively:

- uses the current output layer $\mathbf{W}_{\ell,j}$ to learn the hidden variables $\mathbf{S}_{\ell,j}$ (or equivalently the weights $\mathbf{V}_{\ell,j}, b_{\ell,j}$) to some accuracy— by applying GAN_FirstHidden if $\ell = 1$ or GAN_FowardSuperResolution if $\ell \geq 2$; and
- uses the current hidden variables $\mathbf{S}_{\ell,j}$ to learn the output layer $\mathbf{W}_{\ell,j}$ to an even better accuracy— by applying GAN_OutputLayer.

This alternating process repeats for $T' = \widetilde{O}(1)$ stages, and the global bias parameter $b$ used in the discriminator shrinks by a large (technical) factor after each stage.

Once again, we have broken the learning into multiple parts for analysis purpose, so it becomes clear how the generator can leverage the discriminator at different stages to learn the target distribution. (With more careful choices of learning rates, one can also combine them altogether.)

Please note besides a simple SVD warm-start that is called only once per output layer $\mathbf{W}_{\ell,j}$, all the learning is done using minmax optimization on a generator-discriminator objective. Our main algorithm relies on $\mathsf{GAN_FowardSuperResolution}$, where the higher-level layers learn how to super-resolute the image from lower resolution.

[9]After all, since the hidden variables $\mathbf{S}_{\ell,j}$ at this layer $\ell$— which depend on weights $\{\mathbf{V}_{\ell,j}\}_{j\in[d_\ell]}$— are still *not learned*, at this point, the best one can do is to look at the data covariance and give $\mathbf{W}_{\ell,j}$ a very rough estimate.

---

**Algorithm 4** ($\mathsf{MINMAX\text{-}GAN}$) Minmax optimization, final algorithm

---

**for** $\ell = 1, 2, \cdots, L$ **do**
  For every $j \in [d_\ell]$, call $\mathsf{INIT_OutputLayer}(\ell, j)$. ◇ *see Algorithm 5 and this gives $\mathbf{W}_{\ell,j}$ a warm start*
  $b \leftarrow m_\ell^{-0.3}$.
  **for** $t \in [T']$ **do**
    $b \leftarrow b \times m_\ell^{-0.02}$
    **if** $\ell = 1$:
      call $\mathsf{GAN_FirstHidden}(\mathbf{W}_1, b)$; ◇ *see Algorithm 2 and this learns $\mathbf{V}_{1,j}, b_{1,j}$*
      **if** $t = T'$, then $[b_{1,j}]_r \leftarrow [b_{1,j}]_r + \frac{1}{\mathsf{poly}(d/\varepsilon)}$. ◇ *useless; only for stating theorems more cleanly*
    **if** $\ell > 1$, **then for** every $j \in [d_\ell]$: ◇ *see Algorithm 3 and this learns $\mathbf{V}_{\ell,j}, b_{\ell,j}$*
      call $\mathsf{GAN_FowardSuperResolution}(\mathbf{W}_\ell, \mathbf{W}_{\ell-1}, b, \ell, j)$
    **for** every $j \in [d_\ell]$
      call $\mathsf{GAN_OutputLayer}(\mathbf{W}_{\ell,j}, b, \ell, j)$; ◇ *see Algorithm 1 and this learns $\mathbf{W}_{\ell,j}$*
  **end for**
**end for**

---

# 4 Main Theorem and High-Level Proof Plan

We state our main theorem as follows.

**Theorem 4.1** (main)**.** *For every $d > 0$, every $\varepsilon \in [\frac{1}{d^{\omega(1)}}, \frac{1}{2}]$, letting $G(z) = (X_1(z), \dots, X_L(z))$ be the generator learned after running Algorithm 4 (which runs in time/sample complexity $\mathsf{poly}(d/\varepsilon)$), then w.h.p. there is a column orthonormal matrix $\mathbf{U} \in \mathbb{R}^{m_0 \times m_0'}$ such that*

$$\mathbf{Pr}_{z \sim \mathcal{N}(0, \mathbf{I}_{m_0' \times m_0'})} \left[ \left\| G^\star(\mathbf{U}z) - G(z) \right\|_2 \leq \varepsilon \right] \geq 1 - \frac{1}{(d/\varepsilon)^{\omega(1)}} \ .$$

Theorem 4.1 relies on the following main lemmas. Our Lemma D.1 says that we can easily obtain some warm start on $\mathbf{W}$.

**Lemma D.1** (initialization)**.** *For every $\ell \in [L], j \in [d_\ell]$, w.h.p. Algorithm 5 finishes in at most $T = \mathsf{poly}(k_\ell) \cdot m_\ell^3$ iterations, and let $\mathbf{W}_{\ell,j}$ be the matrix output by Algorithm 5. Then, there is a permutation $P \colon [m_\ell] \to [m_\ell]$ such that for every $p \in [m_\ell]$,*

$$\|[\mathbf{W}^\star_{\ell,j}]_p - [\mathbf{W}_{\ell,j}]_{P(p)}\|_2 \leq \frac{\mathsf{poly}(k_\ell)}{\sqrt{m_\ell}}$$

*(For notation simplicity, in this paper we assume as if $P(p) = p$ is the identity permutation.)*

Our Lemma F.10 shows that, as long as $\mathbf{W}_1$ is moderately $\delta$-accurate, then applying gradient descent ascent using discriminators $D^{(4)}$ and $D^{(5)}$, we can learn $\langle \mathbf{V}^\star_{1,j,r}, \mathbf{V}^\star_{1,j',r'} \rangle$ to some accuracy that depends on $\delta$.

**Lemma F.10** ($D^{(4)} + D^{(5)}$, learn first hidden)**.** *For every $\delta \in (0, m_1^{-0.1}]$, suppose for every $j \in [d_1]$, for every $p \in [m_1]$, $\|[\mathbf{W}^\star_{1,j}]_p - [\mathbf{W}_{1,j}]_p\|_2 \leq \delta$. Then, after running $\mathsf{GAN_FirstHidden}$ with $b = \delta \cdot m_1^{0.02}$, we have:*

- *for every $j \in [d_1]$, $r \neq r' \in [m_1]$: $|\langle \mathbf{V}_{1,j,r}, \mathbf{V}_{1,j,r'} \rangle - \langle \mathbf{V}^\star_{1,j,r}, \mathbf{V}^\star_{1,j,r'} \rangle| \leq \mathsf{poly}(k_1) \delta m_1^{0.02}$*
- *for every $j \neq j' \in [d_1]$, $r, r' \in [m_1]$: $|\langle \mathbf{V}_{1,j,r}, \mathbf{V}_{1,j',r'} \rangle - \langle \mathbf{V}^\star_{1,j,r}, \mathbf{V}^\star_{1,j',r'} \rangle| \leq \mathsf{poly}(m_1) \delta$*

Our Lemma E.2 then shows, if the results in Lemma F.10 are satisfied, then we can further reinforce the accuracy of $\mathbf{W}_1$.

**Lemma E.2** ($D^{(1)}$, reinforce output layer). *For every $\ell \in [L]$, if Lemma F.10 holds (for $\ell = 1$) or Lemma G.2 holds (for $\ell > 1$), then after running* GAN_OutputLayer *with $\mathbb{b} = \delta m_\ell^{0.172}$, we must have: for every $r \in [m_\ell]$:*

$$\|[\mathbf{W}_{\ell,j}]_r - [\mathbf{W}^\star_{\ell,j}]_r\|_2 \leq m^{-0.02}\delta$$

Therefore, if we repeatedly and alternatively apply GAN_FirstHidden and GAN_OutputLayer, then we can keep decreasing $\delta$ (and thus $b, \mathbb{b}$) until they are $\frac{1}{\mathsf{poly}(d/\varepsilon)}$ small. This implies not only $\|[\mathbf{W}_{1,j}]_r - [\mathbf{W}^\star_{1,j}]_r\|_2 \leq \frac{1}{\mathsf{poly}(d/\varepsilon)}$ but also the existence of a column orthonormal matrix $\mathbf{U}$ such that:

$$\|\mathbf{V}_1 - \mathbf{V}_1^\star \mathbf{U}\|_F \leq \frac{1}{\mathsf{poly}(d/\varepsilon)} \enspace .$$

One can show this implies w.h.p.

$$\mathcal{S}_{1,j}(z) \leq \mathcal{S}^\star_{1,j}(\mathbf{U}z) \quad \text{and} \quad \|\mathcal{S}_{1,j}(z) - \mathcal{S}^\star_{1,j}(\mathbf{U}z)\|_2 \leq \frac{1}{\mathsf{poly}(d/\varepsilon)} \enspace .$$

Our next Lemma G.2 shows that, as long as resolution $\ell - 1$ is learned to sufficiently good accuracy, and $\mathbf{W}_\ell$ is learned to *moderate* $\delta$-accuracy, then we can also learn the $\ell$-th hidden layer (namely $\mathbf{S}_\ell$) to some accuracy that depends on $\delta$.

**Lemma G.2** ($D^{(2)}$, super-resolution in deeper layers). *Given $\ell \in \{2, 3, \dots, L\}$, suppose for some column orthonormal matrix $\mathbf{U} \in \mathbb{R}^{m_0 \times m_0'}$, w.h.p.*

$$\mathcal{S}_{\ell-1,j}(z) \leq \mathcal{S}^\star_{\ell-1,j}(\mathbf{U}z) \quad \textit{and} \quad \|\mathcal{S}_{\ell-1,j}(z) - \mathcal{S}^\star_{\ell-1,j}(\mathbf{U}z)\|_2 \leq \frac{1}{\mathsf{poly}(d/\varepsilon)} \enspace .$$

*In addition, suppose for some $\delta \in (0, m_1^{-0.1}]$, for every $j \in [d_\ell]$, for every $p \in [m_\ell]$, $\|[\mathbf{W}^\star_{\ell,j}]_p - [\mathbf{W}_{\ell,j}]_p\|_2 \leq \delta$. Then, after running* GAN_FowardSuperResolution *with $b = \delta m_\ell^{0.02}$, we must have: for every $r \in [m_\ell]$, w.h.p. over $z \sim \mathcal{N}(0, \mathbf{I}_{m_0' \times m_0'})$:*

$$[\mathcal{S}_{\ell,j}(z)]_r \leq [\mathcal{S}^\star_{\ell,j}(\mathbf{U}z)]_r, \quad |[\mathcal{S}_{\ell,j}(z)]_r - [\mathcal{S}^\star_{\ell,j}(\mathbf{U}z)]_r| \leq O(\delta m^{0.02})$$

Finally, if we repeatedly apply GAN_FowardSuperResolution and GAN_OutputLayer, then we can keep decreasing $\delta$ (and thus $b, \mathbb{b}$) until they are $\frac{1}{\mathsf{poly}(d/\varepsilon)}$ small. This implies not only $\|[\mathbf{W}_{\ell,j}]_r - [\mathbf{W}^\star_{\ell,j}]_r\|_2 \leq \frac{1}{\mathsf{poly}(d/\varepsilon)}$ but also that we can learn the $\ell$-th hidden layer to sufficiently good accuracy:

$$[\mathcal{S}_{\ell,j}(z)]_r \leq [\mathcal{S}^\star_{\ell,j}(\mathbf{U}z)]_r, \quad |[\mathcal{S}_{\ell,j}(z)]_r - [\mathcal{S}^\star_{\ell,j}(\mathbf{U}z)]_r| \leq \frac{1}{\mathsf{poly}(d/\varepsilon)}$$

These combined together gives the proof for Theorem 4.1.

# 5 Discussion: Learning Hierarchical Generative Models

In this work, we present, to the best of our knowledge, a first result regarding how to learn continuous, hierarchical generative models efficiently and provably, using GANs. Besides a simple initialization on the output layer only, all of our learning is done via training GAN's generator-discriminator objective using gradient descent ascent. Our key observation is that the **forward super-resolution** structure of the target distribution (such as for images) makes learning much easier. We believe we have made some non-trivial contributions towards understanding how real-world distributions can be efficiently learned using GANs. On the other hand, we point out**our**

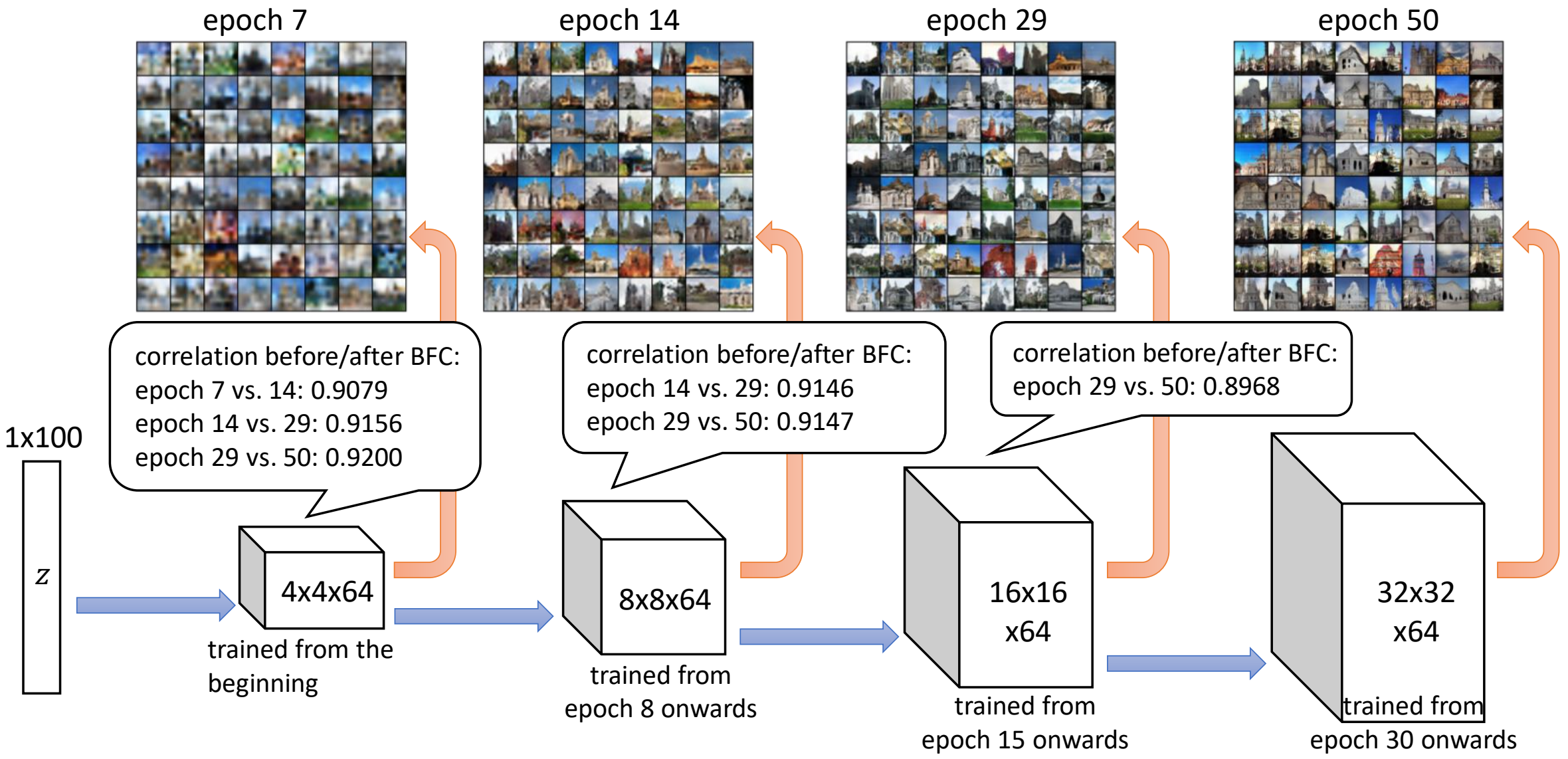

Figure 7: Layer-wise pre-training GANs with backward feature correction (BFC). In this experiment, we first train only lower-level layers for lower resolution images, and then add higher-level layers also into the set of trainable parameters and train against higher resolution images. Through training higher-level layers altogether, the lower-level features get improved; more importantly, the incremental changes (measured by vector corrections) in the neuron weights are small. This is known as backward feature correction in [4].

**result is still very preliminary**, and we can see lots of room for improvement. We summarize some open problems below:

- Training one discriminator from scratch.

  We have used different discriminators at different stages of training. Although it seems likely in practice, it is not clear whether our proof can be easily extended to training a single generator-discriminator objective from the beginning (such as by adding $D^{(1)} + D^{(2)} + D^{(4)} + D^{(5)}$). It is important to understand what would happen if we use just one universal discriminator.

- The learning rates between generator and discriminator.

  We have trained the generator/discriminator at a faster rate at different stages of learning. It is important to understand what if a uniform learning rate is used, or an adaptive gradient algorithm (such as Adam) is used instead of gradient descent ascent, which is typically how GANs are trained in practice. Can an adaptive algorithm balance the learning rates automatically?

- Agnostic learning.

  Our setting focuses on the realizable case, where the (unknown) target network $G^\star$ can generate images perfectly. The immediate next question is what if the output of target network has some error comparing to the true distribution? Although our current result can handle certain level of small noise (such as noise of magnitude $\sim \frac{1}{d}$ using simple Lipschitz-ness bounds), it would be much more interesting to extend the result beyond this level.

To address the last question, we point out that the difficulty of agnostic learning in forward super-resolution is that errors made in lower resolution can propagate to higher levels. Such an error is okay for generating simple images (see Figure 8(top) on Page 16); however, for more complicated images, we expect the network to be able to reduce "over-fitting to such errors" on lower-level layers, through training higher-level layers together. In theory, such process is known as**backward feature correction** [2, 4], where we believe it is a key step towards understanding GANs as well.

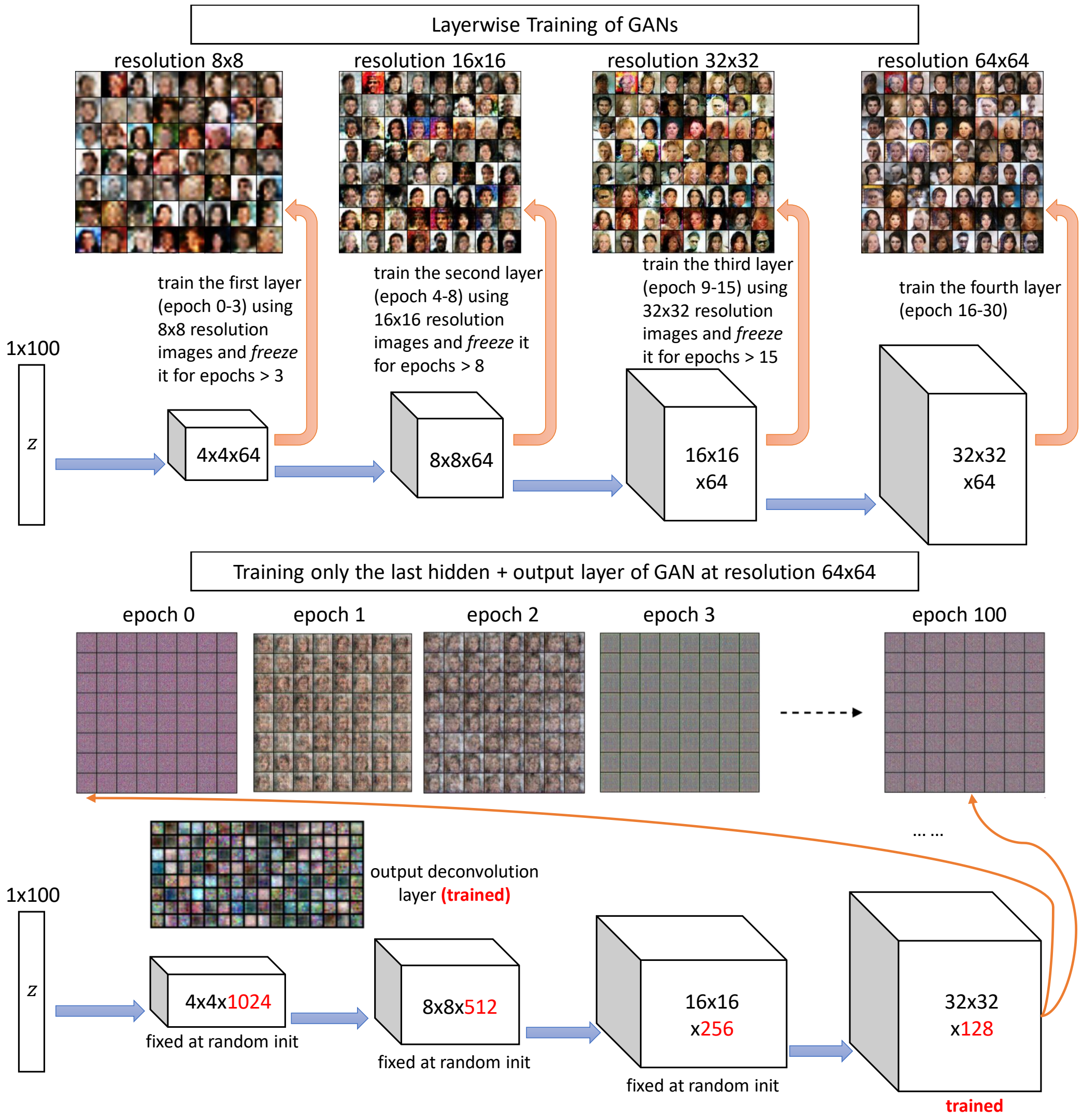

Figure 8: **(Top).** Layer-wise training GANs. We first train the first output layer/hidden layer of the generator to generate images at resolution $8 \times 8$; after that, we freeze the first hidden/output layer (and do not train them anymore). Next, train the second hidden/output layer at resolution $16 \times 16$ (the second hidden layer still takes as input the output of the trained first hidden layer); after that, we freeze both the second and the first hidden/output layers and train the third pair at resolution $32 \times 32$. Eventually, we freeze the 1,2,3 hidden/output layers and only train the last hidden layer at resolution $64 \times 64$.
**(Bottom).** As a comparison, if we only train the last hidden layer and the output layer (we even enlarge the network to incorporate more random features), then the images are very bad (the discriminator loss quickly goes to zero and the generator no longer learns).
**Conclusion: layer-wise forward super-resolution can indeed learn features from lower resolution images that are much better than random features**. We believe that this is a strong evidence that forward super-resolution makes the learning much easier.

As we illustrate in Figure 7 on Page 15, backward feature correction in GANs is also a **local feature correction process**, which makes it possible to analyze in theory.

In the end, we point out that there is a vast body of theory works considering learning neural networks under Gaussian input, but they are all in the supervised learning setting where the goal is to match the labels [5, 21, 27, 30, 48, 49, 62, 66, 67, 69, 69–71, 73, 84, 86, 87, 90, 93, 95, 97, 100]. They are fundamentally different from distribution learning.

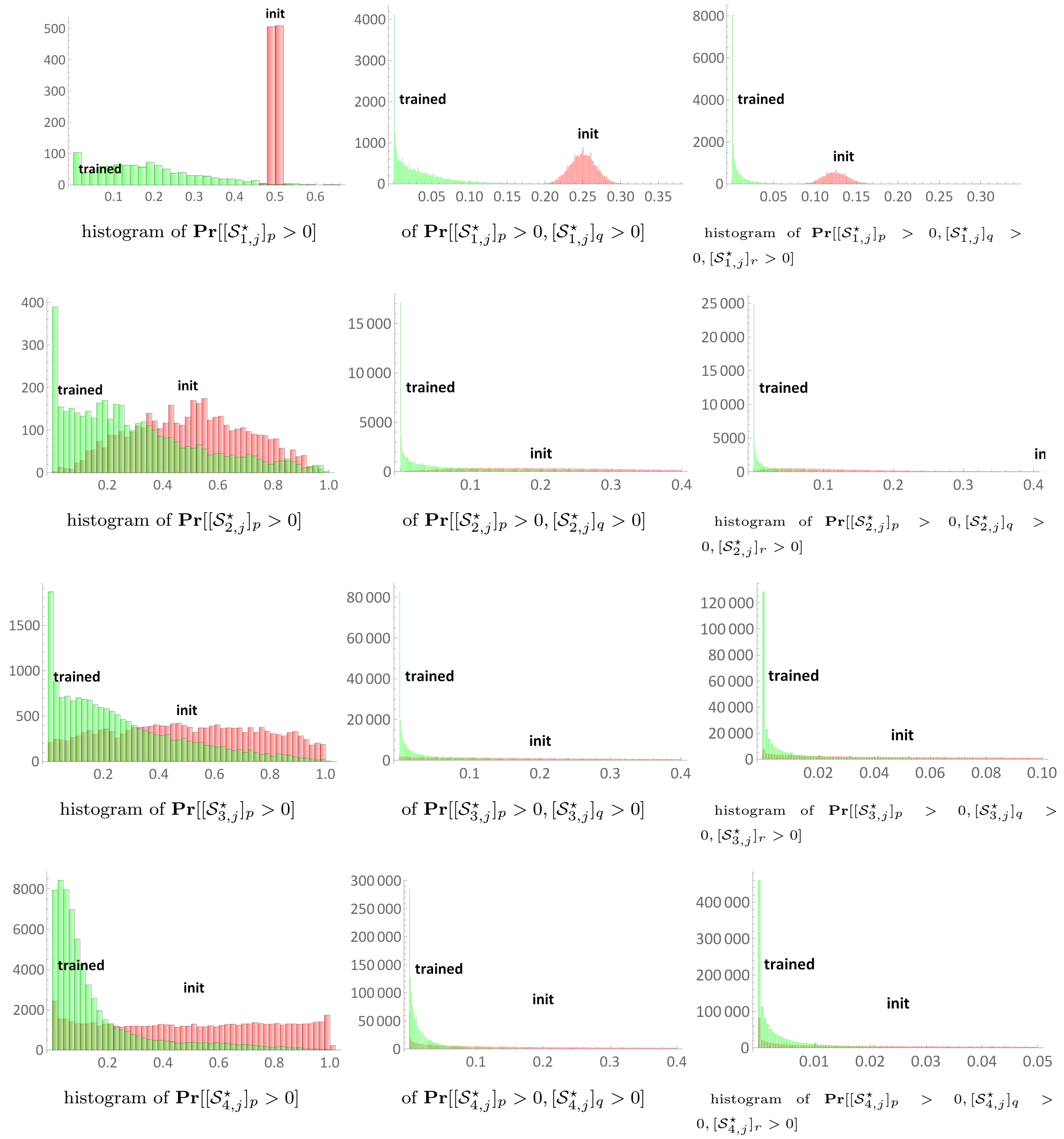

histogram of $\mathbf{Pr}[[\mathcal{S}^\star_{1,j}]_p > 0]$ | of $\mathbf{Pr}[[\mathcal{S}^\star_{1,j}]_p > 0, [\mathcal{S}^\star_{1,j}]_q > 0]$ | histogram of $\mathbf{Pr}[[\mathcal{S}^\star_{1,j}]_p > 0, [\mathcal{S}^\star_{1,j}]_q > 0, [\mathcal{S}^\star_{1,j}]_r > 0]$

histogram of $\mathbf{Pr}[[\mathcal{S}^\star_{2,j}]_p > 0]$ | of $\mathbf{Pr}[[\mathcal{S}^\star_{2,j}]_p > 0, [\mathcal{S}^\star_{2,j}]_q > 0]$ | histogram of $\mathbf{Pr}[[\mathcal{S}^\star_{2,j}]_p > 0, [\mathcal{S}^\star_{2,j}]_q > 0, [\mathcal{S}^\star_{2,j}]_r > 0]$

histogram of $\mathbf{Pr}[[\mathcal{S}^\star_{3,j}]_p > 0]$ | of $\mathbf{Pr}[[\mathcal{S}^\star_{3,j}]_p > 0, [\mathcal{S}^\star_{3,j}]_q > 0]$ | histogram of $\mathbf{Pr}[[\mathcal{S}^\star_{3,j}]_p > 0, [\mathcal{S}^\star_{3,j}]_q > 0, [\mathcal{S}^\star_{3,j}]_r > 0]$

histogram of $\mathbf{Pr}[[\mathcal{S}^\star_{4,j}]_p > 0]$ | of $\mathbf{Pr}[[\mathcal{S}^\star_{4,j}]_p > 0, [\mathcal{S}^\star_{4,j}]_q > 0]$ | histogram of $\mathbf{Pr}[[\mathcal{S}^\star_{4,j}]_p > 0, [\mathcal{S}^\star_{4,j}]_q > 0, [\mathcal{S}^\star_{4,j}]_r > 0]$

Figure 9: Histograms of $\mathbf{Pr}[[\mathcal{S}^\star_{\ell,j}]_p > 0]$ (left), of $\mathbf{Pr}[[\mathcal{S}^\star_{\ell,j}]_p > 0, [\mathcal{S}^\star_{\ell,j}]_q > 0]$ (middle), and of $\mathbf{Pr}[[\mathcal{S}^\star_{\ell,j}]_p > 0, [\mathcal{S}^\star_{\ell,j}]_q > 0, [\mathcal{S}^\star_{\ell,j}]_r > 0]$ (right) at random initialization vs. after training the architecture in Figure 1. Note: All the neurons are still active after training, meaning $\mathbf{Pr}[[\mathcal{S}^\star_{\ell,j}]_p > 0] > 0$, this is mainly due to *batch-normalization* is applied to each neuron.

**Conclusion:** The generator network does learn sparse, not-too-positively correlated hidden activations. Notice that **we did not regularize the sparsity or correlation during the training**, it is rather an intrinsic property that the generator discovered during the training process. Hence, it is reasonable to assume that the activations of the true network can be even sparser.

# Appendix I: Some Missing Details

We discuss more related works in Section A. We include in Section B some missing technical assumptions from Section 2. We include in Section C some missing pseudocodes from Section 3.

## A More on Related Works

**Other related works.** On the efficient (high-dimensional) *distribution learning* side, most of existing theories focus on learning discrete distributions over the Boolean cube or finite sets, including graphical models, Bayesian networks, RBM, LDA etc. [19, 24–26, 28, 29, 36, 41, 55, 63, 77, 83] These are very different from the domain of GANs that target to model real-world distributions over continuous manifold. For continuous distributions, existing theories have efficient learning guarantees for certain simple distributions, linear transformation of simple distributions [13, 14, 67, 70, 89] (mostly just coordinate-wise independent), learning mixture of Gaussians [23, 38, 80] or learning some very simple distributions generated by two-layer neural network [66]. All of the cited works consider distributions that are much simpler, and very far away from the high-complexity distributions generated by GANs (such as the distribution of real images).

On a separate note, it is perhaps not surprising that existing theories are so limited: unlike in supervised learning where the concept class consists of linear functions / low degree polynomials can always be learned efficiently, in generative models, to the best of our knowledge, even *a linear transformation of a known distribution* is not generally known to be efficiently learnable, unless the known distribution is coordinate-wise independent, etc.

**Comparison to sparse coding.** Although in our assumption, we assume within each patch $j$ and resolution $\ell$, the images $X_{\ell,j}$ are generated according to some sparse coding generative model, however, our case is fundamentally different from traditional sparse coding, in the following way:

- In traditional sparse coding, the goal is typically to learn the dictionary (in our case the weights $\mathbf{W}_{\ell,j}$) when the sparse signals (in our case the hidden variables $\mathbf{S}_{\ell,j}$) are pair-wise independent. However, in this paper, the signals are dependent and we do not know any theoretical work that can learn the dictionary efficiently and accurately (i.e., to a $\frac{1}{\mathsf{poly}(d)}$ error). In fact, in this paper, the *more interesting* task is to learn the distribution of those hidden variables $\mathbf{S}_{\ell,j}$—which are generated by a multi-layer neural network. Even at layer $\ell = 1$, across different patches $j$, those hidden variables $\mathbf{S}_{\ell,j}$ can be arbitrarily correlated; thus one has to rely on learning a generator network to model their joint correlations. In sum, we do not believe one can reproduce our result simply via sparse decoding and learning each patch individually.
- Our theory sheds light on the following alternating process. At random initialization, due to the strong mismatch between the distributions of hidden variables $\mathbf{S}_{\ell,j}$ in the learner vs. target generator networks, the output layer $\mathbf{W}_{\ell,j}$ cannot be learned very well. However, this so-so output layer still allows the learning process to perform some non-perfect decoding, good enough to help learn the hidden layers $\mathbf{S}_{\ell,j}$ to some accuracy. After that, with better-learned hidden layers, the output layer $\mathbf{W}_{\ell,j}$ can be further reinforced to a better accuracy, which in turn results in a better accuracy on the distribution of hidden layers. This alternating process repeats during training. (Again, this is how our theory is different from traditional sparse coding.)

- Existing (provable) sparse coding typically relies on algorithms such as sum of squares, linear programming, alternating minimization [12, 22, 88],[10] which *does not shed light* on how neural network *trained by local-search algorithms* can learn such distributions in practice. We show that, besides a mild warm start using SVD, the dictionary (i.e., the output layer weights) *as well as* the hidden variables in the target network can be learned *using the practical method*: stochastic gradient descent ascent (SGDA) over a GAN generator-discriminator training objective.
- Last but not least, we show that GANs, trained using SGDA over a generator-discriminator objective, can simulate a sequence of traditional learning techniques, such as sparse coding, regression (to learn forward super-resolution), moment matching (to learn the first hidden layer) so that they can jointly perform *hierarchical distribution learning for real-world distributions.*

# B Missing Problem Setup for Section 2

**Notations.** Recall $\mathsf{ReLU}(z) = \max\{z, 0\}$. For some sufficiently small $\zeta = \frac{1}{\mathsf{poly}(d/\varepsilon)}$, we define a smoothed (convex) ReLU activation $\widetilde{\mathsf{ReLU}}(z)$ with Lipschitz continuous second-order derivative:

$$\widetilde{\mathsf{ReLU}}(z) = \begin{cases} z - \zeta/2 & \text{if } z \geq \zeta; \\ 0 & \text{if } z < 0. \end{cases} \qquad \text{with } \widetilde{\mathsf{ReLU}}'(z) \in [0,1] \text{ and } \widetilde{\mathsf{ReLU}}''(z) \in \left[0, O(\tfrac{1}{\zeta})\right] \text{ for } z \in [0, \zeta]$$

We also consider a smoothed leaky-ReLU activation $\mathsf{LeakyReLU}(z)$ with Lipschitz continuous second-order derivative, satisfying

$$\mathsf{LeakyReLU}(z) = \begin{cases} z & \text{if } z \geq \zeta; \\ \zeta z & \text{if } z < 0. \end{cases} \text{ with } \mathsf{LeakyReLU}'(z) \in [0,1] \text{ and } |\mathsf{LeakyReLU}''(z)| \leq O(\tfrac{1}{\zeta}) \text{ for } z \in [0, \zeta]$$

The leak and smoothing are *for analysis propose* to remove vanishing gradient and to make sure that the function is sufficiently smooth. One can construct them using Hermite interpolation.

**Assumption B.1.** *We also make the following assumptions for efficient optimization.* [11]

- *Anti-concentration: for every $p \neq q \in [m_\ell]$ and $\delta \geq 0$:*

$$\mathbf{Pr}_z[[\mathcal{S}^\star_{\ell,j}]_p \leq \delta \mid [\mathcal{S}^\star_{\ell,j}]_p > 0] \leq \mathsf{poly}(k_\ell)\delta$$
$$\mathbf{Pr}_z[[\mathcal{S}^\star_{\ell,j}]_p \leq \delta \mid [\mathcal{S}^\star_{\ell,j}]_p > 0, [\mathcal{S}^\star_{\ell,j}]_q > 0] \leq \mathsf{poly}(k_\ell)\delta$$

- *Non-degeneracy: let $\mu = (\mu_{p,q})_{p,q\in[m_\ell]}$ be the matrix such that $\mu_{p,q} = \mathbb{E}_z[\mathbb{1}_{[\mathcal{S}^\star_{\ell,j}]_p > 0}[\mathcal{S}^\star_{\ell,j}]_q]$, then the minimum singular value $\sigma_{\min}(\mu) \geq \frac{1}{m_\ell^{1.15}}$.*
- *Non-degeneracy on $\ell = 1$: for all $(j, r) \neq (j', r') \in [d_1] \times [m_1]$,*

$$\mathbf{Pr}_z[[\mathcal{S}^\star_{1,j'}]_{r'} > 0, [\mathcal{S}^\star_{1,j}]_r > 0] \in \left[\frac{1}{\mathsf{poly}(d)}, 1 - \frac{1}{\mathsf{poly}(d)}\right] \times \min\{\mathbf{Pr}_z[[\mathcal{S}^\star_{1,j'}]_{r'} > 0], \mathbf{Pr}_z[[\mathcal{S}^\star_{1,j}]_r > 0]\}$$

The anti-concentration condition is quite normal for Gaussian-like distributions: for example, it is automatically satisfied in the first hidden layer ($\ell = 1$), and are typically true for Lipschitz functions of Gaussian variables [35] (as in higher layers). The first non-degeneracy condition is almost free since matrix $\mu$ is almost diagonal-dominant: the diagonals are $\geq \frac{1}{\mathsf{poly}(k_\ell)m_\ell}$ but the off-diagonal entries are less than $\varepsilon_2 = \frac{\mathsf{poly}(k_\ell)}{m_\ell^2}$. If for example one strengthens the "not-too-positive correlation" assumption to "negative correlation" where $\varepsilon_2$ is a bit smaller or simply to "independent", then $\sigma_{\min}(\mu) \geq \frac{1}{\mathsf{poly}(k_\ell)m_\ell}$ holds automatically. Again, $\mu$ is defined within a patch $\mathcal{S}^\star_{\ell,j}$, it has no impact on the global structure of the network (across different patches $j$). The second non-degeneracy condition can be removed, but we keep it to simplify the proof by avoiding case analysis.

[10] Note Arora et al. [13] introduces a neural algorithm for dictionary learning, but it uses a customized local-search type of update that is not exactly stochastic gradient descent or its variants.

[11] Again, $\mathsf{poly}(k_\ell)$ can be an arbitrary polynomial such as $(k_\ell)^{100}$, and our final theorem holds for sufficiently large $d$ because $d^{o(1)} > \mathsf{poly}(k_\ell)$. The $\mathsf{poly}(d)$ can also be an arbitrary polynomial, and our final complexity $\mathsf{poly}(d)$ in the theorem shall grow into a larger polynomial according to it. In this paper, we repeatedly use $\mathsf{poly}(k_\ell)$ and $\mathsf{poly}(d)$ without specifying their degrees, and one can easily write down how such polynomials propagate at the expense of complicating the notations.

**Which neural network satisfies these assumptions?** We give a simple and generic set of generator neural networks satisfying all the listed assumptions. Consider the case $m_{\ell+1} \leq m_\ell$ and $d_{\ell+1} \geq d_\ell$. Let each $\mathbf{V}^\star_{\ell,j,j'} \in \mathbb{R}^{m_\ell \times m_{\ell-1}}$ satisfy $\mathrm{supp}([\mathbf{V}^\star_{\ell,j,j'}]_r) \cap \mathrm{supp}([\mathbf{V}^\star_{\ell,j,j'}]_{r'}) = \varnothing$ for distinct $r, r' \in [m_\ell]$. Let the connection graph satisfy $|\mathcal{P}_{\ell,j}| = 1$. Let the norm of each $[\mathbf{V}^\star_{\ell,j,j'}]_r$ and the biases $b^\star_{\ell,j}$ to be set such that $\mathbf{Pr}_z[[\mathcal{S}^\star_{\ell,j}]_p \neq 0], \mathbb{E}_z[[\mathcal{S}^\star_{\ell,j}]_p] \in \left[\frac{1}{\mathsf{poly}(k_\ell)m_\ell}, \frac{\mathsf{poly}(k_\ell)}{m_\ell}\right]$ are satisfied.

With this construction, one can easily conclude that for each $\ell, j$, the quantities $[\mathcal{S}^\star_{\ell,j}]_r$ are independent across $r \in [m_\ell]$. This easily satisfies all the aforementioned assumptions (in particular, the anti-concentration can be shown by induction again using independence). Again, we stress that in the above model, the independence is only *within a patch*; across different patches the the distributions can be arbitrarily correlated. Therefore, the global structure of the images can be preserved.

# C Missing Pseudocodes for Section 3

## C.1 Simple Initialization of Output Layers

We consider a simple SVD warm start to initialize (only) the output layers $\mathbf{W}_{\ell,j}$ of the network. It merely involves a simple estimator of certain truncated covariance of the data. See Algorithm 5.

**Algorithm 5** ($\mathsf{INIT_OutputLayer}$) warm-start of the output layer using simple SVD

**Input:** $\ell, j$

$\mathcal{W} = \varnothing$, $t = 0$, $b = \frac{1}{\mathsf{poly}(k_\ell)}$.
**for** $t \in [T]$ **do**
  Sample 2 data patches $\{[X^\star_{\ell,j}]^{(i)}\}_{i \in \{1,2\}}$, and define $x = \sum_{i \in [2]} [X^\star_{\ell,j}]^{(i)}$.
  Sample $N$ i.i.d. true samples $\mathcal{Z}^\star = \{[X^\star]^{(i)}\}_{i \in [N]}$
  $\mathbf{H} \leftarrow \mathbb{E}_{X^\star \sim \mathcal{Z}^\star}\left[\langle x, X^\star_{\ell,j}\rangle [P_{\mathcal{W}}(X^\star_{\ell,j})]^{\otimes 2}\right]$, where $P_{\mathcal{W}}(X^\star_{\ell,j}) \stackrel{\mathrm{def}}{=} \prod_{w \in \mathcal{W}} (\mathbf{I} - ww^\top \mathbb{1}_{\langle w, X^\star \rangle \geq b}) X^\star_{\ell,j}$.
  $v \leftarrow$ be the (unit norm) top eigenvector of $\mathbf{H}$, and let $\Delta$ be the eigen-gap.
  **if** $\Delta > \frac{1}{m}$ and $\forall w \in \mathcal{W}$, $\langle v, w \rangle \leq 0.5$ **then**
    $v' \leftarrow \mathbb{E}_{X^\star \sim \mathcal{Z}^\star}[\mathbb{1}_{\langle v, X^\star_{\ell,j}\rangle \geq b} X^\star_{\ell,j}]$, $v' \leftarrow \frac{v'}{\|v'\|_2}$.
    $\mathcal{W} \leftarrow \mathcal{W} \cup \{v'\}$.
  **end if**
  If $|\mathcal{W}| = m_\ell$: **Break**.
**end for**
Set $\mathbf{W}_{\ell,j} = (w)_{w \in \mathcal{W}}$.

## C.2 Gradient Descent Ascent for GAN

Given a generator $G(z; \theta_G)$ that takes $z \sim \mathcal{N}(0, \mathbf{I})$ and trainable parameters $\theta_G$, and a discriminator $D(X; \theta_D)$ that takes an image $X$ and trainable parameters $\theta_D$, the min-max objective for generative adversarial network (GAN) is given as

$$\min_{\theta_G} \max_{\theta_D} \mathcal{L}(\theta_D, \theta_G) = \mathop{\mathbb{E}}_{X^\star \sim \mathcal{D}} [D(X^\star; \theta_D)] - \mathop{\mathbb{E}}_{z \sim \mathcal{N}(0, \mathbf{I})} \mathbb{E}[D(G(z; \theta_G); \theta_D)] + R_1(\theta_G) - R_2(\theta_D) \tag{C.1}$$

Above, $R_1(\theta_G)$ and $R_2(\theta_D)$ are (optional) regularizers for the generator and discrminator respectively.

In this paper, we refer (stochastic) gradient descent ascent $\mathsf{SGDA}$ algorithm for (C.1) as Algorithm 6, and note it optimizes over the discriminator at a faster rate.

---

**Algorithm 6** ($\mathsf{SGDA}$) stochastic gradient descent ascent

---

**Input:** a GAN objective (C.1), initial parameters $\theta_G, \theta_D$

$\eta \leftarrow \frac{1}{\mathsf{poly}(d/\varepsilon)}$, $T \leftarrow \frac{\mathsf{poly}(d/\varepsilon)}{\eta}$,
**for** $t \in [T]$ **do**
  Gather $N = \mathsf{poly}(d/\varepsilon)$ samples $\mathcal{Z}^\star = \{[X^\star]^{(i)}\}_{i \in [N]}$ + fake samples $\mathcal{Z} = \{G(z^{(i)}; \theta_G)\}_{i \in [N]}$.
  Define $\mathcal{L} = \mathbb{E}_{X^\star \sim \mathcal{Z}^\star}[D(X^\star; \theta_D)] - \mathbb{E}_{X \sim \mathcal{Z}} \mathbb{E}[D(X; \theta_D)] + R_1(\theta_G) - R_2(\theta_D)$
  **for** $t \in [T]$ **do**
    Update: $\theta_D \leftarrow \theta_D + \eta \nabla_{\theta_D} \mathcal{L}(\theta_G, \theta_D)$.
  **end for**
  Update: $\theta_G \leftarrow \theta_G - \eta \nabla_{\theta_G} \mathcal{L}(\theta_G, \theta_D) - \eta \xi$.
    $\diamond$ $\xi \sim \mathcal{N}(0, \sigma_\xi^2 \mathbf{I})$ *for a small* $\sigma_\xi = \frac{1}{\mathsf{poly}(d/\varepsilon)}$*; only for theoretical analysis purpose to escape saddle points.*
**end for**

---

In one of our algorithms, we have instead written the min-max objective with generator inside:

$$\min_{\theta_D} \max_{\theta_G} \mathcal{L}(\theta_D, \theta_G) = \mathop{\mathbb{E}}_{X^\star \sim \mathcal{D}} [D(X^\star; \theta_D)] - \mathop{\mathbb{E}}_{z \sim \mathcal{N}(0, \mathbf{I})} \mathbb{E}[D(G(z; \theta_G); \theta_D)] - R_1(\theta_G) + R_2(\theta_D)$$

In this case, we refer $\mathsf{SGDA}$ to (stochastic) gradient descent-ascent where the generator moves at a faster rate. We do not repeat the pseudocode here for cleanness.

# Appendix II: Complete Proofs

We now begin the proof, for simplicity, we only focus on the population case here where we have access to the true expectation. It is an easy exercise to move to the finite sample case (with $\mathsf{poly}(d/\varepsilon)$ samples), since our network has $\mathsf{poly}(d)$ many parameters and the weight matrices are bounded by $\mathsf{poly}(d)$ in norm during training. Moreover, during our entire training process, all of our training objectives shall be $\mathsf{poly}(d)$ bounded, $\mathsf{poly}(d/\varepsilon)$-Lipschitz smooth and second-order Lipschitz smooth. Our algorithm is run with fresh sample at every iteration, hence by standard concentration bound we can easily obtain the desired result.

## D Output Layer Warm-Start Initialization

In this section we prove Lemma D.1. We restate it below:

**Lemma D.1** (Algorithm 5, restated)**.** *For every $\ell \in [L], j \in [d_\ell]$ and selecting $b = \frac{1}{\mathsf{poly}(k_\ell)}$, w.h.p. Algorithm 5 finishes in at most $T = \mathsf{poly}(k_\ell) \cdot m_\ell^3$ iterations, and let $\mathbf{W}_{\ell,j}$ be the matrix output by Algorithm 5. Then, there is a permutation $P\colon [m_\ell] \to [m_\ell]$ such that for every $p \in [m_\ell]$,*

$$\|[\mathbf{W}^\star_{\ell,j}]_p - [\mathbf{W}_{\ell,j}]_{P(p)}\|_2 \leq \frac{\mathsf{poly}(k_\ell)}{\sqrt{m_\ell}}$$

(For notation simplicity, in this paper we assume as if $P(p) = p$ is the identity permutation.)

We first establish some critical claims. Recall $X^\star_{\ell,j} = \mathbf{W}^\star_{\ell,j} \mathcal{S}^\star_{\ell,j}$ so we naturally have

**Claim D.2.** *We have that for every $\ell \in [L], j \in [d_\ell]$, let $\mathbf{W}^\star = \mathbf{W}^\star_{\ell,j} \in \mathbb{R}^{d \times m_\ell}$ and $\mathbf{W} = \mathbf{W}_{\ell,j}$, we have:*

$$\begin{aligned} J := \mathbb{E}\left[\left(X^\star_{\ell,j}\right)^{\otimes 3}\right] &= \sum_{r \in [m_\ell]} \alpha_r [\mathbf{W}^\star_r]^{\otimes 3} \\ &+ \sum_{p,q \in [m_\ell], p \neq q} \beta_{p,q} \left([\mathbf{W}^\star_p]^{\otimes 2} \otimes \mathbf{W}^\star_q + \mathbf{W}^\star_p \otimes \mathbf{W}^\star_q \otimes \mathbf{W}^\star_p + \mathbf{W}^\star_q \otimes [\mathbf{W}^\star_p]^{\otimes 2}\right) \\ &+ \sum_{p,q,r \in [m_\ell], p \neq q \neq r} \gamma_{p,q,r} \mathbf{W}^\star_p \otimes \mathbf{W}^\star_q \otimes \mathbf{W}^\star_r \end{aligned}$$

*Where we have:*

$$\alpha_r = \mathbb{E}\left[\left([\mathcal{S}^\star_{\ell,j}]_r\right)^3\right], \beta_{p,q} = \mathbb{E}\left[\left([\mathcal{S}^\star_{\ell,j}]_p\right)^2 [\mathcal{S}^\star_{\ell,j}]_q\right], \gamma_{p,q,r} = \mathbb{E}\left[[\mathcal{S}^\star_{\ell,j}]_p [\mathcal{S}^\star_{\ell,j}]_q [\mathcal{S}^\star_{\ell,j}]_r\right]$$

*Note under Assumption 2.2, we have $0 \leq \beta_{p,q} \leq \mathsf{poly}(k_\ell) \cdot \varepsilon_1$ and $0 \leq \gamma_{p,q,r} \leq \mathsf{poly}(k_\ell) \cdot \varepsilon_2$.*

Now, we have the following Claim:

**Claim D.3** (Tensor to SVD)**.** *For every $\ell \in [L], j \in [d_\ell]$, let $J$ be defined as above, and let $x$ be the sum of constantly many random sampled $X^\star_{\ell,j}$, then w.h.p. we have:*

$$\left\|J(x, *, *) - \mathbf{M}(x)\right\|_2 \leq \mathsf{poly}(k_\ell) \times (\varepsilon_1 \sqrt{m_\ell} + \varepsilon_2 m_\ell)$$

*where $\mathbf{M}(x) := \sum_{r \in [m_\ell]} \alpha_r \langle \mathbf{W}^\star_r, x\rangle [\mathbf{W}^\star_r]^{\otimes 2}$.*

*Proof of Claim D.3.* We know that

$$J(x,*,*) - \mathbf{M}(x) = \sum_{p,q\in[m_\ell],p\neq q} \beta_{p,q}\left([\mathbf{W}^\star_p]^{\otimes 2}\langle \mathbf{W}^\star_q, x\rangle + \mathbf{W}^\star_p \otimes \mathbf{W}^\star_q \langle \mathbf{W}^\star_p, x\rangle + \mathbf{W}^\star_q \otimes \mathbf{W}^\star_p \langle \mathbf{W}^\star_p, x\rangle\right)$$
$$+ \sum_{p,q,r\in[m_\ell],p\neq q\neq r} \gamma_{p,q,r}\langle \mathbf{W}^\star_p, x\rangle \mathbf{W}^\star_q \otimes \mathbf{W}^\star_r \tag{D.1}$$

Since $x = \sum_i [X^\star_{\ell,j}]^{(i)}$ where each sample $[X^\star_{\ell,j}]^{(i)} = \mathbf{W}^\star[\mathcal{S}^\star_{\ell,j}]^{(i)}$ for a vector $\mathcal{S}^\star_{\ell,j}$ that is $k_\ell$-sparse and $\|\mathcal{S}^\star_{\ell,j}\|_\infty \leq \mathsf{poly}(k_\ell)$ with high probability, we derive with high probability that

$$\left|\sum_{q\in[m_\ell]}\langle \mathbf{W}^\star_q, x\rangle\right| \leq \mathsf{poly}(k_\ell) \quad \text{and} \quad \|x\|_2 \leq \mathsf{poly}(k_\ell) \enspace .$$

On the other hand, to bound the spectral norm of the RHS of (D.1), let us take any vector $y \in \mathbb{R}^d$ with $\|y\|_2 = 1$. We have

$$\begin{aligned}
&\left\|\sum_{p\in[m_\ell]}\langle \mathbf{W}^\star_p, x\rangle\langle \mathbf{W}^\star_p, y\rangle \left(\sum_{q\in[m_\ell],q\neq p}\mathbf{W}^\star_q\beta_{p,q}\right)\right\|_2 \\
&\leq \sum_{p\in[m_\ell]}|\langle \mathbf{W}^\star_p, x\rangle\langle \mathbf{W}^\star_p, y\rangle| \left\|\left(\sum_{q\in[m_\ell],q\neq p}\mathbf{W}^\star_q\beta_{p,q}\right)\right\|_2 \\
&\leq \sqrt{\sum_{p\in[m_\ell]}\left(\langle \mathbf{W}^\star_p, y\rangle\right)^2}\sqrt{\sum_{p\in[m_\ell]}|\langle \mathbf{W}^\star_p, x\rangle|^2}\cdot \mathsf{poly}(k_\ell)\sqrt{m_\ell}\varepsilon_1 \leq \mathsf{poly}(k_\ell)\sqrt{m_\ell}\varepsilon_1 \cdot \|x\|_2
\end{aligned}$$

Similarly, we have

$$\left\|\sum_{q\in[m_\ell]}\langle \mathbf{W}^\star_q, x\rangle\langle \mathbf{W}^\star_p, y\rangle \left(\sum_{p\in[m_\ell],q\neq p}\mathbf{W}^\star_p\beta_{p,q}\right)\right\|_2 \leq \mathsf{poly}(k_\ell)\sqrt{m_\ell}\varepsilon_1 \cdot \|x\|_2$$

Finally, we have

$$\begin{aligned}
&\left\|\sum_{q,r\in[m_\ell],q\neq r}\mathbf{W}^\star_r\langle \mathbf{W}^\star_q, y\rangle \sum_{p\in[m_\ell],p\neq q,p\neq r}\left(\gamma_{p,q,r}\langle \mathbf{W}^\star_p, x\rangle\right)\right\|_2 \\
&\leq \left\|\sum_{q,r\in[m_\ell],q\neq r}\mathbf{W}^\star_r\langle \mathbf{W}^\star_q, y\rangle\right\|_2 \cdot \mathsf{poly}(k_\ell)\varepsilon_2 \\
&\leq \sqrt{\sum_{r\in[m_\ell]}\left(\sum_{q\in[m_\ell],q\neq r}|\langle \mathbf{W}^\star_q, y\rangle|\right)^2}\cdot \mathsf{poly}(k_\ell)\varepsilon_2 \leq \mathsf{poly}(k_\ell)\varepsilon_2\cdot m_\ell
\end{aligned}$$

Together they imply the desired spectral norm bound. □

We first present a Claim regarding the matrix $\mathbf{H}$ in Algorithm 5:

**Claim D.4** (SVD initialization)**.** *In Algorithm 5, for every $\ell \in [L], j \in [d_\ell]$, denote by $\mathbf{W}^\star = \mathbf{W}^\star_{\ell,j}$ and suppose there is an $\delta \in (0, m^{-0.005}]$ such that for every $w \in \mathcal{W}$, there is a unique $p \in [m_\ell]$ (so different $w$ uses different $p$) with:*

$$\|\mathbf{W}^\star_p - w\|_2 \leq \delta$$

*We denote by $\mathcal{P}$ the set of $p$ corresponding to all $w \in \mathcal{W}$. Suppose $b > \mathsf{poly}(k_\ell)\delta$. Then, we must have that:*

$$\left\|\mathbf{H} - \sum_{r\in[m_\ell]}\alpha'_r\langle \mathbf{W}^\star_r, x\rangle[\mathbf{W}^\star_r]^{\otimes 2}\right\|_2 \leq \mathsf{poly}(k_\ell)\cdot\left(\frac{\delta}{m_\ell} + \varepsilon_1\sqrt{m_\ell} + \varepsilon_2 m_\ell\right)$$

*where $\alpha'_r = \mathbb{E}\left[([\mathcal{S}^\star_{\ell,j}]_r)^3\right]$ for any $r \notin \mathcal{P}$ and for $\alpha'_r \leq \frac{\mathsf{poly}(k_\ell)b^3}{m_\ell}$ for any $r \in \mathcal{P}$.*

*Proof of Claim D.4.* For any $p \in \mathcal{P}$, let us denote the corresponding $w \in \mathcal{W}$ satisfying $\|w - \mathbf{W}^\star_p\|_2 \leq \delta$ by $w_p$. Let us abbreviate the notation by writing $X^\star_{\ell,j} = \mathbf{W}^\star\mathcal{S}^\star$ for $\mathcal{S}^\star = \mathcal{S}^\star_{\ell,j}$.

Using $\|\mathcal{S}^\star\|_\infty \leq \mathsf{poly}(k_\ell)$ and $\|\mathcal{S}^\star\|_0 \leq k_\ell$ again, we can easily conclude that w.h.p., for every $p \in \mathcal{P} \subseteq [m_\ell]$, recalling that $\|w_p - \mathbf{W}^\star_p\|_2 \leq \delta \leq m^{-0.005}$ and $b > \mathsf{poly}(k_\ell)\delta$,

$$[\mathcal{S}^\star]_p = \langle \mathbf{W}^\star_p, X^\star_{\ell,j}\rangle \geq b + \delta\mathsf{poly}(k_\ell) \implies \langle w_p, X^\star_{\ell,j}\rangle \geq b \tag{D.2}$$

$$\langle w_p, X^\star_{\ell,j}\rangle \geq b \implies [\mathcal{S}^\star]_p = \langle \mathbf{W}^\star_p, X^\star_{\ell,j}\rangle \geq b - \delta\mathsf{poly}(k_\ell) > 0 \tag{D.3}$$

Recall $P_{\mathcal{W}}(X^\star_{\ell,j}) \stackrel{\text{def}}{=} \prod_{w\in\mathcal{W}}(\mathbf{I} - ww^\top \mathbb{1}_{\langle w, X^\star\rangle \geq b})X^\star_{\ell,j}$. Let us denote by $\mathcal{U}(X^\star_{\ell,j}) \subseteq \mathcal{P}$ the set $p \in \mathcal{P}$ such that $\langle w_p, X^\star_{\ell,j}\rangle \geq b$. From the relationship (D.2), one can derive w.h.p. that

$$\Big\|P_{\mathcal{W}}(X^\star_{\ell,j}) - \textstyle\sum_{p\in[m_\ell], p\notin\mathcal{U}(X^\star_{\ell,j})} \mathbf{W}^\star_p \mathcal{S}^\star_p\Big\|_2 \leq \delta\mathsf{poly}(k_\ell)$$

Let us denote $\mathcal{X} = \sum_{p\in[m_\ell], p\notin\mathcal{U}(X^\star_{\ell,j})} \mathbf{W}^\star_p \mathcal{S}^\star_p$ and $\Xi = P_{\mathcal{W}}(X^\star_{\ell,j}) - \mathcal{X}$. Recall we have defined (the expectation is taken over $X^\star$)

$$\mathbf{H} := \mathop{\mathbb{E}}_{X^\star\sim\mathcal{Z}^\star}\left[\langle x, X^\star_{\ell,j}\rangle [P_{\mathcal{W}}(X^\star_{\ell,j})]^{\otimes 2}\right] = \mathbb{E}\left[\langle x, X^\star_{\ell,j}\rangle \mathcal{X}^{\otimes 2}\right] + \mathbb{E}\left[\langle x, X^\star_{\ell,j}\rangle \left(\mathcal{X}\otimes\Xi + \Xi\otimes\mathcal{X} + \Xi^{\otimes 2}\right)\right]$$

Using the above derivation we know w.h.p. $\left\|\left(\mathcal{X}\otimes\Xi + \Xi\otimes\mathcal{X} + \Xi^{\otimes 2}\right)\right\|_2 \leq \mathsf{poly}(k_\ell)\delta$. Also, using the randomness of $X^\star_{\ell,j}$ and Assumption 2.2, we have $\mathbb{E}\left[|\langle x, X^\star_{\ell,j}\rangle|\right] \leq \frac{\mathsf{poly}(k_\ell)}{m_\ell}$. Together, they imply

$$\left\|\mathbf{H} - \mathbb{E}\left[\langle x, X^\star_{\ell,j}\rangle [\mathcal{X}]^{\otimes 2}\right]\right\|_2 \leq \frac{\mathsf{poly}(k_\ell)\delta}{m_\ell}$$

On the other hand, by (a variant of) Claim D.3, we can easily conclude that:[12]

$$\left\|\mathbb{E}\left[\langle x, X^\star_{\ell,j}\rangle [\mathcal{X}]^{\otimes 2}\right] - \sum_{p\in[m_\ell]} \langle \mathbf{W}^\star_p, x\rangle [\mathbf{W}^\star_p]^{\otimes 2}\, \mathbb{E}\left[[\mathcal{S}^\star_p]^3 \mathbb{1}_{p\notin\mathcal{U}(X^\star_{\ell,j})}\right]\right\|_2 \leq \mathsf{poly}(k_\ell)\cdot(\varepsilon_1\sqrt{m_\ell} + \varepsilon_2 m_\ell)$$

Therefore, it remains to calculate $\mathbb{E}\left[[\mathcal{S}^\star_p]^3 \mathbb{1}_{p\notin\mathcal{U}(X^\star_{\ell,j})}\right]$.

For any $p \in \mathcal{P}$, recalling $\mathcal{U}(X^\star_{\ell,j}) \subseteq \mathcal{P}$, we have $p \notin \mathcal{U}(X^\star_{\ell,j})$ implies $\langle w_p, X^\star_{\ell,j}\rangle < b$ so $\mathcal{S}^\star_p = \langle \mathbf{W}^\star_p, X^\star_{\ell,j}\rangle < 2b$. Also, recall that $\mathbf{Pr}[\mathcal{S}^\star_p \neq 0] \leq \frac{\mathsf{poly}(k_\ell)}{m_\ell}$. Together, we have $\left|\mathbb{E}\left[[\mathcal{S}^\star_p]^3 \mathbb{1}_{p\notin\mathcal{U}(X^\star_{\ell,j})}\right]\right| \leq b^3\frac{\mathsf{poly}(k_\ell)}{m_\ell}$. On the other hand, for any $p \notin \mathcal{P}$, recalling $\mathcal{U}(X^\star_{\ell,j}) \subseteq \mathcal{P}$, we have $\mathbb{E}\left[[\mathcal{S}^\star_p]^3 \mathbb{1}_{p\notin\mathcal{U}(X^\star_{\ell,j})}\right] = \mathbb{E}\left[\mathcal{S}^3_p\right]$. This completes the proof. □

Now we have the following Claim that bound the eigenvalue gap of $\mathbf{H}$, we show that:

**Claim D.5.** *In Algorithm 5, for every $\ell \in [L], j \in [d_\ell]$, for every non-empty $\mathcal{R} \subseteq [m_\ell]$, let*

$$r^\star = \arg\max_{r\in\mathcal{R}}\left\{\mathbb{E}\left[\left([\mathcal{S}^\star_{\ell,j}]_r\right)^3\right]\langle \mathbf{W}^\star_r, x\rangle\right\}$$

$$V_1 = \max_{r\in\mathcal{R}}\left\{\mathbb{E}\left[\left([\mathcal{S}^\star_{\ell,j}]_r\right)^3\right]\langle \mathbf{W}^\star_r, x\rangle\right\}$$

$$V_2 = \max_{r\in\mathcal{R}, r\neq r^\star}\left\{\mathbb{E}\left[\left([\mathcal{S}^\star_{\ell,j}]_r\right)^3\right]\langle \mathbf{W}^\star_r, x\rangle\right\}$$

*We have w.p. $\geq \frac{1}{\mathsf{poly}(k_\ell)m_\ell^2}$ over the choice of $x$, it satisfies $V_1 - V_2 > \frac{1}{\mathsf{poly}(k_\ell)m_\ell}$.*

*Proof of Claim D.5.* Recall we have defined $\alpha_r \stackrel{\text{def}}{=} \mathbb{E}\left[([\mathcal{S}^\star_{\ell,j}]_r)^3\right]$.

To begin with, recall Assumption 2.2 implies for every $r \in [m_\ell]$, the probability $\mathbf{Pr}\left[[\mathcal{S}^\star_{\ell,j}]_r \neq 0\right] \geq \frac{1}{\mathsf{poly}(k_\ell)m_\ell}$. As a result, with probability at least $\frac{|\mathcal{R}|}{\mathsf{poly}(k_\ell)m_\ell}$, it satisfies $[\mathcal{S}^\star_{\ell,j}]_r \neq 0$ at least for one $r \in \mathcal{R}$ (by principle of inclusion-exclusion together with $\varepsilon_1 \leq \frac{\mathsf{poly}(k_\ell)}{m_\ell^2}$). Let this event be $\mathcal{E}_1$. By averaging, we know there exists an $r_0 \in \mathcal{R}$ such that

$$\mathbf{Pr}\left[\forall r \in \mathcal{R}, \alpha_r[\mathcal{S}^\star_{\ell,j}]_r \leq \alpha_{r_0}[\mathcal{S}^\star_{\ell,j}]_{r_0} \mid \mathcal{E}_1\right] \geq \frac{1}{|\mathcal{R}|} \ .$$

[12] Of course, if $\mathcal{U}(X^\star_{\ell,j}) \equiv \varnothing$ for any $X^\star_{\ell,j}$ this would exactly be Claim D.3 (this corresponds to the first iteration of Algorithm 5. Otherwise, one has to apply the proof of Claim D.3 which is analogous but more involved in notations.

Note that when this event happens, we necessarily have $[\mathcal{S}^\star_{\ell,j}]_{r_0} > 0$. Therefore, we can write

$$\mathbf{Pr}\left[\forall r \in \mathcal{R}, \alpha_r[\mathcal{S}^\star_{\ell,j}]_r \leq \alpha_{r_0}[\mathcal{S}^\star_{\ell,j}]_{r_0} \bigwedge [\mathcal{S}^\star_{\ell,j}]_{r_0} > 0\right] \geq \frac{1}{\mathsf{poly}(k_\ell) \cdot m_\ell} \ .$$

At the same time, Assumption 2.2 tells us $\mathbf{Pr}\left[[\mathcal{S}^\star_{\ell,j}]_{r_0} > \frac{1}{\mathsf{poly}(k_\ell)} \mid [\mathcal{S}^\star_{\ell,j}]_{r_0} > 0\right] \geq \frac{1}{\mathsf{poly}(k_\ell)}$. This further implies

$$\mathbf{Pr}\left[\mathcal{E}_2\right] \stackrel{\text{def}}{=} \mathbf{Pr}\left[\forall r \in \mathcal{R}, \alpha_r[\mathcal{S}^\star_{\ell,j}]_r \leq \alpha_{r_0}[\mathcal{S}^\star_{\ell,j}]_{r_0} \bigwedge [\mathcal{S}^\star_{\ell,j}]_{r_0} > \frac{1}{\mathsf{poly}(k_\ell)}\right] \geq \frac{1}{\mathsf{poly}(k_\ell) m_\ell}$$

On the other hand, for any $r_1 \in \mathcal{R} \setminus \{r_0\}$, we have $\mathbf{Pr}\left[[\mathcal{S}^\star_{\ell,j}]_{r_1} > 0 \wedge [\mathcal{S}^\star_{\ell,j}]_{r_0} > 0\right] \leq \frac{\mathsf{poly}(k_\ell)}{m_\ell^2}$ and therefore

$$\mathbf{Pr}\left[[\mathcal{S}^\star_{\ell,j}]_{r_1} > 0 \,\middle|\, \forall r \in \mathcal{R}, \alpha_r[\mathcal{S}^\star_{\ell,j}]_r \leq \alpha_{r_0}[\mathcal{S}^\star_{\ell,j}]_{r_0} \bigwedge [\mathcal{S}^\star_{\ell,j}]_{r_0} > \frac{1}{\mathsf{poly}(k_\ell)}\right] \leq \frac{\mathsf{poly}(k_\ell)}{m_\ell}$$

Now, let us recall we have 2 independent samples of $X^\star_{\ell,j}$ (and thus $\mathcal{S}^\star_{\ell,j}$). We know with probability at least $\frac{1}{\mathsf{poly}(k_\ell) m_\ell^2}$, for both $i \in \{1,2\}$, the $i$-th sample $[\mathcal{S}^\star_{\ell,j}]^{(i)}$ satisfies the above event $\mathcal{E}_2$. Let us denote by this event $\mathcal{E}_2^{(1)} \wedge \mathcal{E}_2^{(2)}$, so we have just derived $\mathbf{Pr}\left[[\mathcal{S}^\star_{\ell,j}]^{(i)}_{r_1} > 0 \mid \mathcal{E}_2^{(1)} \wedge \mathcal{E}_2^{(2)}\right] \leq \frac{\mathsf{poly}(k_\ell)}{m_\ell}$ for any $i \in \{1,2\}$. This further implies, for any $r_1 \in \mathcal{R} \setminus \{r_0\}$, with probability

$$\mathbf{Pr}\left[[\mathcal{S}^\star_{\ell,j}]^{(1)}_{r_1} > 0 \wedge [\mathcal{S}^\star_{\ell,j}]^{(2)}_{r_1} > 0 \,\middle|\, \mathcal{E}_2^{(1)} \wedge \mathcal{E}_2^{(2)}\right] \leq \frac{\mathsf{poly}(k_\ell)}{m_\ell^2} \ .$$

By union bound, we have

$$\mathbf{Pr}\left[\forall r_1 \in \mathcal{R} \setminus \{r_0\} \colon [\mathcal{S}^\star_{\ell,j}]^{(1)}_{r_1} = 0 \vee [\mathcal{S}^\star_{\ell,j}]^{(2)}_{r_1} = 0 \,\middle|\, \mathcal{E}_2^{(1)} \wedge \mathcal{E}_2^{(2)}\right] \geq 1 - \frac{\mathsf{poly}(k_\ell)}{m_\ell} \geq 0.5 \ .$$

In sum, we just concluded with probability at least $\frac{1}{\mathsf{poly}(k_\ell) m_\ell^2}$ over the choice of $x$, it satisfies

- $\forall i \in \{1,2\}, \forall r \in \mathcal{R}, \alpha_r[\mathcal{S}^\star_{\ell,j}]^{(i)}_r \leq \alpha_{r_0}[\mathcal{S}^\star_{\ell,j}]^{(i)}_{r_0}$, and $[\mathcal{S}^\star_{\ell,j}]^{(i)}_{r_0} > \frac{1}{\mathsf{poly}(k_\ell)}$
- $\forall r_1 \in \mathcal{R} \setminus \{r_0\}$, either $[\mathcal{S}^\star_{\ell,j}]^{(1)}_r = 0$ or $[\mathcal{S}^\star_{\ell,j}]^{(2)}_r = 0$.

Under these two conditions, one can easily conclude that for every $r_1 \in \mathcal{R} \setminus \{r_0\}$:

$$\alpha_{r_0}\langle \mathbf{W}^\star_{r_0}, x\rangle \geq \alpha_{r_1}\langle \mathbf{W}^\star_{r_1}, x\rangle + \frac{\alpha_{r_0}}{\mathsf{poly}(k_\ell)} \geq \alpha_{r_1}\langle \mathbf{W}^\star_{r_1}, x\rangle + \frac{1}{\mathsf{poly}(k_\ell) m_\ell} \ ,$$

where the last inequality uses $\alpha_{r_0} \geq \frac{1}{\mathsf{poly}(k_\ell) m}$ which can be carefully derived from Assumption 2.2. □

**Claim D.6.** *For every $\ell \in [L], d \in [d_\ell]$, for every $\delta \in (0, m^{-0.005}]$ and $b \in [\mathsf{poly}(k_\ell)\delta, 1/\mathsf{poly}(k_\ell)]$, for every $v$ such that there is $p \in [m_\ell]$ with $\|v - [\mathbf{W}^\star_{\ell,j}]_p\|_2 \leq \delta$, we have: there is $\alpha \geq \frac{1}{\mathsf{poly}(k_\ell) m_\ell}$ such that*

$$\|\,\mathbb{E}[\mathbb{1}_{\langle v, X^\star_{\ell,j}\rangle \geq b} X^\star_{\ell,j}] - \alpha[\mathbf{W}^\star_{\ell,j}]_p\|_2 \leq \frac{\mathsf{poly}(k_\ell)}{m_\ell^{3/2}}$$

*Proof of Claim D.6.* Let us write

$$\mathbb{1}_{\langle v, X^\star_{\ell,j}\rangle \geq b} X^\star_{\ell,j} = \sum_{q\in[m_\ell]} \left(\mathbb{1}_{\langle v, X^\star_{\ell,j}\rangle \geq b} [\mathcal{S}^\star_{\ell,j}]_q\right) [\mathbf{W}^\star_{\ell,j}]_q \enspace .$$

Since $\{[\mathbf{W}^\star_{\ell,j}]_q\}_q$ are orthogonal vectors, it suffices to bound the (non-negative) coefficients.

- When $q = p$, note that $\langle v, X^\star_{\ell,j}\rangle \geq b$ is implied by $[\mathcal{S}^\star_{\ell,j}]_p > 2b > b + \delta\mathsf{poly}(k_\ell)$ (see relationship (D.2)). We therefore have using Assumption 2.2:

$$\mathbb{1}_{\langle v, X^\star_{\ell,j}\rangle \geq b} [\mathcal{S}^\star_{\ell,j}]_p \geq \mathbb{1}_{[\mathcal{S}^\star_{\ell,j}]_p \geq 2b} [\mathcal{S}^\star_{\ell,j}]_p \geq \frac{1}{\mathsf{poly}(k_\ell) m_\ell}$$

- When $q \neq p$, note that $\langle v, X^\star_{\ell,j}\rangle \geq b$ implies $[\mathcal{S}^\star_{\ell,j}]_p > 0$ (see relationship (D.2)). We therefore have using Assumption 2.2:

$$\mathbb{1}_{\langle v, X^\star_{\ell,j}\rangle \geq b} [\mathcal{S}^\star_{\ell,j}]_q \leq \mathbb{1}_{[\mathcal{S}^\star_{\ell,j}]_p > 0} [\mathcal{S}^\star_{\ell,j}]_q \leq \mathsf{poly}(k_\ell)\varepsilon_1 \leq \frac{\mathsf{poly}(k_\ell)}{m_\ell^2}$$

Together we finish the proof. □

## D.1 Proof of Lemma D.1

*Proof of Lemma D.1.* We again abbreviate by writing $\mathbf{W}^\star = \mathbf{W}^\star_{\ell,j}$. Let us prove by induction (by the size $|\mathcal{W}|$) that for every $w \in \mathcal{W}$, there is a *unique* $r \in [m_\ell]$ satisfying

$$\|w - \mathbf{W}^\star_r\|_2 \leq \frac{\mathsf{poly}(k_\ell)}{\sqrt{m_\ell}} \enspace .$$

Suppose at this stage we have $|\mathcal{W}| = i \in \{0, 1, \ldots, m_\ell - 1\}$. Let us denote again by $\mathcal{P}$ the set of indicies $p \in [m_\ell]$ such that there exists $w \in \mathcal{W}$ with $\|\mathbf{W}^\star_p - w\|_2 \leq \delta$, and let us denote by this (unique) $w$ as $w_p$. It is clear that $|\mathcal{W}| = |\mathcal{P}|$.

Applying Claim D.4, we know if we define $\mathbf{H}' \stackrel{\text{def}}{=} \sum_{r\in[m_\ell]} \alpha'_r \langle \mathbf{W}^\star_r, x\rangle [\mathbf{W}^\star_r]^{\otimes 2}$, then

$$\left\|\mathbf{H} - \mathbf{H}'\right\|_2 \leq \mathsf{poly}(k_\ell) \cdot \left(\frac{\delta}{m_\ell} + \varepsilon_1\sqrt{m_\ell} + \varepsilon_2 m_\ell\right) \leq \varepsilon' = \frac{1}{m^{1.009}}$$

where $\alpha'_r = \mathbb{E}\left[([\mathcal{S}^\star_{\ell,j}]_r)^3\right]$ for any $r \notin \mathcal{P}$ and for $\alpha'_r \leq \frac{\mathsf{poly}(k_\ell) b^3}{m_\ell}$ for any $r \in \mathcal{P}$. Note the eigenvectors of $\mathbf{H}'$ are precisely $\mathbf{W}^\star_1, \ldots, \mathbf{W}^\star_{m_\ell}$.

By Claim D.5, we know with probability at least $\frac{1}{\mathsf{poly}(k_\ell) m_\ell^2}$ over the choice of $x$, the eigengap (i.e. gap between the largest and the second largest eigenvalue) of $\mathbf{H}'$ is at least $V_1 - V_2 > \frac{1}{\mathsf{poly}(k_\ell) m_\ell}$; and when this good event on $x$ happens, we necessarily have

- $\mathbf{W}^\star_{r_0}$ is the top eigenvector of $\mathbf{H}'$ for some $r_0 \in \mathcal{P}$ (this uses $b \leq \frac{1}{\mathsf{poly}(k_\ell)}$), and
- the eigengap of $\mathbf{H}$ is at least $V_1 - V_2 - 2\varepsilon' > \frac{1}{\mathsf{poly}(k_\ell) m_\ell}$.

On the forward direction, under this good event on $x$, we can apply the gap-free Wedin's theorem [1, Lemma B.3], which tells us if the algorithm calculates $v$ as the top (unit) eigenvector for $\mathbf{H}$, then it corresponds to eigenvalue at least $V_1 - \varepsilon'$, and

$$\|v - \mathbf{W}^\star_{r_0}\|_2 \leq \frac{\varepsilon'}{V_1 - V_2 - \varepsilon'} \leq \frac{m_\ell^{-1.009}}{\frac{1}{\mathsf{poly}(k_\ell) m_\ell}} \leq \frac{1}{m_\ell^{0.008}}$$

This means $\langle v, w\rangle < 0.5$ for any $w \in \mathcal{W}$, so Algorithm 5 can reach the "inner if" Line 8 with probability at least $\frac{1}{\mathsf{poly}(k_\ell) m_\ell^2}$ (for each generated random $x$). (This gives the iteration complexity $T$ for Algorithm 5.)

On the reverse direction, as long as Algorithm 5 reaches Line 8, we know the eigengap of $\mathbf{H}$ is at least $\frac{1}{\mathsf{poly}(k_\ell) m_\ell}$, and by the gap-free Wedin's theorem again, letting $v$ be the top eigenvector of $\mathbf{H}$ and letting $\mathbf{W}_r^\star$ be the top eigenvector of $\mathbf{H}'$, then

$$\|v - \mathbf{W}_r^\star\|_2 \le \frac{m_\ell^{-1.009}}{\frac{1}{\mathsf{poly}(k_\ell) m_\ell}} \le \frac{1}{m_\ell^{0.008}} \ ;$$

furthermore, this $r$ cannot be inside $\mathcal{P}$ since otherwise $\langle v, w\rangle$ would be greater than 0.5 for some $w \in \mathcal{W}$. Therefore, we can apply Claim D.6 to derive that

$$\|v' - \mathbf{W}_r^\star\|_2 \le \frac{\mathsf{poly}(k_\ell)}{\sqrt{m_\ell}} \ .$$

□

# E Learning the Output Layer

The goal of this section is to prove Lemma E.2 that we restate below. We focus on a fixed layer $\ell \in [L]$ and throughout this section, we assume

**Induction E.1.** *At the beginning of Algorithm 1, there exists some $\delta \in (0, m_\ell^{-0.49}]$ such that for every $j \in [d_\ell], p \in [m_\ell]$, $\|[\mathbf{W}^\star_{\ell,j}]_p - [\mathbf{W}_{\ell,j}]_p\|_2 \le \delta$. We choose $b = \delta \cdot m_\ell^{0.02}$ and $\mathbb{b} = b \cdot m_\ell^{0.152}$, and fix $\mathbf{W}_\ell^D = \mathbf{W}_\ell$ to be this initial value.*

*In addition, either Lemma F.10 holds (for $\ell = 1$) or Lemma G.2 holds ($\ell > 1$). (In particular, this implies w.h.p. $\|\mathsf{ReLU}(\mathcal{S}_{\ell,j})\|_0 \le m_\ell^{0.01}$ and $\|\mathcal{S}_{\ell,j}\|_\infty \le \mathsf{poly}(k_\ell)$ for every $j \in [d_\ell]$.)*

**Lemma E.2** (learn output layer)**.** *For every $\ell \in [L]$, under Induction E.1, after running the gradient descent-ascent process on $D^{(1)}$ in Algorithm 1 to learn $\mathbf{W}_\ell$, we have for every $j \in [d_\ell], r \in [m_\ell]$:*

$$\|[\mathbf{W}_{\ell,j}]_r - [\mathbf{W}^\star_{\ell,j}]_r\|_2 \le m_\ell^{0.152}\mathbb{b}^2 \le m_\ell^{-0.02}\delta$$

Recall for given $\ell \in [L]$ and $j \in [d_\ell]$, the min-max objective for learning the output layer is

$$\min\nolimits_{\mathbf{W}_{\ell,j}} \max\nolimits_{V^D_{\ell,j}} \Big( \Big( \mathbb{E}[D^{(1)}_{\ell,j}(X^\star_\ell)] - \mathbb{E}[D^{(1)}_{\ell,j}(X_\ell)] \Big) - \textstyle\sum_{r\in[m_\ell]} \|V^D_{\ell,j,r}\|_2^{1+c} \Big) \tag{E.1}$$

where

$$D^{(1)}_{\ell,j}(Y) \stackrel{\text{def}}{=} \textstyle\sum_{r\in[m_\ell]} \Big( \mathsf{ReLU}'([(\mathbf{W}^D_{\ell,j})^\top Y_j]_r - \mathbb{b}) \langle Y_j, V^D_{\ell,j,r}\rangle \Big)$$

Note the min-max objective (E.1) with respect to the discriminator (i.e., $V^D_{\ell,j}$) is concave (indeed, linear plus $\|V^D_{\ell,j,r}\|_2^{1+c}$), so during the inner gradient ascent process, the discriminator can converge to an (approximate) maximizer efficiently. For simplicity, we can without loss of generality assume as if the discriminator has reached the (unique) exact maximizer, for reasons that are explained in Footnote 18.

After substituting the exact maximizer for $V^D_{\ell,j}$ into the min-max objective (E.1), we can rewrite it as a minimization problem on the generator side:[13]

$$\mathsf{Obj}^{(1)}_{\ell,j}(\mathbf{W}_{\ell,j}) = \left( \frac{1}{(1+c)^{1/c}} - \frac{1}{(1+c)^{1+1/c}} \right) \times$$

[13] This uses the fact that for every constant $c > 0$, every $a_1, a_2, \cdots, a_m \ge 0$,

$$\min_{c_1, c_2, \cdots, c_m \in \mathbb{R}} \sum_{r\in[m]} \left( c_r a_r + |c_r|^{1+c} \right) = - \sum_{r\in[m]} \left( \frac{1}{(1+c)^{1/c}} - \frac{1}{(1+c)^{1+1/c}} \right) a_r^{1+1/c} \ .$$

$$\sum_{r\in[m_\ell]} \left\| \mathbb{E}\left[\mathsf{ReLU}'(\underbrace{[(\mathbf{W}^D_{\ell,j})^\top X_{\ell,j}]_r}_{\overset{\text{def}}{=} s} - \mathbb{b})X_{\ell,j}\right] - \mathbb{E}\left[\mathsf{ReLU}'(\underbrace{[(\mathbf{W}^D_{\ell,j})^\top X^\star_{\ell,j}]_r}_{\overset{\text{def}}{=} s^\star} - \mathbb{b})X^\star_{\ell,j}\right] \right\|_2^{1+1/c}$$

$$= \left(\frac{1}{(1+c)^{1/c}} - \frac{1}{(1+c)^{1+1/c}}\right) \times \sum_{r\in[m_\ell]} \left\| \underbrace{\mathbb{E}\left[\mathbb{1}_{s_r-\mathbb{b}>0}X_{\ell,j}\right]}_{\overset{\text{def}}{=} R_p} - \underbrace{\mathbb{E}\left[\mathbb{1}_{s^\star_r-\mathbb{b}>0}X^\star_{\ell,j}\right]}_{\overset{\text{def}}{=} R^\star_p} \right\|_2^{1+1/c} \tag{E.2}$$

Above, for notation simplicity, we have defined (note we have dropped the index $j$ for notation simplicty)

$$s \overset{\text{def}}{=} [\mathbf{W}^D_{\ell,j}]^\top X_{\ell,j} = [\mathbf{W}^D_{\ell,j}]^\top \mathbf{W}_{\ell,j}\mathcal{S}_{\ell,j}, \qquad s^\star \overset{\text{def}}{=} [\mathbf{W}^D_{\ell,j}]^\top X^\star_{\ell,j} = [\mathbf{W}^D_{\ell,j}]^\top \mathbf{W}^\star_{\ell,j}\mathcal{S}^\star_{\ell,j}$$
$$R_p := \mathbb{E}\left[\mathbb{1}_{s_p-\mathbb{b}>0}X_{\ell,j}\right], \qquad R^\star_p := \mathbb{E}\left[\mathbb{1}_{s^\star_p-\mathbb{b}>0}X^\star_{\ell,j}\right] \tag{E.3}$$

We shall prove later in Claim E.9 that

**Invariant E.3.** *Throughout the training of $D^{(1)}$, we shall maintain the following invariant property:*[14]

$$\forall j \in [d_\ell], p \in [m_\ell]: \quad \|[\mathbf{W}^\star_{\ell,j}]_p - [\mathbf{W}_{\ell,j}]_p\|_2 \le b\mathsf{poly}(k_\ell)\cdot m_\ell^{0.151} \ .$$

## E.1 Axuliary Claims: Between $\|R^\star - R\|_2$ and $\|\mathbf{W}^\star - \mathbf{W}\|_2$

We first establish some critical relationships Claim E.6 and Claim E.7 between $\left\|R^\star_p - R_p\right\|_2$ (corresponding to our objective (E.2)) and $\left\|[\mathbf{W}^\star_{\ell,j}]_p - [\mathbf{W}^\star_{\ell,j}]_p\right\|_2$ (corresponding to the final error).

**Claim E.4** (decoding)**.** *For every $\ell \in [L], j \in [d_\ell]$, suppose for some $\delta \in (0, m_\ell^{-0.1}]$ it satisfies*

$$\forall p \in [m_\ell]: \quad \|[\mathbf{W}^\star_{\ell,j}]_p - [\mathbf{W}_{\ell,j}]_p\|_2 \le \delta \quad \textit{and} \quad \|[\mathbf{W}^\star_{\ell,j}]_p - [\mathbf{W}^D_{\ell,j}]_p\|_2 \le \delta$$

*Suppose w.h.p. $\|\mathsf{ReLU}(\mathcal{S}_{\ell,j})\|_0 \le m_\ell^{0.01}$, $\|\mathcal{S}_{\ell,j}\|_\infty \le \mathsf{poly}(k_\ell)$. Then, w.h.p.*

$$\forall r \in [m_\ell]: \qquad s^\star_r = [\mathcal{S}^\star_{\ell,j}]_r \pm \delta\mathsf{poly}(k_\ell) \quad \textit{and} \quad s_r = [\mathcal{S}_{\ell,j}]_r \pm \delta m_\ell^{0.02}$$

**Fact E.5.** *Suppose $g \sim \mathcal{N}(0,1)$ satisfies $\mathbf{Pr}[g > b] \le \frac{\mathsf{poly}(k_\ell)}{m_\ell}$, then for any $\delta \le 1$, we have*

$$\mathop{\mathbb{E}}_{g\sim\mathcal{N}(0,1)}[\mathbb{1}_{|g-b|\le\delta}\mathsf{ReLU}(g-b)] \le \frac{\delta^2\mathsf{poly}(k_\ell)}{m_\ell} \ .$$

*Proof.* It is trivial since the probability for $|g-b| \le \delta$ to hold is at most $\frac{\delta\mathsf{poly}(k_\ell)}{m_\ell}$. □

Finally, we have the following Claim:

**Claim E.6** ($\|R^\star - R\|_2$ vs. $\|\mathbf{W}^\star - \mathbf{W}\|_2$)**.** *Under Induction E.1, suppose Invariant E.3 is satisfied at the current point of training for $D^{(1)}$.*

*For every $p \in [m_\ell]$, recall we have defined $s_p, s^\star_p, R_p, R^\star_p$ in (E.3). For every $p, q \in [m_\ell]$, define*

$$\theta_p = \frac{\mathbb{E}\left[\mathsf{ReLU}([\mathcal{S}_{\ell,j}]_p)\right]}{\mathbb{E}\left[\mathsf{ReLU}([\mathcal{S}^\star_{\ell,j}]_p)\right]} \quad \textit{and} \quad \mu'_{p,q} = \mathbb{E}\left[\mathbb{1}_{s^\star_p-\mathbb{b}>0}[\mathcal{S}^\star_{\ell,j}]_q\right]$$

[14] We remark here the factor $m_\ell^{0.15}$ comes from our minimum singular value assumption Assumption B.1.

*Then, we have* $\theta_p = 1 \pm b\mathsf{poly}(k_\ell)$ *and*

$$\left\|R^\star_p - R_p\right\|_2 = \left\|\sum_{q\in[m_\ell]} \mu'_{p,q}\left([\mathbf{W}^\star_{\ell,j}]_q - \theta_q[\mathbf{W}_{\ell,j}]_q\right)\right\|_2 \pm \varepsilon_4 \quad \textit{for } \varepsilon_4 := \frac{\mathbb{b}^2\mathsf{poly}(k_\ell)}{m_\ell} + \frac{\mathbb{b}\mathsf{poly}(k_\ell)}{m_\ell^{1.5}}$$

*Proof of Claim E.6.* First of all, the corollaries in Lemma F.6 (for $\ell = 1$) and Lemma G.2 (for $\ell > 1$) imply

$$\theta_p \in [1 - b\mathsf{poly}(k_\ell), 1 + b\mathsf{poly}(k_\ell)]$$

Now, for every $p \neq q$, Lemma F.10 (for $\ell = 1$) or the corollary in Lemma G.2 (for $\ell > 1$) give

$$\left|\mathbb{E}\left[\mathbb{1}_{s_p - \mathbb{b} > 0}[\mathcal{S}_{\ell,j}]_q\right] - \mathbb{E}\left[\mathbb{1}_{s^\star_p - \mathbb{b} > 0}[\mathcal{S}^\star_{\ell,j}]_q\right]\right| \leq \frac{\mathbb{b}\mathsf{poly}(k_1)}{m_\ell^2}$$

Also, one can derive using Induction E.1 together with w.h.p. $\|\mathsf{ReLU}(\mathcal{S}^\star_{\ell,j})\|_0 \leq k_\ell$ and $\|\mathcal{S}_{\ell,j}\|_\infty \leq \mathsf{poly}(k_\ell)$ (see Assumption 2.2) that $s^\star_p = [\mathcal{S}^\star_{\ell,j}]_p \pm bm^{0.01} \in [\mathcal{S}^\star_{\ell,j}]_p \pm 0.01\mathbb{b}$, and therefore

$$\mathbb{E}\left[\mathbb{1}_{s^\star_p - \mathbb{b} > 0}[\mathcal{S}^\star_{\ell,j}]_q\right] \leq \mathbb{E}\left[\mathbb{1}_{[\mathcal{S}^\star_{\ell,j}]_p > 0}[\mathcal{S}^\star_{\ell,j}]_q\right] \leq \mathsf{poly}(k_\ell)\,\mathbb{E}\left[\mathbb{1}_{[\mathcal{S}^\star_{\ell,j}]_p > 0}\mathbb{1}_{[\mathcal{S}^\star_{\ell,j}]_q > 0}\right] \leq \frac{\mathsf{poly}(k_\ell)}{m_\ell^2}$$

so we also have

$$\left|\mathbb{E}\left[\mathbb{1}_{s_p - \mathbb{b} > 0}[\mathcal{S}_{\ell,j}]_q\right] - \theta_q\,\mathbb{E}\left[\mathbb{1}_{s^\star_p - \mathbb{b} > 0}[\mathcal{S}^\star_{\ell,j}]_q\right]\right| \leq \frac{\mathbb{b}\mathsf{poly}(k_1)}{m_\ell^2} \tag{E.4}$$

As for $p = q$, one can derive using Invariant E.3 together with w.h.p. $\|\mathsf{ReLU}(\mathcal{S}_{\ell,j})\|_0 \leq m_\ell^{0.01}$ and $\|\mathcal{S}_{\ell,j}\|_\infty \leq \mathsf{poly}(k_\ell)$ (see Induction E.1) that $s_p = [\mathcal{S}_{\ell,j}]_p \pm bm_\ell^{0.151}\mathsf{poly}(k_\ell) \in [\mathcal{S}_{\ell,j}]_p \pm 0.01\mathbb{b}$. Together with $s^\star_p = [\mathcal{S}^\star_{\ell,j}]_p \pm 0.01\mathbb{b}$, we can derive the following:

- Using the anti-concentration Assumption B.1, we have

$$\begin{aligned}\mathbb{E}\left[\mathbb{1}_{s^\star_p - \mathbb{b} > 0}[\mathcal{S}^\star_{\ell,j}]_p\right] &= \mathbb{E}\left[\mathbb{1}_{[\mathcal{S}^\star_{\ell,j}]_p > 0}[\mathcal{S}^\star_{\ell,j}]_p\right] \pm \mathbb{E}\left[\mathbb{1}_{|[\mathcal{S}^\star_{\ell,j}]_p| \leq 2\mathbb{b}} \cdot |[\mathcal{S}^\star_{\ell,j}]_p|\right] \\ &= \mathbb{E}\left[\mathbb{1}_{[\mathcal{S}^\star_{\ell,j}]_p > 0}[\mathcal{S}^\star_{\ell,j}]_p\right] \pm \frac{\mathbb{b}^2\mathsf{poly}(k_\ell)}{m_\ell} = \mathbb{E}\left[\mathsf{ReLU}([\mathcal{S}^\star_{\ell,j}]_p)\right](1 \pm \mathbb{b}^2\mathsf{poly}(k_\ell))\end{aligned} \tag{E.5}$$

- In the case of $\ell = 1$, using Fact E.5; and in the case of $\ell > 1$, using the corollary in Lemma G.2, we have

$$\begin{aligned}\mathbb{E}\left[\mathbb{1}_{s_p - \mathbb{b} > 0}[\mathcal{S}_{\ell,j}]_p\right] &= \mathbb{E}\left[\mathbb{1}_{[\mathcal{S}_{\ell,j}]_p > 0}[\mathcal{S}_{\ell,j}]_p\right] \pm \mathbb{E}\left[\mathbb{1}_{|[\mathcal{S}_{\ell,j}]_p| \leq 2\mathbb{b}} \cdot |[\mathcal{S}_{\ell,j}]_p|\right] \\ &= \mathbb{E}\left[\mathbb{1}_{[\mathcal{S}_{\ell,j}]_p > 0}[\mathcal{S}_{\ell,j}]_p\right] \pm \frac{\mathbb{b}^2\mathsf{poly}(k_\ell)}{m_\ell} = \mathbb{E}\left[\mathsf{ReLU}([\mathcal{S}_{\ell,j}]_p)\right](1 \pm \mathbb{b}^2\mathsf{poly}(k_\ell))\end{aligned}$$

They together imply

$$\left|\mathbb{E}\left[\mathbb{1}_{s_p - \mathbb{b} > 0}[\mathcal{S}_{\ell,j}]_p\right] - \theta_p\,\mathbb{E}\left[\mathbb{1}_{s^\star_p - \mathbb{b} > 0}[\mathcal{S}^\star_{\ell,j}]_p\right]\right| \leq \frac{\mathbb{b}^2\mathsf{poly}(k_\ell)}{m_\ell} \tag{E.6}$$

Now, combining (E.4) and (E.6), and abbreviating $\mathbf{W} = \mathbf{W}_{\ell,j}$ and $\mathbf{W}^\star = \mathbf{W}^\star_{\ell,j}$, we have

$$\begin{aligned}\left\|R^\star_p - R_p\right\|_2 &= \left\|\mathbb{E}\textstyle\sum_{q\in[m_\ell]} \mathbb{1}_{s^\star_p - \mathbb{b} > 0}\mathbf{W}^\star_q[\mathcal{S}^\star_{\ell,j}]_q - \mathbb{1}_{s_p - \mathbb{b} > 0}\mathbf{W}_q[\mathcal{S}_{\ell,j}]_q\right\|_2 \\ &= \left\|\textstyle\sum_{q\in[m_\ell]} \mu'_{p,q}\left(\mathbf{W}^\star_q - \theta_q\mathbf{W}_q\right)\right\|_2 \pm \left\|\textstyle\sum_{q\in[m_\ell]} \mathbb{E}\left[\theta_q\mathbb{1}_{s^\star_p - \mathbb{b} > 0}[\mathcal{S}^\star_{\ell,j}]_q - \mathbb{1}_{s_p - \mathbb{b} > 0}[\mathcal{S}_{\ell,j}]_q\right]\mathbf{W}_q\right\|_2 \\ &= \left\|\textstyle\sum_{q\in[m_\ell]} \mu'_{p,q}\left(\mathbf{W}^\star_q - \theta_q\mathbf{W}_q\right)\right\|_2 \pm \frac{\mathbb{b}^2\mathsf{poly}(k_\ell)}{m_\ell} \pm \left\|\textstyle\sum_{q\in[m_\ell], q\neq p} \mathbf{W}_q \cdot (\pm\frac{\mathbb{b}\mathsf{poly}(k_\ell)}{m_\ell^2})\right\|_2\end{aligned}$$

$$\overset{①}{=} \left\| \sum_{q\in[m_\ell]} \mu'_{p,q} \left( \mathbf{W}^\star_q - \theta_q \mathbf{W}_q \right) \right\|_2 \pm \frac{\mathbb{b}^2 \mathsf{poly}(k_\ell)}{m_\ell} \pm \frac{\mathbb{b}\mathsf{poly}(k_\ell)}{m_\ell^{1.5}}$$

Above, ① uses the property that $\|\mathbf{W}_p - \mathbf{W}^\star_p\| \leq \delta$ and $\{\mathbf{W}^\star_q\}_q$ forms an orthonormal basis, which implies for any sequence of $\alpha_p \in [-1, 1]$, it satisfies $\| \sum_{p\in[m_\ell]} \alpha_p \mathbf{W}_p\|_2 \leq O(\sqrt{m_\ell}) + O(\delta m_\ell)$

□

**Claim E.7** (from $\|R^\star - R\|_2$ to $\|\mathbf{W}^\star - \mathbf{W}\|_2$)**.** *In the same setting as Claim E.6, suppose for every $p \in [m_\ell]$*

$$\left\|R^\star_p - R_p\right\|_2 \leq \varepsilon_5$$

*for any $\varepsilon_5 > \varepsilon_4 = \frac{\mathbb{b}^2\mathsf{poly}(k_\ell)}{m_\ell} + \frac{\mathbb{b}\mathsf{poly}(k_\ell)}{m_\ell^{1.5}}$ and $\varepsilon_5 < \frac{1}{m_\ell^{0.99}}$, Then we have:*[15]

$$\|[\mathbf{W}^\star_{\ell,j}]_p - \theta_p[\mathbf{W}_{\ell,j}]_p\|_2 \ , \quad \|[\mathbf{W}^\star_{\ell,j}]_p - \frac{[\mathbf{W}_{\ell,j}]_p}{\|[\mathbf{W}_{\ell,j}]_p\|_2}\|_2 \leq O(\mathsf{poly}(k_\ell) m_\ell^{1.15} \varepsilon_5)$$

*Proof of Claim E.7.* Again, we abbreviate by writing $\mathbf{W} = \mathbf{W}_{\ell,j}$ and $\mathbf{W}^\star = \mathbf{W}^\star_{\ell,j}$. By Claim E.6, we know

$$\left\| \sum_{q\in[m_\ell]} \mu'_{p,q} \left( \mathbf{W}^\star_q - \theta_q \mathbf{W}_q \right) \right\|_2 \leq O(\varepsilon_5) \tag{E.7}$$

This implies

$$\sum_{p\in[m_\ell]} \left\| \sum_{q\in[m_\ell]} \mu'_{p,q} \left( \mathbf{W}^\star_q - \theta_q \mathbf{W}_q \right) \right\|_2^2 = O(\varepsilon_5^2 m_\ell)$$

This means if we write $\mu' = (\mu'_{p,q})_{p,q\in[m_\ell]}$ as a matrix, and write $\Theta = \mathbf{diag}\{\theta_1, \dots, \theta_{m_\ell}\}$, then

$$\|\mu'(\mathbf{W}^\star - \Theta\mathbf{W})\|_F^2 \leq O(\varepsilon_5^2 m_\ell) \tag{E.8}$$

If we also define $\mu_{p,q} = \mathbb{E}[\mathbb{1}_{[\mathcal{S}^\star_{\ell,j}]_p>0}[\mathcal{S}^\star_{\ell,j}]_q]$ and matrix $\mu = (\mu_{p,q})_{p,q\in[m_\ell]}$, then we know

- When $p \neq q$,

$$|\mu_{p,q} - \mu'_{p,q}| \overset{①}{\leq} \mathbb{E}\left[\mathbb{1}_{|[\mathcal{S}^\star_{\ell,j}]_p| \leq 2\mathbb{b}}[\mathcal{S}^\star_{\ell,j}]_q\right] \overset{②}{\leq} \mathbb{b}\mathsf{poly}(k_\ell)\, \mathbf{Pr}[[\mathcal{S}^\star_{\ell,j}]_p > 0, [\mathcal{S}^\star_{\ell,j}]_q > 0] \leq \frac{\mathbb{b}\mathsf{poly}(k_\ell)}{m_\ell^2}$$

  Above, inequality ① uses the definition $\mu'_{p,q} = \mathbb{E}\left[\mathbb{1}_{s^\star_p - \mathbb{b}>0}[\mathcal{S}^\star_{\ell,j}]_q\right]$ in which $s^\star_p = [\mathcal{S}^\star_{\ell,j}]_p \pm \delta\mathsf{poly}(k_\ell))$; an inequality ② uses Proposition F.25b.

- When $p = q$, (E.5) tells us

$$|\mu_{p,p} - \mu'_{p,p}| \leq \mathbb{b}^2\mathsf{poly}(k_\ell) \cdot \mu_{p,p} \leq \frac{\mathbb{b}^2\mathsf{poly}(k_\ell)}{m_\ell}$$

In other words, we have

$$\mu' = \mu + \Lambda + \Delta$$

for some diagonal matrix $\Lambda$ with entries no more than $\frac{\mathbb{b}^2\mathsf{poly}(k_\ell)}{m_\ell}$, and matrix $\Delta$ with $\|\Delta\|_F \leq \frac{\mathbb{b}\mathsf{poly}(k_\ell)}{m_\ell}$.

This means, using $\mathbb{b} < m_\ell^{-0.2}$ from Induction E.1, our assumption on $\sigma_{\min}(\mu) \geq \frac{1}{m_\ell^{1.15}}$ also implies $\sigma_{\min}(\mu') \geq \frac{1}{2m_\ell^{1.15}}$. Letting $\rho := \sigma_{\min}(\mu')$, we can thus derive from (E.8) that

$$\|\mathbf{W}^\star - \Theta\mathbf{W}\|_F^2 \leq O(\varepsilon_5^2 m_\ell/\rho^2)$$

[15]We remark here the factor $m_\ell^{1.15}$ comes from our minimum singular value assumption Assumption B.1.

Hence, applying (E.7), we have that for every $p \in [m_\ell]$,

$$\begin{aligned}
\|\mathbf{W}^\star_p - \theta_p \mathbf{W}_p\|_2 &\leq \frac{O(\varepsilon_5) + \left\|\sum_{q\in[m_\ell],q\neq p} \mu'_{p,q}\left(\mathbf{W}^\star_q - \theta_q \mathbf{W}_q\right)\right\|_2}{\mu'_{p,p}} \\
&\leq \frac{O(\varepsilon_5) + \sqrt{\sum_{q\in[m_\ell],q\neq p}(\mu'_{p,q})^2}\sqrt{\sum_{q\in[m_\ell],q\neq p}\left\|\mathbf{W}^\star_q - \theta_q\mathbf{W}_q\right\|_2^2}}{\mu'_{p,p}} \\
&\leq \frac{O(\varepsilon_5) + \frac{\mathsf{poly}(k_\ell)}{m_\ell^{1.5}} \cdot \frac{\varepsilon_5\sqrt{m_\ell}}{\rho}}{\frac{1}{\mathsf{poly}(k_\ell) m_\ell}} \\
&\leq \mathsf{poly}(k_\ell) m_\ell \cdot O(\varepsilon_5)(1 + \mathsf{poly}(k_\ell) m_\ell^{-1}/\rho) \leq O(\mathsf{poly}(k_\ell) m_\ell^{1.15}\varepsilon_5)
\end{aligned}$$

Note since $\|\mathbf{W}^\star_p\|_2 = 1$, we also have the same bound for $\|\mathbf{W}^\star_p - \frac{\mathbf{W}_p}{\|\mathbf{W}_p\|}\|_2$.

□

## E.2 Proof of Lemma E.2: Optimization using Gradient Descent

At the beginning of gradient descent on the generator's objective (E.2) (thus at the beginning of Algorithm 1), we start with $\mathbf{W}_{\ell,j} = \mathbf{W}^D_{\ell,j}$ and it satisfies

$$\forall \ell \in [d_\ell] \quad , \quad \forall p \in [m_\ell] \colon \quad \|[\mathbf{W}^\star_{\ell,j}]_p - [\mathbf{W}_{\ell,j}]_p\|_2 \leq \delta$$

Applying Claim E.6 we know

$$\left\|R^\star_p - R_p\right\|_2 = \left\|\sum_{q\in[m_\ell]} \mu'_{p,q}\left([\mathbf{W}^\star_{\ell,j}]_q - \theta_q[\mathbf{W}_{\ell,j}]_q\right)\right\|_2 \pm \varepsilon_4$$

It is easy to see that $\|[\mathbf{W}^\star_{\ell,j}]_q - \theta_q[\mathbf{W}_{\ell,j}]_q\|_2 \leq b\mathsf{poly}(k_\ell)$. Simple derivations (see the proof of Claim E.6) also implies $\mu'_{p,q} \leq \frac{\mathsf{poly}(k_\ell)}{m_\ell^2}$ (for $p \neq q$) and $\mu'_{p,p} \leq \frac{\mathsf{poly}(k_\ell)}{m_\ell}$. Combining them, we know:

**Fact E.8.** *At initialization, it satisfies*

$$\forall p \in [m_\ell] \colon \qquad \left\|R^\star_p - R_p\right\|_2 \leq \frac{b\mathsf{poly}(k_\ell)}{m_\ell} \implies \mathsf{Obj}^{(1)}_{\ell,j}(\mathbf{W}_{\ell,j}) \leq O(1)\cdot m_\ell \cdot \left(\frac{b\mathsf{poly}(k_\ell)}{m_\ell}\right)^{1+1/c} .$$

Next, by the explicit formula of the minimization objective (E.2), we know as long as the objective (E.2) does not exceed any constant multiple of the initialization— namely $\mathsf{Obj}^{(1)}_{\ell,j}(\mathbf{W}_{\ell,j}) \leq O(1)\cdot m_\ell \cdot \left(\frac{b\mathsf{poly}(k_\ell)}{m_\ell}\right)^{1+1/c}$— by choosing $c = 0.001$, it satisfies

$$\forall p \in [m_\ell] \colon \qquad \left\|R^\star_p - R_p\right\|_2 \leq \frac{b\mathsf{poly}(k_\ell)}{m_\ell}\cdot m_\ell^{c/(1+c)} \leq \frac{b\mathsf{poly}(k_\ell)}{m_\ell}\cdot m_\ell^{0.001}$$

and according to Claim E.7, this implies $\|[\mathbf{W}^\star_{\ell,j}]_p - \theta_p[\mathbf{W}_{\ell,j}]_p\|_2 \leq b\mathsf{poly}(k_\ell)\cdot m_\ell^{0.151}$. Using $|1-\theta_p| \leq b\mathsf{poly}(k_\ell)$ from Claim E.6 we immediately conclude that

**Claim E.9.** *As long as the objective $\mathsf{Obj}^{(1)}_{\ell,j}(\mathbf{W}_{\ell,j})$ does not exceed a constant multiple of the initialization, it satisfies $\|[\mathbf{W}^\star_{\ell,j}]_p - [\mathbf{W}_{\ell,j}]_p\|_2 \leq b\mathsf{poly}(k_\ell)\cdot m_\ell^{0.151}$. In other words, Invariant E.3 is maintained.*

To show the convergence of the generator, notice that at every step $t$, suppose we are currently at $\mathbf{W}_{\ell,j} = \mathbf{W}_{\ell,j}^{(t)}$, we can view the generator as optimizing over the convex function:

$$g_t(\mathbf{W}_{\ell,j}) = \left(\frac{1}{(1+c)^{1/c}} - \frac{1}{(1+c)^{1+1/c}}\right) \times$$
$$\sum_{r\in[m_\ell]} \left\| \mathbb{E}\left[\mathsf{ReLU}'([(\mathbf{W}_{\ell,j}^D)^\top \mathbf{W}_{\ell,j}^{(t)} \mathcal{S}_{\ell,j}]_r - b)\mathbf{W}_{\ell,j}\mathcal{S}_{\ell,j}\right] - \mathbb{E}\left[\mathsf{ReLU}'([(\mathbf{W}_{\ell,j}^D)^\top X_{\ell,j}^\star]_r - b) X_{\ell,j}^\star\right]\right\|_2^{1+1/c}$$

Obviously we have

$$\nabla \mathsf{Obj}_{\ell,j}^{(1)}(\mathbf{W}_{\ell,j})\Big|_{\mathbf{W}_{\ell,j}=\mathbf{W}_{\ell,j}^{(t)}} = \nabla g_t(\mathbf{W}_{\ell,j})\Big|_{\mathbf{W}_{\ell,j}=\mathbf{W}_{\ell,j}^{(t)}}$$

Using a similar derivation of Claim E.6, one can show that for $\mathbf{W}^\sharp$ defined as $[\mathbf{W}^\sharp]_q = \frac{1}{\theta_q}[\mathbf{W}_{\ell,j}^\star]_q$, it satisfies $g_t(\mathbf{W}^\sharp) \leq O(m_\ell)\varepsilon_4^{1+1/c}$. Therefore, using standard mirror-descent analysis of non-convex (but one-point convex) functions [3, 5, 7], we have[16].

$$\begin{aligned}
\mathbf{W}_{\ell,j}^{(t+1)} \leftarrow \mathbf{W}_{\ell,j}^{(t)} - \eta \nabla g_t(\mathbf{W}_{\ell,j}^{(t)}) \implies & \\
g_t(\mathbf{W}_{\ell,j}^{(t)}) - g_t(\mathbf{W}^\sharp) &\leq \langle \nabla g_t(\mathbf{W}_{\ell,j}^{(t)}), \mathbf{W}_{\ell,j}^{(t)} - \mathbf{W}^\sharp \rangle \\
&= \frac{\eta}{2}\|\nabla g_t(\mathbf{W}_{\ell,j}^{(t)})\|_F^2 + \frac{1}{2\eta}\left(\|\mathbf{W}_{\ell,j}^{(t)} - \mathbf{W}^\sharp\|_F^2 - \|\mathbf{W}_{\ell,j}^{(t+1)} - \mathbf{W}^\sharp\|_F^2\right)
\end{aligned}$$

Therefore, after telescoping, for sufficiently small $\eta = \frac{1}{\mathsf{poly}(d/\varepsilon)}$, after $T = \mathsf{poly}(d/\varepsilon)$ iterations, we have

$$\frac{1}{T}\sum_{t=0}^{T-1} g_t(\mathbf{W}_{\ell,j}^{(t)}) - g_t(\mathbf{W}^\sharp) \leq \eta \mathsf{poly}(d) + \frac{\|\mathbf{W}_{\ell,j}^{(0)} - \mathbf{W}^\sharp\|_F^2}{\eta T} \leq \frac{1}{\mathsf{poly}(d/\varepsilon)}$$

Thus, for at least $1 - \frac{1}{\mathsf{poly}(d/\varepsilon)}$ fraction of the iterations $t$, it satisfies

$$\mathsf{Obj}_{\ell,j}^{(1)}(\mathbf{W}_{\ell,j}^{(t)}) = g_t(\mathbf{W}_{\ell,j}^{(t)}) \leq g_t(\mathbf{W}^\sharp) + \frac{1}{\mathsf{poly}(d/\varepsilon)} \leq O(m_\ell)\varepsilon_4^{1+1/c}$$

This means $\|R_p - R_p^\star\|_2 \leq m_\ell^{0.001}\varepsilon_4$ so applying Claim E.7 again, we have

$$\left\| \frac{[\mathbf{W}_{\ell,j}^{(t)}]_p}{\|[\mathbf{W}_{\ell,j}^{(t)}]_p\|_2} - [\mathbf{W}_{\ell,j}^\star]_p \right\|_2 \leq m_\ell^{0.151} \cdot \left(\mathbb{b}^2 \mathsf{poly}(k_\ell) + \frac{\mathbb{b}\mathsf{poly}(k_\ell)}{m_\ell^{0.5}}\right)$$

This finishes the proof of Lemma E.2. ■

# F Learning the First Hidden Layer

In this section, we establish claims that, when combined together, imply Lemma F.10.

Throughout this section, we assume

**Induction F.1.** *There exists some $\delta \in (0, m_1^{-0.1}]$ such that for every $j \in [d_1]$, for every $p \in [m_1]$, $\|[\mathbf{W}_{1,j}^\star]_p - [\mathbf{W}_{1,j}]_p\|_2 \leq \delta$. We choose $b = \delta \cdot m_1^{0.02}$ and fix $\mathbf{W}_{1,j}^D = \mathbf{W}_{1,j}$.*

Note that Induction F.1 may either come from the warm-start initialization of $\mathbf{W}_{1,j}$ (see Lemma D.1), or from the learning process of the output layer (see Lemma E.2).

[16] Background: in the Euclidean space, gradient descent is also mirror descent, but the mirror-descent analysis (regarding potential function decrease) can be very different from the gradient-descent analysis (regarding per-step objective decrease). For interested readers, see [6]

**Invariant F.2.** *Throughout the gradient descent-ascent process for $D^{(4)}$ and $D^{(5)}$, we shall maintain the following invariant properties: for every $j \in [d_1], r \in [m_1]$:*

*(a)* $\mathbf{Pr}[[\mathcal{S}_{1,j}]_r > 0] \in \big[\frac{1}{\mathsf{poly}(k_1)m_1}, \frac{\mathsf{poly}(k_1)}{m_1}\big]$;

*(b)* $\frac{1}{\mathsf{poly}(k_1)} < [b_{1,j}]_r < \Theta(\log m_1) \cdot \alpha_{1,j,r} < \mathsf{poly}(k_1)$ *(recall $\alpha_{1,j,r} = \|\mathbf{V}_{1,j}\|_2$)*

*(c)* $s_{r,r'} \stackrel{\text{def}}{=} \mathbb{E}\left[\widetilde{\mathsf{ReLU}}\,(s_r - b)\,\widetilde{\mathsf{ReLU}}\,(s_{r'} - b)\right] \le \frac{\mathsf{poly}(k_1)}{m_1^2}$ *for every $r, r' \in [m_1]$ and $r' \neq r$; where* $s_r \stackrel{\text{def}}{=} [\mathbf{W}_{1,j}^\top X_{1,j}]_r$

*(d) w.h.p.* $\|\mathsf{ReLU}(\mathcal{S}_{1,j})\|_0 \le m^{0.01}$

We shall prove in Section F.1 that these invariant properties are satisfied at initialization, and shall prove in Section F.2 that the first three invariants together imply Invariant F.2d.

During the gradient descent-ascent optimization of $D^{(4)}$, we shall prove (see Claim F.7 in Section F.3) that Invariant F.2a and Invariant F.2b are maintained. Note Invariant F.2c is automatically satisfied thanks to our initialization (see Proposition F.4 in Section F.1)

During the gradient descent-ascent optimization of $D^{(5)}$, we shall show (see Claim F.13 in Section F.4) that Invariant F.2c is maintained. Note Invariant F.2a and Invariant F.2b are automatically satisfied since they are unchanged during the optimization of $D^{(5)}$.

## F.1 Initialization

For each $j \in [d_1]$, $r \in [m_1]$, we initialize $\mathbf{V}_1$ by setting $[\mathbf{V}_{1,j}]_r = \alpha\left(\sqrt{1-\gamma^2} \cdot \mathbf{e}_{j*[m_1]+r} + \gamma \mathbf{e}_1\right)$. Here, $\{\mathbf{e}_p\}_{p \in [2d_1 m_1]}$ is an arbitrary set of basis vectors. We initialize $[b_{1,j}]_r = \beta$.

**Proposition F.3.** *For any $\Lambda = \mathsf{poly}(k_1)$, by choosing $\alpha = \mathsf{poly}(k_1)\Lambda$ and $\beta = \alpha \cdot \sqrt{2\log(m_1 \mathsf{poly}(k_1))}$, it satisfies $\mathbb{E}_{g\sim\mathcal{N}(0,1)}[\mathsf{ReLU}(\alpha g - \beta)] = \frac{\mathsf{poly}(k_1)\Lambda}{m_1}$ and $\mathbb{E}_{g\sim\mathcal{N}(0,1)}[\mathsf{ReLU}'(\alpha g - \beta)] = \frac{\mathsf{poly}(k_1)\Lambda}{m_1}$.*

*As a corollary, before applying gradient descent ascent on $D^{(4)}$ in Algorithm 2, we have*

$$\mathbb{E}\left[\mathsf{ReLU}([\mathbf{V}_{1,j}z]_r - [b_{1,j}]_r)\right] = \frac{\mathsf{poly}(k_1)}{m_1} \ge \mathsf{poly}(k_1) \cdot \mathbb{E}\left[\mathsf{ReLU}([\mathbf{V}^\star_{1,j}z]_r - [b^\star_{1,j}]_r)\right]$$
$$\mathbb{E}\left[\mathsf{ReLU}'([\mathbf{V}_{1,j}z]_r - [b_{1,j}]_r)\right] = \frac{\mathsf{poly}(k_1)}{m_1} \ge \mathsf{poly}(k_1) \cdot \mathbb{E}\left[\mathsf{ReLU}'([\mathbf{V}^\star_{1,j}z]_r - [b^\star_{1,j}]_r)\right]$$

*This also means Invariant F.2a and Invariant F.2b are satisfied at initialization.*

*Proof of Proposition F.3.* By simple Gaussian tail bound, one can verify that when $t = \sqrt{2\log(m_1\mathsf{poly}(k_1))}$, it satisfies $\mathbf{Pr}_{g\sim\mathcal{N}(0,1)}[g > t] = \frac{\mathsf{poly}(k_1)}{m_1}$. Thus, setting $\alpha = \Lambda\mathsf{poly}(k_1)$ and $\beta = \alpha \cdot \sqrt{2\log(m_1\mathsf{poly}(k_1))}$, using Fact F.19 which gives a relationship between $\mathsf{ReLU}(\cdot)$ and $\mathsf{ReLU}'(\cdot)$, we immediately have $\mathbb{E}_{g\sim\mathcal{N}(0,1)}[\mathsf{ReLU}'(\alpha g - \beta)] = \frac{\mathsf{poly}(k_1)\Lambda}{m_1}$ and $\mathbb{E}_{g\sim\mathcal{N}(0,1)}[\mathsf{ReLU}'(\alpha g - \beta)] = \frac{\mathsf{poly}(k_1)\Lambda}{m_1}$. □

**Proposition F.4.** *For any $\Lambda = \mathsf{poly}(k_1)$, there is a choice $\gamma \in (0,1)$ satisfying, given dependent Gaussian $g_1, g_2 \sim \mathcal{N}(0,1)$ with $\mathbb{E}[\langle g_1, g_2\rangle] = \gamma$, for any $b_1, b_2 > 0$ satisfying $\mathbf{Pr}[g_1 > b_1], \mathbf{Pr}[g_2 > b_2] \in \big[\frac{1}{\mathsf{poly}(k_1)m_1}, \frac{\mathsf{poly}(k_1)}{m_1}\big]$, we have the joint distribution*

$$\mathbf{Pr}[g_1 > b_1, g_2 > b_2] \in \big[\frac{\Lambda}{m_1^2}, \frac{\mathsf{poly}(k_1)}{m_1^2}\big]$$

*As a corollary,*[17] *before applying gradient descent ascent on $D^{(5)}$ in Algorithm 2, we have for*

[17] Note we shall guarantee (see Lemma F.6) that after gradient descent-ascent on $D^{(4)}$, it satisfies $\mathbb{E}\left[\mathsf{ReLU}([\mathbf{V}_{1,j}z]_r - [b_{1,j}]_r)\right] \in \big[\frac{1}{\mathsf{poly}(k_1)m_1}, \frac{\mathsf{poly}(k_1)}{m_1}\big]$ and $\mathbb{E}\left[\mathsf{ReLU}'([\mathbf{V}_{1,j}z]_r - [b_{1,j}]_r)\right] \in \big[\frac{1}{\mathsf{poly}(k_1)m_1}, \frac{\mathsf{poly}(k_1)}{m_1}\big]$.

*every* $(j,r) \neq (j',r') \in [d_1] \times [m_1]$,

$$\mathbb{E}\left[\mathsf{ReLU}'([\mathbf{V}_{1,j}z]_r - [b_{1,j}]_r)\mathsf{ReLU}'([\mathbf{V}_{1,j'}z]_{r'} - [b_{1,j'}]_{r'})\right] = \frac{\mathsf{poly}(k_1)}{m_1^2}$$
$$\geq \mathsf{poly}(k_1) \cdot \mathbb{E}\left[\mathsf{ReLU}'([\mathbf{V}^\star_{1,j}z]_r - [b^\star_{1,j}]_r)\mathsf{ReLU}'([\mathbf{V}^\star_{1,j'}z]_{r'} - [b^\star_{1,j'}]_{r'})\right]$$

*As another corollary, we know Invariant F.2c are satisfied from the initialization and throughout the optimization of* $D^{(4)}$.

*Proof of Proposition F.4.* Let us first decrease $b_1$ to $b_1'$ and decrease $b_2$ to $b_2'$ so that $\mathbf{Pr}[g_1 > b_1'] = \mathbf{Pr}[g_2 > b_2'] = \frac{1}{\mathsf{poly}(k_1)m_1}$ touches the lower bound in the assumption. It is easy to verify that $b_1', b_2' = \sqrt{2\log m_1}(1 - o(1))$ and $|b_1 - b_1'|, |b_2 - b_2'| \leq O(\frac{\log k_1}{\sqrt{\log m_1}})$.

At this point, we know $\mathbf{Pr}[g_1 > b_1, g_2 > b_2]$ is an increasing function in $\gamma > 0$, so let $\gamma$ be the smallest such value so that

$$\mathbf{Pr}[g_1 > b_1', g_2 > b_2'] = \frac{\Lambda}{m_1^2} \enspace .$$

By Proposition F.22, we know $\gamma \leq O(\frac{\log k_1}{\log m_1}) \leq o(1)$.

Let us write $g_2 = \gamma g_1 + \sqrt{1-\gamma^2} g$ for $g \sim \mathcal{N}(0,1)$ being independent of $g_1$. Note

$$g_2 > b_2' + |b_2 - b_2'| \iff g > \frac{b_2' - \gamma g_1 + |b_2 - b_2'|}{\sqrt{1-\gamma^2}}$$

We have w.p. $\geq 1 - \frac{1}{\mathsf{poly}(d)}$ it satisfies $|g_1| \leq O(\sqrt{\log m_1})$. When this happens,

$$\frac{b_2' - \gamma g_1}{\sqrt{1-\gamma^2}} \geq b_2' - O(\frac{\log k_1}{\log m_1}) \cdot \sqrt{\log m_1} \geq \sqrt{2\log m_1}(1 - o(1)) \quad \text{and} \quad \frac{|b_2 - b_2'|}{\sqrt{1-\gamma^2}} \leq O(\frac{\log k_1}{\sqrt{\log m_1}})$$

and thus

$$\mathbf{Pr}\left[g_2 > b_2 \mid |g_1| \leq O(\sqrt{\log m_1})\right] = \mathsf{poly}(k_1)\, \mathbf{Pr}\left[g_2 > b_2' \mid |g_1| \leq O(\sqrt{\log m_1})\right] \enspace .$$

Integrating over $g_1 > b_1$, we have

$$\mathbf{Pr}[g_1 > b_1, g_2 > b_2] \leq \mathsf{poly}(k_1)\, \mathbf{Pr}[g_1 > b_1, g_2 > b_2'] + \frac{1}{\mathsf{poly}(d)}$$

Applying this derivation again finishes the proof that

$$\mathbf{Pr}[g_1 > b_1, g_2 > b_2] \leq \mathsf{poly}(k_1)\, \mathbf{Pr}[g_1 > b_1', g_2 > b_2'] + \frac{1}{\mathsf{poly}(d)} \leq \frac{\mathsf{poly}(k_1)}{m_1^2} \enspace .$$

□

## F.2 Sparsity

The following claim ensures that Invariant F.2d is maintained throughout Algorithm 2. (Note, a slightly modified version of it also ensures Invariant F.2d is satisfied at initialization, i.e., before min-max optimization of $D^{(4)}$. We ignore it for simplicity.)

**Claim F.5** (sparsity)**.** *Under Induction F.1, suppose Invariant F.2c, Invariant F.2b, Invariant F.2a are satisfied, and suppose for some* $j \in [d_1]$, $\|\mathsf{ReLU}(\mathcal{S}_{1,j})\|_0 \leq m^{0.03}$ *Then, w.h.p.*

$$\|\mathsf{ReLU}(\mathcal{S}_{1,j})\|_0 \leq m^{0.01}$$

*In other words, during the optimization process of* $D^{(4)}$ *and* $D^{(5)}$*, Invariant F.2d is maintained.*

*Proof of Claim F.5.* Recall $[\mathcal{S}_{1,j}]_r = \mathsf{LeakyReLU}\big(\alpha_{1,j,r}\frac{[\mathbf{V}_{1,j}]_r}{\|[\mathbf{V}_{1,j}]_r\|_2}z - [b_{1,j}]_r\big)$. Using $\mathbf{Pr}[[\mathcal{S}_{1,j}]_r > 0] \in \big[\frac{1}{\mathsf{poly}(k_1)m_1}, \frac{\mathsf{poly}(k_1)}{m_1}\big]$ and the tail bound of Gaussian, we know $\frac{[b_{1,j}]_r}{\alpha_{1,j,r}} = (1 \pm o(1)) \cdot \sqrt{2\log m_1}$ for every $r \in [m_1]$. Also, Invariant F.2b says

$$\alpha_{1,j,r} \leq \mathsf{poly}(k_1) \ .$$

One can also derive using the tail distribution of Gaussian that

$$\mathbf{Pr}\left[\frac{[\mathbf{V}_{1,j}]_r}{\|[\mathbf{V}_{1,j}]_r\|_2}z - \frac{[b_{1,j}]_r}{\alpha_{1,j,r}} - \frac{0.1}{\sqrt{\log m_1}} > 0\right] = \Theta(1) \cdot \mathbf{Pr}[[\mathcal{S}_{1,j}]_r > 0] \in \big[\frac{1}{\mathsf{poly}(k_1)m_1}, \frac{\mathsf{poly}(k_1)}{m_1}\big] \tag{F.1}$$

Recall $X_{1,j} = \mathbf{W}_{1,j}\mathcal{S}_{1,j}$. Using $\|[\mathbf{W}^\star_{1,j}]_p - [\mathbf{W}_{1,j}]_p\|_2 \leq \delta$, $\|\mathsf{ReLU}(\mathcal{S}_{1,j})\|_0 \leq m_1^{0.03}$ and $\|\mathcal{S}_{1,j}\|_\infty \leq \mathsf{poly}(k_1)$, one can derive

$$s_r = [\mathbf{W}_{1,j}^\top X_{1,j}]_r = [\mathcal{S}_{1,j}]_r \pm \delta m_1^{0.04}$$

This implies

$$s_{r,r'} \geq \Omega(\alpha_{1,j,r}\alpha_{1,j,r'}\log m_1)\times \mathbf{Pr}\left[\frac{[\mathbf{V}_{1,j}]_r}{\|[\mathbf{V}_{1,j}]_r\|_2}z - \frac{[b_{1,j}]_r}{\alpha_{1,j,r}} - \frac{0.1}{\sqrt{\log m_1}} > 0 \bigwedge \frac{[\mathbf{V}_{1,j}]_{r'}}{\|[\mathbf{V}_{1,j}]_{r'}\|_2}z - \frac{[b_{1,j}]_{r'}}{\alpha_{1,j,r'}} - \frac{0.1}{\sqrt{\log m_1}} > 0\right]$$

Using the assumption on $s_{r,r'}$ and $\alpha_{1,j,r} \leq \mathsf{poly}(k_1)$, we immediately have

$$\mathbf{Pr}\left[\frac{[\mathbf{V}_{1,j}]_r}{\|[\mathbf{V}_{1,j}]_r\|_2}z - \frac{[b_{1,j}]_r}{\alpha_{1,j,r}} - \frac{0.1}{\sqrt{\log m_1}} > 0 \bigwedge \frac{[\mathbf{V}_{1,j}]_{r'}}{\|[\mathbf{V}_{1,j}]_{r'}\|_2}z - \frac{[b_{1,j}]_{r'}}{\alpha_{1,j,r'}} - \frac{0.1}{\sqrt{\log m_1}} > 0\right] \leq \frac{\mathsf{poly}(k_1)}{m_1^2}$$

Combining this with (F.1), and invoking Proposition F.22, we have

$$\langle v_r, v_{r'}\rangle \leq O\left(\frac{\log k_1}{\log m_1}\right) = o(1) \qquad \text{for } v_r := \frac{[\mathbf{V}_{1,j}]_r}{\|[\mathbf{V}_{1,j}]_r\|_2}.$$

To prove the final sparsity upper bound, let us denote by $g_r = \frac{[\mathbf{V}_{1,j}]_r}{\|[\mathbf{V}_{1,j}]_r\|_2}z$ and $b_r = \frac{[b_{1,j}]_r}{\alpha_{1,j,r}}$. In this notation, $[\mathcal{S}_{1,j}]_r > 0$ if and only if $g_r > b_r$.

Divide $[m_1]$ into into $m_1/q$ chunks of size $q$ each, for $q$ to be chosen later. Within each chunk $\Lambda \subseteq [m_1]$, let us calculate the probability that for at least $n = \omega(1)$ coordinates $r \in \Lambda$, it satisfies $g_r > b_r$, where the choice of $n$ comes from Proposition F.23. This probability is at most

$$\binom{q}{c} \cdot O\left(\exp\left\{-\frac{(1-o(1))\cdot(2\log m_1)\cdot c}{2}\right\}\right) \leq \big(\frac{q}{c}\big)^c \cdot \frac{1}{m_1^{c(1-o(1))}}$$

If we choose $q = c \cdot m_1^{0.99}$, then the above probability is at most $\frac{1}{m_1^{\omega(1)}}$. By union bound, with high probability, within each chunk there are at most $c$ coordinates $r$ satisfying $g_r > b_r$. Since there are at most $m_1/q = m_1^{0.01} \cdot \frac{1}{c}$ chunks, this means in total there are at most $m_1^{0.01}$ coordinates $r \in [m_1]$ with $g_r > b_r$. This finishes the proof. □

### F.3 Objective $D^{(4)}$

Our goal of this subsection is to show the following:

**Lemma F.6** ($D^{(4)}$ final). *Under Induction F.1, after the min-max training of $D^{(4)}$ in Algorithm 2, for every $j \in [d_1], r \in [m_1]$ the solution $\alpha_{1,j,r}$ and $b_{1,j}$ satisfies:*

$$\begin{aligned}
|\alpha_{1,j,r} - \|[\mathbf{V}^\star_{1,j}]_r\|_2| &\leq b\mathsf{poly}(k_1) \\
|[b_{1,j}]_r - [b^\star_{1,j}]_r| &\leq b\mathsf{poly}(k_1) \\
\mathbb{E}[\mathsf{ReLU}([\mathcal{S}_{1,j}]_r)] &= \mathbb{E}[\mathsf{ReLU}([\mathcal{S}^\star_{1,j}]_r)] \cdot (1 \pm b\mathsf{poly}(k_1)) \\
\mathbb{E}[\mathsf{ReLU}'([\mathcal{S}_{1,j}]_r)] &= \mathbb{E}[\mathsf{ReLU}'([\mathcal{S}^\star_{1,j}]_r)] \cdot (1 \pm b\mathsf{poly}(k_1))
\end{aligned}$$

Recall from Section 3.2 that for given $j \in [d_1]$ and $r \in [m_1]$, we have the following min-max objective

$$\min_{\alpha_{1,j,r},[b_{1,j}]_r} \max_{v^D_{j,r},w^D_{j,r}} \left(\mathbb{E}[D^{(4)}_{j,r}(X^\star_1)] - \mathbb{E}[D^{(4)}_{j,r}(X_1)]\right) - |v^D_{j,r}|^2 - |w^D_{j,r}|^2$$

where

$$D^{(4)}_{j,r}(Y) \stackrel{\text{def}}{=} \widetilde{\mathsf{ReLU}}\left([\mathbf{W}_{1,j}^\top Y_j]_r - 2b\right) o_k(v^D_{j,r}) + \widetilde{\mathsf{ReLU}}'\left([\mathbf{W}_{1,j}^\top Y_j]_r - b\right) o_k(w^D_{j,r})$$

for $o_k(z) := \mathsf{poly}(k)\mathsf{ReLU}(z) - \mathsf{ReLU}(-z)$.

On the discriminator side, $D^{(4)}_{j,r}(Y)$ is piecewise linear (recall the trainable parameters for the discriminator are $v^D_{j,r}$ and $w^D_{j,r}$) so the objective on the discriminator side is strongly concave. Hence, the (approximate) maximizer on the discriminator side can be found efficiently.

In this section, we can assume for simplicity that on the the discriminator side, the (unique) exact maximizer is obtained.[18] At this global maximizer of the discriminator, the generator's objective is given by:

$$\begin{aligned}
\mathsf{Obj}^{(4)}_{j,r}(\alpha_{1,j,r},[b_{1,j}]_r) \stackrel{\text{def}}{=} &\left[\mathbb{E}\left[\widetilde{\mathsf{ReLU}}\left([\mathbf{W}_{1,j}^\top X_{1,j}]_r - 2b\right)\right] - \mathbb{E}\left[\widetilde{\mathsf{ReLU}}\left([\mathbf{W}_{1,j}^\top X^\star_{1,j}]_r - 2b\right)\right]\right]_+^2 \\
&+ \left[\mathbb{E}\left[\widetilde{\mathsf{ReLU}}\left([\mathbf{W}_{1,j}^\top X_{1,j}]_r - 2b\right)\right] - \mathbb{E}\left[\widetilde{\mathsf{ReLU}}\left([\mathbf{W}_{1,j}^\top X^\star_{1,j}]_r - 2b\right)\right]\right]_-^2 \times \mathsf{poly}(k_1) \\
&+ \left[\mathbb{E}\left[\widetilde{\mathsf{ReLU}}'\left([\mathbf{W}_{1,j}^\top X_{1,j}]_r - b\right)\right] - \mathbb{E}\left[\widetilde{\mathsf{ReLU}}'\left([\mathbf{W}_{1,j}^\top X^\star_{1,j}]_r - b\right)\right]\right]_+^2 \\
&+ \left[\mathbb{E}\left[\widetilde{\mathsf{ReLU}}'\left([\mathbf{W}_{1,j}^\top X_{1,j}]_r - b\right)\right] - \mathbb{E}\left[\widetilde{\mathsf{ReLU}}'\left([\mathbf{W}_{1,j}^\top X^\star_{1,j}]_r - b\right)\right]\right]_-^2 \times \mathsf{poly}(k_1)
\end{aligned} \tag{F.2}$$

Since we have freezed $\mathbf{W}^D_{1,j} = \mathbf{W}_{1,j} \in \mathbb{R}^{d\times m_1}$ before applying gradient descent ascent (see the beginning of Algorithm 2), in this subsection, we simplify write $\mathbf{W}^D_{1,j}$ as $\mathbf{W}_{1,j}$.

### F.3.1 Invariants from Self-Regularization

We first show that Invariant F.2b and Invariant F.2a are maintained during the minimization process of $\mathsf{Obj}^{(4)}_{j,r}$ (thus of $D^{(4)}$), as long as the objective does not increase by much from the initialization. It says the trainable parameters $\alpha_{1,j,r}, [b_{1,j}]_r$ are within a moderate range, so can be viewed as a self-regularization behavior on $D^{(4)}$.

[18] This is because, for any function $f(x,y)$ that (1) is strongly concave in $y$ and (2) $\frac{\partial f(x,y)}{\partial x}$ is $\mathsf{poly}(d)$-Lipschitz continous in $y$, suppose $y^* = \arg\max_y\{f(x_1,y)\}$ is the exact maximizer and $y_1$ is a sufficiently accurate maximizer $f(x_1,y_1) \geq f(x_1,y^*) - \varepsilon$, then one can derive $\|y_1 - y^*\|^2 \leq \varepsilon \cdot \mathsf{poly}(d)$ using strong concavity, which then implies $\left\|\frac{\partial f(x,y_1)}{\partial x}\Big|_{x=x_1} - \frac{\partial f(x,y^*)}{\partial x}\Big|_{x=x_1}\right\| \leq \sqrt{\varepsilon}\cdot\mathsf{poly}(d)$. In sum, as long as one sufficiently maximizes on the $y$ side, there is no difference in terms of the gradient descent process on the minimization (i.e. $x$) side.

**Claim F.7** ($D^{(4)}$ invariants). *Under Induction F.1, suppose Invariant F.2d is satisfied. Then, during the the training process of* $\mathsf{Obj}^{(4)}_{j,r}$ *(thus of* $D^{(4)}$*), it always satisfies*

- $\frac{1}{\mathsf{poly}(k_1)} < [b_{1,j}]_r < \Theta(\log m_1) \cdot \alpha_{1,j,r} < \mathsf{poly}(k_1)$ *and*
- $\mathbf{Pr}[[\mathcal{S}_{1,j}]_r > 0] \in \left[\frac{1}{\mathsf{poly}(k_1) m_1}, \frac{\mathsf{poly}(k_1)}{m_1}\right]$ .

*In other words, Invariant F.2b and Invariant F.2a are maintained.*

*Proof of Claim F.7.* Using the assumption $\|[\mathbf{W}_{1,j}]_p - [\mathbf{W}^\star_{1,j}]_p\|_2 \leq \delta$, together with the sparsity from Assumption 2.2, we have $[(\mathbf{W}_{1,j})^\top X^\star_{1,j}]_r = [\mathcal{S}^\star_{1,j}]_r \pm 0.01b$. Therefore, applying Assumption 2.2 again we have

$$\mathbf{Pr}\left[[(\mathbf{W}_{1,j})^\top X^\star_{1,j}]_r - b > 0\right] \in \left[\frac{1}{\mathsf{poly}(k_1) m_1}, \frac{\mathsf{poly}(k_1)}{m_1}\right] .$$

Conditioning on this happening, $[\mathcal{S}^\star_{1,j}]_r > \frac{1}{\mathsf{poly}(k_1)}$ with probability at least $\frac{1}{\mathsf{poly}(k_1)}$. Together, we have (note this also holds when $b$ is replaced with $2b$)

$$\frac{1}{\mathsf{poly}(k_1) m_1} \leq \mathbb{E}\left[\widetilde{\mathsf{ReLU}}\left([(\mathbf{W}_{1,j})^\top X^\star_{1,j}]_r - 2b\right)\right] \leq \frac{\mathsf{poly}(k_1)}{m_1}$$
$$\frac{1}{\mathsf{poly}(k_1) m_1} \leq \mathbb{E}\left[\widetilde{\mathsf{ReLU}}'\left([(\mathbf{W}_{1,j})^\top X^\star_{1,j}]_r - b\right)\right] \leq \frac{\mathsf{poly}(k_1)}{m_1}$$

During the training process, since the objective does not increase by more than a constant factor (when learning rate is sufficiently small), we claim that[19]

$$\frac{1}{\mathsf{poly}(k_1) m_1} \overset{①}{\leq} \mathbb{E}\left[\widetilde{\mathsf{ReLU}}\left([(\mathbf{W}_{1,j})^\top X_{1,j}]_r - 2b\right)\right] \overset{②}{\leq} \frac{\mathsf{poly}(k_1)}{m_1}$$
$$\frac{1}{\mathsf{poly}(k_1) m_1} \overset{③}{\leq} \mathbb{E}\left[\widetilde{\mathsf{ReLU}}'\left([(\mathbf{W}_{1,j})^\top X_{1,j}]_r - b\right)\right] \overset{④}{\leq} \frac{\mathsf{poly}(k_1)}{m_1}$$

Using $\|[\mathbf{W}^\star_{1,j}]_p - [\mathbf{W}_{1,j}]_p\|_2 \leq \delta$, and using Invariant F.2d that (denote by "$-r$" the sub-vector except the $r$-th coordinate) $\|\mathsf{ReLU}([\mathcal{S}_{1,j}]_{-r})\|_0 \leq m_1^{0.01}$ and $\|[\mathcal{S}_{1,j}]_{-r}\|_\infty \leq \mathsf{poly}(k_1)$, one can derive

$$[(\mathbf{W}_{1,j})^\top X_{1,j}]_r = (1 \pm \delta)[\mathcal{S}_{1,j}]_r \pm \delta m_1^{0.01} \mathsf{poly}(k_1)$$

By $|\widetilde{\mathsf{ReLU}}(x) - \mathsf{ReLU}(x)| \leq O(\zeta)$, $\widetilde{\mathsf{ReLU}}'(x) \geq 0$ and $\widetilde{\mathsf{ReLU}}'(x) \leq O(1)\mathsf{ReLU}'(x)$, and our choice $b \geq \delta m^{0.02}$ and $\zeta \ll \frac{1}{\mathsf{poly}(m_1)}$, we have

$$② \implies \mathbb{E}\left[\mathsf{ReLU}\left((1-\delta)[\mathcal{S}_{1,j}]_r - 2.1b\right)\right] \overset{②'}{\leq} \frac{\mathsf{poly}(k_1)}{m_1}$$
$$④ \implies \mathbb{E}\left[\mathsf{ReLU}'\left((1-\delta)[\mathcal{S}_{1,j}]_r - 1.1b\right)\right] \overset{④'}{\leq} \frac{\mathsf{poly}(k_1)}{m_1}$$
$$① \implies \frac{1}{\mathsf{poly}(k_1) m_1} \overset{①'}{\leq} \mathbb{E}\left[\mathsf{ReLU}\left((1+\delta)[\mathcal{S}_{1,j}]_r - 1.9b\right)\right]$$
$$③ \implies \frac{1}{\mathsf{poly}(k_1) m_1} \overset{③'}{\leq} \mathbb{E}\left[\mathsf{ReLU}'\left((1+\delta)[\mathcal{S}_{1,j}]_r - 0.9b\right)\right]$$

---

[19] This is because at initialization, we have according to Proposition F.3 we can ensure

$$\mathbb{E}\left[\widetilde{\mathsf{ReLU}}\left([(\mathbf{W}_{1,j})^\top X^\star_{1,j}]_r - 2b\right)\right] \leq \mathbb{E}\left[\widetilde{\mathsf{ReLU}}\left([(\mathbf{W}_{1,j})^\top X_{1,j}]_r - b\right)\right] \leq \frac{\mathsf{poly}(k_1)}{m_1}$$
$$\mathbb{E}\left[\widetilde{\mathsf{ReLU}}'\left([(\mathbf{W}_{1,j})^\top X^\star_{1,j}]_r - 2b\right)\right] \leq \mathbb{E}\left[\widetilde{\mathsf{ReLU}}'\left([(\mathbf{W}_{1,j})^\top X_{1,j}]_r - b\right)\right] \leq \frac{\mathsf{poly}(k_1)}{m_1}$$

Recall

$$[\mathcal{S}_{1,j}]_r = \mathsf{LeakyReLU}(\alpha_{1,j,r} g - [b_{1,j}]_r)$$

for a standard Gaussian $g \sim \mathcal{N}(0,1)$. We must have $[b_{1,j}]_r > -3b$ because otherwise with constant probability $[\mathcal{S}_{1,j}]_r > 2.5b$ and this violates ④′. Since w.h.p. we have $\alpha_{1,j,r} g - [b_{1,j}]_r > -\alpha_{1,j,r} \cdot \mathsf{poly}(k_1) - 3b$, we have w.h.p.

$$\begin{aligned} \mathsf{ReLU}(\alpha_{1,j,r} g - [b_{1,j}]_r) - O(\mathsf{poly}(k_1)\alpha_{1,j,r} + b + 1) \cdot \zeta &\le \mathsf{LeakyReLU}(\alpha_{1,j,r} g - [b_{1,j}]_r) \\ &\le \mathsf{ReLU}(\alpha_{1,j,r} g - [b_{1,j}]_r) + O(\zeta) \quad \text{(F.3)} \end{aligned}$$

Using (F.3) we have

$$\text{②}' \implies \mathbb{E}\left[\mathsf{ReLU}\left(\alpha_{1,j,r} g - [b_{1,j}]_r - 2.2b\right)\right] \overset{\text{②}''}{\le} \frac{\mathsf{poly}(k_1)}{m_1} + O(\mathsf{poly}(k_1)\alpha_{1,j,r}) \cdot \zeta$$

$$\text{④}' \implies \mathbb{E}\left[\mathsf{ReLU}'\left(\alpha_{1,j,r}(g - \mathsf{poly}(k_1)\zeta) - [b_{1,j}]_r - 1.2b\right)\right] \overset{\text{④}''}{\le} \frac{\mathsf{poly}(k_1)}{m_1} \implies \alpha_{1,j,r} \le [b_{1,j}]_r + 1.2b$$

$$\text{①}' \implies \frac{1}{\mathsf{poly}(k_1) m_1} \overset{\text{①}''}{\le} \mathbb{E}\left[\mathsf{ReLU}\left(\alpha_{1,j,r} g - [b_{1,j}]_r - 1.8b\right)\right]$$

$$\text{③}' \implies \frac{1}{\mathsf{poly}(k_1) m_1} \overset{\text{③}''}{\le} \mathbb{E}\left[\mathsf{ReLU}'\left(\alpha_{1,j,r} g - [b_{1,j}]_r - 0.8b\right)\right] \implies \frac{[b_{1,j}]_r}{\alpha_{1,j,r}} \le O(\log m_1)$$

We make two important claims.

- First, it must satisfy $\alpha_{1,j,r} < \mathsf{poly}(k_1)$.

  This is because if not, then $\frac{[b_{1,j}]_r}{\alpha_{1,j,r}} \le O(\log m_1)$ and $\frac{b}{\alpha_{1,j,r}} < \frac{1}{\mathsf{poly}(k_1)} \ll \frac{1}{\log m_1}$ imply we also have

  $$\text{③}'' \implies \mathbb{E}\left[\mathsf{ReLU}'\left(\alpha_{1,j,r} g - [b_{1,j}]_r - 2.2b\right)\right] \ge \frac{1}{\mathsf{poly}(k_1) m_1}$$

  At the same time, Fact F.19 gives us the relationship

  $$\mathbb{E}\left[\mathsf{ReLU}\left(\alpha_{1,j,r} g - [b_{1,j}]_r - 2.2b\right)\right] = \Theta(\alpha_{1,j,r}) \cdot \mathbb{E}\left[\mathsf{ReLU}'\left(\alpha_{1,j,r} g - [b_{1,j}]_r - 2.2b\right)\right] \frac{\alpha_{1,j,r}}{[b_{1,j}]_r + 2.2b}$$

  Putting them together, and plugging in ②″ and $\frac{[b_{1,j}]_r}{\alpha_{1,j,r}} \le O(\log m_1)$ shows that $\alpha_{1,j,r} \le \mathsf{poly}(k_1)$.

- Second, it must satisfy $\alpha_{1,j,r} > \frac{1}{\mathsf{poly}(k_1)}$.

  This is because Fact F.19 gives us the relationship

  $$\mathbb{E}\left[\mathsf{ReLU}\left(\alpha_{1,j,r} g - [b_{1,j}]_r - 1.8b\right)\right] \le \Theta(\alpha_{1,j,r}) \cdot \mathbb{E}\left[\mathsf{ReLU}'\left(\alpha_{1,j,r} g - [b_{1,j}]_r - 1.2b\right)\right] \frac{\alpha_{1,j,r}}{[b_{1,j}]_r + 1.2b}$$

  Plugging in ④″ and $\alpha_{1,j,r} \le [b_{1,j}]_r + 1.2b$ gives us $\alpha_{1,j,r} > \frac{1}{\mathsf{poly}(k_1)}$.

Together, we have just proven the relationship

$$\frac{1}{\mathsf{poly}(k_1)} < [b_{1,j}]_r < \Theta(\log m_1) \cdot \alpha_{1,j,r} < \mathsf{poly}(k_1) \enspace .$$

As for the second invariant, its is only a corollary of the above analysis. Recall from the above derivation that

$$\text{④}' \implies \mathbb{E}\left[\mathsf{ReLU}'\left(\alpha_{1,j,r}(g - \mathsf{poly}(k_1)\zeta) - [b_{1,j}]_r - 1.2b\right)\right] \overset{\text{④}''}{\le} \frac{\mathsf{poly}(k_1)}{m_1}$$

$$\text{③}' \implies \frac{1}{\mathsf{poly}(k_1) m_1} \overset{\text{③}''}{\le} \mathbb{E}\left[\mathsf{ReLU}'\left(\alpha_{1,j,r} g - [b_{1,j}]_r - 0.8b\right)\right]$$

Using the derived property that $\alpha_{1,j,r} \in \left[\frac{1}{\mathsf{poly}(k_1)}, \mathsf{poly}(k_1)\right]$, together with $b \ll m^{-0.01}$, we can safely conclude that

$$④'' \implies \mathbb{E}\left[\mathsf{ReLU}'\left(\alpha_{1,j,r} g - [b_{1,j}]_r\right)\right] \leq \frac{\mathsf{poly}(k_1)}{m_1}$$

$$③'' \implies \frac{1}{\mathsf{poly}(k_1) m_1} \leq \mathbb{E}\left[\mathsf{ReLU}'\left(\alpha_{1,j,r} g - [b_{1,j}]_r\right)\right]$$

which exactly means $\mathbf{Pr}[[\mathcal{S}_{1,j}]_r > 0] \in \left[\frac{1}{\mathsf{poly}(k_1) m_1}, \frac{\mathsf{poly}(k_1)}{m_1}\right]$ . □

### F.3.2 Existence of Descent Direction

Now we show the following Claim regarding the optimization of $\alpha_{1,j,r}$ and $b_{1,j}$ in Algorithm 2 using $D^{(4)}$.

To show that this objective can be minimized (approximately) efficiently, have the following claim:

**Claim F.8.** *Under Induction F.1 and suppose Invariant F.2 is satisfied. There exists $\eta_0 = \frac{1}{\mathsf{poly}(d)}$ such that for every $\eta \in [-\eta_0, \eta_0]$, if we update*

$$\alpha'_{1,j,r} \leftarrow \alpha_{1,j,r} + \eta\alpha \quad \text{and} \quad [b_{1,j}]'_r \leftarrow [b_{1,j}]_r + \eta\beta \ ,$$

*and let $X'_{1,j}$ be the image patch after the update, we then have:*

*(a) When $\alpha = \alpha_{1,j,r}$, $\beta = [b_{1,j}]_r$, we have:*

$$\mathbb{E}\left[\widetilde{\mathsf{ReLU}}\left([(\mathbf{W}_{1,j})^\top X'_{1,j}]_r - b\right)\right] = \mathbb{E}\left[\widetilde{\mathsf{ReLU}}\left([(\mathbf{W}_{1,j})^\top X_{1,j}]_r - b\right)\right](1 + \eta \cdot \Theta(1)) \tag{F.4}$$

$$\mathbb{E}\left[\widetilde{\mathsf{ReLU}}'\left([(\mathbf{W}_{1,j})^\top X'_{1,j}]_r - b\right)\right] = \mathbb{E}\left[\widetilde{\mathsf{ReLU}}'\left([(\mathbf{W}_{1,j})^\top X_{1,j}]_r - b\right)\right](1 \pm b\mathsf{poly}(k_1)\eta) \tag{F.5}$$

*(b) When $\alpha = \alpha_{1,j,r}$, $\beta = 0$, we have:*

$$\mathbb{E}\left[\widetilde{\mathsf{ReLU}}\left([(\mathbf{W}_{1,j})^\top X'_{1,j}]_r - b\right)\right] = \mathbb{E}\left[\widetilde{\mathsf{ReLU}}\left([(\mathbf{W}_{1,j})^\top X_{1,j}]_r - b\right)\right](1 + \eta\theta_1) \tag{F.6}$$

$$\mathbb{E}\left[\widetilde{\mathsf{ReLU}}'\left([(\mathbf{W}_{1,j})^\top X'_{1,j}]_r - b\right)\right] = \mathbb{E}\left[\widetilde{\mathsf{ReLU}}'\left([(\mathbf{W}_{1,j})^\top X_{1,j}]_r - b\right)\right](1 + \eta\theta_2) \tag{F.7}$$

*for some $\theta_1, \theta_2 \in [\frac{1}{\mathsf{poly}(k_1)}, \mathsf{poly}(k_1)]$.*

*Proof of Claim F.8.* Without loss of generality we only consider the case for $\eta > 0$. Recall

$$X_{1,j} = \mathbf{W}_{1,j}\mathcal{S}_{1,j} \quad \text{for} \quad [\mathcal{S}_{1,j}]_r = \mathsf{LeakyReLU}\left(\alpha_{1,j,r}\frac{[\mathbf{V}_{1,j}]_r}{\|[\mathbf{V}_{1,j}]_r\|_2} z - [b_{1,j}]_r\right)$$

and from Claim F.7, we have

$$\frac{1}{\mathsf{poly}(k_1)} < [b_{1,j}]_r < \Theta(\log m_1) \cdot \alpha_{1,j,r} < \mathsf{poly}(k_1) \quad \text{and} \quad \mathbf{Pr}[[\mathcal{S}_{1,j}]_r > 0] \in \left[\frac{1}{\mathsf{poly}(k_1) m_1}, \frac{\mathsf{poly}(k_1)}{m_1}\right] \ .$$

1. In the case of $\alpha = \alpha_{1,j,r}$, $\beta = [b_{1,j}]_r$, the update satisfies $[\mathcal{S}'_{1,j}]_r = (1 + \eta)[\mathcal{S}_{1,j}]_r \pm O(\eta\zeta)$ and therefore $X'_{1,j} = X_{1,j} + \eta[\mathbf{W}_{1,j}]_r\left([\mathcal{S}_{1,j}]_r \pm O(\zeta)\right)$. This implies

$$[(\mathbf{W}_{1,j})^\top X'_{1,j}]_r = [(\mathbf{W}_{1,j})^\top X_{1,j}]_r + \eta\left\|[\mathbf{W}_{1,j}]_r\right\|_2^2\left([\mathcal{S}_{1,j}]_r \pm O(\zeta)\right) \tag{F.8}$$

Using again $\|[\mathbf{W}_{1,j}]_p - [\mathbf{W}^\star_{1,j}]_p\|_2 \leq \delta$ which gives approximate orthogonality for $\{\mathbf{W}_{1,j}\}_{p\in[m_1]}$, together with $\|\mathcal{S}_{1,j}\|_0 \leq m^{0.01}$ and $\|\mathcal{S}_{1,j}\|_\infty \leq \mathsf{poly}(k_1)$, we have

$$[(\mathbf{W}_{1,j})^\top X_{1,j}]_r = [\mathcal{S}_{1,j}]_r + \delta_r \quad \text{where} \quad \text{w.h.p.} \quad |\delta_r| \leq \delta \cdot m_1^{0.02} \ll b \tag{F.9}$$

This means

$$[(\mathbf{W}_{1,j})^\top X_{1,j}]_r \geq 1.1b \implies [\mathcal{S}_{1,j}]_r = \Theta\Big([(\mathbf{W}_{1,j})^\top X_{1,j}]_r - b\Big) \quad \text{and}$$
$$[(\mathbf{W}_{1,j})^\top X_{1,j}]_r \leq 0.9b \implies [(\mathbf{W}_{1,j})^\top X'_{1,j}]_r < b$$

Together, they imply

$$\mathbb{E}\Big[\widetilde{\mathsf{ReLU}}\Big([(\mathbf{W}_{1,j})^\top X'_{1,j}]_r - b\Big)\Big] = \mathbb{E}\Big[\widetilde{\mathsf{ReLU}}\Big([(\mathbf{W}_{1,j})^\top X_{1,j}]_r - b\Big)\Big](1+\Theta(\eta))$$
$$\pm O(\eta b)\cdot \mathbf{Pr}\Big[[(\mathbf{W}_{1,j})^\top X_{1,j}]_r \in [0.9b, 1.1b]\Big] \tag{F.10}$$

This last probability error can be derived as follows:

$$\begin{aligned}
&\mathbf{Pr}\Big[[(\mathbf{W}_{1,j})^\top X_{1,j}]_r \in [0.9b, 1.1b]\Big] \\
&= \mathbf{Pr}\Big[\alpha_{1,j,r}\frac{[\mathbf{V}_{1,j}]_r}{\|[\mathbf{V}_{1,j}]_r\|_2}z - [b_{1,j}]_r - b + \delta_r \in \big[-0.1b, 0.1b\big]\Big] \\
&\leq \mathbf{Pr}\Big[\alpha_{1,j,r}\frac{[\mathbf{V}_{1,j}]_r}{\|[\mathbf{V}_{1,j}]_r\|_2}z - [b_{1,j}]_r - b \in \big[-0.2b, 0.2b\big]\Big] \\
&\overset{①}{\leq} b\mathsf{poly}(k_1)\,\mathbf{Pr}\Big[\alpha_{1,j,r}\frac{[\mathbf{V}_{1,j}]_r}{\|[\mathbf{V}_{1,j}]_r\|_2}z - [b_{1,j}]_r > 0\Big] \leq \frac{b\mathsf{poly}(k_1)}{m_1}
\end{aligned}$$

Above, ① uses the property $\mathbf{Pr}\Big[\alpha_{1,j,r}\frac{[\mathbf{V}_{1,j}]_r}{\|[\mathbf{V}_{1,j}]_r\|_2}z - [b_{1,j}]_r > 0\Big] \in \big[\frac{1}{\mathsf{poly}(k_1)m_1}, \frac{\mathsf{poly}(k_1)}{m_1}\big]$ from Claim F.7, as well as $\frac{b}{\alpha_{1,j,r}} \ll \frac{1}{\mathsf{poly}(k_1)} \leq \frac{1}{\mathsf{polylog} m_1}$.

Putting this back to (F.10) finishes the proof of (F.4).

2. In the case of $\alpha = \alpha_{1,j,r}$, $\beta = [b_{1,j}]_r$, let us see the expectation of the derivative.

   Note when $[(\mathbf{W}_{1,j})^\top X_{1,j}]_r > 1.1b$, then we have both $\widetilde{\mathsf{ReLU}}'\big([(\mathbf{W}_{1,j})^\top X_{1,j}]_r - b\big) = 1$ and $\widetilde{\mathsf{ReLU}}'\Big([(\mathbf{W}_{1,j})^\top X'_{1,j}]_r - b\Big) = 1$. Therefore, to calculate the difference of $\widetilde{\mathsf{ReLU}}'$, it suffices to consider $[(\mathbf{W}_{1,j})^\top X_{1,j}]_r \leq 1.1b$; note this also implies $|[\mathcal{S}_{1,j}]_r| \leq 1.2b$ so applying (F.8) we have

$$|[(\mathbf{W}_{1,j})^\top X'_{1,j}]_r - [(\mathbf{W}_{1,j})^\top X_{1,j}]_r| \leq O(\eta b)$$

   Applying Taylor expansion with Cauchy remainder, we immediately have

$$\begin{aligned}
&\Big|\widetilde{\mathsf{ReLU}}'\Big([(\mathbf{W}_{1,j})^\top X'_{1,j}]_r - b\Big) - \widetilde{\mathsf{ReLU}}'\Big([(\mathbf{W}_{1,j})^\top X_{1,j}]_r - b\Big)\Big| \\
&\leq O(\eta b)\widetilde{\mathsf{ReLU}}''\Big([(\mathbf{W}_{1,j})^\top X_{1,j}]_r - b + \theta\cdot O(\eta b)\Big)
\end{aligned}$$

   for some $\theta \in [-1,1]$. Note when this second-order derivative $\widetilde{\mathsf{ReLU}}''(\cdot)$ is non-zero, it must satisfy $[(\mathbf{W}_{1,j})^\top X_{1,j}]_r - b \in \big[-\frac{\zeta}{4}, \frac{5\zeta}{4}\big]$ (and when this happens, $\mathsf{ReLU}''(\cdot) \leq O(\frac{1}{\zeta})$). This gives upper bound

$$\begin{aligned}
&\Big|\mathbb{E}\,\widetilde{\mathsf{ReLU}}'\Big([(\mathbf{W}_{1,j})^\top X'_{1,j}]_r - b\Big) - \mathbb{E}\,\widetilde{\mathsf{ReLU}}'\Big([(\mathbf{W}_{1,j})^\top X_{1,j}]_r - b\Big)\Big| \\
&\leq O(\frac{\eta b}{\zeta})\,\mathbf{Pr}\Big[[(\mathbf{W}_{1,j})^\top X_{1,j}]_r - b \in \big[-\frac{\zeta}{4}, \frac{5\zeta}{4}\big]\Big] \\
&= O(\frac{\eta b}{\zeta})\,\mathbf{Pr}\Big[\alpha_{1,j,r}\frac{[\mathbf{V}_{1,j}]_r}{\|[\mathbf{V}_{1,j}]_r\|_2}z - [b_{1,j}]_r - b + \delta_r \in \big[-\frac{\zeta}{4}, \frac{5\zeta}{4}\big]\Big] \\
&\overset{①}{\leq} O(\frac{\eta b}{\zeta})\,\mathbf{Pr}\Big[\alpha_{1,j,r}\frac{[\mathbf{V}_{1,j}]_r}{\|[\mathbf{V}_{1,j}]_r\|_2}z - [b_{1,j}]_r - b \in \big[-\frac{\zeta}{4}, \frac{5\zeta}{4}\big]\Big]
\end{aligned}$$

$$\overset{②}{\leq} O(\frac{\eta b}{\zeta})\zeta \mathsf{poly}(k_1)\, \mathbf{Pr}\left[\alpha_{1,j,r}\frac{[\mathbf{V}_{1,j}]_r}{\|[\mathbf{V}_{1,j}]_r\|_2}z - [b_{1,j}]_r > 0\right] \leq \frac{\eta b \mathsf{poly}(k_1)}{m_1}$$

Above, ① uses Proposition F.24; while ② uses the property $\mathbf{Pr}\left[\alpha_{1,j,r}\frac{[\mathbf{V}_{1,j}]_r}{\|[\mathbf{V}_{1,j}]_r\|_2}z - [b_{1,j}]_r > 0\right] \in \left[\frac{1}{\mathsf{poly}(k_1)m_1}, \frac{\mathsf{poly}(k_1)}{m_1}\right]$ from Claim F.7, as well as $\frac{b}{\alpha_{1,j,r}} \ll \frac{1}{\mathsf{poly}(k_1)} \leq \frac{1}{\mathsf{polylog}(m_1)}$.

This finishes the proof of (F.5).

3. In the case of $\alpha = \alpha_{1,j,r}$, $\beta = 0$, let us see the expectation of the derivative.

This time, let us be more precise and write

$$[(\mathbf{W}_{1,j})^\top X_{1,j}]_r = (1-\delta_{r,0})[\mathcal{S}_{1,j}]_r + \delta_{-r}$$
$$\text{where} \quad \delta_{r,0} = 1 - \langle [\mathbf{W}_{1,j}]_r, [\mathbf{W}^\star_{1,j}]_r\rangle \quad \text{and} \quad \text{w.h.p.} \quad |\delta_{-r}| \leq \delta \cdot m_1^{0.02} \tag{F.11}$$

In this formula (F.11), we note that $\delta_{-r}$ does *not* depend on $\alpha_{1,j,r}$. We thus have

$$\widetilde{\mathsf{ReLU}}'\Big([(\mathbf{W}_{1,j})^\top X_{1,j}]_r - b\Big) = \widetilde{\mathsf{ReLU}}'\Big((1-\delta_{r,0})\mathsf{LeakyReLU}\big(\alpha_{1,j,r}\frac{[\mathbf{V}_{1,j}]_r}{\|[\mathbf{V}_{1,j}]_r\|_2}z - [b_{1,j}]_r\big) + \delta_{-r} - b\Big)$$

If we denote by $\mathbf{g}_r = \frac{[\mathbf{V}_{1,j}]_r}{\|[\mathbf{V}_{1,j}]_r\|_2}z$ for abbreviation, then using the relationship between $\widetilde{\mathsf{ReLU}}$, $\mathsf{LeakyReLU}$ and the original $\mathsf{ReLU}$, and $\zeta = \frac{1}{\mathsf{poly}(d)}$ together with $|\delta_{r,0}|, |\delta_{-r}| \ll b$, it is not hard to observe that with high probability

$$\diamondsuit \overset{\text{def}}{=} \widetilde{\mathsf{ReLU}}'\Big([(\mathbf{W}_{1,j})^\top X_{1,j}]_r - b\Big) = \widetilde{\mathsf{ReLU}}'\left((1-\delta_{r,0})\alpha_{1,j,r}\mathbf{g}_r - (1-\delta_{r,0})[b_{1,j}]_r + \delta_{-r} - b\right)$$

Similarly, after the update, we have

$$\diamondsuit' \overset{\text{def}}{=} \widetilde{\mathsf{ReLU}}'\Big([(\mathbf{W}_{1,j})^\top X'_{1,j}]_r - b\Big) = \widetilde{\mathsf{ReLU}}'\left((1-\delta_{r,0})(1+\eta)\alpha_{1,j,r}\mathbf{g}_r - (1-\delta_{r,0})[b_{1,j}]_r + \delta_{-r} - b\right)$$

Observe that if $\alpha_{1,j,r}\mathbf{g}_r < 0.9[b_{1,j}]_r$ then both $\diamondsuit = \diamondsuit' = 0$; if $\alpha_{1,j,r}\mathbf{g}_r > 1.1[b_{1,j}]_r$ then both $\diamondsuit = \diamondsuit' = 1$. Therefore, it suffices to consider $\alpha_{1,j,r}\mathbf{g}_r \in \big[0.9[b_{1,j}]_r, 1.1[b_{1,j}]_r\big]$.

Using Taylor expansion with Cauchy remainder, we have

$$\diamondsuit' - \diamondsuit = \eta\alpha_{1,j,r}\mathbf{g}_r \cdot \widetilde{\mathsf{ReLU}}''\left((1-\delta_{r,0})(1+\theta\eta)\alpha_{1,j,r}\mathbf{g}_r - (1-\delta_{r,0})[b_{1,j}]_r + \delta_{-r} - b\right) \tag{F.12}$$

for some $\theta \in [0,1]$. To prove a lower bound to this, let us note whenever

$$(1-\delta_{r,0})\alpha_{1,j,r}\mathbf{g}_r - (1-\delta_{r,0})[b_{1,j}]_r + \delta_{-r} - b \in [\frac{\zeta}{4}, \frac{3\zeta}{4}]$$

it must satisfy

$$\widetilde{\mathsf{ReLU}}''\left((1-\delta_{r,0})(1+\theta\eta)\alpha_{1,j,r}\mathbf{g}_r - (1-\delta_{r,0})[b_{1,j}]_r + \delta_{-r} - b\right) \geq \Omega(\frac{1}{\zeta})$$

Putting this to the Cauchy remainder (F.12), we can derive

$$\begin{aligned}\mathbb{E}[\diamondsuit'] - \mathbb{E}[\diamondsuit] &\geq \Omega(\frac{\eta[b_{1,j}]_r}{\zeta}) \cdot \mathbf{Pr}\left[(1-\delta_{r,0})\alpha_{1,j,r}\mathbf{g}_r - (1-\delta_{r,0})[b_{1,j}]_r + \delta_{-r} - b \in [\frac{\zeta}{4}, \frac{3\zeta}{4}]\right] \\ &\overset{①}{=} \Omega(\frac{\eta[b_{1,j}]_r}{\zeta}) \cdot \mathbf{Pr}\left[\alpha_{1,j,r}\mathbf{g}_r - [b_{1,j}]_r - b \in [\frac{\zeta}{4}, \frac{3\zeta}{4}]\right] \\ &\overset{②}{\geq} \Omega(\frac{\eta[b_{1,j}]_r}{\zeta}) \cdot \frac{\zeta}{\mathsf{poly}(k_1)}\mathbf{Pr}\left[\alpha_{1,j,r}\mathbf{g}_r - [b_{1,j}]_r > 0\right] \geq \frac{\eta}{\mathsf{poly}(k_1)m_1}\end{aligned}$$

Above, ① uses Proposition F.24; ② uses $\mathbf{Pr}\left[\alpha_{1,j,r}\mathbf{g}_r - [b_{1,j}]_r > 0\right] \in [\frac{1}{\mathsf{poly}(k_1)m_1}, \frac{\mathsf{poly}(k_1)}{m_1}]$ which says $\frac{[b_{1,j}]_r}{\alpha_{1,j,r}} = (1 \pm o(1))\sqrt{2\log m_1}$, as well as $\frac{b}{\alpha_{1,j,r}} \ll \frac{1}{\mathsf{poly}(k_1)}$.

For a similar reason, one can note in the Cauchy remainder (F.12), the second-order derivative $\widetilde{\mathsf{ReLU}}''(\cdot) \in [0, O(\frac{1}{\zeta})]$, and it is non-zero only when

$$(1-\delta_{r,0})\alpha_{1,j,r}\mathbf{g}_r - (1-\delta_{r,0})[b_{1,j}]_r + \delta_{-r} - b \in \big[-\frac{\zeta}{4}, \frac{5\zeta}{4}\big]$$

This yields an upper bound

$$\begin{aligned}
\mathbb{E}[\diamondsuit'] - \mathbb{E}[\diamondsuit] &\le O(\frac{\eta[b_{1,j}]_r}{\zeta}) \cdot \mathbf{Pr}\left[\alpha_{1,j,r}\mathbf{g}_r - [b_{1,j}]_r - b \in \big[-\frac{\zeta}{4}, \frac{5\zeta}{4}\big]\right] \\
&\le O(\frac{\eta[b_{1,j}]_r}{\zeta}) \cdot \zeta \mathsf{poly}(k_1)\, \mathbf{Pr}\left[\alpha_{1,j,r}\mathbf{g}_r - [b_{1,j}]_r > 0\right] \le \frac{\eta \mathsf{poly}(k_1)}{m_1}
\end{aligned}$$

4. In the case of $\alpha = \alpha_{1,j,r}$, $\beta = 0$, the calculation for the expectation of $\widetilde{\mathsf{ReLU}}$ is analogous to that of $\widetilde{\mathsf{ReLU}}'$, so we ignore the details here.

□

### F.3.3 Proof of Lemma F.6: Optimization using Gradient Descent

**Claim F.9.** *Under Induction F.1 and suppose Invariant F.2 is satisfied. There exists some $\eta_1 = \frac{1}{\mathsf{poly}(d)}$ satisfying that, as long as $\mathsf{Obj}^{(4)}_{j,r}(\alpha_{1,j,r}, [b_{1,j}]_r) \ge \frac{b^2\mathsf{poly}(k_1)}{m_1^2}$, suppose we apply gradient descent*

$$(\alpha''_{1,j,r}, [b_{1,j}]''_r) \leftarrow (\alpha_{1,j,r}, [b_{1,j}]_r) - \eta \nabla \mathsf{Obj}^{(4)}_{j,r}(\alpha_{1,j,r}, [b_{1,j}]_r)$$

*for any $\eta \in (0, \eta_1)$, then it satisfies*

$$\mathsf{Obj}^{(4)}_{j,r}(\alpha''_{1,j,r}, [b_{1,j}]''_r) \le \mathsf{Obj}^{(4)}_{j,r}(\alpha_{1,j,r}, [b_{1,j}]_r) - \frac{\eta}{\mathsf{poly}(d)} \ .$$

*Proof of Claim F.9.* As long as $\mathsf{Obj}^{(4)}_{j,r}(\alpha_{1,j,r}, [b_{1,j}]_r) \ge \frac{b^2\mathsf{poly}(k_1)}{m_1^2}$, we know from the objective in (F.2) that

$$\text{either} \quad \Bigg| \underbrace{\mathbb{E}\left[\widetilde{\mathsf{ReLU}}\left([(\mathbf{W}_{1,j})^\top X_{1,j}]_r - 2b\right)\right]}_{\diamondsuit} - \underbrace{\mathbb{E}\left[\widetilde{\mathsf{ReLU}}\left([(\mathbf{W}_{1,j})^\top X^\star_{1,j}]_r - 2b\right)\right]}_{\diamondsuit^\star} \Bigg| \ge \frac{b\mathsf{poly}(k_1)}{m_1}$$

$$\text{or} \quad \Bigg| \underbrace{\mathbb{E}\left[\widetilde{\mathsf{ReLU}}'\left([(\mathbf{W}_{1,j})^\top X_{1,j}]_r - b\right)\right]}_{\heartsuit} - \underbrace{\mathbb{E}\left[\widetilde{\mathsf{ReLU}}'\left([(\mathbf{W}_{1,j})^\top X^\star_{1,j}]_r - b\right)\right]}_{\heartsuit^\star} \Bigg| \ge \frac{b\mathsf{poly}(k_1)}{m_1}$$

Recall from the proof of Claim F.7 that $\diamondsuit, \heartsuit, \diamondsuit^\star, \heartsuit^\star \in \big[\frac{1}{\mathsf{poly}(k_1)m_1}, \frac{\mathsf{poly}(k_1)}{m_1}\big]$.

At this point, we can

- Either apply Claim F.8a to increase or decrease $\diamondsuit$ by a factor of $1+\Theta(\eta)$ or $1-\Theta(\eta)$, while keeping $\heartsuit$ roughly unchanged (by a factor of $(1 \pm b\mathsf{poly}(k_1)\eta)$.
- Or apply Claim F.8b first and apply Claim F.8a next to increase or decrease $\heartsuit$ by a factor $1 + (1 \pm b\mathsf{poly}(k_1))\eta$ or $1 - (1 \pm b\mathsf{poly}(k_1))\eta$, while keeping $\diamondsuit$ unchanged.

In other words, as long as $\mathsf{Obj}^{(4)}_{j,r}(\alpha_{1,j,r}, [b_{1,j}]_r) \ge \frac{b^2\mathsf{poly}(k_1)}{m_1^2}$, there always exists a descent direction $(\alpha, \beta)$ with $|\alpha|, |\beta| \le \mathsf{poly}(k_1)$ and $\eta_0 = \frac{1}{\mathsf{poly}(d)}$ such that for every $\eta \in [0, \eta_0]$, if we update

$$\alpha'_{1,j,r} \leftarrow \alpha_{1,j,r} + \eta\alpha \quad \text{and} \quad [b_{1,j}]'_r \leftarrow [b_{1,j}]_r + \eta\beta \ ,$$

it satisfies

$$\mathsf{Obj}^{(4)}_{j,r}\big(\alpha'_{1,j,r}, [b_{1,j}]'_r\big) \leq \mathsf{Obj}^{(4)}_{j,r}\big(\alpha_{1,j,r}, [b_{1,j}]_r\big) - \eta \frac{b}{m_1^2 \mathsf{poly}(k_1)}$$

This implies

$$\|\nabla \mathsf{Obj}^{(4)}_{j,r}\big(\alpha_{1,j,r}, [b_{1,j}]_r\big)\|_2 \geq \frac{b}{m_1^2 \mathsf{poly}(k_1)}$$

Finally, since $\mathsf{Obj}^{(4)}_{j,r}$ is $\mathsf{poly}(d)$-Lipschitz smooth in the parameters, we know that for sufficiently small learning rate $\eta < \eta_1$ (for some $\eta_1 = \frac{1}{\mathsf{poly}(d)}$, applying gradient descent (instead of following the existential direction), we also have[20]

$$\mathsf{Obj}^{(4)}_{j,r}\big(\alpha''_{1,j,r}, [b_{1,j}]''_r\big) \leq \mathsf{Obj}^{(4)}_{j,r}\big(\alpha_{1,j,r}, [b_{1,j}]_r\big) - \frac{\eta}{\mathsf{poly}(d)} \enspace .$$

This proves that gradient descent can continue to make progress and at least until $\mathsf{Obj}^{(4)}_{j,r}\big(\alpha_{1,j,r}, [b_{1,j}]_r\big) \leq \frac{b^2 \mathsf{poly}(k_1)}{m_1^2}$. □

*Proof of Lemma F.6.* Using again the property (see e.g. (F.9))

$$[(\mathbf{W}_{1,j})^\top X_{1,j}]_r = [\mathcal{S}_{1,j}]_r \pm 0.01b = \mathsf{LeakyReLU}\Big(\alpha_{1,j,r} \frac{[\mathbf{V}_{1,j}]_r}{\|[\mathbf{V}_{1,j}]_r\|_2} z - [b_{1,j}]_r\Big) \pm 0.01b$$

together with $\frac{1}{\mathsf{poly}(k_1)} \leq \alpha_{1,j,r}, [b_{1,j}]_r \leq \mathsf{poly}(k_1)$ and $\mathbf{Pr}[[\mathcal{S}_{1,j}]_r > 0] \in \big[\frac{1}{\mathsf{poly}(k_1) m_1}, \frac{\mathsf{poly}(k_1)}{m_1}\big]$ from Claim F.7, by standard analysis of Gaussian we have

$$\mathbb{E}\Big[\widetilde{\mathsf{ReLU}}\Big([(\mathbf{W}_{1,j})^\top X_{1,j}]_r - 2b\Big)\Big] = \mathbb{E}\left[\mathsf{ReLU}\left([\mathcal{S}_{1,j}]_r\right)\right] \cdot (1 \pm b\mathsf{poly}(k_1))$$

$$\mathbb{E}\Big[\widetilde{\mathsf{ReLU}}'\Big([(\mathbf{W}_{1,j})^\top X_{1,j}]_r - b\Big)\Big] = \mathbb{E}\left[\mathsf{ReLU}'\left([\mathcal{S}_{1,j}]_r\right)\right] \cdot (1 \pm b\mathsf{poly}(k_1))$$

Similarly, the sparsity of $\mathcal{S}^\star_{1,j}$ in Assumption 2.2 implies

$$[(\mathbf{W}_{1,j})^\top X^\star_{1,j}]_r = [\mathcal{S}^\star_{1,j}]_r \pm 0.01b = \mathsf{ReLU}\Big(\|[\mathbf{V}^\star_{1,j}]_r\|_2 \frac{[\mathbf{V}^\star_{1,j}]_r}{\|[\mathbf{V}^\star_{1,j}]_r\|_2} z - [b^\star_{1,j}]_r\Big) \pm 0.01b \enspace ,$$

from Fact F.20 we have $\frac{1}{\mathsf{poly}(k_1)} \leq \|[\mathbf{V}^\star_{1,j}]_r\|_2 \leq [b^\star_{1,j}]_r \leq \mathsf{poly}(k_1)$, from Assumption 2.2 we have $\mathbf{Pr}[[\mathcal{S}^\star_{1,j}]_p > 0] \in \big[\frac{1}{\mathsf{poly}(k_1) m_1}, \frac{\mathsf{poly}(k_1)}{m_1}\big]$, so by manipulating properties of Gaussian distribution we can also derive

$$\mathbb{E}\Big[\widetilde{\mathsf{ReLU}}\Big([(\mathbf{W}_{1,j})^\top X^\star_{1,j}]_r - 2b\Big)\Big] = \mathbb{E}\left[\mathsf{ReLU}\left([\mathcal{S}^\star_{1,j}]_r\right)\right] \cdot (1 \pm b\mathsf{poly}(k_1))$$

$$\mathbb{E}\Big[\widetilde{\mathsf{ReLU}}'\Big([(\mathbf{W}_{1,j})^\top X^\star_{1,j}]_r - b\Big)\Big] = \mathbb{E}\left[\mathsf{ReLU}'\left([\mathcal{S}^\star_{1,j}]_r\right)\right] \cdot (1 \pm b\mathsf{poly}(k_1))$$

Now, for every $j$ and $r$, after applying gradient descent-ascent for sufficiently many (i.e., $\mathsf{poly}(d)$) iterations, we must have

$$\mathsf{Obj}^{(4)}_{j,r}\big(\alpha_{1,j,r}, [b_{1,j}]_r\big) \leq \frac{b^2 \mathsf{poly}(k_1)}{m_1^2}$$

[20]This comes from standard analysis of gradient descent, namely, given a (not necessarily convex) function $f(x)$ that is $L$-Lipschitz smooth, then for every $\eta \leq \frac{1}{2L}$, if we update $x \leftarrow x - \eta \nabla f(x)$, we can decrease the objective $f(x)$ by at least $\frac{\eta}{2}\|\nabla f(x)\|_2^2$.

since otherwise gradient descent can continue to decrease the objective (see Claim F.9). This means

$$\left|\mathbb{E}\left[\widetilde{\mathsf{ReLU}}\left([(\mathbf{W}_{1,j})^\top X_{1,j}]_r - 2b\right)\right] - \mathbb{E}\left[\widetilde{\mathsf{ReLU}}\left([(\mathbf{W}_{1,j})^\top X^\star_{1,j}]_r - 2b\right)\right]\right| \le \frac{b\mathsf{poly}(k_1)}{m_1}$$

$$\text{and} \quad \left|\mathbb{E}\left[\widetilde{\mathsf{ReLU}}'\left([(\mathbf{W}_{1,j})^\top X_{1,j}]_r - b\right)\right] - \mathbb{E}\left[\widetilde{\mathsf{ReLU}}'\left([(\mathbf{W}_{1,j})^\top X^\star_{1,j}]_r - b\right)\right]\right| \le \frac{b\mathsf{poly}(k_1)}{m_1}$$

By the previous analysis, we immediately have

$$\mathbb{E}[\mathsf{ReLU}([\mathcal{S}_{1,j}]_r)] = \mathbb{E}[\mathsf{ReLU}([\mathcal{S}^\star_{1,j}]_r)] \cdot (1 \pm b\mathsf{poly}(k_1))$$
$$\mathbb{E}[\mathsf{ReLU}'([\mathcal{S}_{1,j}]_r)] = \mathbb{E}[\mathsf{ReLU}'([\mathcal{S}^\star_{1,j}]_r)] \cdot (1 \pm b\mathsf{poly}(k_1))$$

This immediately implies (by simple properties of Gaussian, see Fact F.21)

$$|\alpha_{1,j,r} - \|[\mathbf{V}^\star_{1,j}]_r\|_2| \le b\mathsf{poly}(k_1)$$
$$|[b_{1,j}]_r - [b^\star_{1,j}]_r| \le b\mathsf{poly}(k_1)$$

□

## F.4 Objective $D^{(5)}$

Our goal of this subsection is to prove the following:

**Lemma F.10** ($D^5$ final)**.** *Under Induction F.1, after the min-max training of $D^{(5)}$ in Algorithm 2, the solution $\mathbf{V}_1$ satisfies*

- *for every $j \in [d_1]$, $r \ne r' \in [m_1]$, $|s^\star_{j,j,r,r'} - s_{j,j,r,r'}| \le \frac{b\mathsf{poly}(k_1)}{m_1^2}$ and this implies*
$$\left|\langle [\mathbf{V}^\star_{1,j}]_r, [\mathbf{V}^\star_{1,j'}]_{r'}\rangle - \langle [\mathbf{V}_{1,j}]_r, [\mathbf{V}_{1,j'}]_{r'}\rangle\right| \le \mathsf{poly}(k_1)b$$
$$\forall \mathbb{b} \in [b, m^{-0.01}]\colon \left|\mathbb{E}\left[\mathsf{ReLU}'(s_r - \mathbb{b})[\mathcal{S}_{1,j}]_{r'}\right] - \mathbb{E}\left[\mathsf{ReLU}'(s^\star_r - \mathbb{b})[\mathcal{S}^\star_{1,j}]_{r'}\right]\right| \le \frac{\mathbb{b}\mathsf{poly}(k_1)}{m_1^2}$$
- *for every $(j,r) \ne (j',r') \in [d_1] \times [m_1]$, $|s^\star_{j,j',r,r'} - s_{j,j',r,r'}| \le b\mathsf{poly}(m_1)$ and this implies*
$$\left|\langle [\mathbf{V}^\star_{1,j}]_r, [\mathbf{V}^\star_{1,j'}]_{r'}\rangle - \langle [\mathbf{V}_{1,j}]_r, [\mathbf{V}_{1,j'}]_{r'}\rangle\right| \le \mathsf{poly}(m_1)b$$
- *w.h.p. $\|\mathsf{ReLU}(\mathcal{S}_{1,j})\|_0 \le m_1^{0.01}$ and $\|\mathcal{S}_{1,j}\|_\infty \le \mathsf{poly}(k_1)$.*

We recall the min-max objective for $D^{(5)}$.

$$\min_{\{\mathbf{V}_{1,j}\}_{j\in[d_1]}} \max_{u^D} \left(\mathbb{E}[D^{(5)}(X^\star_1)] - \mathbb{E}[D^{(5)}(X_1)]\right) + R^{(5)}(\mathbf{V}_{1,j}) - H(u^D)$$

On the discriminator side, we choose the regularizer $H(u)$ to be the (negative) entropy regularizer, that is, $H(u) = \sum_i u_i \log u_i$, and assume that the discriminator is optimized over the simplex: $\{u \ge 0\colon \sum_i u_i = 1\}$.[21]

One can verify that the min-max objective is strongly concave in $u^D$, so the (approximate) maximizer on the discriminator side can be found efficiently. In this subsection, we assume for simplicity that on the the discriminator side, an exact maximizer is already obtained (for reasons similar to Footnote 18).

At this exact maximizer, we can write the generator's objective as

$$\min_{\mathbf{V}_1 \text{ s.t. } \|[\mathbf{V}_{1,\ell}]_r\|_2 = \alpha_{1,\ell,r}} \left\{Obj^{(5)}(\mathbf{V}_1) \stackrel{\text{def}}{=} H(\mathbf{V}_1) + R^{(5)}(\mathbf{V}_1)\right\} \tag{F.13}$$

[21]Alternatively, one can re-parameterize the discriminator parameters by setting $u_i \leftarrow \frac{u_i}{\|u\|_1}$; this can avoid having a bounded region for the discriminator.

where

$$H(\mathbf{V}_1) \stackrel{\text{def}}{=} \log\Bigg( \sum_{(j,r)\neq(j',r')\in[d_1]\times[m_1]} \exp\Big\{C_{j,j',r,r'}\left(s_{j,j',r,r'} - s^\star_{j,j',r,r'}\right)\Big\} + \exp\Big\{C_{j,j',r,r'}\left(s^\star_{j,j',r,r'} - s_{j,j',r,r'}\right)\Big\}\Bigg)$$

for $C_{j,j,r,r'} = \frac{m_1^2}{b\mathsf{poly}(k_1)}$ and $C_{j,j',r,r'} = \frac{1}{\mathsf{poly}(m_1)}$ when $j \neq j'$. In this subsection we denote by

$$s_{j,r} \stackrel{\text{def}}{=} [\mathbf{W}_{1,j}^\top X_{1,j}]_r \qquad s_{j,j',r,r'} \stackrel{\text{def}}{=} \mathbb{E}\left[\widetilde{\mathsf{ReLU}}\left(s_{j,r} - b\right)\widetilde{\mathsf{ReLU}}\left(s_{j',r'} - b\right)\right]$$

$$s^\star_{j,r} \stackrel{\text{def}}{=} [\mathbf{W}_{1,j}^\top X^\star_{1,j}]_r \qquad s^\star_{j,j',r,r'} \stackrel{\text{def}}{=} \mathbb{E}\left[\widetilde{\mathsf{ReLU}}\left(s^\star_{j,r} - b\right)\widetilde{\mathsf{ReLU}}\left(s^\star_{j',r'} - b\right)\right]$$

On the generator side, we choose the regularizer $R^{(5)}(\mathbf{V}_1)$ as

$$R^{(5)}(\mathbf{V}_1) \stackrel{\text{def}}{=} \sum_{(j,r)\neq(j',r')\in[d_1]\times[m_1]} \exp\left\{\mathsf{poly}(k_1)\cdot\left(-(1-\tau) - \left\langle \frac{[\mathbf{V}_{1,j}]_r}{\|[\mathbf{V}_{1,j}]_r\|_2}, \frac{[\mathbf{V}_{1,j'}]_{r'}}{\|[\mathbf{V}_{1,j'}]_{r'}\|_2}\right\rangle\right)\right\} \exp\left\{\mathsf{poly}(k_1)\cdot\left(-(1-\tau) + \left\langle \frac{[\mathbf{V}_{1,j}]_r}{\|[\mathbf{V}_{1,j}]_r\|_2}, \frac{[\mathbf{V}_{1,j'}]_{r'}}{\|[\mathbf{V}_{1,j'}]_{r'}\|_2}\right\rangle\right)\right\} \tag{F.14}$$

**Fact F.11.** *There is an absolute constant $\tau > 0$ so that $\left\langle \frac{[\mathbf{V}^\star_{1,j}]_r}{\|[\mathbf{V}^\star_{1,j}]_r\|_2}, \frac{[\mathbf{V}^\star_{1,j'}]_{r'}}{\|[\mathbf{V}^\star_{1,j'}]_{r'}\|_2}\right\rangle \in [-1+0.5\tau, 1-0.5\tau]$ and $R^{(5)}(\mathbf{V}^\star_1) \leq \exp^{-\mathsf{poly}(k_1)}$.*

*Proof.* Recall Assumption B.1 tells us

$$\frac{1}{\mathsf{poly}(d)} \leq \mathbf{Pr}[[\mathcal{S}^\star_{\ell,j'}]_{r'} > 0, [\mathcal{S}^\star_{\ell,j}]_r > 0] \leq \min\Big\{\mathbf{Pr}[[\mathcal{S}^\star_{\ell,j'}]_{r'} > 0], \mathbf{Pr}[[\mathcal{S}^\star_{\ell,j}]_r > 0]]\Big\}\cdot(1 - \frac{1}{\mathsf{poly}(d)})$$

One can verify that this implies

$$\left\langle \frac{[\mathbf{V}^\star_{1,j}]_r}{\|[\mathbf{V}^\star_{1,j}]_r\|_2}, \frac{[\mathbf{V}^\star_{1,j'}]_{r'}}{\|[\mathbf{V}^\star_{1,j'}]_{r'}\|_2}\right\rangle \in [-1+v, 1-v]$$

for some constant $v < 0.5$ (the larger $\mathsf{poly}(d)$ is in Assumption B.1, the smaller this constant $v$ is). Simply choosing $\tau \stackrel{\text{def}}{=} 2v$ finishes the proof. □

Our regularizer $R^{(5)}$ in (F.14) ensures that $\frac{[\mathbf{V}_{1,j'}]_{r'}}{\|[\mathbf{V}_{1,j'}]_{r'}\|_2}$ does not go too close to $-1$ or 1, so is an exponential barrier function.

### F.4.1 Initialization and Invariant

**Fact F.12** (initialization for $D^{(5)}$)**.** *Before we start to apply gradient descent ascent on $D^{(5)}$, it satisfies*

$$H(\mathbf{V}_1) \leq \frac{\mathsf{poly}(k_1)}{b} \quad \textit{and} \quad R^{(5)}(\mathbf{V}_1) \leq d^{-\mathsf{poly}(k_1)}$$

*Proof.* This is because, after gradient descent ascent on $D^{(4)}$, using $\langle[\mathbf{V}_{1,j}]_r, [\mathbf{V}_{1,j'}]_{r'}\rangle = 0$, we have

$$s_{j,j',r,r'} = \mathbb{E}\left[\widetilde{\mathsf{ReLU}}\left(s_{j,r} - b\right)\right]\mathbb{E}\left[\widetilde{\mathsf{ReLU}}\left(s_{j',r'} - b\right)\right] \leq \frac{\mathsf{poly}(k_1)}{m_1^2}$$

From Assumption 2.2 and Proposition F.25a, we also have

$$s^\star_{j,j,r,r'} \leq \frac{\mathsf{poly}(k_1)}{m_1^2}$$

They together give the desired bound on $H(\mathbf{V}_1)$. $\square$

**Claim F.13** ($D^{(5)}$ invariants)**.** *Under Induction F.1, suppose Invariant F.2 is satisfied and the optimization for $D^{(4)}$ is over (i.e., Lemma F.6 holds).*

*During the optimization of $D^{(5)}$, suppose the objective does not increase by much from the initialization (that is, $H(\mathbf{V}_1) + R^{(5)}(\mathbf{V}_1) \leq \frac{\mathsf{poly}(k_1)}{b}$), then it satisfies for every $(j,r) \neq (j',r') \in [d_1] \times [m_1]$:*

- $\left\langle \frac{[\mathbf{V}_{1,j}]_r}{\|[\mathbf{V}_{1,j}]_r\|_2}, \frac{[\mathbf{V}_{1,j'}]_{r'}}{\|[\mathbf{V}_{1,j'}]_{r'}\|_2} \right\rangle \in [-1+0.5\tau, 1-0.5\tau]$
- $s_{j,j,r,r'} \leq s^\star_{j,j,r,r'} + \frac{\mathsf{poly}(k_1)}{m_1^2} \leq \frac{\mathsf{poly}(k_1)}{m_1^2}$ *(thus Invariant F.2c is satisfied)*

### F.4.2 Existence of Hessian Update

**Definition F.14** (update direction)**.** *During the training process of $D^{(5)}$, at any point $\mathbf{V}_1$, given $\eta \in (0,1)$, let us consider an update direction*

$$\mathbf{V}'_{1,j} \leftarrow \sqrt{1-\eta}\mathbf{V}_{1,j} + \sqrt{\eta}\mathbf{S}_j$$

*satisfying*

- $\mathbf{V}_{1,i}^\top \mathbf{S}_j = \mathbf{0}$ *for every $i,j \in [d_1]$;*
- $\|\mathbf{S}_j\|_F \leq \mathsf{poly}(d)$ *for every $j \in [d_1]$;*
- $[\mathbf{S}_j \mathbf{S}_{j'}]_{r,r'} = \left\langle \frac{[\mathbf{V}^\star_{1,j}]_r}{\|[\mathbf{V}^\star_{1,j}]_r\|_2}, \frac{[\mathbf{V}^\star_{1,j'}]_{r'}}{\|[\mathbf{V}^\star_{1,j'}]_{r'}\|_2} \right\rangle \cdot \alpha_{1,j,r}\alpha_{1,j',r'}$ *for every $j,j' \in [d_1]$, $r,r' \in [m_1]$.*

Note an explicit construction of $\{\mathbf{S}_j\}_{j\in[d_1]}$ can be found in Claim F.29. Recall also the notation $\alpha_{1,j,r} = \|[\mathbf{V}_{1,j}]_r\|_2$ implies

$$\|[\mathbf{V}'_{1,j}]_r\|_2 = \|[\mathbf{V}_{1,j}]_r\|_2 = \alpha_{1,j,r}$$

after the update.

**Claim F.15.** *In the same setting as Claim F.13, during the training process of $D^{(5)}$, there exists $\eta_0 = \frac{1}{\mathsf{poly}(d)}$ such that for every $\eta \in [-\eta_0, \eta_0]$, the update direction $\mathbf{V}'_1$ in Definition F.14 satisfies for every $j = j'$, $r \neq r' \in [m_1]$:*

- *If $s_{j,j,r,r'} < s^\star_{j,j,r,r'} - \frac{b\mathsf{poly}(k_1)}{m_1^2}$, then after the update $s'_{j,j,r,r'} \geq s_{j,j,r,r'} + \frac{\eta}{\mathsf{poly}(d)}$;*
- *If $s_{j,j,r,r'} > s^\star_{j,j,r,r'} + \frac{b\mathsf{poly}(k_1)}{m_1^2}$, then after the update $s'_{j,j,r,r'} \leq s_{j,j,r,r'} - \frac{\eta}{\mathsf{poly}(d)}$;*
- *If $s_{j,j,r,r'} \in s^\star_{j,j,r,r'} \pm \frac{b\mathsf{poly}(k_1)}{m_1^2}$, then after the update $s'_{j,j,r,r'} = s_{j,j,r,r'} \pm \eta\frac{b\mathsf{poly}(k_1)}{m_1^2}$.*

*and for every $j \neq j'$, $r,r' \in [m_1]$:*

- *If $s_{j,j',r,r'} < s^\star_{j,j',r,r'} - b\mathsf{poly}(m_1)$, then after the update $s'_{j,j',r,r'} \geq s_{j,j',r,r'} + \frac{\eta}{\mathsf{poly}(d)}$;*
- *If $s_{j,j',r,r'} > s^\star_{j,j',r,r'} + b\mathsf{poly}(m_1)$, then after the update $s'_{j,j',r,r'} \leq s_{j,j',r,r'} - \frac{\eta}{\mathsf{poly}(d)}$;*
- *If $s_{j,j',r,r'} \in s^\star_{j,j',r,r'} \pm b\mathsf{poly}(m_1)$, then after the update $s'_{j,j',r,r'} = s_{j,j',r,r'} \pm \eta b\mathsf{poly}(m_1)$.*

*Proof of Claim F.15.* In this proof we focus on proving the first three items (for $j = j'$), and towards the end we discuss later in Remark F.16 how to generalize it to the case of $j \neq j'$.

Recall from Lemma F.6 that we already have for every $j \in [d_1], r \in [m_1]$:

$$\|[\mathbf{V}_{1,j}]_r\|_2 = \alpha_{1,j,r} = \|[\mathbf{V}^\star_{1,j}]\|_2(1 \pm b\mathsf{poly}(k_1))$$

Let us define

$$[\mathbf{M}^\star]_{r,r'} \stackrel{\text{def}}{=} \left\langle \frac{[\mathbf{V}^\star_{1,j}]_r}{\|[\mathbf{V}^\star_{1,j}]_r\|_2}, \frac{[\mathbf{V}^\star_{1,j}]_{r'}}{\|[\mathbf{V}^\star_{1,j}]_{r'}\|_2} \right\rangle \cdot \alpha_{1,j,r}\alpha_{1,j,r'}$$

Now, Claim F.29 gives us matrix $\mathbf{S}_j \in \mathbb{R}^{m_1 \times m_0'}$ with $\mathbf{S}_j\mathbf{S}_j^\top = \mathbf{M}^\star$ satisfying

$$\left(\sqrt{1-\eta}\mathbf{V}_{1,j} + \sqrt{\eta}\mathbf{S}_j\right)\left(\sqrt{1-\eta}\mathbf{V}_{1,j} + \sqrt{\eta}\mathbf{S}_j\right)^\top = (1-\eta)\mathbf{V}_{1,j}\mathbf{V}_{1,j}^\top + \eta\mathbf{M}^\star$$

With the above notions, we can consider update

$$\mathbf{V}'_{1,j} = \sqrt{1-\eta}\mathbf{V}_{1,j} + \sqrt{\eta}\mathbf{\Delta}_2 \stackrel{\text{def}}{=} \mathbf{V}_{1,j} + \eta\mathbf{\Delta}_1 + \sqrt{\eta}\mathbf{\Delta}_2 \stackrel{\text{def}}{=} \mathbf{V}_{1,j} + \mathbf{\Delta} \in \mathbb{R}^{m_1 \times m_0'}$$

for $\mathbf{\Delta}_1 = \frac{\sqrt{1-\eta}-1}{\eta}\mathbf{V}_{1,j}$ and $\mathbf{\Delta}_2 = \pm\mathbf{S}_j$ (with a $50\% - 50\%$ random sign). Using the orthogonality $\mathbf{V}_{1,j}\mathbf{S}_j^\top = \mathbf{0}_{m_1 \times m_1}$ from Claim F.29, we know that $\mathbf{\Delta}_2 z \in \mathbb{R}^{m_1}$ is independent of $[\mathbf{V}_{1,j}]z \in \mathbb{R}^{m_1}$.

Note although we set $\mathbf{V}'_{1,j}$ as a random matrix depending on the random sign in $\mathbf{\Delta}_2$, in fact, the quantity $s'_{j,j,r,r'}$ after the update is *independent* of this random sign. (Indeed, using $\mathbf{V}_{1,j}\mathbf{S}_j^\top = \mathbf{0}_{m_1 \times m_1}$, we know the distribution of $\mathbf{V}'_{1,j}z = \sqrt{1-\eta}\mathbf{V}_{1,j}z + \sqrt{\eta}\mathbf{S}_j z$ is completely unchanged after replacing $\mathbf{S}_j$ with $-\mathbf{S}_j$.) However, we shall use this randomness in $\mathbf{V}'_{1,j}$ to simplify proofs.

Let us denote$g_r = [\mathbf{V}_{1,j}z]_r$ and $\beta_r = [b_{1,j}]_r$, and we have using the sparsity Invariant F.2d

$$s_r = [(\mathbf{W}_{1,j})^\top X_{1,j}]_r = \mathsf{LeakyReLU}(g_r - \beta_r) + \delta_r \quad \text{where} \quad \text{w.h.p.} \quad |\delta_r| \leq \delta \cdot m_1^{0.02} \ll b$$
$$\widetilde{\mathsf{ReLU}}\left(s_r - b\right) = \widetilde{\mathsf{ReLU}}\left(g_r - \beta_r + \delta_r - b\right)$$

Let us also denote by

$$P_{r,r'} \stackrel{\text{def}}{=} \mathbb{E}\left[\widetilde{\mathsf{ReLU}}\left(s_r - b\right)\widetilde{\mathsf{ReLU}}\left(s_{r'} - b\right)\right] m_1^2$$

Using w.h.p. $s_r, s_{r'} \leq \mathsf{poly}(k_1)$, we have

$$\mathbb{E}\left[\widetilde{\mathsf{ReLU}}\left(s_r - b\right)\widetilde{\mathsf{ReLU}}'\left(s_{r'} - b\right)\right] \ , \ \mathbb{E}\left[\widetilde{\mathsf{ReLU}}'\left(s_r - b\right)\widetilde{\mathsf{ReLU}}'\left(s_{r'} - b\right)\right] \leq \frac{P_{r,r'}\mathsf{poly}(k_1)}{m_1^2} \tag{F.15}$$

Applying Proposition F.25c, we also have

$$\mathbb{E}\left[\widetilde{\mathsf{ReLU}}\left(s_r - b\right)\widetilde{\mathsf{ReLU}}''\left(s_{r'} - b\right)\right] \leq O(\frac{1}{\zeta})\,\mathbb{E}\left[\widetilde{\mathsf{ReLU}}\left(s_r - b\right)\mathbb{1}_{|s_{r'} - b| \leq \zeta}\right] \leq \frac{P_{r,r'}\mathsf{poly}(k_1)}{m_1^2} \tag{F.16}$$

Recall

$$X_{1,j} = \mathbf{W}_{1,j}\mathcal{S}_{1,j} \quad \text{for} \quad \mathcal{S}_{1,j} = \mathsf{LeakyReLU}(\mathbf{V}_{1,j}z - b_{1,j})$$

Now, denote $\mathcal{S} := \mathcal{S}_{1,j}$ and $\mathcal{S}'$ be its value after we update $\mathbf{V}_{1,j}$ to $\mathbf{V}'_{1,j}$ (without changing $z \in \mathbb{R}^{m_0'}$). We know for every $p \in [m_1]$, for sufficiently small $\eta > 0$, by Taylor expansion (recalling w.h.p. $|[\mathbf{\Delta}z]_p| \leq \sqrt{\eta}\mathsf{poly}(k_1) + O(\eta) \leq \sqrt{\eta}\mathsf{poly}(k_1)$ which uses Claim F.7):

$$[\mathcal{S}']_p = [\mathcal{S}]_p + \theta_p \stackrel{\text{def}}{=} [\mathcal{S}]_p + \mathsf{LeakyReLU}'([\mathbf{V}_{1,j}z - b_{1,j}]_p)[\mathbf{\Delta}z]_p \pm \frac{\eta\mathsf{poly}(k_1)}{\zeta}\mathbb{1}_{|[\mathbf{V}_{1,j}z - b_{1,j}]_p| \leq 2\zeta}$$

We denote by $s'_r, s'_{r'}$ the values $s_r, s_{r'}$ after the update from $\mathbf{V}_{1,j}$ to $\mathbf{V}'_{1,j}$. We have

$$s'_r = s_r + \xi_r \stackrel{\text{def}}{=} s_r + \textstyle\sum_{p \in [m_1]}\langle[\mathbf{W}_{1,j}]_r, [\mathbf{W}_{1,j}]_p\rangle\theta_p$$
$$\widetilde{\mathsf{ReLU}}\left(s'_r - b\right) = \widetilde{\mathsf{ReLU}}\left(s_r - b\right) + \widetilde{\mathsf{ReLU}}'\left(s_r - b\right)\xi_r + \tfrac{1}{2}\widetilde{\mathsf{ReLU}}''(s_r - b)\xi_r^2 + O(\eta^{1.5})$$

Hence we can calculate that:

$$\mathbb{E}\left[\widetilde{\mathsf{ReLU}}\left(s'_r - b\right)\widetilde{\mathsf{ReLU}}\left(s'_{r'} - b\right)\right] = \mathbb{E}\left[\widetilde{\mathsf{ReLU}}\left(s_r - b\right)\widetilde{\mathsf{ReLU}}\left(s_{r'} - b\right)\right] + \text{additional terms}$$

where the additional terms are given as:

1. $\mathbb{E}_z\left[\widetilde{\mathsf{ReLU}}\left(s_r-b\right)\widetilde{\mathsf{ReLU}}'\left(s_{r'}-b\right)\xi_{r'}\right]$, this term can be calculated as:

$$
\begin{aligned}
&\mathop{\mathbb{E}}_z\left[\widetilde{\mathsf{ReLU}}\left(s_r-b\right)\widetilde{\mathsf{ReLU}}'\left(s_{r'}-b\right)\xi_{r'}\right]\\
&=\eta\mathop{\mathbb{E}}_z\left[\widetilde{\mathsf{ReLU}}\left(s_r-b\right)\widetilde{\mathsf{ReLU}}'\left(s_{r'}-b\right)\left(\sum_{p\in[m_1]}\langle[\mathbf{W}_{1,j}]_{r'},[\mathbf{W}_{1,j}]_p\rangle\mathsf{LeakyReLU}'([\mathbf{V}_{1,j}z-b_{1,j}]_p)[\mathbf{\Delta}_1 z]_p\right)\right]\\
&\quad\pm\frac{\eta\mathsf{poly}(k_1)}{\zeta}\mathop{\mathbb{E}}_z\left[\widetilde{\mathsf{ReLU}}\left(s_r-b\right)\widetilde{\mathsf{ReLU}}'\left(s_{r'}-b\right)\left(\sum_{p\in[m_1]}\langle[\mathbf{W}_{1,j}]_{r'},[\mathbf{W}_{1,j}]_p\rangle\mathbb{1}_{|[\mathbf{V}_{1,j}z-b_{1,j}]_p|\le 2\zeta}\right)\right]\\
&=\eta\mathop{\mathbb{E}}_z\left[\widetilde{\mathsf{ReLU}}\left(s_r-b\right)\widetilde{\mathsf{ReLU}}'\left(s_{r'}-b\right)\left(\sum_{p\in[m_1]}\langle[\mathbf{W}_{1,j}]_{r'},[\mathbf{W}_{1,j}]_p\rangle\mathbb{1}_{[\mathbf{V}_{1,j}z-b_{1,j}]_p\ge\zeta}[\mathbf{\Delta}_1 z]_p\right)\right]\\
&\quad\pm\frac{\eta\mathsf{poly}(k_1)}{\zeta}\mathop{\mathbb{E}}_z\left[\widetilde{\mathsf{ReLU}}\left(s_r-b\right)\widetilde{\mathsf{ReLU}}'\left(s_{r'}-b\right)\left(\sum_{p\in[m_1]}\langle[\mathbf{W}_{1,j}]_{r'},[\mathbf{W}_{1,j}]_p\rangle\mathbb{1}_{|[\mathbf{V}_{1,j}z-b_{1,j}]_p|\le 2\zeta}\right)\right]
\end{aligned}
$$

where the second inequality uses $|[\mathbf{\Delta}_1 z]_p|\le\mathsf{poly}(k_1)$ with high probability.

Notice that

- We have for $p\neq r,r'$:

$$
\begin{aligned}
&\mathop{\mathbb{E}}_z\left[\widetilde{\mathsf{ReLU}}\left(s_r-b\right)\widetilde{\mathsf{ReLU}}'\left(s_{r'}-b\right)\mathbb{1}_{|[\mathbf{V}_{1,j}z-b_{1,j}]_p\le\zeta}\right]\\
&\le\mathsf{poly}(k_1)\mathop{\mathbb{E}}_z\left[\widetilde{\mathsf{ReLU}}'\left(s_r-b\right)\widetilde{\mathsf{ReLU}}'\left(s_{r'}-b\right)\mathbb{1}_{|[\mathbf{V}_{1,j}z-b_{1,j}]_p\le\zeta}\right]\\
&\le\mathsf{poly}(k_1)\mathop{\mathbb{E}}_z\left[\mathbb{1}_{g_r-\beta_r\ge-2b}\mathbb{1}_{g_{r'}-\beta_{r'}\ge-2b}\mathbb{1}_{|g_p-\beta_p|\le\zeta}\right]\le\frac{\zeta\mathsf{poly}(k_1)}{m_1^3}P_{r,r'}
\end{aligned}
$$

  where the last inequality uses Proposition F.25e.

- When $p=r'$ or $r$ we naively have $\mathbb{E}_z\left[\widetilde{\mathsf{ReLU}}\left(s_r-b\right)\widetilde{\mathsf{ReLU}}'\left(s_{r'}-b\right)\mathbb{1}_{|[\mathbf{V}_{1,j}z-b_{1,j}]_p\le\zeta}\right]=0$

They imply

$$
\begin{aligned}
&\mathop{\mathbb{E}}_z\left[\widetilde{\mathsf{ReLU}}\left(s_r-b\right)\widetilde{\mathsf{ReLU}}'\left(s_{r'}-b\right)\xi_{r'}\right]\\
&=\eta\mathop{\mathbb{E}}_z\left[\widetilde{\mathsf{ReLU}}\left(s_r-b\right)\widetilde{\mathsf{ReLU}}'\left(s_{r'}-b\right)\left(\sum_{p\in[m_1]}\langle[\mathbf{W}_{1,j}]_{r'},[\mathbf{W}_{1,j}]_p\rangle\mathbb{1}_{[\mathbf{V}_{1,j}z-b_{1,j}]_p\ge\zeta}[\mathbf{\Delta}_1 z]_p\right)\right]\\
&\quad\pm\eta\mathsf{poly}(k_1)\frac{1}{m_1^2}\delta P_{r,r'}
\end{aligned}
$$

Using the property that w.h.p. $|[\mathbf{\Delta}_1 z]_p|\le\mathsf{poly}(k_1)$ and $\delta\cdot\sum_{p\in[m_1]}\mathbb{1}_{[\mathbf{V}_{1,j}z-b_{1,j}]_p>0}\ll b$, together with (F.15), we further have

$$
\begin{aligned}
&\mathop{\mathbb{E}}_z\left[\widetilde{\mathsf{ReLU}}\left(s_r-b\right)\widetilde{\mathsf{ReLU}}'\left(s_{r'}-b\right)\xi_{r'}\right]\\
&=\eta\mathop{\mathbb{E}}_z\left[\widetilde{\mathsf{ReLU}}\left(s_r-b\right)\widetilde{\mathsf{ReLU}}'\left(s_{r'}-b\right)\left(\mathbb{1}_{[\mathbf{V}_{1,j}z-b_{1,j}]_{r'}\ge\zeta}[\mathbf{\Delta}_1 z]_{r'}\right)\right]\pm\eta\mathsf{poly}(k_1)\frac{1}{m_1^2}\delta P_{r,r'}\\
&=\frac{\eta}{2}\mathop{\mathbb{E}}_z\left[\widetilde{\mathsf{ReLU}}\left(g_r-\beta_r\right)\widetilde{\mathsf{ReLU}}'\left(g_{r'}-\beta_{r'}\right)g_{r'}\right]\pm\eta\mathsf{poly}(k_1)\frac{1}{m_1^2}bP_{r,r'} \qquad \text{(F.17)}
\end{aligned}
$$

where the last equality uses Proposition F.27, and $\mathbf{\Delta}_1 = \frac{1-\sqrt{1-\eta}}{\eta}\mathbf{V}_{1,j}$ for $\frac{1-\sqrt{1-\eta}}{\eta} = \frac{1}{2} \pm O(\eta)$.

2. $\frac{1}{2}\mathbb{E}_z\left[\widetilde{\mathsf{ReLU}}\,(s_r - b)\,\widetilde{\mathsf{ReLU}}''(s_{r'} - b)\xi_{r'}^2\right]$: for this term, we know that ignoring any $\eta^{1.5}$ or higher-order terms, we have

$$\begin{aligned}
&\frac{1}{2}\mathop{\mathbb{E}}_z\left[\widetilde{\mathsf{ReLU}}\,(s_r - b)\,\widetilde{\mathsf{ReLU}}''(s_{r'} - b)\xi_{r'}^2\right]\\
&= \frac{\eta}{2}\mathop{\mathbb{E}}_z\left[\widetilde{\mathsf{ReLU}}\,(s_r - b)\,\widetilde{\mathsf{ReLU}}''(s_r - b)\left(\sum_{p\in[m_1]}\langle[\mathbf{W}_{1,j}]_{r'}, [\mathbf{W}_{1,j}]_p\rangle \mathsf{LeakyReLU}'([\mathbf{V}_{1,j}z - b_{1,j}]_p)[\mathbf{\Delta}_2 z]_p\right)^2\right]\\
&\overset{①}{=} \frac{\eta}{2}\mathop{\mathbb{E}}_z\left[\widetilde{\mathsf{ReLU}}\,(s_r - b)\,\widetilde{\mathsf{ReLU}}''(s_r - b)\left(\mathsf{LeakyReLU}'([\mathbf{V}_{1,j}z - b_{1,j}]_{r'})[\mathbf{\Delta}_2 z]_{r'} \pm b\mathsf{poly}(k_1)\right)^2\right]\\
&\overset{②}{=} \frac{\eta}{2}\mathop{\mathbb{E}}_z\left[\widetilde{\mathsf{ReLU}}\,(s_r - b)\,\widetilde{\mathsf{ReLU}}''(s_{r'} - b)[\mathbf{\Delta}_2 z]_{r'}^2\right] \pm \eta\frac{b\mathsf{poly}(k_1)}{m_1^2}P_{r,r'}\\
&\overset{③}{=} \frac{\eta}{2}\mathop{\mathbb{E}}_z\left[\widetilde{\mathsf{ReLU}}\,(g_r - \beta_r)\,\widetilde{\mathsf{ReLU}}''(g_{r'} - \beta_{r'})\right]\mathbb{E}[g_{r'}^2] \pm \eta\frac{b\mathsf{poly}(k_1)}{m_1^2}P_{r,r'} \qquad \text{(F.18)}
\end{aligned}$$

Above, ① uses the property that w.h.p. $|[\mathbf{\Delta}_2 z]_p| \leq \mathsf{poly}(k_1)$ and $\delta\cdot\sum_{p\in[m_1]}\mathbb{1}_{[\mathbf{V}_{1,j}z - b_{1,j}]_p > 0} \ll b$; ② uses (F.16); and ③ uses Proposition F.27 together with the property that the randomness in $\mathbf{\Delta}_2 z \in \mathbb{R}^{m_1}$ is independent of $[\mathbf{V}_{1,j}]z \in \mathbb{R}^{m_1}$, together with

$$\mathbb{E}[[\mathbf{\Delta}_2 z]_{r'}^2] = [\mathbf{M}^\star]_{r,r} = \mathbb{E}[[\mathbf{V}_{1,j}z]_{r'}^2] = \mathbb{E}[g_{r'}^2]\ .$$

3. $\mathbb{E}_z\left[\widetilde{\mathsf{ReLU}}'\,(s_r - b)\,\widetilde{\mathsf{ReLU}}'\,(s_{r'} - b)\,\xi_r\xi_{r'}\right]$, this term can be bounded as (ignoring terms higher than $\eta^{1.5}$ order):

$$\begin{aligned}
&\mathop{\mathbb{E}}_z\left[\widetilde{\mathsf{ReLU}}'\,(s_r - b)\,\widetilde{\mathsf{ReLU}}'\,(s_{r'} - b)\,\xi_r\xi_{r'}\right]\\
&\overset{①}{=} \mathop{\mathbb{E}}_z\Big[\widetilde{\mathsf{ReLU}}'\,(s_r - b)\,\widetilde{\mathsf{ReLU}}'\,(s_{r'} - b)\left(\mathsf{LeakyReLU}'([\mathbf{V}_{1,j}z - b_{1,j}]_r)[\mathbf{\Delta}_2 z]_r \pm b\mathsf{poly}(k_1)\right)\\
&\qquad\qquad\left(\mathsf{LeakyReLU}'([\mathbf{V}_{1,j}z - b_{1,j}]_{r'})[\mathbf{\Delta}_2 z]_{r'} \pm b\mathsf{poly}(k_1)\right)\Big]\\
&\overset{②}{=} \eta\mathop{\mathbb{E}}_z\left[\widetilde{\mathsf{ReLU}}'\,(s_r - b)\,\widetilde{\mathsf{ReLU}}'\,(s_{r'} - b)\,[\mathbf{\Delta}_2 z]_r[\mathbf{\Delta}_2 z]_{r'}\right] \pm \eta\frac{b\mathsf{poly}(k_1)}{m_1^2}P_{r,r'}\\
&\overset{③}{=} \eta\mathop{\mathbb{E}}_z\left[\widetilde{\mathsf{ReLU}}'\,(g_r - \beta_r)\,\widetilde{\mathsf{ReLU}}'\,(g_{r'} - \beta_{r'})\right]\langle[\mathbf{V}^\star_{1,\ell}]_r, [\mathbf{V}^\star_{1,\ell}]_{r'}\rangle \pm \eta\frac{b\mathsf{poly}(k_1)}{m_1^2}P_{r,r'} \qquad \text{(F.19)}
\end{aligned}$$

Above, ① again uses the property that w.h.p. $|[\mathbf{\Delta}_2 z]_p| \leq \mathsf{poly}(k_1)$ and $\delta\cdot\sum_{p\in[m_1]}\mathbb{1}_{[\mathbf{V}_{1,j}z - b_{1,j}]_p > 0} \ll b$; ② uses (F.15); and ③ uses Proposition F.27 together with the property that the randomness in $\mathbf{\Delta}_2 z \in \mathbb{R}^{m_1}$ is independent of $[\mathbf{V}_{1,j}]z \in \mathbb{R}^{m_1}$, as well as

$$\mathbb{E}[[\mathbf{\Delta}_2 z]_r[\mathbf{\Delta}_2 z]_{r'}] = [\mathbf{M}^\star]_{r,r'} = \langle[\mathbf{V}^\star_{1,\ell}]_r, [\mathbf{V}^\star_{1,\ell}]_{r'}\rangle\cdot\left(1 \pm b\mathsf{poly}(k_1)\right)$$

Combining Eq (F.17), Eq (F.18), Eq (F.19), we have

$$\begin{aligned}
\mathbb{E}\left[\widetilde{\mathsf{ReLU}}\,(s'_r - b)\,\widetilde{\mathsf{ReLU}}\,(s'_{r'} - b)\right] &= \mathbb{E}\left[\widetilde{\mathsf{ReLU}}\,(s_r - b)\,\widetilde{\mathsf{ReLU}}\,(s_{r'} - b)\right] \pm \eta\frac{b\mathsf{poly}(k_1)}{m_1^2}P_{r,r'}\\
&+ \mathop{\mathbb{E}}_z\left[\widetilde{\mathsf{ReLU}}\,(g_r - \beta_r)\,\widetilde{\mathsf{ReLU}}'\,(g_{r'} - \beta_{r'})\,g_{r'}\right] + \mathop{\mathbb{E}}_z\left[\widetilde{\mathsf{ReLU}}'\,(g_r - \beta_r)\,\widetilde{\mathsf{ReLU}}\,(g_{r'} - \beta_{r'})\,g_r\right]\\
&+ \eta\mathop{\mathbb{E}}_z\left[\widetilde{\mathsf{ReLU}}'\,(g_r - \beta_r)\,\widetilde{\mathsf{ReLU}}'\,(g_{r'} - \beta_{r'})\right]\langle[\mathbf{V}^\star_{1,\ell}]_r, [\mathbf{V}^\star_{1,\ell}]_{r'}\rangle\\
&+ \frac{\eta}{2}\mathop{\mathbb{E}}_z\left[\widetilde{\mathsf{ReLU}}\,(g_r - \beta_r)\,\widetilde{\mathsf{ReLU}}''(g_{r'} - \beta_{r'})\right]\mathbb{E}[g_{r'}^2] + \frac{\eta}{2}\mathop{\mathbb{E}}_z\left[\widetilde{\mathsf{ReLU}}''\,(g_r - \beta_r)\,\widetilde{\mathsf{ReLU}}(g_{r'} - \beta_{r'})\right]\mathbb{E}[g_r^2]
\end{aligned}$$

Applying Claim F.28, we can further simplify it as

$$\mathbb{E}\left[\widetilde{\mathsf{ReLU}}\left(s'_r - b\right)\widetilde{\mathsf{ReLU}}\left(s'_{r'} - b\right)\right] = \mathbb{E}\left[\widetilde{\mathsf{ReLU}}\left(s_r - b\right)\widetilde{\mathsf{ReLU}}\left(s_{r'} - b\right)\right] \pm \eta\frac{b\mathsf{poly}(k_1)}{m_1^2}P_{r,r'}$$
$$+ \eta\left[\widetilde{\mathsf{ReLU}}'\left(g_r - \beta_r\right)\widetilde{\mathsf{ReLU}}'\left(g_{r'} - \beta_{r'}\right)\right]\left(\langle[\mathbf{V}^\star_{1,\ell}]_r, [\mathbf{V}^\star_{1,\ell}]_{r'}\rangle - \langle[\mathbf{V}_{1,j}]_r, [\mathbf{V}_{1,j}]_{r'}\rangle\right)$$
$$= \mathbb{E}\left[\widetilde{\mathsf{ReLU}}\left(s_r - b\right)\widetilde{\mathsf{ReLU}}\left(s_{r'} - b\right)\right] \pm \eta\frac{b\mathsf{poly}(k_1)}{m_1^2}P_{r,r'}$$
$$+ \eta\Theta\left(\frac{P_{r,r'}}{m_1^2}\right)\left(\langle[\mathbf{V}^\star_{1,\ell}]_r, [\mathbf{V}^\star_{1,\ell}]_{r'}\rangle - \langle[\mathbf{V}_{1,j}]_r, [\mathbf{V}_{1,j}]_{r'}\rangle\right) \tag{F.20}$$

Finally, we consider three cases.

- Suppose $\mathbb{E}\left[\widetilde{\mathsf{ReLU}}\left(s_r - b\right)\widetilde{\mathsf{ReLU}}\left(s_{r'} - b\right)\right] < \mathbb{E}\left[\widetilde{\mathsf{ReLU}}\left(s^\star_r - b\right)\widetilde{\mathsf{ReLU}}\left(s^\star_{r'} - b\right)\right] - \frac{b\mathsf{poly}(k_1)}{m_1^2}$.

  Recall from Assumption 2.2 and Proposition F.25a it satisfies

  $$\frac{1}{\mathsf{poly}(d)} \le \mathbb{E}\left[\widetilde{\mathsf{ReLU}}\left(s^\star_r - b\right)\widetilde{\mathsf{ReLU}}\left(s^\star_{r'} - b\right)\right] \le \frac{\mathsf{poly}(k_1)}{m_1^2}\ .$$

  This implies

  $$\mathbb{E}\left[\widetilde{\mathsf{ReLU}}\left(s_r - b\right)\widetilde{\mathsf{ReLU}}\left(s_{r'} - b\right)\right] < (1 - b\mathsf{poly}(k_1))\,\mathbb{E}\left[\widetilde{\mathsf{ReLU}}\left(s^\star_r - b\right)\widetilde{\mathsf{ReLU}}\left(s^\star_{r'} - b\right)\right] - \frac{1}{\mathsf{poly}(d)}$$

  and thus by applying Proposition F.25d, we have

  $$\langle[\mathbf{V}_{1,j}]_r, [\mathbf{V}_{1,j}]_{r'}\rangle \le \langle[\mathbf{V}^\star_{1,j}]_r, [\mathbf{V}^\star_{1,j}]_{r'}\rangle - b\mathsf{poly}(k_1)$$

  Therefore, in this case, using $\gamma^2 = b\mathsf{poly}(k_1)$ for a sufficiently large $\mathsf{poly}(k_1)$ factor, we can derive from (F.20) that

  $$\mathbb{E}\left[\widetilde{\mathsf{ReLU}}\left(s'_r - b\right)\widetilde{\mathsf{ReLU}}\left(s'_{r'} - b\right)\right] \ge \mathbb{E}\left[\widetilde{\mathsf{ReLU}}\left(s_r - b\right)\widetilde{\mathsf{ReLU}}\left(s_{r'} - b\right)\right] + \eta\frac{b\mathsf{poly}(k_1)}{m_1^2}P_{r,r'}$$

  On the other hand, our choice of the regularizer $R^{(5)}$ in (F.14) ensures that $\left\langle \frac{[\mathbf{V}_{1,j}]_r}{\|[\mathbf{V}_{1,j}]_r\|_2}, \frac{[\mathbf{V}_{1,j'}]_{r'}}{\|[\mathbf{V}_{1,j'}]_{r'}\|_2} \right\rangle \ge -0.99$. One can easily verify that this implies $\frac{P_{r,r'}}{m_1^2} = \mathbb{E}\left[\widetilde{\mathsf{ReLU}}\left(s_r - b\right)\widetilde{\mathsf{ReLU}}\left(s_{r'} - b\right)\right] \ge \frac{1}{\mathsf{poly}(d)}$ by applying Proposition F.25a. This finishes the proof of the first case.

- Suppose $\mathbb{E}\left[\widetilde{\mathsf{ReLU}}\left(s_r - b\right)\widetilde{\mathsf{ReLU}}\left(s_{r'} - b\right)\right] > \mathbb{E}\left[\widetilde{\mathsf{ReLU}}\left(s^\star_r - b\right)\widetilde{\mathsf{ReLU}}\left(s^\star_{r'} - b\right)\right] + \frac{b\mathsf{poly}(k_1)}{m_1^2}$.

  In this case, together with the Invariant F.2c, one can similarly derive $\langle[\mathbf{V}_{1,j}]_r, [\mathbf{V}_{1,j}]_{r'}\rangle > \langle[\mathbf{V}^\star_{1,j}]_r, [\mathbf{V}^\star_{1,j}]_{r'}\rangle + b\mathsf{poly}(k_1)$ and therefore from (F.20) we have

  $$\mathbb{E}\left[\widetilde{\mathsf{ReLU}}\left(s'_r - b\right)\widetilde{\mathsf{ReLU}}\left(s'_{r'} - b\right)\right] \le \mathbb{E}\left[\widetilde{\mathsf{ReLU}}\left(s_r - b\right)\widetilde{\mathsf{ReLU}}\left(s_{r'} - b\right)\right] - \eta\frac{b\mathsf{poly}(k_1)}{m_1^2}P_{r,r'}$$

  This finishes the proof of the second case.

- In the remaining case, for similar reason (F.20) tells us

  $$\mathbb{E}\left[\widetilde{\mathsf{ReLU}}\left(s'_r - b\right)\widetilde{\mathsf{ReLU}}\left(s'_{r'} - b\right)\right] = (1 \pm \eta b\mathsf{poly}(k_1))\,\mathbb{E}\left[\widetilde{\mathsf{ReLU}}\left(s_r - b\right)\widetilde{\mathsf{ReLU}}\left(s_{r'} - b\right)\right]$$

Above, we have finished the proof of Claim F.15 for the case of $j = j'$.

□

*Remark* F.16. The proof for $j \neq j'$ for Claim F.15 is very analogous to $j = j'$, and the main difference is that we no longer have $s_{j,j,r,r'}, s^\star_{j,j,r,r'} \leq \frac{\mathsf{poly}(k_1)}{m_1^2}$ (which implies

$$\left\langle \frac{[\mathbf{V}_{1,j}]_r}{\|[\mathbf{V}_{1,j}]_r\|_2}, \frac{[\mathbf{V}_{1,j}]_{r'}}{\|[\mathbf{V}_{1,j}]_{r'}\|_2} \right\rangle, \left\langle \frac{[\mathbf{V}^\star_{1,j}]_r}{\|[\mathbf{V}^\star_{1,j}]_r\|_2}, \frac{[\mathbf{V}^\star_{1,j}]_{r'}}{\|[\mathbf{V}^\star_{1,j}]_{r'}\|_2} \right\rangle \leq o(1)$$

by Proposition F.22).

Instead, this time we only know $\left\langle \frac{[\mathbf{V}_{1,j}]_r}{\|[\mathbf{V}_{1,j}]_r\|_2}, \frac{[\mathbf{V}_{1,j'}]_{r'}}{\|[\mathbf{V}_{1,j'}]_{r'}\|_2} \right\rangle \leq 1 - \tau/2$ thanks to the regularizer (see Claim F.13) and $\left\langle \frac{[\mathbf{V}^\star_{1,j}]_r}{\|[\mathbf{V}^\star_{1,j}]_r\|_2}, \frac{[\mathbf{V}^\star_{1,j'}]_{r'}}{\|[\mathbf{V}^\star_{1,j'}]_{r'}\|_2} \right\rangle \leq 1 - \tau/2$ thanks to the non-degenerate assumption Assumption B.1 (see Fact F.11).

This shall affect all the propositions about joint Gaussian (namely, Proposition F.25, Proposition F.26 and Proposition F.27); however, the three propositions remain to hold once we relax $\mathsf{poly}(k_1)$ with $\mathsf{poly}(m_1)$. For such reason, the the last three items of Claim F.15 can be derived in the same way as the first three items, after replacing $\mathsf{poly}(k-1)$ with $\mathsf{poly}(m_1)$.

### F.4.3 Proof of Lemma F.10: Optimization by Hessian Update

In this subsection, we verify that Claim F.15 gives a valid Hessian descent direction, which shows that during the training of $D^{(5)}$, either the gradient is large or the Hessian has a sufficiently negative direction. This shall be sufficient to guarantee convergence on the minimization objective $Obj^{(5)}(\mathbf{V}_1)$ (F.13) until we find a point $\mathbf{V}_1$ satisfying $Obj^{(5)}(\mathbf{V}_1) \leq \mathsf{poly}(k_1)$.

**Background on Hessian Descent.** For readers not familiar with this standard routine, we explain it below. For any function $f(x) \in \mathbb{R}^d \times \mathbb{R}$ that is $\mathsf{poly}(d)$ Lipschitz smooth and Lipschitz second-order smooth

$$\text{i.e., } \|\nabla^2 f(x)\|_{spectral} \leq \mathsf{poly}(d) \text{ and } \|\nabla^2 f(x) - \nabla^2 f(y)\|_{spectral} \leq \mathsf{poly}(d)\|x-y\|_2,$$

after choosing $\eta = \frac{1}{\mathsf{poly}(d/\varepsilon)}$ as the learning rate for the (perturbed) gradient descent, and $T = \frac{\mathsf{poly}(d/\varepsilon)}{\eta}$ be the number if iterations, then we can guarantee to find a point $x$ satisfying $\|\nabla f(x)\|_2 \leq \varepsilon$ and $\lambda_{\min}(\nabla^2 f(x)) \geq -\varepsilon$. This can be found in well-known literature such as [58] for (perturbed) gradient descent and [46, 59] for (perturbed) stochastic gradient descent.

Furthermore, as long as there exists vector $x_1 \in \mathbb{R}^d$ and a random vector $x_2 \in \mathbb{R}^d$ with $\mathbb{E}[x_2] = 0$ satisfying $\|x_1\|_2 \leq \mathsf{poly}(d)$, $\mathbb{E}[\|x_2\|_2^2] \leq \mathsf{poly}(d)$ and

$$\mathbb{E}_{x_2}[f(x + \eta x_1 + \sqrt{\eta} x_2)] \leq f(x) - \eta \frac{1}{\mathsf{poly}(d)} \quad . \tag{F.21}$$

then we have either $\|\nabla f(x)\|_2 > \frac{1}{\mathsf{poly}(d)}$ or $\lambda_{\min}(\nabla^2 f(x)) > -\frac{1}{\mathsf{poly}(d)}$ (see Fact F.30 or [4, 7]). In other words, it suffices for us to prove the *existence* of a Hessian update direction (F.21), and if (F.21) is satisfied, then (perturbed) gradient descent can continue to decrease the objective.

Finally, as we shall see, our minimization problem $Obj^{(5)}(\mathbf{V}_1)$ (F.13) corresponds to optimizing over a function $g(x) = f(x/\|x\|_2)$ over the unit ball. This requires us to construct a Hessian update direction satisfying (F.21) together with $\|x+\eta x_1+\sqrt{\eta}x_2\|_2 = \|x\|_2$. When this strengthened version of (F.21) is satisfied, then we also have

$$\mathbb{E}_{x_2}[g(x + \eta x_1 + \sqrt{\eta} x_2)] \leq g(x) - \eta \frac{1}{\mathsf{poly}(d)} \quad .$$

so the same conclusion holds on $g(x) = f(x/\|x\|_2)$.[22]

**Claim F.17.** *In the same setting as Claim F.13, during the training process of $D^{(5)}$, consider the (normalized) objective $Obj^{(5)}(\mathbf{V}_1) \stackrel{\text{def}}{=} H(\{[\mathbf{V}_{1,\ell}]_r \cdot \frac{\alpha_{1,\ell,r}}{[\mathbf{V}_{1,\ell}]_r}\}_{j\in[d_1],r\in[m_1]}) + R^{(5)}(\mathbf{V}_1) \geq \mathsf{poly}(k_1)$, it satisfies*

$$\text{either} \quad \|\nabla Obj^{(5)}(\mathbf{V}_1)\|_F \geq \frac{1}{\mathsf{poly}(d)} \quad \text{or} \quad \lambda_{\min}\left(\nabla^2 Obj^{(5)}(\mathbf{V}_1)\right) \leq \frac{1}{\mathsf{poly}(d)}$$

*Proof of Claim F.17.* Let us consider the update defined in Definition F.14. One can verify that, as long as $H(\mathbf{V}_1) \geq \mathsf{poly}(k_1)$, it necessarily satisfies that

$$\begin{aligned} &\text{either for some } j, r \neq r'\text{:} && |s_{j,j,r,r'} - s^\star_{j,j,r,r'}| > \frac{b\mathsf{poly}(k_1)}{m_1^2} \\ &\text{or for some } (j,r) \neq (j',r')\text{:} && |s_{j,j',r,r'} - s^\star_{j,j',r,r'}| > b\mathsf{poly}(m_1) \end{aligned}$$

In both cases, Claim F.15 guarantees that $H(\mathbf{V}_1') \leq H(\mathbf{V}_1) - \frac{\eta}{\mathsf{poly}(d)}$ so we can make progress.

As for the regularizer, recall from Definition F.14 that $\|[\mathbf{V}'_{1,j}]_r\|_2 = \|[\mathbf{V}_{1,j}]_r\|_2$ is unchanged. Thus one can verify

$$\left\langle \frac{[\mathbf{V}'_{1,j}]_r}{\|[\mathbf{V}'_{1,j}]_r\|_2}, \frac{[\mathbf{V}'_{1,j'}]_{r'}}{\|[\mathbf{V}'_{1,j'}]_{r'}\|_2} \right\rangle = (1-\eta)\left\langle \frac{[\mathbf{V}_{1,j}]_r}{\|[\mathbf{V}_{1,j}]_r\|_2}, \frac{[\mathbf{V}_{1,j'}]_{r'}}{\|[\mathbf{V}_{1,j'}]_{r'}\|_2} \right\rangle + \eta \left\langle \frac{[\mathbf{V}^\star_{1,j}]_r}{\|[\mathbf{V}^\star_{1,j}]_r\|_2}, \frac{[\mathbf{V}^\star_{1,j'}]_{r'}}{\|[\mathbf{V}^\star_{1,j'}]_{r'}\|_2} \right\rangle$$

Recalling from Fact F.11 that $\left\langle \frac{[\mathbf{V}^\star_{1,j}]_r}{\|[\mathbf{V}^\star_{1,j}]_r\|_2}, \frac{[\mathbf{V}^\star_{1,j'}]_{r'}}{\|[\mathbf{V}^\star_{1,j'}]_{r'}\|_2} \right\rangle \in [-1+0.5\tau, 1-0.5\tau]$, so this tells us

$$R^{(5)}(\mathbf{V}_1') \leq (1-\eta) R^{(5)}(\mathbf{V}_1') + \frac{\eta}{d^{\mathsf{poly}(k_1)}}$$

In sum, one can conclude that as long as $Obj^{(5)}(\mathbf{V}_1) \geq \mathsf{poly}(k_1)$, it satisfies

$$Obj^{(5)}(\mathbf{V}_1') \leq Obj^{(5)}(\mathbf{V}_1) - \frac{\eta}{\mathsf{poly}(d)}$$

Furthermore, note from Definition F.14 that negating $\mathbf{S}_j \leftarrow -\mathbf{S}_j$ does not change $Obj^{(5)}(\mathbf{V}_1')$. Therefore, applying standard Hessian analysis on a second-order smooth objective (see Fact F.30), we can conclude that

$$\text{either} \quad \|\nabla Obj^{(5)}(\mathbf{V}_1)\|_F \geq \frac{1}{\mathsf{poly}(d)} \quad \text{or} \quad \lambda_{\min}\left(\nabla^2 Obj^{(5)}(\mathbf{V}_1)\right) \leq \frac{1}{\mathsf{poly}(d)}$$

□

Applying Claim F.17, using standard Hessian update analysis (see the aforementioned "Background" paragraph), we arrive at the Lemma F.10.

(Note the implication in the first item of Lemma F.10 can be derived using Proposition F.25d; and the implication in the second item can be derived using Proposition F.25d after replacing $\mathsf{poly}(k_1)$ with $\mathsf{poly}(m_1)$, see Remark F.16.)

[22] There is a minor issue here. In principle, when $\|x\|_2$ is close to zero, the smoothness property of $g(x)$ shall be violated. However, if we start with $x = x_0$ satisfying $\|x_0\|_2 \geq \frac{1}{\mathsf{poly}(d)}$, then one can show during the training process, when $\eta = \frac{1}{\mathsf{poly}(d)}$ is sufficiently small, it always satisfies $\|x\|_2 \geq 0.99\|x_0\|_2$. This is because, during each gradient descent step, we have $x' \leftarrow x - \eta \nabla g(x)$ but $\nabla g(x)$ is *orthogonal* to $x$. In other words, $x$ moves by $\eta$ but $\|x'\|_2 = \|x\|_2 \pm O(\eta^2)$; so if $\eta$ is sufficiently small and $T = \frac{\mathsf{poly}(d/\varepsilon)}{\eta}$, then $\|x'\|_2$ does not change by much during the minimization process.

## F.5 Putting All Together

Suppose we

- initialize $\mathbf{W}_1$ using warm start (see Algorithm 5 and Lemma D.1),
- then repeatedly (and alternatively)
  - learn $\mathbf{V}_1$ (see Algorithm 2 and Lemma F.10),
  - learn $\mathbf{W}_1$ (see Algorithm 1 and Lemma E.2),
  - decrease $b \leftarrow bm^{-0.02}$;

Then, combining all the cited lemmas, we have

**Lemma F.18** ($D^{(1)} + D^{(4)} + D^{(5)}$)**.** *After iteration $\ell = 1$ is finished in Algorithm 4, there is a column orthonormal matrix $\mathbf{U} \in \mathbb{R}^{m_0 \times m_0'}$ satisfying, for every $j \in [d_1]$:*

- $\|\mathbf{V}_{1,j} - \mathbf{V}^\star_{1,j}\mathbf{U}\|_F \leq \frac{1}{\mathsf{poly}(d/\varepsilon)}$
- *w.h.p. over $z \sim \mathcal{N}(0, \mathbf{I}_{m_0' \times m_0'})$:* $[\mathcal{S}^\star_{1,j}(\mathbf{U}z)]_r - \frac{1}{\mathsf{poly}(d/\varepsilon)} \leq [\mathcal{S}_{1,j}(z)]_r \leq [\mathcal{S}^\star_{1,j}(\mathbf{U}z)]_r$.

*Proof of Lemma F.18.* First note after recursion, we have $b = \frac{1}{\mathsf{poly}(d/\varepsilon)}$ and Lemma F.10 tells us for all $(j, r), (j', r') \in [d_1] \times [m_1]$:

$$\left|\langle [\mathbf{V}^\star_{1,j}]_r, [\mathbf{V}^\star_{1,j'}]_{r'}\rangle - \langle [\mathbf{V}_{1,j}]_r, [\mathbf{V}_{1,j'}]_{r'}\rangle\right| \leq \frac{1}{\mathsf{poly}(d/\varepsilon)}$$

This implies there is a column orthonormal matrix $\mathbf{U} \in \mathbb{R}^{m_0 \times m_0'}$ such that:

$$\forall j \in [d_1]: \quad \|\mathbf{V}_{1,j} - \mathbf{V}^\star_{1,j}\mathbf{U}\|_F \leq \frac{1}{\mathsf{poly}(d/\varepsilon)}$$

Recall $\mathcal{S}^\star_{1,j}(z) = \mathsf{ReLU}(\mathbf{V}^\star_{1,j}z - b^\star_{1,j})$ and $\mathcal{S}_{1,j}(z) = \mathsf{LeakyReLU}(\mathbf{V}_{1,j}z - b_{1,j})$. And, recall from Lemma F.6 we can have

$$|[b_{1,j}]_r - [b^\star_{1,j}]_r| \leq \frac{1}{\mathsf{poly}(d/\varepsilon)}$$

This implies for every $r \in [m_1]$, w.h.p. over $z \sim \mathcal{N}(0, \mathbf{I}_{m_0' \times m_0'})$:

$$[\mathcal{S}^\star_{1,j}(\mathbf{U}z)]_r| \leq \frac{1}{\mathsf{poly}(d/\varepsilon)}$$

At this point, if we increase each $[b_{1,j}]_r$ slightly by $\frac{1}{\mathsf{poly}(d/\varepsilon)}$ (see Line 8 in Algorithm 4), we have

$$[\mathcal{S}_{1,j}(z)]_r \leq [\mathcal{S}^\star_{1,j}(\mathbf{U}z)]_r, \quad |[\mathcal{S}_{1,j}(z)]_r - [\mathcal{S}^\star_{1,j}(\mathbf{U}z)]_r| \leq \frac{1}{\mathsf{poly}(d/\varepsilon)}$$

□

## F.6 Missing Math Propositions

### F.6.1 Simple Properties on ReLU

Below we state some simple propositions that are manipulations of properties of Gaussian distributions. We call some of the facts since they are straightforward to prove.

**Fact F.19.** *When $\beta > 0$, we have* $\mathbb{E}_{g \sim \mathcal{N}(0,1)}[\mathsf{ReLU}(g - \beta)] = \Theta(1) \cdot \mathbf{Pr}[g > \beta] \cdot \min\{\frac{1}{\beta}, 1\}$.

**Fact F.20.** *Under Assumption 2.2, we have* $\frac{1}{\mathsf{poly}(k_1)} \leq \|[\mathbf{V}^\star_{1,j}]_r\|_2 \leq [b^\star_{1,j}]_r \leq \mathsf{poly}(k_1)$ *for every $j \in [d_1], r \in [m_1]$.*

*Proof of Fact F.20.* Recall we can write $[\mathcal{S}^\star_{1,j}]_r = \mathsf{ReLU}(\alpha g - \beta) = \alpha \cdot \mathsf{ReLU}(g - \frac{\beta}{\alpha})$ for $\alpha = \|[\mathbf{V}^\star_{1,j}]_r\|_2$ and $\beta = [b^\star_{1,j}]_r$, and $g \sim \mathcal{N}(0,1)$.

- By the sparsity upper bound in Assumption 2.2, we have $\frac{\beta}{\alpha} \geq 1$.
- Conditioning on $g > \frac{\beta}{\alpha}$, with at least constant probability it satisfies $g > \frac{\beta}{\alpha} + \frac{\alpha}{2\beta}$. Since Assumption 2.2 says w.h.p. $|[\mathcal{S}^\star_{1,j}]_r| \leq \mathsf{poly}(k_1)$, we conclude that have $\alpha \cdot \frac{\alpha}{2\beta} \leq \mathsf{poly}(k_1)$.
- Assumption 2.2 implies $\mathbf{Pr}\left[g > \frac{\beta}{\alpha}\right] \geq \Omega(\frac{1}{\mathsf{poly}(k_1) m_1})$. This means $\frac{\beta}{\alpha} \leq O(\log m_1)$.
- By the expectation lower bound $\mathbb{E}[[\mathcal{S}^\star_{1,j}]_r] \geq \frac{1}{\mathsf{poly}(k_1) m_1}$ in Assumption 2.2, we can derive (using Fact F.19) $\alpha \cdot \mathbf{Pr}\left[g > \frac{\beta}{\alpha}\right] \cdot \frac{\alpha}{\beta} \geq \Omega(\frac{1}{\mathsf{poly}(k_1) m_1})$. Combining this with the sparsity upper bound $\mathbf{Pr}\left[g > \frac{\beta}{\alpha}\right] \leq O(\frac{\mathsf{poly}(k_1)}{m_1})$, we have $\frac{\alpha^2}{\beta} \geq \Omega(\frac{1}{\mathsf{poly}(k_1)})$.

Combining all of these, we have $\frac{1}{\mathsf{poly}(k_1)} \leq \alpha \leq \beta \leq \mathsf{poly}(k_1)$. □

**Fact F.21.** *Let $g \sim \mathcal{N}(0,1)$ and consider $\gamma, \gamma^\star, \beta, \beta^\star > 0$. Define*

$$a^\star = \mathbb{E}[\mathsf{ReLU}(\gamma^\star g - \beta^\star)], \qquad a = \mathbb{E}[\mathsf{ReLU}(\gamma g - \beta)]$$
$$c^\star = \mathbb{E}[\mathsf{ReLU}'(\gamma^\star g - \beta^\star)], \qquad c = \mathbb{E}[\mathsf{ReLU}'(\gamma g - \beta)] \ .$$

*Suppose $a, a^\star, c, c^\star \in \left[\frac{1}{m_1 \mathsf{poly}(k_1)}, \frac{\mathsf{poly}(k_1)}{m_1}\right]$, and suppose*

$$|a - a^\star| + |c - c^\star| \leq \frac{\tau}{m_1} \quad \textit{for some } \tau \in (1/\mathsf{poly}(d), m_1^{-0.01}] \ .$$

*Then, we have:*

$$|\gamma - \gamma^\star|, |\beta - \beta^\star| \leq \mathsf{poly}(k_1)\tau$$

*Proof of Fact F.21.* One can easily conclude (anyways see Fact F.20) that $\gamma, \gamma^\star, \beta, \beta^\star \in [1/\mathsf{poly}(k_1), \mathsf{poly}(k_1)]$. Now, we can write

$$c = \mathbb{E}\left[\mathsf{ReLU}'(g - \beta/\gamma)\right] \quad \text{and} \quad c^\star = \mathbb{E}\left[\mathsf{ReLU}'(g - \beta^\star/\gamma^\star)\right]$$

Using $|c - c^\star| \leq \frac{\tau}{m_1}$ we have derive $\left|\frac{\beta}{\gamma} - \frac{\beta^\star}{\gamma^\star}\right| \leq \tau \mathsf{poly}(k_1)$.

Next, when $\beta/\gamma$ is fixed, we know that $a$ is determined by $\gamma$ in the sense that

$$a = \gamma \cdot \mathbb{E}\left[(g - \beta/\gamma)\mathsf{ReLU}'(g - \beta/\gamma)\right]$$
$$a^\star = \gamma^\star \cdot \mathbb{E}\left[(g - \beta^\star/\gamma^\star)\mathsf{ReLU}'(g - \beta^\star/\gamma^\star)\right]$$

By $|a - a^\star| \leq \frac{\tau}{m_1}$ and $\left|\frac{\beta}{\gamma} - \frac{\beta^\star}{\gamma^\star}\right| \leq \tau \mathsf{poly}(k_1)$, we can derive

$$|\gamma - \gamma^\star| \leq \tau \mathsf{poly}(k_1) \ .$$ □

### F.6.2 Gaussian Correlation

**Proposition F.22** (Gaussian correlation)**.** *Suppose $g_1, g_2 \sim \mathcal{N}(0,1)$ are two dependent standard Gaussian variables and their joint distribution is also Gaussian. Suppose for $b_1, b_2 > 0$, $\gamma \in (0, 0.02]$ and $\kappa > 1$, it satisfies*

$$\mathbf{Pr}[g_1 > b_1] \geq \gamma, \quad \mathbf{Pr}[g_2 > b_2] \geq \gamma, \quad \mathbf{Pr}[g_1 > b_1, g_2 > b_2] \leq \gamma^2 \kappa \ .$$

*Then, we have:* $\mathbb{E}[\langle g_1, g_2 \rangle] \leq O\left(\frac{\log \kappa}{\log(1/\gamma)}\right)$.

*Proof of Proposition F.22.* First of all, if $b_2 < 2$ then we can increase $b_2$ to 2: this still ensures $\mathbf{Pr}[g_2 > b_2] > 0.02 \geq \gamma$ while at the same time only decreases $\mathbf{Pr}[g_1 > b_1, g_2 > b_2]$. This is similar for $b_1$ so we can wlog. assume $b_1, b_2 \geq 2$. Also without loss of generality, let us assume $b_1 \geq b_2$ and $\mathbb{E}[\langle g_1, g_2 \rangle] > 0$.

We write $g_2$ as $g_2 = \alpha g_1 + \sqrt{1-\alpha^2} g_3$, where $g_3$ is standard Gaussian independent of $g_1$ with $\alpha = \mathbb{E}[\langle g_1, g_2 \rangle] > 0$. We calculate that

$$\mathbf{Pr}[g_1 > b_1, g_2 > b_2] \geq \gamma \,\mathbf{Pr}[g_2 > b_2 \mid g_1 > b_1] \geq \gamma \,\mathbf{Pr}\left[g_3 \geq \frac{b_2 - \alpha b_1}{\sqrt{1-\alpha^2}}\right] \geq \gamma \,\mathbf{Pr}\left[g_3 \geq b_2 \left(\frac{1-\alpha}{\sqrt{1-\alpha^2}}\right)\right]$$

so by our assumption we have

$$\mathbf{Pr}\left[g_2 \geq b_2\right] \geq \gamma \quad \text{and} \quad \mathbf{Pr}\left[g_3 \geq b_2 \left(\frac{1-\alpha}{\sqrt{1-\alpha^2}}\right)\right] \leq \gamma\kappa$$

Using the gaussian tail bound which says for any $g \sim \mathcal{N}(0,1)$ and $b > 0$:

$$\frac{\exp\{-b^2/2\}}{\sqrt{2\pi}} \left(\frac{1}{b} - \frac{1}{b^3}\right) \leq \mathbf{Pr}[g > b] \leq \frac{\exp\{-b^2/2\}}{\sqrt{2\pi} b}$$

we respectively have

$$\exp\left\{-\frac{b_2^2}{2}\right\} = \Omega\left(b_2 \gamma\right) \quad \text{and} \quad \exp\left\{-\frac{b_2^2(1-\alpha)^2}{2(1-\alpha^2)}\right\} = O\left(b_2 \gamma \kappa\right)$$

Above, the first inequality implies $b_2 \geq \Omega(\sqrt{\log(1/\gamma)})$, and dividing the two inequalities we have

$$\exp\left\{b_2^2 \frac{\alpha - \alpha^2/2}{1-\alpha^2}\right\} = O(\kappa)$$

Together, they imply $\alpha \leq O\left(\frac{\log \kappa}{\log(1/\gamma)}\right)$. □

### F.6.3 Anti-Concentration for Not-So-Correlated Gaussians

We also have the following proposition of anti-concentration for not-so-correlated standard Gaussian variables.

**Proposition F.23** (anti-correlation)**.** *Suppose we have (dependent) standard Gaussian random variables $g_1, g_2, \cdots, g_n$ in $\mathbb{R}$ and their joint distribution is also Gaussian. Suppose for all $i, i' \in [n], i \neq i'$: $\mathbb{E}[\langle g_i, g_{i'} \rangle] \leq \xi \leq o(1)$ (can be arbitrarily negative, but not too positive). Suppose we are given $b_1, b_2, \cdots, b_n \in [0.5B, 2B]$ for $B = \omega(1)$. Then, as long as $n < o(\xi^{-2/3})$, it satisfies*

$$\mathbf{Pr}[\forall i, g_i \geq b_i] = O\left(\exp\left\{-\frac{(1-o(1))\sum_{i\in[n]} b_i^2}{2}\right\}\right)$$

*Proof of Proposition F.23.* Let $g = (g_1, g_2, \cdots, g_n)$ where $g = \mathbf{G} g'$ for $g' \sim \mathcal{N}(0, \mathbf{I}_{2n\times 2n})$ and $\mathbf{G} \in \mathbb{R}^{n\times 2n}$, where we have that $\mathbb{R}^{n\times n} \ni \mathbf{G}\mathbf{G}^\top := \mathbf{M} := \mathbf{I} + \mathbf{R} - \mathbf{E} \succeq 0$, where $\mathbf{R}, \mathbf{E} \geq 0$ (element-wise) and by $\mathbb{E}[\langle g_i, g_{i'} \rangle] = \xi \leq o(1)$, we have $\|\mathbf{R}\|_2 = \xi\sqrt{n}$, $\|\mathbf{E}\|_2 \leq \sqrt{n}$. Thus, setting $\gamma = \xi\sqrt{n}$ we have $\gamma \mathbf{I} - (1-\gamma)\mathbf{R} \succeq 0$. We can write

$$\mathbf{I} - (1-\gamma)\mathbf{E} = (1-\gamma)(\mathbf{I} + \mathbf{R} - \mathbf{E}) + (\gamma \mathbf{I} - (1-\gamma)\mathbf{R})$$

Hence, there exists (see Claim F.29) matrix $\mathbf{S} \in \mathbb{R}^{n\times 2n}$ such that $\|\mathbf{S}\|_2 \leq \sqrt{\xi\sqrt{n}}$ and

$$(\sqrt{1-\gamma}\mathbf{G} + \mathbf{S})(\sqrt{1-\gamma}\mathbf{G} + \mathbf{S})^\top = \mathbf{I} - (1-\gamma)\mathbf{E} \succeq \mathbf{G}\mathbf{G}^\top$$

Now, denote by $\vec{b}$ the vector $\vec{b} = (b_1, b_2, \cdots, b_n)$, we have either $\|g'\|_2^2 \geq \sum_{i\in[n]} b_i^2$; or $\|g'\|_2^2 \leq \sum_{i\in[n]} b_i^2 \leq 2n \min_{i\in[n]} b_i^2$ and therefore

$$g = \mathbf{G}g' \geq \vec{b} \implies [\sqrt{1-\gamma}\mathbf{G} + \mathbf{S}]g' \geq \vec{b}(1 - O(\gamma) - O(\sqrt{n}\|\mathbf{S}\|_2)) \geq \vec{b}(1 - O(\sqrt{\xi}n^{0.75}))$$

In the later case, we have for every $y \geq 0$ (element-wise) and using $\mathbf{E} \geq 0$ (element-wise), it satisfies

$$y^\top[\mathbf{I} - (1-\gamma)\mathbf{E}]^{-1}y = y^\top \sum_{i=0}^{\infty}(1-\gamma)^i \mathbf{E}^i y \geq \|y\|_2^2$$

Plugging $y = [\sqrt{1-\gamma}\mathbf{G}+\mathbf{S}]g'$, and using the formula that $\mathbf{I} \succeq \mathbf{P}^\top(\mathbf{P}\mathbf{P}^\top)^{-1}\mathbf{P}$ for $\mathbf{P} = \sqrt{1-\gamma}\mathbf{G}+\mathbf{S}$, we have $\|g'\|_2^2 \geq \|y\|_2^2 \geq \|\vec{b}\|_2^2(1 - O(\sqrt{\xi}n^{0.75}))$.

In either case, this implies (using tail bound for chi-square distribution)

$$\mathbf{Pr}[\forall i, g_i \geq b_i] \leq \mathbf{Pr}\left[\|g'\|_2^2 \geq \|\vec{b}\|_2^2(1 - O(\sqrt{\xi}n^{0.75}))\right] \leq O\left(\exp\left\{-\frac{\|\vec{b}\|_2^2(1 - O(\sqrt{\xi}n^{0.75}))}{2} + \frac{n}{4}\right\}\right)$$
$$\leq O\left(\exp\left\{-\frac{(1-o(1))\sum_{i\in[n]} b_i^2}{2}\right\}\right)$$

where the last inequality uses the upper bound assumption on $n$. □

#### F.6.4 Sensitivity of Gaussian

We have the following proposition to bound the sensitivity of Gaussian random variable.

**Proposition F.24** (Inside out). *Let $g_1 \sim \mathcal{N}(0,1)$ be standard Gaussian and $b_1$ be positive value such that*

$$\mathbf{Pr}[g_1 > b_1] \geq \frac{1}{m_1^{100}}$$

*We then have, for every value $\delta \in (0, m^{-0.1}]$, every variable $\delta_1 \in [-\delta, \delta]$ (whose values can depend on $g_1$) such that:*

$$\left|\frac{\partial \delta_1}{\partial g_1}\right| \leq \delta \mathsf{poly}(k_1) \quad \textit{for every } |g_1| \leq O(\sqrt{\log m_1})$$

*We must have (probability density is Lipschitz):*

$$p(g_1 = \delta_1 + b_1) = p(g_1 = b_1)(1 \pm \mathsf{poly}(k_1)\delta)$$

*and as long as $a < \delta$,*

$$\mathbf{Pr}[|g_1 - \delta_1 - b_1| \leq a] = \mathbf{Pr}[|g_1 - b_1| \leq a](1 \pm \mathsf{poly}(k_1)\delta)$$

*Proof of Proposition F.24.* We first notice that $b_1 \in (0, O(\sqrt{\log m_1})]$. Let us write $\delta_1 = \delta_1(g_1, \xi)$ as a function of $g_1$ and other randomness $\xi$ independent of $g_1$. Now we have for every $a \in [0, \delta]$,

$$g_1 + \delta_1 \in [b_1 + a, b_1 - a] \implies g_1 \in [b_1 - 2\delta, b_1 + 2\delta]$$

Thus,

$$\mathbf{Pr}[g_1 + \delta_1 \in [b_1 + a, b_1 - a]] = \int_{x\in[-2\delta,+2\delta],\xi} \mathbb{1}_{\delta_1(b_1+x,\xi)+x\in[-a,a]} p_{g_1}(b_1 + x) p_\xi(\xi) dx d\xi$$
$$\overset{①}{=} \delta \int_{s\in[-2,2],\xi} \mathbb{1}_{\delta_1(b_1+s\delta,\xi)+s\delta\in[-a,a]} p_{g_1}(b_1 + s\delta) p_\xi(\xi) ds d\xi$$

$$\overset{②}{=} \delta\left(1 \pm \delta \mathsf{poly}(k_1)\right) p_{g_1}(b_1) \cdot \int_{\xi} \underbrace{\left(\int_{s\in[-2,2]} \mathbb{1}_{\delta_1(b_1+s\delta,\xi)+s\delta\in[-a,a]} ds\right)}_{♣} p_{\xi}(\xi) d\xi$$

Above, ① is simply by changing variable $x = s\delta$, and ② uses $b_1 \leq O(\sqrt{\log m_1})$ which implies $p_{g_1}(b_1 + s\delta) = (1 \pm O(\log m_1)\delta) p_{g_1}(b_1)$.

Next, we wish to bound ♣ for a fixed $\xi$.

Let $s^*$ be any $s \in [-2, 2]$ such that $\delta_1(b_1 + s\delta, \xi) + s\delta \in [-a, a]$. (If there is no such $s^*$ then ♣ would be zero.) By the condition of the claim:

- We have for any $s > s^* + \frac{2a}{\delta}$ or $s < s^* - \frac{2a}{\delta}$, it satisfies $\delta_1(b_1 + s\delta, \xi) + s\delta \notin [-a, a]$.
- For the the remaining $s \in [s^* - \frac{2a}{\delta}, s^* + \frac{2a}{\delta}]$, it satisfies

$$\delta_1(b_1 + s\delta, \xi) = \delta_1(b_1 + s^*\delta, \xi) \pm Err_\xi \qquad \text{for } Err_\xi \stackrel{\text{def}}{=} \delta a \mathsf{poly}(k_1)$$

  Note the event $\mathbb{1}_{\delta_1(b_1+s\delta,\xi)+s\delta\in[-a,a]} \neq \mathbb{1}_{\delta_1(b_1+s^*\delta,\xi)+s\delta\in[-a,a]}$ happens only when $\delta_1(b_1 + s^*\delta, \xi) + s\delta \in [a - Err_\xi, a + Err_\xi] \cup [-a - Err_\xi, -a + Err_\xi]$, but the probability for this to happen (over the randomness of $s$) is at most $O(\frac{Err_\xi}{\delta})$.

Therefore, we can derive that

$$♣ = \int_{s\in[-2,2]} \mathbb{1}_{\delta_1(b_1+s\delta,\xi)+s\delta\in[-a,a]} ds = \int_{s\in[-2,2]} \mathbb{1}_{\delta_1(b_1+s^*\delta,\xi)+s\delta\in[-a,a]} ds \pm O(\frac{Err_\xi}{\delta}) = \frac{2a}{\delta} \pm a\mathsf{poly}(k_1)$$

Putting this back we have just derived

$$\mathbf{Pr}[g_1 + \delta_1 \in [b_1 + a, b_1 - a]] = 2a \cdot p_{g_1}(b_1) \cdot (1 \pm \delta \mathsf{poly}(k_1))$$

This proves that $p(g_1 = \delta_1 + b_1) = p(g_1 = b_1)(1 \pm \mathsf{poly}(k_1)\delta)$. It finishes the proof of Proposition F.24. □

### F.6.5 Sensitivity of Joint Gaussian

**Proposition F.25** (Probability to inner product)**.** *Let $g_1, g_2, g_3, g_4, g_5$ be dependent standard Gaussian random variable in $\mathbb{R}$ (the joint distribution is also Gaussian), $b_1, b_2, b_3, b_4, b_5$ be four positive values such that for $\delta \in [1/\mathsf{poly}(d), m^{-0.1}]$. Suppose it satisfies*

$$\mathbf{Pr}[g_1 > b_1], \mathbf{Pr}[g_2 > b_2] \in \Big[\frac{1}{\mathsf{poly}(k_1)m_1}, \frac{\mathsf{poly}(k_1)}{m_1}\Big], \quad \mathbf{Pr}[g_1 > b_1, g_2 > b_2] \in \Big[\frac{1}{d^{\log d}}, \frac{\mathsf{poly}(k_1)}{m_1^2}\Big]$$

*Then the following holds:*

*(a) For every $\delta', \delta'' \in [-\delta, \delta]$:*

$$\mathbf{Pr}[g_1 > b_1 + \delta', g_2 > b_2 + \delta''] = (1 \pm \delta \mathsf{poly}(k_1)) \mathbf{Pr}[g_1 > b_1, g_2 > b_2] \pm \frac{1}{\mathsf{poly}(d)}$$

*(b) For every $\delta'' \in [-\delta, \delta]$, if $|g_1| \leq O(\sqrt{\log m_1})$, then*

$$\mathbf{Pr}\left[g_2 > b_2 + \delta'' \mid g_1\right] = (1 \pm \delta \mathsf{poly}(k_1)) \mathbf{Pr}\left[g_2 > b_2 \mid g_1\right]$$

*(c) For every $\delta' \in [0, \delta]$,*

$$\mathbf{Pr}\left[g_2 \in [b_2 - \delta', b_2 + \delta'], g_1 > b_1\right] \leq \delta' \mathsf{poly}(k_1) \mathbf{Pr}\left[g_2 > b_2, g_1 > b_1\right] + \frac{1}{\mathsf{poly}(d)}$$

(d) *Suppose we also have* $\mathbf{Pr}[g_3 > b_1, g_4 > b_2] \in \big[\frac{1}{d^{\log d}}, \frac{\mathsf{poly}(k_1)}{m_1^2}\big]$*, then*

$$\begin{aligned}
&\mathbb{E}[g_3 g_4] - \mathbb{E}[g_1 g_2] \in [0, \delta \mathsf{poly}(k_1)] \\
\iff &\mathbf{Pr}[g_3 > b_1, g_4 > b_2] = (1 + \delta \mathsf{poly}(k_1))\, \mathbf{Pr}[g_1 > b_1, g_2 > b_2] \pm \frac{1}{\mathsf{poly}(d)} \\
\iff &\mathbb{E}[\mathsf{ReLU}'(g_3 - b_1)\mathsf{ReLU}(g_4 - b_2)] = (1 + \delta \mathsf{poly}(k_1))\, \mathbb{E}[\mathsf{ReLU}'(g_1 - b_1)\mathsf{ReLU}(g_2 - b_2)] \pm \frac{1}{\mathsf{poly}(d)} \\
\iff &\mathbb{E}[\mathsf{ReLU}(g_3 - b_1)\mathsf{ReLU}(g_4 - b_2)] = (1 + \delta \mathsf{poly}(k_1))\, \mathbb{E}[\mathsf{ReLU}(g_1 - b_1)\mathsf{ReLU}(g_2 - b_2)] \pm \frac{1}{\mathsf{poly}(d)}
\end{aligned}$$

(e) *Moreover, if in addition*

$$\mathbf{Pr}[g_5 > b_5] \in \big[\frac{1}{\mathsf{poly}(k_1) m_1}, \frac{\mathsf{poly}(k_1)}{m_1}\big], \quad \mathbf{Pr}[g_1 > b_1, g_5 > b_5] \in \big[\frac{1}{d^{\log d}}, \frac{\mathsf{poly}(k_1)}{m_1^2}\big]$$

$$\mathbf{Pr}[g_2 > b_2, g_5 > b_5] \in \big[\frac{1}{d^{\log d}}, \frac{\mathsf{poly}(k_1)}{m_1^2}\big]$$

*we have for every* $\delta' \in [-\delta, \delta]$*:*

$$\mathbf{Pr}[g_1 > b_1, g_2 > b_2 \mid g_5 = b_5 + \delta'] \le \mathbf{Pr}[g_1 > b_1, g_2 > b_2]\mathsf{poly}(k_1) + \frac{1}{\mathsf{poly}(d)}$$

*and integrating over* $g_5 \in [b_5 - \delta, b_5 + \delta]$ *we also have*

$$\mathbf{Pr}[g_1 > b_1, g_2 > b_2, |g_5 - b_5| \le \delta] \le \delta \mathsf{poly}(k_1)\, \mathbf{Pr}[g_1 > b_1, g_2 > b_2]\, \mathbf{Pr}[g_5 > b_5] + \frac{1}{\mathsf{poly}(d)}$$

*Proof of Proposition F.25.*

(a) Let us prove Proposition F.25a. Let us write $g_2 = \alpha g_1 + \sqrt{1-\alpha^2} g$ for $g \sim \mathcal{N}(0,1)$ being independent of $g_1$. Proposition F.22 tells us $\alpha \le O(\frac{\log k_1}{\log m_1})$; using $\mathbf{Pr}[g_1 > b_1, g_2 > b_2] \ge \frac{1}{d^{\log d}}$ one can easily derive $\alpha \ge -1 + \Omega(\frac{1}{\log d})$. Note

$$g_2 > b_2 + \delta'' \iff g > \frac{b_2 - \alpha g_1 + \delta''}{\sqrt{1-\alpha^2}}$$

We have w.p. $\ge 1 - \frac{1}{\mathsf{poly}(d)}$ it satisfies $|g_1| \le O(\sqrt{\log m_1})$. When this happens,

$$\frac{b_2 - \alpha g_1}{\sqrt{1-\alpha^2}} \ge b_2 - O(\frac{\log k_1}{\log m_1}) \cdot \sqrt{\log m_1} \ge \Omega(\sqrt{\log m_1}) \quad \text{and} \quad \frac{\delta''}{\sqrt{1-\alpha^2}} \le O(\delta \log m_1)$$

and thus

$$\mathbf{Pr}\left[g_2 > b_2 + \delta'' \mid |g_1| \le O(\sqrt{\log m_1})\right] = (1 \pm \delta \mathsf{poly}(k_1))\, \mathbf{Pr}\left[g_2 > b_2 \mid |g_1| \le O(\sqrt{\log m_1})\right] .$$

Integrating over $g_1 > b_1 + \delta'$, we have

$$\mathbf{Pr}[g_1 > b_1 + \delta', g_2 > b_2 + \delta''] = (1 \pm \delta \mathsf{poly}(k_1))\, \mathbf{Pr}[g_1 > b_1 + \delta', g_2 > b_2] \pm \frac{1}{\mathsf{poly}(d)}$$

Applying this derivation again to get ride of $\delta'$ finishes the proof of Proposition F.25a.

(b) The above derivation of Proposition F.25a also proves Proposition F.25b.

(c) The same derivation of Proposition F.25a also implies, for the conditional probability density function, for any $|\delta''| \le \delta'$,

$$p\left[g_2 = b_2 + \delta'' \mid |g_1| \le O(\sqrt{\log m_1})\right] = (1 \pm \delta' \mathsf{poly}(k_1)) \cdot p\left[g_2 = b_2 \mid |g_1| \le O(\sqrt{\log m_1})\right] .$$

After integration we have

$$\mathbf{Pr}\left[g_2 \in [b_2-\delta', b_2+\delta'] \mid |g_1| \le O(\sqrt{\log m_1})\right] \le O(\delta') \times p\left[g_2 = b_2 \mid |g_1| \le O(\sqrt{\log m_1})\right]$$
$$\le \delta' \mathsf{poly}(k_1)\, \mathbf{Pr}\left[g_2 > b_2 \mid |g_1| \le O(\sqrt{\log m_1})\right] \ .$$

Integrating over all $g_1 > b_1$ finishes the proof.

(d) Again let us write $g_2 = \alpha g_1 + \sqrt{1-\alpha^2} g$ for $g \sim \mathcal{N}(0,1)$ being independent of $g_1$. By the derivations above we already know $\alpha = \mathbb{E}[g_1 g_2] \in [-1+o(1), o(1)]$. Let us couple the randomness by assuming $g_3 = g_1$, and assume $\mathbb{E}[g_3 g_4] = \alpha + \tau$.

There are three " $\Longleftrightarrow$ " to prove and let us first focus on the first one. We have

$$\mathbf{Pr}\left[g_2 > b_2 \mid |g_1| \le O(\sqrt{\log m_1})\right] = \mathop{\mathbf{Pr}}_{g\sim\mathcal{N}(0,1)}\left[g > \frac{b_2 - \alpha g_1}{\sqrt{1-\alpha^2}}\right]$$
$$\mathbf{Pr}\left[g_4 > b_2 \mid |g_1| \le O(\sqrt{\log m_1})\right] = \mathop{\mathbf{Pr}}_{g\sim\mathcal{N}(0,1)}\left[g > \frac{b_2 - (\alpha+\tau) g_1}{\sqrt{1-(\alpha+\tau)^2}}\right]$$

By Taylor expansion, we have

$$\frac{b_2 - (\alpha+\tau) g_1}{\sqrt{1-(\alpha+\tau)^2}} = \frac{b_2 - \alpha g_1}{\sqrt{1-\alpha^2}} + \frac{\alpha b_2 - g_1}{(1-\alpha)^{3/2}}\tau \pm O(\tau^2 \mathsf{poly}(k_1))$$

Note since $\alpha \in [-1+o(1), o(1)]$, we have $\frac{b_2-\alpha g_1}{\sqrt{1-\alpha^2}} \ge b_2 - o(g_1) \ge \sqrt{2\log m_1}(1-o(1))$; at the same time, $\alpha b_2 \le o(\sqrt{\log m_1})$ and $g_1 \ge b_1 = \Omega(\sqrt{\log m_1})$ imply $\frac{\alpha b_2 - g_1}{(1-\alpha)^{3/2}} \in \left[-\Omega(\sqrt{\log m_1}), -\mathsf{poly}(k_1)\Omega(\sqrt{\log m_1})\right]$. This further implies, conditioning on $g_1 \in [b_1, O(\sqrt{\log m_1})]$, it satisfies

$$\mathbf{Pr}\left[g > \frac{b_2 - (\alpha+\tau) g_1}{\sqrt{1-(\alpha+\tau)^2}} \mid g_1 \in \left[b_1, O(\sqrt{\log m_1})\right]\right]$$
$$= (1+\tau\theta)\, \mathbf{Pr}\left[g > \frac{b_2 - \alpha g_1}{\sqrt{1-\alpha^2}} \mid g_1 \in \left[b_1, O(\sqrt{\log m_1})\right]\right] \tag{F.22}$$

for some $\theta \in [\frac{1}{\mathsf{poly}(k_1)}, \mathsf{poly}(k_1)]$. Therefore,

$$\mathbb{E}[g_3 g_4] - \mathbb{E}[g_1 g_2] \in [0, \tau\mathsf{poly}(k_1)]$$
$$\Longleftrightarrow \mathbf{Pr}[g_3 > b_1, g_4 > b_2] = (1+\tau\mathsf{poly}(k_1))\, \mathbf{Pr}[g_1 > b_1, g_2 > b_2] \pm \frac{1}{\mathsf{poly}(d)}$$

As for the second " $\Longleftrightarrow$ ", one can use a similar derivation but replacing (F.22) with the following:

$$0 \le \mathbb{E}\left[\mathsf{ReLU}\Big(g - \frac{b_2 - (\alpha+\tau) g_1}{\sqrt{1-(\alpha+\tau)^2}}\Big) - \mathsf{ReLU}\Big(g - \frac{b_2 - \alpha g_1}{\sqrt{1-\alpha^2}}\Big) \,\Big|\, g_1 \in \left[b_1, O(\sqrt{\log m_1})\right]\right]$$
$$\le \tau\mathsf{poly}(k_1)\, \mathbf{Pr}\left[g > \frac{b_2 - \alpha g_1}{\sqrt{1-\alpha^2}} \,\Big|\, g_1 \in \left[b_1, O(\sqrt{\log m_1})\right]\right]$$
$$\le \tau\mathsf{poly}(k_1)\, \mathbb{E}\left[\mathsf{ReLU}\Big(g - \frac{b_2 - \alpha g_1}{\sqrt{1-\alpha^2}}\Big) \,\Big|\, g_1 \in \left[b_1, O(\sqrt{\log m_1})\right]\right]$$

As for the third " $\Longleftrightarrow$ ", one can multiply the above calculation of expectation by $\mathsf{ReLU}(g_1 - b_1)$.

(e) Let us prove Proposition F.25e. Let us write $g_1$, $g_2$ as: $g_2 = \alpha g_5 + \sqrt{1-\alpha^2} g$, where $g$ is a standard Gaussian independent of $g_5$. $g_1 = \gamma g_5 + \beta g + \sqrt{1-\gamma^2-\beta^2} h$, where $h$ is a standard Gaussian independent of $g, g_5$.

Now, by Proposition F.22, we know that

$$\alpha, \gamma, \alpha\gamma + \sqrt{1-\alpha^2}\beta \leq O\left(\frac{\log k_1}{\log m_1}\right) \tag{F.23}$$

We consider two cases:

- $\alpha < 0, \gamma < 0$, in this case, we know that $\sqrt{1-\alpha^2}\beta \leq O\left(\frac{\log k_1}{\log m_1}\right)$. Moreover, by the condition of this claim, we know that $b_5 = \Theta(\sqrt{\log m_1})$, thus:

$$g_5 = b_5 + \delta', g_1 > b_1, g_2 > b_2 \implies \sqrt{1-\alpha^2}g > b_2, \beta g + \sqrt{1-\gamma^2-\beta^2}h > b_1 \tag{F.24}$$

  Now, a simple calculation of standard Gaussian shows that

$$\mathbf{Pr}[g_1 > b_1, g_2 > b_2 \mid g_5 = b_5 + \delta'] \leq \mathbf{Pr}[g_1 > b_1, g_2 > b_2]\mathsf{poly}(k_1) + \frac{1}{d^{\log d}} \tag{F.25}$$

- At least one of $\alpha, \gamma \geq 0$, without loss of generality, we can assume that $\alpha > 0$. In this case, we know that $|\alpha| \leq O\left(\frac{\log k_1}{\log m_1}\right)$, which again implies that $\sqrt{1-\alpha^2}\beta \leq O\left(\frac{\log k_1}{\log m_1}\right)$, hence $\beta \leq O\left(\frac{\log k_1}{\log m_1}\right)$. Thus, we have: (notice that $b_5 = \Theta(\sqrt{\log m_1})$)

$$g_5 = b_5 + \delta', g_1 > b_1, g_2 > b_2 \tag{F.26}$$

$$\implies \sqrt{1-\alpha^2}g > b_2 - O\left(\frac{\log k_1}{\sqrt{\log m_1}}\right), \text{ and } \beta g + \sqrt{1-\gamma^2-\beta^2}h > b_1 - O\left(\frac{\log k_1}{\sqrt{\log m_1}}\right) \tag{F.27}$$

  Now, we have that for sufficiently large $\nu = O(1)$: w.p. $1 - \frac{1}{\mathsf{poly}(d)}$ we have: $|g_2| \leq \nu\sqrt{\log m_1}$, notice that $\gamma \leq O\left(\frac{\log k_1}{\log m_1}\right)$, so we have:

$$\mathbf{Pr}[g_1 > b_1, g_2 > b_2] = \mathbf{Pr}\left[\gamma g_5 + \beta g + \sqrt{1-\gamma^2-\beta^2}h > b_1, \alpha g_5 + \sqrt{1-\alpha^2}g > b_2\right] \tag{F.28}$$

$$\geq \mathbf{Pr}\left[\gamma g_5 + \beta g + \sqrt{1-\gamma^2-\beta^2}h > b_1, \sqrt{1-\alpha^2}g > b_2 - \alpha\nu\sqrt{\log m_1}\right] + \frac{1}{\mathsf{poly}(d)} \tag{F.29}$$

$$\geq \frac{1}{2}\mathbf{Pr}\left[\gamma g_5 + \beta g + \sqrt{1-\gamma^2-\beta^2}h > b_1, \sqrt{1-\alpha^2}g > b_2 - \alpha\nu\sqrt{\log m_1} \mid g_5 \geq 0\right] + \frac{1}{\mathsf{poly}(d)} \tag{F.30}$$

$$\geq \frac{1}{2}\mathbf{Pr}\left[\beta g + \sqrt{1-\gamma^2-\beta^2}h > b_1 - \nu\frac{\log k_1}{\sqrt{\log m_1}}, \sqrt{1-\alpha^2}g > b_2 - \alpha\nu\sqrt{\log m_1}\right] + \frac{1}{\mathsf{poly}(d)} \tag{F.31}$$

  Together, we can easily conclude that:

$$\mathbf{Pr}[g_1 > b_1, g_2 > b_2 \mid g_5 = b_5 + \delta'] \leq \mathbf{Pr}[g_1 > b_1, g_2 > b_2]\mathsf{poly}(k_1) + \frac{1}{\mathsf{poly}(d)} \tag{F.32}$$

□

We have the following bound on the sensitivity of dependent Gaussian random variables. (One can compare this to Proposition F.24 on a similar result for a single Gaussian.)

**Proposition F.26** (Inside out)**.** *Let $g_1, g_2$ be two dependent standard Gaussian random variable in $\mathbb{R}$, $b_1, b_2$ be two positive values such that*

$$\mathbf{Pr}[g_1 > b_1], \mathbf{Pr}[g_2 > b_2] \in [\frac{1}{\mathsf{poly}(k_1) m_1}, \frac{\mathsf{poly}(k_1)}{m_1}], \quad \mathbf{Pr}[g_1 > b_1, g_2 > b_2] \in [\frac{1}{d^{\log d}}, \frac{\mathsf{poly}(k_1)}{m_1^2}]$$

*We then have: for every value $\delta \in [1/d^{\log d}, m^{-0.1}]$, every variable $\delta_1, \delta_2 \in [-\delta, \delta]$ (whose values can depend on $g_1, g_2$) such that*

$$\left|\frac{\partial \delta_1}{\partial g_1}\right|, \left|\frac{\partial \delta_1}{\partial g_2}\right|, \left|\frac{\partial \delta_2}{\partial g_1}\right|, \left|\frac{\partial \delta_2}{\partial g_2}\right| \leq \delta \mathsf{poly}(k_1) \quad \text{for every } |g_1|, |g_2| \leq O(\sqrt{\log m_1})$$

*We must have that*

*(a) Cumulative probability density is Lipschitz:*

$$\mathbf{Pr}[g_1 \geq b_1 + \delta_1 \mid g_2 = b_2 + \delta_2] = \mathbf{Pr}[g_1 \geq b_1 \mid g_2 = b_2](1 \pm \mathsf{poly}(k_1)\delta) \pm \frac{1}{\mathsf{poly}(d)} \tag{F.33}$$

*(b) Expectation is Lipschitz:*

$$\mathbb{E}[\widetilde{\mathsf{ReLU}}(g_2 + \delta_2 - b_2) \mid g_1 = b_1 - \delta_1] = \mathbb{E}[\widetilde{\mathsf{ReLU}}(g_2 - b_2) \mid g_1 = b_1](1 \pm \mathsf{poly}(k_1)\delta) \pm \frac{1}{\mathsf{poly}(d)}$$

*Proof of Proposition F.26.* We first prove Proposition F.26a.

By Proposition F.22, we can write $g_1 = \alpha g_2 + \sqrt{1-\alpha^2} g$ where $g \sim \mathcal{N}(0,1)$ is independent of $g_2$, for $\alpha \in [-1, o(1)]$. Using $\mathbf{Pr}[g_1 > b_1, g_2 > b_2] \geq \frac{1}{d^{\log d}}$ one can easily derive $\alpha \geq -1 + \Omega(\frac{1}{\sqrt{\log d}})$ which implies $\sqrt{1-\alpha^2} \geq \Omega(\frac{1}{\log d})$.

Let us write $\delta_1 = \delta_1(g, g_2, \xi)$ where $\xi$ is independent of $g, g_2$. We have for $\nu = \Theta(\log d)$,

$$g_1 + \delta_1 \in [b_1 + a, b_1 - a] \implies g \in [b' - 2\nu\delta, b' + 2\nu\delta] \quad \text{where} \quad b' = \frac{b_1 - \alpha g_2}{\sqrt{1-\alpha^2}}$$

Thus, given any $g_2$ with $|g_2| \leq O(\sqrt{\log m_1})$,

$$\begin{aligned}
&\mathbf{Pr}[g_1 + \delta_1 \in [b_1 + a, b_1 - a] \mid g_2] \\
&= \int_{y,\xi} \mathbb{1}_{\delta_1(y, g_2, \xi) + \sqrt{1-\alpha^2} y + \alpha g_2 \in [b_1 - a, b_1 + a]} p_g(y) p_\xi(\xi) dx d\xi \\
&= \int_{x \in [-2\nu\delta, +2\nu\delta], \xi} \mathbb{1}_{\delta_1(b' + x, g_2, \xi) + \sqrt{1-\alpha^2} x \in [-a, a]} p_g(b' + x) p_\xi(\xi) dx d\xi \\
&\overset{①}{=} \nu\delta \int_{s \in [-2,2], \xi} \mathbb{1}_{\delta_1(b' + s\nu\delta, g_2, \xi) + \sqrt{1-\alpha^2} s\nu\delta \in [-a, a]} p_g(b' + s\nu\delta) p_\xi(\xi) ds d\xi \\
&\overset{②}{=} \nu\delta \left(1 \pm \delta \mathsf{poly}(k_1)\right) p_g(b') \cdot \int_\xi \underbrace{\left( \int_{s \in [-2,2]} \mathbb{1}_{\delta_1(b' + s\nu\delta, g_2, \xi) + \sqrt{1-\alpha^2} s\nu\delta \in [-a, a]} ds \right)}_{\clubsuit} p_\xi(\xi) d\xi
\end{aligned}$$

Above, ① again is by changing variable $x = s\nu\delta$, and ② uses $b' \geq (1 - o(1))\sqrt{2 \log m_1}$ which implies $p_g(b' + s\nu\delta) = (1 \pm O(\log m_1)\nu\delta) p_g(b')$. A similar derivation now gives

$$\clubsuit = \int_{s \in [-2,2]} \mathbb{1}_{\delta_1(b' + s^*\nu\delta, g_2, \xi) + \sqrt{1-\alpha^2} s\nu\delta \in [-a, a]} ds \pm \frac{\delta a \mathsf{poly}(k_1)}{\delta} = \frac{2a}{\delta\nu\sqrt{1-\alpha^2}} \pm a\mathsf{poly}(k_1)$$

Putting this back we have just derived

$$\mathbf{Pr}[g_1 + \delta_1 \in [b_1 + a, b_1 - a] \mid g_2] = \frac{2a}{\sqrt{1-\alpha^2}} \cdot p_g(b') \cdot (1 \pm \delta \mathsf{poly}(k_1))$$

$$= \frac{2a}{\sqrt{1-\alpha^2}} \cdot p_g\Big(\frac{b_1 - \alpha g_2}{\sqrt{1-\alpha^2}}\Big) \cdot (1 \pm \delta \mathsf{poly}(k_1))$$

Letting $a \to 0$ we have for every $|g_2| \le O(\sqrt{\log m_1})$:

$$p[g_1 = b_1 - \delta_1 \mid g_2] = (1 \pm \delta \mathsf{poly}(k_1)) p[g_1 = b_1 \mid g_2] \tag{F.34}$$

Integrating over $g_2 \ge b_2 - \delta$ we have

$$p[g_1 = b_1 - \delta_1 \mid g_2 \ge b_2 - \delta] = (1 \pm \delta \mathsf{poly}(k_1)) p[g_1 = b_1 \mid g_2 \ge b_2 - \delta] \pm \frac{1}{\mathsf{poly}(d)}$$

Note

$$\begin{aligned}
\mathbf{Pr}[g_2 + \delta_2 \ge b_2 \mid g_1 = b_1 - \delta_1] &\le \mathbf{Pr}[g_2 \ge b_2 - \delta \mid g_1 = b_1 - \delta_1] \\
&= \frac{p[g_1 = b_1 - \delta_1 \mid g_2 \ge b_2 - \delta] \, \mathbf{Pr}[g_2 \ge b_2 - \delta]}{p[g_1 = b_1 - \delta_1]} \\
&\le \frac{(1 + \delta \mathsf{poly}(k_1)) p[g_1 = b_1 \mid g_2 \ge b_2 - \delta] \, \mathbf{Pr}[g_2 \ge b_2 - \delta]}{(1 - \delta \mathsf{poly}(k_1)) p[g_1 = b_1]} \pm \frac{1}{\mathsf{poly}(d)} \\
&\le (1 + \delta \mathsf{poly}(k_1)) \frac{p[g_1 = b_1 \mid g_2 \ge b_2 - \delta] \, \mathbf{Pr}[g_2 \ge b_2 - \delta]}{p[g_1 = b_1]} \pm \frac{1}{\mathsf{poly}(d)} \\
&= (1 + \delta \mathsf{poly}(k_1)) \, \mathbf{Pr}[g_2 \ge b_2 - \delta \mid g_1 = b_1] \pm \frac{1}{\mathsf{poly}(d)} \\
&\le (1 + \delta \mathsf{poly}(k_1)) \, \mathbf{Pr}[g_2 \ge b_2 \mid g_1 = b_1] \pm \frac{1}{\mathsf{poly}(d)}
\end{aligned}$$

This finishes the proof of Proposition F.26a. We next look at Proposition F.26b. For a sufficiently large constant $c > 100$, we have

$$\begin{aligned}
&\mathbb{E}[\widetilde{\mathsf{ReLU}}(g_2 + \delta_2 - b_2) \mid g_1 = b_1 - \delta_1] \\
&= \int_{t \ge 0}^{c\sqrt{\log m_1}} \widetilde{\mathsf{ReLU}}'(t) \, \mathbf{Pr}\Big[g_2 + \delta_2 - b_2 \in \big[t, c\sqrt{\log m_1}\big] \mid g_1 = b_1 - \delta_1\Big] \, dt \pm \frac{1}{\mathsf{poly}(d)} \\
&\le \int_{t \ge 0}^{c\sqrt{\log m_1}} \widetilde{\mathsf{ReLU}}'(t) \, \mathbf{Pr}\Big[g_2 - b_2 \in \big[t - \delta, c\sqrt{\log m_1}\big] \mid g_1 = b_1 - \delta_1\Big] \, dt + \frac{1}{\mathsf{poly}(d)} \\
&\overset{①}{=} \int_{t \ge 0}^{c\sqrt{\log m_1}} \widetilde{\mathsf{ReLU}}'(t) \frac{p\Big[g_1 = b_1 - \delta_1 | g_2 - b_2 \in \big[t - \delta, c\sqrt{\log m_1}\big]\Big] \, \mathbf{Pr}[g_2 - b_2 \in \big[t - \delta, c\sqrt{\log m_1}\big]]}{p[g_1 = b_1 - \delta_1]} dt + \frac{1}{\mathsf{poly}(d)} \\
&\overset{②}{\le} (1 + \delta \mathsf{poly}(k_1)) \int_{t \ge 0}^{c\sqrt{\log m_1}} \widetilde{\mathsf{ReLU}}'(t) \frac{p\Big[g_1 = b_1 | g_2 - b_2 \in \big[t - \delta, c\sqrt{\log m_1}\big]\Big] \, \mathbf{Pr}[g_2 - b_2 \in \big[t - \delta, c\sqrt{\log m_1}\big]]}{p[g_1 = b_1]} dt + \frac{1}{\mathsf{poly}(d)} \\
&\overset{③}{=} (1 + \delta \mathsf{poly}(k_1)) \int_{t \ge 0}^{c\sqrt{\log m_1}} \widetilde{\mathsf{ReLU}}'(t) \, \mathbf{Pr}\Big[g_2 - b_2 \in \big[t - \delta, c\sqrt{\log m_1}\big] \mid g_1 = b_1\Big] \, dt + \frac{1}{\mathsf{poly}(d)} \\
&= (1 + \delta \mathsf{poly}(k_1)) \int_{t \ge 0}^{c\sqrt{\log m_1}} \widetilde{\mathsf{ReLU}}'(t) \, \mathbf{Pr}\left[g_2 - b_2 \ge t - \delta \mid g_1 = b_1\right] dt + \frac{1}{\mathsf{poly}(d)} \\
&\overset{④}{\le} (1 + \delta \mathsf{poly}(k_1)) \int_{t \ge 0}^{c\sqrt{\log m_1}} \widetilde{\mathsf{ReLU}}'(t) \, \mathbf{Pr}\left[g_2 - b_2 \ge t \mid g_1 = b_1\right] dt + \frac{1}{\mathsf{poly}(d)} \\
&= (1 + \delta \mathsf{poly}(k_1)) \, \mathbb{E}[\widetilde{\mathsf{ReLU}}(g_2 - b_2) \mid g_1 = b_1] + \frac{1}{\mathsf{poly}(d)}
\end{aligned}$$

Above, ① uses Bayes' theorem; ② is by integrating (F.34) for the numerator and applying Proposition F.24 for the denominator; ③ is by Bayes' theorem again; and ④ uses Proposition F.25b.

Note the other direction can be proved similarly. This finishes the proof of Proposition F.26b.

□

We have the following Claim:

**Proposition F.27.** *In the same setting as Proposition F.26, we have:*

*(a)*

$$\mathbb{E}\left[\widetilde{\mathsf{ReLU}}(g_1 - b_1 + \delta_1 - b)\widetilde{\mathsf{ReLU}}'(g_2 - b_2 + \delta_1 - b)g_1\right]$$
$$= \mathbb{E}\left[\widetilde{\mathsf{ReLU}}(g_1 - b_1)\widetilde{\mathsf{ReLU}}'(g_2 - b_2)g_1\right](1 \pm \mathsf{poly}(k_1)b) \pm \frac{1}{\mathsf{poly}(d)}$$

*(b)*

$$\mathbb{E}\left[\widetilde{\mathsf{ReLU}}(g_1 - b_1 + \delta_1 - b)\widetilde{\mathsf{ReLU}}(g_2 - b_2 + \delta_1 - b)\right]$$
$$= \mathbb{E}\left[\widetilde{\mathsf{ReLU}}(g_1 - b_1)\widetilde{\mathsf{ReLU}}(g_2 - b_2)\right](1 \pm \mathsf{poly}(k_1)b) \pm \frac{1}{\mathsf{poly}(d)}$$

*(c)*

$$\mathbb{E}\left[\widetilde{\mathsf{ReLU}}(g_1 - b_1 + \delta_1 - b)\widetilde{\mathsf{ReLU}}''(g_2 - b_2 + \delta_1 - b)\right]$$
$$= \mathbb{E}\left[\widetilde{\mathsf{ReLU}}(g_1 - b_1)\widetilde{\mathsf{ReLU}}''(g_2 - b_2)\right](1 \pm \mathsf{poly}(k_1)b) \pm \frac{1}{\mathsf{poly}(d)}$$

*(d)*

$$\mathbb{E}\left[\widetilde{\mathsf{ReLU}}'(g_1 - b_1 + \delta_1 - b)\widetilde{\mathsf{ReLU}}'(g_2 - b_2 + \delta_1 - b)\right]$$
$$= \mathbb{E}\left[\widetilde{\mathsf{ReLU}}'(g_1 - b_1)\widetilde{\mathsf{ReLU}}'(g_2 - b_2)\right](1 \pm \mathsf{poly}(k_1)b) \pm \frac{1}{\mathsf{poly}(d)}$$

*Proof of Proposition F.27.*

(a) Let us look at Proposition F.27a. Using $\delta_1 \le b$, and the monotonicity of $\widetilde{\mathsf{ReLU}}'(x)$, we know

$$\widetilde{\mathsf{ReLU}}(g_1 - b_1 + \delta_1 - b) = \widetilde{\mathsf{ReLU}}(g_1 - b_1) \pm b\widetilde{\mathsf{ReLU}}'(g_1 - b_1)$$

Using this and w.h.p. $0 \le g_1 \le \mathsf{poly}(k_1)$, we calculate that

$$\begin{aligned}
&\mathbb{E}\left[\widetilde{\mathsf{ReLU}}(g_1 - b_1 + \delta_1 - b)\widetilde{\mathsf{ReLU}}'(g_2 - b_2 + \delta_1 - b)g_1\right] \\
&= \mathbb{E}\left[\widetilde{\mathsf{ReLU}}(g_1 - b_1)\widetilde{\mathsf{ReLU}}'(g_2 - b_2 + \delta_1 - b)g_1\right] \\
&\quad \pm b\mathsf{poly}(k_1)\,\mathbb{E}\left[\widetilde{\mathsf{ReLU}}'(g_1 - b_1)\widetilde{\mathsf{ReLU}}'(g_2 - b_2 + \delta_1 - b)\right] \pm \frac{1}{d^{\omega(1)}} \\
&= \mathbb{E}\left[\widetilde{\mathsf{ReLU}}(g_1 - b_1)\widetilde{\mathsf{ReLU}}'(g_2 - b_2 + \delta_1 - b)g_1\right] \pm b\mathsf{poly}(k_1)\,\mathbf{Pr}[g_1 \ge b_1, g_2 \ge b_2] \pm \frac{1}{\mathsf{poly}(d)}
\end{aligned} \tag{F.35}$$

where the last inequality uses Proposition F.25a. At the same time, we have

$$\begin{aligned}
&\left|\mathbb{E}\left[\widetilde{\mathsf{ReLU}}(g_1 - b_1)\left(\widetilde{\mathsf{ReLU}}'(g_2 - b_2 + \delta_1 - b) - \widetilde{\mathsf{ReLU}}'(g_2 - b_2)\right)g_1\right]\right| \\
&\le \mathsf{poly}(k_1)\,\mathbf{Pr}[g_1 \ge b_1, g_2 \in [b_2 - 2b, b_2 + 2b]] + \frac{1}{d^{\omega(1)}} \\
&\le b\mathsf{poly}(k_1)\,\mathbf{Pr}[g_1 \ge b_1, g_2 \ge b_2] + \frac{1}{d^{\omega(1)}}
\end{aligned}$$

where the last inequality uses Proposition F.25c. Finally, notice that

$$\mathbb{E}\left[\widetilde{\mathsf{ReLU}}(g_1 - b_1)\widetilde{\mathsf{ReLU}}'(g_2 - b_2)g_1\right] \geq \frac{1}{\mathsf{poly}(k_1)}\,\mathbb{E}\left[\widetilde{\mathsf{ReLU}}'(g_1 - b_1 - \frac{1}{\mathsf{poly}(k_1)})\widetilde{\mathsf{ReLU}}'(g_2 - b_2)\right]$$
$$\geq \frac{1}{\mathsf{poly}(k_1)}\,\mathbf{Pr}[g_1 \geq b_1, g_2 \geq b_2] - \frac{1}{\mathsf{poly}(d)}$$

where the last inequality uses Proposition F.25a again. Combining everything so far finishes the proof of Proposition F.27a.

(b) The proof of Proposition F.27b is completely analogous to Proposition F.27a: it suffices to prove an analogous version of (F.35) and then apply it twice.

(c) Let us look at Proposition F.27c. Proposition F.26b directly gives

$$\mathbb{E}\left[\widetilde{\mathsf{ReLU}}(g_1 - b_1 + \delta_1 - b)\widetilde{\mathsf{ReLU}}''(g_2 - b_2 + \delta_2 - b)\right]$$
$$= \int_{t\in[0,\zeta]} \widetilde{\mathsf{ReLU}}''(t)\,\mathbb{E}\left[\widetilde{\mathsf{ReLU}}(g_1 - b_1 + \delta_1 - b) \mid g_2 - b_2 + \delta_2 - b = t\right] p[g_2 - b_2 + \delta_2 - b = t]dt$$
$$\overset{①}{=} (1 \pm b\mathsf{poly}(k_1)) \int_{t\in[0,\zeta]} \widetilde{\mathsf{ReLU}}''(t)\,\mathbb{E}\left[\widetilde{\mathsf{ReLU}}(g_1 - b_1) \mid g_2 - b_2 = t\right] p[g_2 - b_2 = 0]dt \pm \frac{1}{\mathsf{poly}(d)}$$
$$= (1 \pm b\mathsf{poly}(k_1))\,\mathbb{E}\left[\widetilde{\mathsf{ReLU}}(g_1 - b_1)\widetilde{\mathsf{ReLU}}''(g_2 - b_2)\right] \pm \frac{1}{\mathsf{poly}(d)}$$

Above, ① uses Proposition F.26b together with Proposition F.24.

(d) To show Proposition F.27d, we have

$$\mathbb{E}\left[\widetilde{\mathsf{ReLU}}'(g_1 - b_1 + \delta_1 - b)\widetilde{\mathsf{ReLU}}'(g_2 - b_2 + \delta_1 - b)\right]$$
$$= \mathbb{E}\left[\widetilde{\mathsf{ReLU}}'(g_1 - b_1)\widetilde{\mathsf{ReLU}}'(g_2 - b_2)\right] \pm 2\left(\mathbf{Pr}[g_1 \geq b_1, |g_2 - b_2| \leq 2b] + \mathbf{Pr}[g_2 \geq b_2, |g_1 - b_1| \leq 2b]\right)$$
$$= \mathbb{E}\left[\widetilde{\mathsf{ReLU}}'(g_1 - b_1)\widetilde{\mathsf{ReLU}}'(g_2 - b_2)\right] \pm b\mathsf{poly}(k_1)\,\mathbf{Pr}[g_1 \geq b_1, g_2 \geq b_2]$$

where the last equality uses Proposition F.25c again. At the same time, we have

$$\mathbb{E}\left[\widetilde{\mathsf{ReLU}}'(g_1 - b_1)\widetilde{\mathsf{ReLU}}'(g_2 - b_2)\right] \geq \Omega(1)\,\mathbf{Pr}[g_1 \geq b_1, g_2 \geq b_2] - \frac{1}{\mathsf{poly}(d)}$$

which uses Proposition F.25a. Combining them finishes the proof of Proposition F.27d.

□

### F.6.6 Functions under Gaussian Variable

Now we use an equality of smoothed function under Gaussian variable:

**Claim F.28.** *Let $g_1, g_2$ be (dependent) standard Gaussian in $\mathbb{R}$. Now, for every third order differentiable function $f : \mathbb{R} \to \mathbb{R}$, for every $\alpha_1, \alpha_2, \beta_1, \beta_2$, we have:*

$$\mathbb{E}\left[f'(\alpha_1 g_1 - \beta_1)f(\alpha_2 g_2 - \beta_2)\alpha_1 g_1\right] + \mathbb{E}\left[f(\alpha_1 g_1 - \beta_1)f'(\alpha_2 g_2 - \beta_2)\alpha_2 g_2\right]$$
$$= \mathbb{E}\left[f''(\alpha_1 g_1 - \beta_1)f(\alpha_2 g_2 - \beta_2)\alpha_1^2\right] + \mathbb{E}\left[f(\alpha_1 g_1 - \beta_1)f''(\alpha_2 g_2 - \beta_2)\alpha_2^2\right]$$
$$+ 2\alpha_1\alpha_2\,\mathbb{E}\left[f'(\alpha_1 g_1 - \beta_1)f'(\alpha_2 g_2 - \beta_2)\right]\mathbb{E}[\langle g_1, g_2\rangle]$$

*Proof of Claim F.28.* Let $g_3, g_4$ be independent copy of $g_1, g_2$, then for every $\eta > 0$,

$$\mathbb{E}[f(\sqrt{1-\eta}\alpha_1 g_1 + \sqrt{\eta}\alpha_1 g_3 - \beta_1)f(\sqrt{1-\eta}\alpha_2 g_2 + \sqrt{\eta}\alpha_2 g_4 - \beta_2)]$$
$$= \mathbb{E}[f(\alpha_1 g_1 - \beta_1)f(\alpha_2 g_2 - \beta_2)]$$

Doing taylor expansion w.r.t. $\eta$ and comparing the term with $\eta$, we complete the proof. $\square$

### F.6.7 Basics for Hessian Update

**Claim F.29.** *Given $n$ matrices $\mathbf{V}_1, \dots, \mathbf{V}_n \in \mathbb{R}^{d \times d'}$ for $d' \geq nd + m$, and matrices $\mathbf{V}_1^\star, \dots, \mathbf{V}_n^\star \in \mathbb{R}^{d \times m}$, for every $\eta \in [0,1]$, there exist matrices $\mathbf{S}_1, \dots, \mathbf{S}_n \in \mathbb{R}^{d \times d'}$ such that*

$$\left(\sqrt{1-\eta}\mathbf{V}_i + \sqrt{\eta}\mathbf{S}_i\right)\left(\sqrt{1-\eta}\mathbf{V}_j + \sqrt{\eta}\mathbf{S}_j\right)^\top = (1-\eta)\mathbf{V}_i\mathbf{V}_j^\top + \eta\mathbf{V}_i^\star(\mathbf{V}_j^\star)^\top$$

*and $\mathbf{V}_i\mathbf{S}_j^\top = \mathbf{0}_{d\times d}$ and $\mathbf{S}_i\mathbf{S}_j^\top = \mathbf{V}_i^\star(\mathbf{V}_j^\star)^\top$.*

*Proof of Claim F.29.* This claim is trivial. Let $\{s_1, \dots, s_m \in \mathbb{R}^{d'}\}$ be an arbitrary set of unit orthonormal vectors that is not in the row span of any $\mathbf{V}_1, \dots, \mathbf{V}_n$ (we can do so because $d' \geq nd + m$). Denote by this row-orthonormal matrix $\mathbf{U} = [s_1, \dots, s_m]^\top \in \mathbb{R}^{m \times d'}$ and $\mathbf{U}^\top \mathbf{V}_i = \mathbf{0}$ for every $i$. Then, we can set $\mathbf{S}_i = \mathbf{V}_i^\star \mathbf{U}$. $\square$

Moreover, we have the following claim which can be found in [4, 7].

**Fact F.30.** *For every $B$-second-order smooth function $f : \mathbb{R}^d \to \mathbb{R}$, every $\varepsilon > 0$, every fixed vectors $x \in \mathbb{R}^d$, suppose for every sufficiently small $\eta > 0$, there exists vector $x_1 \in \mathbb{R}^d$ and a random vector $x_2 \in \mathbb{R}^d$ with $\mathbb{E}[x_2] = 0$ satisfying $\|x_1\|_2 \leq Q_1$, $\mathbb{E}[\|x_2\|_2^2] \leq Q_2$ and*

$$\mathop{\mathbb{E}}_{x_2}[f\left(x + \eta x_1 + \sqrt{\eta} x_2\right)] \leq f(x) - \eta\varepsilon \ .$$

*Then, either $\|\nabla f(x)\| \geq \frac{\varepsilon}{2Q_1}$ or $\lambda_{\min}(\nabla^2 f(x)) \leq -\frac{\varepsilon}{Q_2}$, where $\lambda_{\min}$ is the minimal eigenvalue.*

*Proof of Fact F.30.* We know that

$$\begin{aligned} & f\left(x + \eta x_1 + \sqrt{\eta} x_2\right) \\ &= f(x) + \langle \nabla f(x), \eta x_1 + \sqrt{\eta} x_2 \rangle + \frac{1}{2}\left(\eta x_1 + \sqrt{\eta} x_2\right)^\top \nabla^2 f(x) \left(\eta x_1 + \sqrt{\eta} x_2\right) \pm O(B\eta^{1.5}). \end{aligned}$$

Taking expectation, we know that

$$\mathbb{E}[f\left(x + \sqrt{\eta} x_2\right)] = f(x) + \eta\langle \nabla f(x), x_1\rangle + \eta\frac{1}{2}\mathbb{E}\left[x_2^\top \nabla^2 f(x) x_2\right] \pm O(B\eta^{1.5})$$

Thus, either $\langle \nabla f(x), x_1 \rangle \leq -\varepsilon/2$ or $\mathbb{E}\left[x_2^\top \nabla^2 f(x) x_2\right] \leq -\varepsilon$, which completes the proof. $\square$

# G Learning Other Hidden Layers

In this section, we fix a layer $\ell = 2, 3, \dots, L$ and prove Lemma G.2. We restate it below together with a few corollaries.

Throughout this section, we assume

**Induction G.1.** *There exists some $\delta \in (0, m_\ell^{-0.1}]$ such that for every $j \in [d_\ell]$, for every $p \in [m_\ell]$, $\|[\mathbf{W}_{\ell,j}^\star]_p - [\mathbf{W}_{\ell,j}]_p\|_2 \leq \delta$. We choose $b = \delta \cdot m_\ell^{0.02}$ and fix $\mathbf{W}_{\ell,j}^D = \mathbf{W}_{\ell,j}$.*

*In addition, we layer $\ell - 1$ is sufficiently optimized: it satisfies*

- *$\|\mathbf{W}_{\ell-1,j}^\star - \mathbf{W}_{\ell-1,j}\|_2 \leq \frac{1}{\mathsf{poly}(d/\varepsilon)}$ and*
- *for some column orthonormal matrix $\mathbf{U} \in \mathbb{R}^{m_0 \times m_0'}$, w.h.p.*

$$\mathcal{S}_{\ell-1,j}(z) \leq \mathcal{S}_{\ell-1,j}^\star(\mathbf{U}z) \quad \text{and} \quad \|\mathcal{S}_{\ell-1,j}(z) - \mathcal{S}_{\ell-1,j}^\star(\mathbf{U}z)\|_2 \leq \frac{1}{\mathsf{poly}(d/\varepsilon)} \ .$$

**Lemma G.2** ($D^{(2)}$ final)**.** *Under Induction G.1, right after the gradient descent-ascent process in Algorithm 3, with probability at least $1 - \frac{1}{\mathsf{poly}(d/\varepsilon)}$, it satisfies*

$$\left|[\mathcal{S}^\star_{\ell,j}(\mathbf{U}z)]_r - [\mathcal{S}_{\ell,j}(z)]_r\right| \leq 7b \ .$$

*Now, after we* **increase the bias** $[b_{\ell,j}]_r$ *by $7b$ in the last line of Algorithm 3, then we have*

$$[\mathcal{S}_{\ell,j}(z)]_r \leq [\mathcal{S}^\star_{\ell,j}(\mathbf{U}z)]_r, \quad |[\mathcal{S}_{\ell,j}(z)]_r - [\mathcal{S}^\star_{\ell,j}(\mathbf{U}z)]_r| \leq O(b) \tag{G.1}$$

*As corollaries, we know*

- $\mathbb{E}\left[\mathsf{ReLU}([\mathcal{S}_{\ell,j}]_r)\right] = \mathbb{E}\left[\mathsf{ReLU}([\mathcal{S}^\star_{\ell,j}]_r)\right] \cdot (1 \pm b\mathsf{poly}(k_\ell))$
- *If we denote by $s_r = \left(\mathbf{W}_{\ell,j}^\top X_{\ell,j}\right)_r$ and $s^\star_r = \left(\mathbf{W}_{\ell,j}^\top X^\star_{\ell,j}\right)_r$, then for every $\mathbb{b} \in [b, m^{-0.01}]$,*

$$\left|\mathbb{E}\left[\mathbb{1}_{s_p - \mathbb{b} > 0}[\mathcal{S}_{\ell,j}]_q\right] - \mathbb{E}\left[\mathbb{1}_{s^\star_p - \mathbb{b} > 0}[\mathcal{S}^\star_{\ell,j}]_q\right]\right| \leq \frac{\mathbb{b}\mathsf{poly}(k_1)}{m_\ell^2}$$

- $\mathbb{E}\left[\mathbb{1}_{|[\mathcal{S}_{\ell,j}]_p| \leq 2\mathbb{b}} \cdot |[\mathcal{S}_{\ell,j}]_p|\right] \leq \frac{\mathbb{b}^2\mathsf{poly}(k_\ell)}{m_\ell}$ *for every $\mathbb{b} \in [b, m^{-0.01}]$.*

Recall for given $j \in [d_\ell]$ and $r \in [m_\ell]$, the min-max objective for $D^{(2)}$ is:

$$\min_{\{[\mathbf{V}^D_{\ell,j,j'}]_r\}_{j' \in \mathcal{P}_{\ell,j}}, [b^D_{\ell,j}]_r;} \ \max_{\{[\mathbf{V}_{\ell,j,j'}]_r\}_{j' \in \mathcal{P}_{\ell,j}}, [b_{\ell,j}]_r} \left(\mathbb{E}[D^{(2)}_{\ell,j,r}(X^\star_\ell, X^\star_{\ell-1})] - \mathbb{E}[D^{(2)}_{\ell,j,r}(X_\ell, X_{\ell-1})]\right)$$
$$- \lambda_G \|\mathbf{V}_\ell\|_F^2 + \lambda_D \|\mathbf{V}^D_\ell\|_F^2 \tag{G.2}$$

where $\lambda_D = \frac{1}{\mathsf{poly}(d/\varepsilon)}$, $\lambda_G = \frac{\lambda_D}{\mathsf{poly}(d/\varepsilon)}$, and[23]

$$\begin{aligned}
D^{(2)}_{\ell,j,r}(Y_1, Y_2) &\stackrel{\text{def}}{=} \widetilde{\mathsf{abs}}\left(s_r - \mathsf{LeakyReLU}(\mathfrak{s}_r)\right) \\
\widetilde{\mathsf{abs}}(x) &= \widetilde{\mathsf{ReLU}}(x - b) + \widetilde{\mathsf{ReLU}}(-x - b) \\
s_r &:= \left([\mathbf{W}^D_{\ell,j}]^\top Y_{1,j}\right)_r \\
\mathfrak{s}_r &:= \left(\textstyle\sum_{j' \in \mathcal{P}_{\ell,j}} \mathbf{V}^D_{\ell,j,j'} \mathsf{LeakyReLU}\left([\mathbf{W}^D_{\ell-1,j'}]^\top Y_{2,j'}\right) - b^D_{\ell,j}\right)_r
\end{aligned}$$

**Fact G.3.** *Under Induction G.1, for every $j \in [d_{\ell-1}]$, if we denote by*

$$\begin{aligned}
(\mathcal{S}^\star_{\ell-1,j})' &\stackrel{\text{def}}{=} \mathsf{LeakyReLU}\left([\mathbf{W}_{\ell-1,j}]^\top X^\star_{\ell-1,j}\right) \\
\mathcal{S}'_{\ell-1,j} &\stackrel{\text{def}}{=} \mathsf{LeakyReLU}\left([\mathbf{W}_{\ell-1,j}]^\top X_{\ell-1,j}\right)
\end{aligned}$$

*then*

$$\|(\mathcal{S}^\star_{\ell-1,j})' - \mathcal{S}^\star_{\ell-1,j}\|_2 \leq \frac{1}{\mathsf{poly}(d/\varepsilon)} \quad \text{and} \quad \|(\mathcal{S}_{\ell-1,j})' - \mathcal{S}_{\ell-1,j}\|_2 \leq \frac{1}{\mathsf{poly}(d/\varepsilon)}$$

*Proof.* Trivial because $X^\star_{\ell-1,j} = \mathbf{W}^\star_{\ell-1,j}\mathcal{S}^\star_{\ell-1,j}$ and $X_{\ell-1,j} = \mathbf{W}_{\ell-1,j}\mathcal{S}_{\ell-1,j}$. □

**Invariant G.4.** *Throughout the gradient descent-ascent process for $D^{(2)}_{\ell,j,r}$, we shall maintain the following invariant properties:*

*(a)* $\|\mathsf{ReLU}(\mathcal{S}_{\ell,j})\|_0 \leq O(k_\ell)$, $\|([\mathcal{S}_{\ell,j}]_{r'})_{r' \neq r}\|_\infty \leq \mathsf{poly}(k_\ell)$.

*(b)* $\|[\mathbf{V}^D_{\ell,j}]_r\|_F^2 \leq \frac{\mathsf{poly}(k_\ell)}{\lambda_D}$

[23] To be precise, suppose $\|\mathbf{V}^\star_{\ell,j}\|_F^2 \leq \Lambda$ in the target network, then we choose the regularizer weight $\lambda_D = \frac{1}{\Lambda\mathsf{poly}(d/\varepsilon)}$ and $\lambda_G = \frac{\lambda_D}{\mathsf{poly}(d/\varepsilon)}$.

We shall prove later in Claim G.8 that Invariant G.4b is satisfied (thanks to our choice of the $\lambda_D$ regularizer.

As for Invariant G.4a, we first note it is satisfied at initiation. Indeed, recall

$$\mathcal{S}_{\ell,j} = \mathsf{LeakyReLU}\left(\sum_{j'\in\mathcal{P}_{\ell,j}} \mathbf{V}_{\ell,j,j'}\mathcal{S}_{\ell-1,j'} - b_{\ell,j}\right)$$

Since we have initialized $\mathbf{V}_{\ell,j,j'} = \mathbf{I}$ for one of the $j' \in \mathcal{P}_{\ell,j}$ and others $\mathbf{V}_{\ell,j,j''}$ as zero, as well as $b_{\ell,j} = 0$, the desired property

$$\|\mathsf{ReLU}(\mathcal{S}_{\ell,j})\|_0 \le O(k_\ell), \|\mathcal{S}_{\ell,j}\|_\infty \le \mathsf{poly}(k_\ell)$$

holds at the beginning of training using Induction G.1 (which says that $\mathcal{S}_{\ell-1,j}$ satisfies the desired property). As for why Invariant G.4a is maintained throughout the training, note that Invariant G.4a is defined with respect to $r' \in [m_\ell] \setminus \{r\}$. Therefore, it suffices to show that Invariant G.4a holds at the *end* of the training for $D^{(4)}_{\ell,j,r}$. We shall do so in Lemma G.2.

## G.1 Generator

On the generator side, the maximization process is over $\{[\mathbf{V}_{\ell,j,j'}]_r\}_{j'\in\mathcal{P}_{\ell,j}}$ and $[b_{\ell,j}]_r$ in (G.2). Since $D^{(2)}_{\ell,j,r}(X^\star_\ell, X^\star_{\ell-1})$ does not depend on these parameters, we see that the maximization in (G.2) becomes

$$\max_{\{[\mathbf{V}_{\ell,j,j'}]_r\}_{j'\in\mathcal{P}_{\ell,j}},[b_{\ell,j}]_r} -\left(\mathbb{E}[D^{(2)}_{\ell,j,r}(X_\ell, X_{\ell-1})]\right) - \lambda_G\|[\mathbf{V}_{\ell,j}]_r\|_F^2 \tag{G.3}$$

where

$$\begin{aligned}
D^{(2)}_{\ell,j,r}(X_\ell, X_{\ell-1}) &= \widetilde{\mathsf{abs}}\left(s_r - \mathsf{LeakyReLU}(\mathfrak{s}_r)\right) \\
s_r &:= \left(\mathbf{W}_{\ell,j}^\top X_{\ell,j}\right)_r = \left(\mathbf{W}_{\ell,j}^\top \mathbf{W}_{\ell,j}\mathsf{LeakyReLU}\left(\sum_{j'\in\mathcal{P}_{\ell,j}} \mathbf{V}_{\ell,j,j'}\mathcal{S}_{\ell-1,j'} - b_{\ell,j}\right)\right)_r \\
\mathfrak{s}_r &:= \left(\sum_{j'\in\mathcal{P}_{\ell,j}} \mathbf{V}^D_{\ell,j,j'} \underbrace{\mathsf{LeakyReLU}\left(\mathbf{W}_{\ell-1,j'}^\top X_{\ell-1,j'}\right)}_{\stackrel{\text{def}}{=}\mathcal{S}'_{\ell-1,j'}} - b^D_{\ell,j}\right)_r
\end{aligned} \tag{G.4}$$

Note $\mathfrak{s}_r$ is fixed and only $s_r$ depends on the training variables $\{[\mathbf{V}_{\ell,j,j'}]_r\}_{j'\in\mathcal{P}_{\ell,j}}, [b_{\ell,j}]_r$.

For notation abbreviation, let us write

$$v^G_{j'} := [\mathbf{V}_{\ell,j,j'}]_r \ , \quad b^G := [b_{\ell,j}]_r \ , \quad v^D_{j'} := [\mathbf{V}^D_{\ell,j,j'}]_r \ , \quad b^D := [b^D_{\ell,j}]_r \ .$$

### G.1.1 Existence of Ascent Direction

We have the following claim which states that the exists an ascent direction for the generator objective (G.3).

**Claim G.5.** *Under Induction G.1, suppose Invariant G.4 holds. For every $j \in [d_\ell], r \in [m_\ell]$, for $R$ defined as (c.f. (G.4)):*

$$R(\{v^G_{j'}\}_{j'\in\mathcal{P}_{\ell,j}}, b^G) \stackrel{\text{def}}{=} D^{(2)}_{\ell,j,r}(X_\ell, X_{\ell-1}) = \widetilde{\mathsf{abs}}\left(s_r - \mathsf{LeakyReLU}(\mathfrak{s}_r)\right)$$

*there exists absolute constant $\eta_0 = \frac{1}{\mathsf{poly}(d/\varepsilon)}$ such that for every $\eta \in [0, \eta_0]$, consider update*

$$v^G_{j'} \leftarrow (1-\eta)v^G_{j'} + \eta v^D_{j'}, \quad b^G \leftarrow (1-\eta)b^G + \eta b^D$$

*then w.h.p.*

$$R\left(\{(1-\eta)v^G_{j'} + \eta v^D_{j'}\}_{j'\in\mathcal{P}_{\ell,j}}, (1-\eta)b^G + \eta b^D\right) \le (1 - 0.4\eta\zeta)R(\{v^G_{j'}\}_{j'\in\mathcal{P}_{\ell,j}}, b^G) \ .$$

*Proof of Claim G.5.* Recall

$$X_{\ell,j} = \mathbf{W}_{\ell,j}\mathcal{S}_{\ell,j} = \mathbf{W}_{\ell,j}\mathsf{LeakyReLU}\left(\sum_{j'\in\mathcal{P}_{\ell,j}} \mathbf{V}_{\ell,j,j'}\mathcal{S}_{\ell-1,j'} - b_{\ell,j}\right)$$

so we can write

$$s_r = \left[\mathbf{W}_{\ell,j}^\top X_{\ell,j}\right]_r = \|[\mathbf{W}_{\ell,j}]_r\|_2^2 \cdot \mathsf{LeakyReLU}\left(\sum_{j'\in\mathcal{P}_{\ell,j}} v_{j'}^G \mathcal{S}_{\ell-1,j'} - b^G\right) + const \tag{G.5}$$

for $\|[\mathbf{W}_{\ell,j}]_r\|_2 = 1$ and some constant *const* that does not depend on $\{v_{j'}^G\}_{j'\in\mathcal{P}_{\ell,j}}$ or $b^G$, and by Invariant G.4a and $\|[\mathbf{W}^\star_{\ell,j}]_p - [\mathbf{W}_{\ell,j}]_p\|_2 \leq \delta$ from Induction G.1, we know w.h.p. $|const| \leq 0.1b$ and also $s_r \geq -0.1b$.

At the same time, for analysis purpose define

$$\widetilde{\mathfrak{s}}_r := \sum_{j'\in\mathcal{P}_{\ell,j}} v_{j'}^D \mathcal{S}_{\ell-1,j'} - b^D \enspace ,$$

and Fact G.3 tells us $|\widetilde{\mathfrak{s}}_r - \mathfrak{s}_r| \leq \frac{1}{\mathsf{poly}(d/\varepsilon)}$.

Now we consider three separate cases. Let $s'_r$ be the value of $s_r$ after the update.

1. When $s_r > 0.2b$, we know w.h.p. $\sum_{j'\in\mathcal{P}_{\ell,j}} v_{j'}^G \mathcal{S}_{\ell-1,j'} - b^G \geq 0.1b$. Therefore, for any sufficiently small $\eta < \frac{1}{\mathsf{poly}(d/\varepsilon)}$, we know after the update,

$$s'_r = (1-\eta)s_r + \eta\widetilde{\mathfrak{s}}_r > 0$$

   - If $\widetilde{\mathfrak{s}}_r > -0.5b$ then by the convexity of $\widetilde{\mathsf{abs}}(\cdot)$ (which comes from the convexity of $\widetilde{\mathsf{ReLU}}(\cdot)$) we have

$$\begin{aligned}\widetilde{\mathsf{abs}}(s'_r - \mathsf{LeakyReLU}(\widetilde{\mathfrak{s}}_r)) &\leq (1-\eta)\widetilde{\mathsf{abs}}(s_r - \mathsf{LeakyReLU}(\mathfrak{s}_r)) + \eta\widetilde{\mathsf{abs}}(\widetilde{\mathfrak{s}}_r - \mathsf{LeakyReLU}(\mathfrak{s}_r)) \\ &= (1-\eta)\widetilde{\mathsf{abs}}(s_r - \mathsf{LeakyReLU}(\mathfrak{s}_r)) \enspace ;\end{aligned}$$

   - If $\widetilde{\mathfrak{s}}_r < -0.5b$ then $s'_r = (1-\eta)s_r + \eta\widetilde{\mathfrak{s}}_r \leq (1-\eta)s_r + \eta\mathsf{LeakyReLU}(\mathfrak{s}_r)$, so using $\widetilde{\mathsf{abs}}((1-\eta)t) \leq (1-\eta)\widetilde{\mathsf{abs}}(t)$, we have

$$\begin{aligned}\widetilde{\mathsf{abs}}(s'_r - \mathsf{LeakyReLU}(\mathfrak{s}_r)) &\leq \widetilde{\mathsf{abs}}\Big(\big((1-\eta)s_r + \eta\mathsf{LeakyReLU}(\mathfrak{s}_r)\big) - \mathsf{LeakyReLU}(\mathfrak{s}_r)\Big) \\ &\leq (1-\eta)\widetilde{\mathsf{abs}}(s_r - \mathsf{LeakyReLU}(\mathfrak{s}_r)) \enspace .\end{aligned}$$

2. When $\widetilde{\mathfrak{s}}_r \leq 0.6b$ and $s_r < 0.2b$, using w.h.p. $s_r \geq -0.1b$ and $\widetilde{\mathfrak{s}}_r > -\mathsf{poly}(d/\varepsilon)$, we have

$$\widetilde{\mathsf{abs}}\left(s_r - \mathsf{LeakyReLU}(\widetilde{\mathfrak{s}}_r)\right) = \widetilde{\mathsf{abs}}\left(s'_r - \mathsf{LeakyReLU}(\widetilde{\mathfrak{s}}_r)\right) = 0$$

3. When $\widetilde{\mathfrak{s}}_r > 0.6b$ and $s_r < 0.2b$. By (G.5) we have

$$s'_r - s_r = \mathsf{LeakyReLU}\left((1-\eta)\left(\sum_{j'\in\mathcal{P}_{\ell,j}} v_{j'}^G\mathcal{S}_{\ell-1,j'} - b^G\right) + \eta\widetilde{\mathfrak{s}}_r\right) - \mathsf{LeakyReLU}\left(\sum_{j'\in\mathcal{P}_{\ell,j}} v_{j'}^G\mathcal{S}_{\ell-1,j'} - b^G\right)$$

   Now, since $s_r \leq 0.2b$, by (G.5) we know w.h.p. $\mathsf{LeakyReLU}\left(\sum_{j'\in\mathcal{P}_{\ell,j}} v_{j'}^G\mathcal{S}_{\ell-1,j'} - b^G\right) \leq 0.3b$, which means that

$$\begin{aligned}(1-\eta)\left(\sum_{j'\in\mathcal{P}_{\ell,j}} v_{j'}^G\mathcal{S}_{\ell-1,j'} - b^G\right) + \eta\widetilde{\mathfrak{s}}_r &\geq \sum_{j'\in\mathcal{P}_{\ell,j}} v_{j'}^G\mathcal{S}_{\ell-1,j'} - b^G + 0.5\eta\widetilde{\mathfrak{s}}_r \\ \implies s'_r &\geq s_r + 0.4\eta\zeta\mathfrak{s}_r\end{aligned}$$

This implies by the monotonicity of $\widetilde{\mathsf{abs}}(t)$ for $t<0$ and $\widetilde{\mathsf{abs}}((1-\eta)t) \leq (1-\eta)\widetilde{\mathsf{abs}}(t)$,

$$\widetilde{\mathsf{abs}}(s'_r - \mathsf{LeakyReLU}(\mathfrak{s}_r)) \leq \widetilde{\mathsf{abs}}(s_r - \mathsf{LeakyReLU}(\mathfrak{s}_r) + 0.5\eta\zeta\mathfrak{s}_r)$$
$$\leq (1-0.5\eta\zeta)\widetilde{\mathsf{abs}}(s_r - \mathsf{LeakyReLU}(\mathfrak{s}_r)) \ . \qquad \square$$

### G.1.2 Optimization

**Lemma G.6** ($D^{(2)}$ generator)**.** *Suppose $\|[\mathbf{V}^D_{\ell,j}]_r\|_F^2 \leq \Lambda$ for some $\Lambda = \mathsf{poly}(d/\varepsilon)$ and the regularizer weight $\lambda_G = \frac{1}{\mathsf{poly}(d/\varepsilon)\Lambda}$, then, by applying $\frac{\mathsf{poly}(d/\varepsilon)}{\eta}$ steps of gradient ascent on the generator side with learning rate $\eta = \frac{1}{\mathsf{poly}(d/\varepsilon)}$, it satisfies*

$$\mathbb{E}[D^{(2)}_{\ell,j,r}(X_\ell, X_{\ell-1})] + \lambda_G \|[\mathbf{V}_{\ell,j}]_r\|_F^2 \leq \frac{1}{\mathsf{poly}(d/\varepsilon)} \quad \text{and} \quad \|[\mathbf{V}_{\ell,j}]_r\|_F^2 \leq 2\|[\mathbf{V}^D_{\ell,j}]_r\|_F^2 \ .$$

*As a corollary, w.p. $\geq 1 - \frac{1}{\mathsf{poly}(d/\varepsilon)}$, it satisfies*

$$\left| [\mathcal{S}_{\ell,j}]_r - \mathsf{LeakyReLU}\left(\sum_{j'\in\mathcal{P}_{\ell,j}} \mathbf{V}^D_{\ell,j,j'}\mathcal{S}_{\ell-1,j'} - b^D_{\ell,j}\right)_r \right| \leq 3b \ .$$

*Proof.* Let us write re-write (G.3) as

$$-\mathsf{Obj}_G\left(\{v^G_{j'}\}_{j'\in\mathcal{P}_{\ell,j}}, b^G\right) \stackrel{\text{def}}{=} \mathbb{E}[D^{(2)}_{\ell,j,r}(X_\ell, X_{\ell-1})] + \lambda_G \sum_{j'\in\mathcal{P}_{\ell,j}} \|v^G_{j'}\|_2^2$$
$$= \mathbb{E}[R(\{v^G_{j'}\}_{j'\in\mathcal{P}_{\ell,j}}, b^G)] + \lambda_G \sum_{j'\in\mathcal{P}_{\ell,j}} \|v^G_{j'}\|_2^2$$

Now, Claim G.5 and the convexity of $\|\cdot\|_2^2$ implies

$$- \mathsf{Obj}_G\left(\{(1-\eta)v^G_{j'} + \eta v^D_{j'}\}_{j'\in\mathcal{P}_{\ell,j}}, (1-\eta)b^G + \eta b^D\right)$$
$$\leq -\mathsf{Obj}_G\left(\{v^G_{j'}\}_{j'\in\mathcal{P}_{\ell,j}}, b^G\right) - 0.4\eta\zeta\,\mathbb{E}[R(\{v^G_{j'}\}_{j'\in\mathcal{P}_{\ell,j}}, b^G)] - \eta\lambda_G \sum_{j'\in\mathcal{P}_{\ell,j}} (\|v^G_{j'}\|_2^2 - \|v^D_{j'}\|_2^2) + \frac{1}{(d/\varepsilon)^{\omega(1)}}$$

In other words,

- As long as $\mathbb{E}[R(\{v^G_{j'}\}_{j'\in\mathcal{P}_{\ell,j}}, b^G)] \geq \frac{1}{\mathsf{poly}(d/\varepsilon)}$, it satisfies for every $\eta < \eta_0$:
$$- \mathsf{Obj}_G\left(\{(1-\eta)v^G_{j'} + \eta v^D_{j'}\}_{j'\in\mathcal{P}_{\ell,j}}, (1-\eta)b^G + \eta b^D\right) \leq -\mathsf{Obj}_G\left(\{v^G_{j'}\}_{j'\in\mathcal{P}_{\ell,j}}, b^G\right) - \frac{\eta}{\mathsf{poly}(d/\varepsilon)} \tag{G.6}$$
- As long as $\lambda_G \sum_{j'\in\mathcal{P}_{\ell,j}} \|v^G_{j'}\|_2^2 \geq \frac{1}{\mathsf{poly}(d/\varepsilon)}$, then we know $\sum_{j'\in\mathcal{P}_{\ell,j}} \|v^G_{j'}\|_2^2 \geq 2\Lambda^2 \geq 2\|[\mathbf{V}^D_{\ell,j}]_r\|_F^2$, and therefore we can also derive (G.6).

This immediately gives $\left\|\nabla \mathsf{Obj}_G\left(\{v^G_{j'}\}_{j'\in\mathcal{P}_{\ell,j}}, b^G\right)\right\|_2 \geq \frac{1}{\mathsf{poly}(d/\varepsilon)}$. Using the fact that $\mathsf{Obj}_G$ is $\mathsf{poly}(d)$-Lipschitz smooth, we know by applying gradient ascent over $\mathsf{Obj}_G$ with sufficiently small learning rate $\eta = \frac{1}{\mathsf{poly}(d/\varepsilon)}$, it guarantees to increase the objective by at least $\frac{1}{\mathsf{poly}(d/\varepsilon)}$.

By repeatedly applying gradient ascent, after $T = \frac{\mathsf{poly}(d/\varepsilon)}{\eta}$ iterations, we can increase the generator's objective at least until $\mathbb{E}[R(\{v^G_{j'}\}_{j'\in\mathcal{P}_{\ell,j}}, b^G)] \leq \frac{1}{\mathsf{poly}(d/\varepsilon)}$, $\lambda_G \sum_{j'\in\mathcal{P}_{\ell,j}} \|v^G_{j'}\|_2^2 \leq \frac{1}{\mathsf{poly}(d/\varepsilon)}$ and $\sum_{j'\in\mathcal{P}_{\ell,j}} \|v^G_{j'}\|_2^2 \leq 2\Lambda^2$ are all satisfied.

As for the corollary, recall from (G.5)

$$s_r = \mathsf{LeakyReLU}\left(\textstyle\sum_{j'\in\mathcal{P}_{\ell,j}} v^G_{j'}\mathcal{S}_{\ell-1,j'} - b^G\right) \pm 0.1b$$

and from Fact G.3

$$\mathfrak{s}_r = \sum_{j'\in\mathcal{P}_{\ell,j}} v^D_{j'} \mathcal{S}_{\ell-1,j'} - b^D \pm \frac{1}{\mathsf{poly}(d/\varepsilon)}$$

After optimization we have

$$\mathbb{E}\left[\widetilde{\mathsf{abs}}(s_r - \mathsf{LeakyReLU}(\mathfrak{s}_r))\right] \leq \frac{1}{\mathsf{poly}(d/\varepsilon)}$$

which implies w.p. $\geq 1 - \frac{1}{\mathsf{poly}(d/\varepsilon)}$ it satisfies

$$\left|[\mathcal{S}_{\ell,j}]_r - \mathsf{LeakyReLU}(\mathfrak{s}_r)\right| \leq 2b \ . \qquad \square$$

## G.2 Discriminator

On the discriminator side, the optimization is over $\{[\mathbf{V}^D_{\ell,j,j'}]_r\}_{j'\in\mathcal{P}_{\ell,j}}$ and $[b^D_{\ell,j}]_r$.

In the min-max objective (G.2), since we have optimized the (inner) generator to sufficiently good accuracy, we have $\mathbb{E}[D^{(2)}_{\ell,j,r}(X_\ell, X_{\ell-1})] \leq \frac{1}{\mathsf{poly}(d/\varepsilon)}$ from Lemma G.6. Using the fact that $\mathbb{E}[D^{(2)}_{\ell,j,r}(X_\ell, X_{\ell-1})]$ is $\mathsf{poly}(d/\varepsilon)$-Lipschitz smooth with respect to $\mathbf{V}^D$, this implies its gradient with respect to $\{[\mathbf{V}^D_{\ell,j,j'}]_r\}_{j'\in\mathcal{P}_{\ell,j}}$ is also of norm at most $\frac{1}{\mathsf{poly}(d/\varepsilon)}$.[24] Therefore, we can assume as if $\mathbb{E}[D^{(2)}_{\ell,j,r}(X_\ell, X_{\ell-1})]$ does not exist, when focusing on the minimization process for the discriminator.

In other words, when the (inner) generator side is sufficiently maximized, the minimization in (G.2) becomes

$$\min_{\{\mathbf{V}^D_{\ell,j,j'}\}_{j'\in\mathcal{P}_{\ell,j}}, b^D_{\ell,j}} \left(\mathbb{E}[D^{(2)}_{\ell,j,r}(X^\star_\ell, X^\star_{\ell-1})]\right) + \lambda_D \|[\mathbf{V}^D_{\ell,j}]_r\|_F^2 \tag{G.7}$$

where

$$\begin{aligned} D^{(2)}_{\ell,j,r}(X^\star_\ell, X^\star_{\ell-1}) &= \widetilde{\mathsf{abs}}\left(s^\star_r - \mathsf{LeakyReLU}(\mathbb{s}_r)\right) \\ s^\star_r &:= \left(\mathbf{W}_{\ell,j}^\top X^\star_{\ell,j}\right)_r = \left(\mathbf{W}_{\ell,j}^\top \mathbf{W}^\star_{\ell,j} \mathsf{ReLU}\left(\sum_{j'\in\mathcal{P}_{\ell,j}} \mathbf{V}^\star_{\ell,j,j'} \mathcal{S}^\star_{\ell-1,j'} - b^\star_{\ell,j}\right)\right)_r \\ \mathbb{s}_r &:= \left(\sum_{j'\in\mathcal{P}_{\ell,j}} \mathbf{V}^D_{\ell,j,j'} \underbrace{\mathsf{LeakyReLU}\left([\mathbf{W}_{\ell-1,j'}]^\top X^\star_{\ell-1,j'}\right)}_{\stackrel{\text{def}}{=}(\mathcal{S}^\star_{\ell-1,j})'} - b^D_{\ell,j}\right)_r \end{aligned} \tag{G.8}$$

Note this time $s^\star_r$ is fixed and only $\mathbb{s}_r$ depends on the training parameters $\{\mathbf{V}^D_{\ell,j,j'}\}_{j'\in\mathcal{P}_{\ell,j}}, b^D_{\ell,j}$.

Let us abbreviate the notations by letting

$$v^\star_{j'} := [\mathbf{V}^\star_{\ell,j,j'}]_r \ , \quad b^\star := [b^\star_{\ell,j}]_r \ , \quad v^D_{j'} := [\mathbf{V}^D_{\ell,j,j'}]_r \ , \quad b^D := [b^D_{\ell,j}]_r \ .$$

### G.2.1 Existence of Descent Direction

We have the following claim which states that the exists a descent direction for the discriminator objective (G.7).

**Claim G.7.** *Under Induction G.1, suppose Invariant G.4 holds. For every $j \in [d_\ell]$ and every $r \in [m_\ell]$, let $R^\star$ be defined as (c.f. (G.8))*

$$R^\star(\{v^D_{j'}\}_{j'\in\mathcal{P}_{\ell,j}}, b^D) := D^{(2)}_{\ell,j,r}(X^\star_\ell, X^\star_{\ell-1}) = \widetilde{\mathsf{abs}}\left(s^\star_r - \mathsf{LeakyReLU}(\mathbb{s}_r)\right)$$

[24] Indeed, for any non-negative function $f(x)$ that is $L$-Lipschitz smooth (and not necessarily convex), we have $f(x) \geq \frac{1}{2L}\|\nabla f(x)\|_2^2$.

*There exists absolute constant* $\eta_0 = \frac{1}{\mathsf{poly}(d/\varepsilon)}$ *such that for every* $\eta \in [0, \eta_0]$*, consider update:*

$$v_{j'}^D \leftarrow (1-\eta) v_{j'}^D + \eta v_{j'}^\star, \quad b^D \leftarrow (1-\eta) b^D + \eta b^\star$$

*we have: w.h.p.*

$$R^\star\left(\{(1-\eta)v_{j'}^D + \eta v_{j'}^\star\}_{j' \in \mathcal{P}_{\ell,j}}, (1-\eta)b^D + \eta b^\star\right) \leq (1 - 0.3\eta\zeta) R^\star(\{v_{j'}^D\}_{j' \in \mathcal{P}_{\ell,j}}, b^D)$$

*Proof of Claim G.7.* Note the update does not change $s_r^\star$ and only changes $\mathbb{s}_r$. Let

$$\mathbb{s}_r' = \left( \sum_{j' \in \mathcal{P}_{\ell,j}} \left((1-\eta) v_{j'}^D + \eta v_{j'}^\star\right) (\mathcal{S}_{\ell-1,j'}^\star)' - \left((1-\eta) b^D + \eta b^\star\right) \right)_r$$

be the value of $\mathbb{s}_r$ after after the update, and let $\mathbb{s}_r^\star := \left(\sum_{j' \in \mathcal{P}_{\ell,j}} v_{j'}^\star (\mathcal{S}_{\ell-1,j'}^\star)' - b^\star\right)_r$. We have

$$\mathbb{s}_r' = (1-\eta)\mathbb{s}_r + \eta \mathbb{s}_r^\star$$

Furthermore, recall $X_{\ell,j}^\star = \mathbf{W}_{\ell,j}^\star \mathsf{ReLU}\left(\sum_{j' \in \mathcal{P}_{\ell,j}} \mathbf{V}_{\ell,j,j'}^\star \mathcal{S}_{\ell-1,j'}^\star - b_{\ell,j}^\star\right)$, so we have

$$s_r^\star = \left[\mathbf{W}_{\ell,j}^\top X_{\ell,j}^\star\right]_r = \langle [\mathbf{W}_{\ell,j}]_r, [\mathbf{W}_{\ell,j}^\star]_r \rangle \cdot \underbrace{\mathsf{ReLU}\left( \sum_{j' \in \mathcal{P}_{\ell,j}} v_{j'}^\star \mathcal{S}_{\ell-1,j'}^\star - b^\star \right)}_{=[\mathcal{S}_{\ell,j}^\star]_r} + const \tag{G.9}$$

Above, since $\|[\mathbf{W}_{\ell,j}]_r\|_2 = \|[\mathbf{W}_{\ell,j}^\star]_r\|_2 = 1$ and $\|[\mathbf{W}_{\ell,j}]_r - [\mathbf{W}_{\ell,j}^\star]_r\|_2 \leq \delta$, we have $\langle [\mathbf{W}_{\ell,j}]_r, [\mathbf{W}_{\ell,j}^\star]_r \rangle \in [1 - \delta^2/2, 1] \subseteq [1 - 0.01b, 1]$; and at the same time, *const* is some constant that does not depend on $\{v_{j'}^D\}_{j' \in \mathcal{P}_{\ell,j}}$ or $b^D$, and by Assumption 2.2 it is easy to derive that w.h.p. $|const| \leq 0.05b$ and also $s_r^\star \geq -0.05b$. In sum, we can derive that

$$s_r^\star = \mathbb{s}_r^\star \pm 0.1b \geq -0.05b$$

Now, we consider three cases:

1. When $\mathbb{s}_r > 0.2b$, we know for sufficiently small $|\delta|$, it satisfies $\mathsf{LeakyReLU}(\mathbb{s}_r + \delta) = \mathbb{s}_r + \delta$. Therefore, by the convexity of $\widetilde{\mathsf{abs}}(\cdot)$ we have

$$\begin{aligned}\widetilde{\mathsf{abs}}(s_r^\star - \mathsf{LeakyReLU}(\mathbb{s}_r')) &\leq (1-\eta)\widetilde{\mathsf{abs}}(s_r^\star - \mathsf{LeakyReLU}(\mathbb{s}_r)) + \eta \widetilde{\mathsf{abs}}(s_r^\star - \mathbb{s}_r^\star) \\ &\leq (1-\eta)\widetilde{\mathsf{abs}}(s_r^\star - \mathsf{LeakyReLU}(\mathbb{s}_r)) \ .\end{aligned}$$

2. When $\mathbb{s}_r < 0.2b$ and $s_r^\star < 0.6b$, recall we have derived $s_r^\star > -0.05b$ and we also have $\mathbb{s}_r > -\mathsf{poly}(d/\varepsilon)$. In this case, we obviously have

$$\widetilde{\mathsf{abs}}\left(s_r^\star - \mathsf{LeakyReLU}(\mathbb{s}_r)\right) = \widetilde{\mathsf{abs}}\left(s_r^\star - \mathsf{LeakyReLU}(\mathbb{s}_r')\right) = 0$$

3. When $\mathbb{s}_r < 0.2b$ and $s_r^\star > 0.6b$, we know $\mathbb{s}_r^\star \geq 0.5b$ and therefore

$$\begin{aligned} & (1-\eta)\mathbb{s}_r + \eta \mathbb{s}_r^\star \geq \mathbb{s}_r + 0.5\eta \mathbb{s}_r^\star \\ \implies & \mathsf{LeakyReLU}(\mathbb{s}_r') = \mathsf{LeakyReLU}\left((1-\eta)\mathbb{s}_r + \eta \mathbb{s}_r^\star\right) \\ & \qquad \geq \mathsf{LeakyReLU}(\mathbb{s}_r) + 0.5\eta\zeta \mathbb{s}_r^\star \end{aligned}$$

This implies by the monotonicity of $\widetilde{\mathsf{abs}}(t)$ for $t > 0$,

$$\begin{aligned}\widetilde{\mathsf{abs}}(s_r^\star - \mathsf{LeakyReLU}(\mathbb{s}_r')) &\leq \widetilde{\mathsf{abs}}(s_r^\star - \mathsf{LeakyReLU}(\mathbb{s}_r) + 0.5\eta\zeta \mathbb{s}_r^\star) \\ &\leq (1 - 0.3\eta\zeta)\widetilde{\mathsf{abs}}(s_r^\star - \mathsf{LeakyReLU}(\mathbb{s}_r))\end{aligned}$$

□

### G.2.2 Optimization

One thing we need to check is on proving the invariant.

**Claim G.8.** *As long as the objective (G.7) is not increased by more than constant factor times the initialization, then Invariant G.4b is satisfied.*

*Proof.* It is easy to verify that at initialization, to our choice of $\mathbf{V}^D_{\ell,j,j'} = \mathbf{I}$ for exactly one $j' \in [d_\ell]$ implies

$$D^{(2)}_{\ell,j,r}(X^\star_\ell, X^\star_{\ell-1}) = \widetilde{\mathsf{abs}}\left(s^\star_r - \mathsf{LeakyReLU}(\mathbb{s}_r)\right) \leq \mathsf{poly}(k_\ell)$$

Therefore, as long as the objective does not increase by much, we should have

$$\lambda_D \|[\mathbf{V}^D_{\ell,j}]_r\|_F^2 \leq \mathsf{poly}(k_\ell)$$

and thus Invariant G.4b is satisfied. □

**Lemma G.9** ($D^{(2)}$ discriminator)**.** *Suppose $\|[\mathbf{V}^\star_{\ell,j}]_r\|_F \leq \Lambda$ for some $\Lambda = \mathsf{poly}(d/\varepsilon)$ and the regularizer weight $\lambda_D \leq \frac{1}{\mathsf{poly}(d/\varepsilon)\Lambda^2}$, then, by applying $\frac{\mathsf{poly}(d/\varepsilon)}{\eta}$ steps of gradient-descent ascent with learning rate $\eta = \frac{1}{\mathsf{poly}(d/\varepsilon)}$, it satisfies*[25]

$$\mathbb{E}[D^{(2)}_{\ell,j,r}(X^\star_\ell, X^\star_{\ell-1})] + \lambda_D \|[\mathbf{V}^D_{\ell,j}]_r\|_F^2 \leq \frac{1}{\mathsf{poly}(d/\varepsilon)} \quad \textit{and} \quad \|[\mathbf{V}^D_{\ell,j}]_r\|_F^2 \leq 2\|[\mathbf{V}^\star_{\ell,j}]_r\|_F^2 \ .$$

*As a corollary, w.p. $\geq 1 - \frac{1}{\mathsf{poly}(d/\varepsilon)}$, it satisfies*

$$\left| [\mathcal{S}^\star_{\ell,j}]_r - \left( \sum_{j' \in \mathcal{P}_{\ell,j}} \mathbf{V}^D_{\ell,j,j'} \mathcal{S}^\star_{\ell-1,j'} - b^D_{\ell,j} \right)_r \right| \leq 3b \ .$$

*Proof of Lemma G.9.* The proof of convergence is completely analogous to that of Lemma G.6. That is, by repeatedly applying gradient descent, we can increase the discriminator's objective at least until $\mathbb{E}[R^*(\{v^G_{j'}\}_{j' \in \mathcal{P}_{\ell,j}}, b^G)] \leq \frac{1}{\mathsf{poly}(d/\varepsilon)}$, $\lambda_D \sum_{j' \in \mathcal{P}_{\ell,j}} \|v^D_{j'}\|_2^2 \leq \frac{1}{\mathsf{poly}(d/\varepsilon)}$, and $\sum_{j' \in \mathcal{P}_{\ell,j}} \|v^D_{j'}\|_2^2 \leq 2\Lambda^2$ are all satisfied.

As for the corollary, recall

$$s^\star_r := \left( [\mathbf{W}_{\ell,j}]^\top X^\star_{\ell,j} \right)_r = \mathcal{S}^\star_{\ell,j} \pm 0.1b$$

$$\mathbb{s}_r := \left( \textstyle\sum_{j' \in \mathcal{P}_{\ell,j}} \mathbf{V}^D_{\ell,j,j'} (\mathcal{S}^\star_{\ell-1,j'})' - b^D_{\ell,j} \right)_r = \left( \textstyle\sum_{j' \in \mathcal{P}_{\ell,j}} \mathbf{V}^D_{\ell,j,j'} \mathcal{S}^\star_{\ell-1,j'} - b^D_{\ell,j} \right)_r \pm \tfrac{1}{\mathsf{poly}(d/\varepsilon)}$$

After optimization, we have

$$\mathbb{E}\left[ \widetilde{\mathsf{abs}}(s^\star_r - \mathsf{LeakyReLU}(\mathbb{s}_r)) \right] \leq \frac{1}{\mathsf{poly}(d/\varepsilon)}$$

which implies w.p. $\geq 1 - \frac{1}{\mathsf{poly}(d/\varepsilon)}$ it satisfies

$$\left| [\mathcal{S}^\star_{\ell,j}]_r - \mathsf{LeakyReLU}(\mathbb{s}_r) \right| \leq 2b \ .$$

□

[25] This means exactly Algorithm 2, that is, apply $\mathsf{poly}(d/\varepsilon)$ many steps of gradient descent on the discriminator side with learning rate $\eta = \frac{1}{\mathsf{poly}(d/\varepsilon)}$, and after *each* descent step, apply $\mathsf{poly}(d/\varepsilon)$ many steps of gradient ascent on the generator side with learning rate $\eta = \frac{1}{\mathsf{poly}(d/\varepsilon)}$.

### G.3 Proof of Lemma G.2: Putting All Together

*Proof of Lemma G.2.* Note our assumption on layer $\ell - 1$ in Induction G.1 together with the corollaries in Lemma G.6 and Lemma G.9 imply, after gradient descent ascent, with probability at least $1 - \frac{1}{\mathsf{poly}(d/\varepsilon)}$, it satisfies

$$\left|[\mathcal{S}^\star_{\ell,j}(\mathbf{U}z)]_r - [\mathcal{S}_{\ell,j}(z)]_r\right| \leq 7b$$

Therefore, if we increase the bias $[b_{\ell,j}]_r$ in

$$\mathcal{S}_{\ell,j} = \mathsf{LeakyReLU}\left(\sum_{j'\in\mathcal{P}_{\ell,j}} \mathbf{V}_{\ell,j,j'}\mathcal{S}_{\ell-1,j'} - b_{\ell,j}\right)$$

by $7b$, then equations in (G.1) are satisfied:

$$[\mathcal{S}_{\ell,j}(z)]_r \leq [\mathcal{S}^\star_{\ell,j}(\mathbf{U}z)]_r, \quad |[\mathcal{S}_{\ell,j}(z)]_r - [\mathcal{S}^\star_{\ell,j}(\mathbf{U}z)]_r| \leq O(b)$$

For the corollaries on expectations, let us recall from Lemma G.6 and Lemma G.9 that after optimization, $\|[\mathbf{V}_{\ell,j}]_r\|_F^2 \leq 4\|[\mathbf{V}^\star_{\ell,j}]_r\|_F^2 \leq \mathsf{poly}(d)$, which implies $|[\mathcal{S}_{\ell,j}]_r| \leq \mathsf{poly}(d)$ is always satisfied. Since the failure event's probability for (G.1) is sufficiently small $\leq \frac{1}{\mathsf{poly}(d/\varepsilon)}$, we can ignore it when deriving the corollaries.

For the **first corollary**, we notice whenever $[\mathcal{S}^\star_{\ell,j}(\mathbf{U}z)]_r = 0$ it satisfies $[\mathcal{S}_{\ell,j}(z)]_r \leq [\mathcal{S}^\star_{\ell,j}(\mathbf{U}z)]_r = 0$. When $[\mathcal{S}^\star_{\ell,j}(\mathbf{U}z)]_r > 0$— which happens with probability at most $\frac{\mathsf{poly}(k_\ell)}{m_\ell}$ by Assumption 2.2) — we know $|[\mathcal{S}_{\ell,j}(z)]_r - [\mathcal{S}^\star_{\ell,j}(\mathbf{U}z)]_r| \leq O(b)$. Together, they imply $\mathbb{E}\left[\mathsf{ReLU}([\mathcal{S}_{\ell,j}]_r)\right] = \mathbb{E}\left[\mathsf{ReLU}([\mathcal{S}^\star_{\ell,j}]_r)\right] \pm \frac{b\mathsf{poly}(k_\ell)}{m_\ell}$ which gives the first corollary.

For the **second corollary**, recall from previous proofs that $s_r = [\mathcal{S}_{\ell,j}]_r \pm 0.01b$ and $s^\star_r = [\mathcal{S}^\star_{\ell,j}]_r \pm 0.01b$. We wish to calculate the expectation difference between $\mathbb{E}\left[\mathbb{1}_{s_p - \mathbb{b} > 0}[\mathcal{S}_{\ell,j}]_q\right]$ and $\mathbb{E}\left[\mathbb{1}_{s^\star_p - \mathbb{b} > 0}[\mathcal{S}^\star_{\ell,j}]_q\right]$.

- Whenever $[\mathcal{S}^\star_{\ell,j}]_q = 0$ or $[\mathcal{S}^\star_{\ell,j}]_p = 0$, we also have $[\mathcal{S}_{\ell,j}]_q \leq 0$ or $[\mathcal{S}_{\ell,j}]_p \leq 0$, and in this case
$$\mathbb{1}_{s_p - \mathbb{b} > 0}[\mathcal{S}_{\ell,j}]_q = 0 = \mathbb{1}_{s^\star_p - \mathbb{b} > 0}[\mathcal{S}^\star_{\ell,j}]_q$$
- Therefore, it suffices to focus on $[\mathcal{S}^\star_{\ell,j}]_q > 0$ and $[\mathcal{S}^\star_{\ell,j}]_p > 0$ (which happens with probability at most $\frac{\mathsf{poly}(k_\ell)}{m_\ell^2}$ by Assumption 2.2).
  - Suppose $\mathbb{1}_{s^\star_p - \mathbb{b} > 0} = \mathbb{1}_{s_p - \mathbb{b} > 0}$, then using $|[\mathcal{S}_{\ell,j}(z)]_q - [\mathcal{S}^\star_{\ell,j}(\mathbf{U}z)]_q| \leq O(b)$, this contributes to the expectation difference by at most $\frac{b\mathsf{poly}(k_\ell)}{m_\ell^2}$.
  - Suppose $\mathbb{1}_{s^\star_p - \mathbb{b} > 0} \neq \mathbb{1}_{s_p - \mathbb{b} > 0}$, this happens only when $\left|[\mathcal{S}^\star_{\ell,j}]_p\right| \leq O(\mathbb{b})$. By the anti-concentration Assumption B.1, we know this happens with probability at most $\frac{\mathbb{b}\mathsf{poly}(k_\ell)}{m_\ell^2}$.

Putting them together we get the desired difference in expectation for the second corollary.

For the **third corollary**, we wish to bound $\mathbb{E}\left[\mathbb{1}_{|[\mathcal{S}_{\ell,j}]_p| \leq 2\mathbb{b}} \cdot |[\mathcal{S}_{\ell,j}]_p|\right]$.

- First note that as long as $[\mathcal{S}^\star_{\ell,j}]_p = 0$, we also have $|[\mathcal{S}_{\ell,j}]_p| \leq \frac{1}{\mathsf{poly}(d)}$ so is negligible.
- Next, whenever $[\mathcal{S}^\star_{\ell,j}]_p \geq \Omega(\mathbb{b})$, we also have $[\mathcal{S}_{\ell,j}]_p > 3\mathbb{b}$ so $\mathbb{1}_{|[\mathcal{S}_{\ell,j}]_p| \leq 2\mathbb{b}} = 0$.
- In the remainder case, we have $0 < [\mathcal{S}^\star_{\ell,j}]_p \leq O(\mathbb{b})$, and this happens with probability at most $\frac{\mathbb{b}\mathsf{poly}(k_\ell)}{m_\ell}$ by Assumption B.1. Under this event, we have $\mathbb{1}_{|[\mathcal{S}_{\ell,j}]_p| \leq 2\mathbb{b}} \cdot |[\mathcal{S}_{\ell,j}]_p| \leq 2\mathbb{b}$.

Putting them together we finish the proof of the third corollary. □

## G.4 Proof of Theorem 4.1

Finally,

- initialize $\mathbf{W}_\ell$ using warm start (see Algorithm 5 and Lemma D.1),
- then repeatedly (and alternatively)
  - learn $\mathbf{V}_\ell$ (see Algorithm 3 and Lemma G.2),
  - learn $\mathbf{W}_1$ (see Algorithm 1 and Lemma E.2),
  - decrease $b \leftarrow bm^{-0.02}$;

Then, combining all the cited lemmas, we have

**Lemma G.10** $(D^{(1)} + D^{(2)})$**.** *After iteration $\ell > 1$ is finished in Algorithm 4, there is a column orthonormal matrix $\mathbf{U} \in \mathbb{R}^{m_0 \times m_0'}$ satisfying, for every $j \in [d_1]$:*

- $\|\mathbf{V}_{\ell,j} - \mathbf{V}^\star_{\ell,j}\mathbf{U}\|_F \leq \frac{1}{\mathsf{poly}(d/\varepsilon)}$
- *w.h.p. over* $z \sim \mathcal{N}(0, \mathbf{I}_{m_0' \times m_0'})$*:* $[\mathcal{S}^\star_{\ell,j}(\mathbf{U}z)]_r - \frac{1}{\mathsf{poly}(d/\varepsilon)} \leq [\mathcal{S}_{\ell,j}(z)]_r \leq [\mathcal{S}^\star_{\ell,j}(\mathbf{U}z)]_r$.

Recursing Lemma G.10 for layers $\ell = 2, 3, \dots, L$ finishes the proof of Theorem 4.1.

*Remark* G.11. Suppose we learn the $\ell$-th layer to accuracy $\varepsilon_\ell$, then the recursion above requires $\varepsilon_\ell \geq \varepsilon_{\ell-1} \cdot \mathsf{poly}(d/\varepsilon)$. This is why we can only support constantly many layeres $L = O(1)$.

To be precise, this also requires us to change the smoothing parameter $\zeta = \frac{1}{\mathsf{poly}(d/\varepsilon)}$ in $\widetilde{\mathsf{ReLU}}$ and LeakyReLU: the lower layers need to use a smaller $\zeta_\ell$ comparing to higher layers. However, in this paper, we have slightly abused the notation and been using the same $\zeta$ throughout.